\documentclass[lettersize,journal]{IEEEtran}

\usepackage[utf8]{inputenc} 
\usepackage[T1]{fontenc}    
\usepackage{hyperref}       
\usepackage{url}            
\usepackage{booktabs}       
\usepackage{amsfonts}       
\usepackage{nicefrac}       
\usepackage{microtype}      
\usepackage{xcolor}         
\usepackage[numbers]{natbib}
\usepackage{pifont}
\usepackage{graphicx}
\usepackage{svg}
\usepackage{bm}
\usepackage{subcaption}
\usepackage{float}
\usepackage{multirow}
\usepackage{amssymb}
\usepackage[utf8]{inputenc}
\usepackage[dvipsnames]{xcolor}
\usepackage{array}
\usepackage{arydshln}
\usepackage{graphicx}
\usepackage{tikz}
\usepackage{svg}
\usepackage{amsmath}
\usepackage{algorithm}
\usepackage{algpseudocode}
\usepackage{booktabs} 

\usepackage{caption} 
\newcommand{\xmark}{\ding{55}}%

\title{Environment-Aware Satellite Image Generation with Diffusion Models}
\author{
\begin{tabular}{cccc}
\IEEEauthorblockN{Nikos Kostagiolas} &
\IEEEauthorblockN{Pantelis Georgiades} &
\IEEEauthorblockN{Yannis Panagakis} &
\IEEEauthorblockN{Mihalis A. Nicolaou}\\

\IEEEauthorblockA{\small CaSToRC} & \IEEEauthorblockA{\small CARE-C} & \IEEEauthorblockA{\small Dept. of Informatics \& Telecommunications} & \IEEEauthorblockA{\small CaSToRC}\\
\small Cyprus Institute & \small Cyprus Institute & \small National and Kapodistrian University of Athens & \small Cyprus Institute\\
\small Nicosia, Cyprus 2121 & \small Nicosia, Cyprus 2121 & \small Athens, Greece 16122 & \small Nicosia, Cyprus 2121\\
\small n.kostagiolas@cyi.ac.cy & \small p.georgiades@cyi.ac.cy & \small yannisp@di.uoa.gr & \small m.nicolaou@cyi.ac.cy\\

\end{tabular}
}

\begin{document}

\maketitle

\begin{abstract}
Diffusion-based foundation models have recently garnered much attention in the field of generative modeling due to their ability to generate images of high quality and fidelity. Although not straightforward, their recent application to the field of remote sensing signaled the first successful trials towards harnessing the large volume of publicly available datasets containing multimodal information. Despite their success, existing methods face considerable limitations: they rely on limited environmental context, struggle with missing or corrupted data, and often fail to reliably reflect user intentions in generated outputs. 
In this work, we propose a novel diffusion model conditioned on environmental context, that is able to generate satellite images by conditioning from any combination of three different control signals: a) text, b) metadata (both static and dynamic), and c) visual data. In contrast to previous works, the proposed method is i) to our knowledge, the first of its kind to condition satellite image generation on dynamic environmental conditions as part of its control signals (e.g., wind, precipitation, solar radiation), and ii) incorporating a metadata fusion strategy that models attribute embedding interactions to account for partially corrupt and/or missing observations. Our method outperforms previous methods both qualitatively (robustness to missing metadata, higher responsiveness to control inputs) and quantitatively (higher fidelity, accuracy, and quality of generations measured using 6 different metrics) in the trials of single-image and temporal generation. The reported results support our hypothesis that conditioning on environmental context can improve the performance of foundation models for satellite imagery, and render our model a promising candidate for usage in downstream tasks. The collected 3-modal dataset is to our knowledge, the first publicly-available dataset to combine data from these three different mediums.
\end{abstract}

\section{Introduction}

\looseness-1Recent advances in AI for Earth Observation have immense potential in addressing  global environmental challenges \citep{climatechange}, with applications spanning methane source detection \citep{ognet,oil}, forest carbon quantification \citep{forest_carbon}, extreme weather prediction \citep{weather}, and crop monitoring \citep{crop1,crop2}. Traditionally, these applications have relied heavily on supervised learning approaches, which require large amounts of labeled data and often struggle with generalization across diverse conditions. This data dependency is particularly challenging in earth observation, where acquiring labeled datasets is expensive, time-consuming, and often requires domain expertise \citep{cell}. Furthermore, the diversity of Earth Observation data, spanning different sensors, resolutions, geographic locations, and temporal conditions makes it more challenging to build models that generalize well across all scenarios.

\looseness-1A solution to this data and computational bottleneck has emerged through the widespread adoption of pre-trained models. Leveraging pre-trained models has led to significant reductions in labeled data requirements \citep{chen2020simpleframeworkcontrastivelearning} while enhancing generalization capabilities, even in out-of-distribution scenarios \citep{hendrycks2019usingselfsupervisedlearningimprove}. Scaling up either the data or the model itself has been shown to directly correlate with substantial performance improvements across a broad range of metrics \citep{kaplan2020scalinglawsneurallanguage,radford2021learningtransferablevisualmodels}. {\it Foundation models}, that is models pre-trained on large unlabeled datasets in a self-supervised manner and then adapted to specific tasks of interest, have recently emerged as a generalist family of models capable of tackling diverse downstream applications, unlocking new levels of performance across various domains (e.g. NLP \citep{bert,gpt}).
Among these, diffusion-based variants have achieved outstanding performance through a series of non-linear transformations that convert noise into meaningful data representations. These models have demonstrated success across multiple domains, including realistic image generation \citep{glide,unknown,hierarchical,article}, art style transformation \citep{Wallace_2023_CVPR,zhang2023inversionbasedstyletransferdiffusion}, super-resolution \citep{wu2023hsrdiffhyperspectralimagesuperresolutionconditional,li2021srdiffsingleimagesuperresolution,saharia2021imagesuperresolutioniterativerefinement}, and video generation \citep{ldm3}.


\looseness-1Despite earth observation being a natural candidate for foundation model applications, the field has only recently begun to benefit from these advances, as most foundation models remain focused on natural image generation. This limited adoption stems from the unique characteristics that distinguish satellite imagery from natural RGB images used in typical computer vision tasks. Remote sensing images present several distinct challenges: they are captured from overhead perspectives with varying spatial resolutions, cover geographically diverse regions (e.g. ranging from urban and rural environments to forests and seas) spanning different climate zones and land cover types, and often include multiple spectral bands beyond the visible. Unlike natural images, satellite imagery requires models that can maintain geographic and temporal consistency across large-scale areas while accurately representing subtle surface features and environmental conditions. Furthermore, remote sensing datasets introduce additional complexities not commonly encountered in traditional computer vision. These include irregular temporal sampling due to satellite revisit patterns, frequent occlusions from cloud coverage, and the integration of heterogeneous metadata sources (coordinates, timestamps, weather and climate data, elevation maps etc.) The sparse and often incomplete nature of this multimodal information creates significant challenges for effectively leveraging foundation models at scale. These factors collectively necessitate specialized approaches that can handle the unique demands of satellite imagery generation while maintaining the benefits of large-scale pre-training.



Currently, only a handful of diffusion model-based approaches have focused on remote sensing applications \citep{Sebaq_2024,10105619,diffusionsat,DBLP:journals/tgrs/LiuCCZS23,crsdiff}, with most relying primarily on textual descriptions for generation control. While text provides valuable contextual information, it proves inadequate for capturing the complex geographic, atmospheric, and environmental conditions that characterize satellite imagery, often resulting in generated images that fail to accurately reflect specific geographic or temporal contexts.
To address these shortcomings, recent approaches have expanded control mechanisms by incorporating additional image modalities or multi-view representations \citep{crsdiff,scalemae,satmae,dino_mc}, or by integrating associated metadata such as geolocation information \citep{diffusionsat}. 
However, existing approaches suffer from a number of key limitations. Firstly, a narrow selection of metadata types is employed, focusing on static coordinate descriptions and timestamps, hence relying on text descriptions to act as primary control - leading to less accurate details in the resulting images, potentially limiting benefit to downstream tasks. Furthermore, dynamic environmental conditions, that significantly impact the appearance of satellite images, are largely ignored. Finally, the conditioning mechanisms for incorporating metadata representations can not effectively handle missing or corrupt metadata, and fail to reliably generate the target images.
These limitations restrict the practical applicability of existing models, as real-world satellite imagery applications frequently involve incomplete metadata and require sensitivity to environmental conditions such as weather patterns, seasonal variations, and atmospheric phenomena that current approaches fail to capture.

Aiming to address the above limitations, in this paper we provide three key contributions:
\begin{enumerate}
    \item We propose a novel, environment-aware Diffusion Model able to distill information from three radically different control input modalities: a) satellite imagery along with its b) corresponding textual caption, and c) associated metadata of both static (e.g. coordinates) and dynamic (e.g. wind conditions) nature. This model instills greater control over the generation process as evidenced by its state-of-the-art performance both qualitatively and quantitatively across an established selection  of metrics and settings, including time series prediction.
    \item We introduce a simple yet effective approach that can capture interactions between metadata embeddings and, thus, enhance robustness in case of missing or omitted attributes from the overall control signal. Our model is shown to  reliably generate images corresponding to dynamic conditions under missing attributes.
    \item We present the first, to our knowledge, publicly-available dataset -- aggregated from public sources \citep{era5,fmow} -- comprising data from these three aforementioned modalities. 
\end{enumerate}

\section{Related Work}

\subsection{Diffusion Models}

A learning paradigm that has recently emerged and has come to dominate the field of generative modeling is that of diffusion models \citep{diffusion1,diffusion2,diffusion3}. Their popularity is largely due to their ability to generate samples from complex data distributions as well as their applicability across various domains, including speech \cite{speech1,speech2}, 3D geometry \citep{3dgeometry1,3dgeometry2,3dgeometry3}, and graphics \citep{graphics1,graphics2,graphics3}. Via a series of iterative denoising steps, diffusion models are able to transform simple noise distributions into complex ones. The most influential variant of these methods, Latent Diffusion Models (LDMs) \citep{ldm1,ldm2,ldm3} has garnered much attention, due to their ability to work within a compressed latent space, which greatly reduces computational demands while preserving high-quality generation. Additionally, the cross-attention layers in LDMs allow the models to generate content conditioned on various inputs, namely text and bounding boxes. This advancement paved the way for a range of novel applications such as subject customization \citep{subject1,subject2,subject3} and text-to-3D generation \citep{graphics2,text3d1,text3d2}. Additional works over the past two years have also demonstrated the impressive adaptability of these models which, given appropriate finetuning, showcase their ability to generate highly realistic images based on a broad selection of inputs, such as text, sketches, or specific attributes. By appropriately expanding their trainable parameters, Controllable Diffusion Models (CDMs) \citep{controllable1,controllable2,controllable3} enable pre-trained diffusion networks to be equipped with control signals of choice.

\subsection{Generative Models for Satellite Imagery}
The application of generative models to satellite imagery has evolved significantly over the past decade, with each new model class addressing key limitations of its predecessors and enabling new use cases in remote sensing, such as super-resolution, data augmentation, domain adaptation, and synthetic scene generation. Generative Adversarial Networks (GANs) were among the first deep generative models applied to satellite imagery. Early works, such as SRGAN \citep{superresolution2} and its variants \cite{sisr1,sisr2,sisr3,sisr4,sisr5,sisr6,sisr7}, leveraged GANs for super-resolving satellite images, thus enabling the enhancement of low-resolution data from sensors like Sentinel-2 and Landsat. Conditional GAN (cGAN) models such as Pix2Pix \citep{pix2pix} and CycleGAN \citep{cyclegan} became popular for image-to-image translation tasks, such as cloud removal, pan-sharpening, or transforming synthetic aperture radar (SAR) to optical images and vice versa \citep{cyclegansar}. As GANs matured, their use expanded to data augmentation in limited-data regimes, where synthetic samples were used to improve model generalization for land cover classification and object detection tasks. Despite these advances, GANs were often criticized for training instability, mode collapse, and difficulty in modeling complex, high-resolution remote sensing scenes. The introduction of Diffusion Models, particularly Denoising Diffusion Probabilistic Models (DDPMs) \citep{diffusion1}, marked a new phase in generative modeling. These models demonstrated remarkable sample quality and diversity, gradually outperforming GANs in image synthesis. In satellite imagery, diffusion models have been used for cloud inpainting, synthetic data generation, and resolution enhancement, offering more stable and diverse outputs than GANs. Moreover, the iterative refinement process of diffusion models better preserves spatial structures crucial for geospatial analysis. This, coupled with the introduction of a novel 3D ControlNet which allowed for providing diffusion models with multimodal control input (text, metadata, or image) support, thus equipping Diffusion Models with more guidance over the generation process, has led to the advent of models like DiffusionSAT \citep{diffusionsat} -- a large-scale generative foundation model trained on high-resolution remote sensing datasets, capable of tasks like temporal generation and super-resolution. Despite being flexible enough, the control input breadth of these models is, still, quite limited, especially as far as their metadata selection is concerned. By extending the control input capabilities of such models to a broader selection of environmental metadata (namely wind and precipitation conditions) this work aims to address these shortcomings while being able to provide better quality and fidelity generations both qualitatively and quantitatively.

\subsection{Multimodal Datasets for Remote Sensing Applications}

There is a great variety of remote sensing tasks that, in some way or form, demand the leveraging of cross-modal data. These tasks are related (but not limited) to a) remote scene classification \citep{ucmerced, aid, resisc}, b) remote image captioning \citep{sydney, rsicd}, c) remote sensing visual question answering (VQA) \citep{rsvqa}, d) multi-label remote scene classification and temporal reasoning \citep{xview2} or e) geolocation-based applications, either using natural \citep{inat, yfcc} or remote sensing images \citep{satclip}. Despite this diversity, most multimodal remote sensing applications remain focused primarily on the established image-text modality pair \citep{rsicd, ucmerced, aid, resisc, sydney, rsvqa}. Even when a third modality is included—usually in the form of associated metadata—it rarely goes beyond location- and time-related information, such as geographic coordinate pairs \citep{inat, yfcc, satclip, xview2, gaia, satlas} and image capture timestamps \citep{xview2, gaia, satlas}. Last but not least, although limited dataset cases that enjoy a great breadth of information in terms of metadata included do exist \citep{mmearth, metaearth} they lack any associated textual-specific information. A detailed overview of the state-of-the-art with respect to multimodal datasets for remote sensing applications can be found in Tbl.\ref{tab:geodatasets}.

\begin{table*}[t]
\centering
\begin{tabular}{c||c|c|c}
                 & \multicolumn{3}{c}{\textbf{Modalities}}                                                                      \\\hline
                 & \textbf{Image} & \textbf{Text}                                    & \textbf{Metadata}                        \\
\textbf{Dataset} &                &                                                  &                                          \\\hline
iNat\citep{inat}  & \checkmark (natural)  & \checkmark (category labels) & \checkmark (coordinates)\\
YFCC100M\citep{yfcc} & \checkmark (natural) & \checkmark (category labels) & \checkmark (coordinates)\\
MP-16            & \checkmark (natural)  & \checkmark (geotag) & \checkmark (coordinates)\\
fMoW\citep{fmow}            & \checkmark (SI)       & \checkmark (category labels) & \checkmark (coordinates + timestamp)\\
S2-100K\citep{satclip} & \checkmark (SI) & \xmark  & \checkmark (coordinates) \\
RSICD\citep{rsicd}           & \checkmark (SI)       & \checkmark (category labels) & \xmark\\
UC-Merced\citep{ucmerced}      & \checkmark (SI)       & \checkmark (category labels)                            & \xmark\\
AID\citep{aid}             & \checkmark (SI)       & \checkmark (category labels)                            & \xmark\\
Resisc-45\citep{resisc}        & \checkmark (SI)       & \checkmark (category labels)                            & \xmark\\
Sydney-Captions\citep{sydney}  & \checkmark (SI)       & \checkmark (scene caption)                              & \xmark\\
TLC              & \checkmark (SI)       & \checkmark (scene caption)                              & \xmark \\
RSVQA\citep{rsvqa}          & \checkmark (SI)       & \checkmark (QA pairs)                                   & \xmark\\
xView2\cite{xview2}           & \checkmark (SI)       & \checkmark (disaster + damage labels)                   & \checkmark (coordinates)\\
GeoPile\cite{geopile}           & \checkmark (SI)       & \xmark                  & \xmark                           \\
SatlasPretrain\cite{satlas}           & \checkmark (SI)       & \checkmark (category labels)                   & \checkmark (coordinates + timestamp)\\
MMEarth\cite{mmearth} & \checkmark (SI) & \xmark & \checkmark (coordinates + timestamp + environmental (limited))\\
GAIA\cite{gaia}           & \checkmark (SI)       & \checkmark (GPT-4o descriptions)                   & \checkmark (coordinates + timestamp)\\       
MetaEarth\cite{metaearth} & \checkmark (SI) & \xmark & \checkmark (coordinates + resolution)\\
\textbf{Ours}             & \checkmark (SI)       & \checkmark (scene caption) & \checkmark (coordinates + timestamp + environmental)\\
\end{tabular}
\caption{Overview of existing multimodal datasets tailored to satellite imagery applications, including the one introduced in this study. Notably, even in the rare cases where three modalities are present, prior datasets have been limited to information about location and time. In contrast, our dataset is the first to accompany image and text information with environment-based metadata of both static and dynamic nature.}
\label{tab:geodatasets}
\end{table*}

\subsection{multimodal Data Fusion for Foundation Models with Contextual Variables}

\looseness-1Equipping foundation models with contextual information is a relatively recent advancement. Models such as GeoCLIP \citep{geoclip} and SatCLIP \citep{satclip}  learn joint embeddings between  geographic coordinates and satellite images - either natural (in the case of GeoCLIP) or remote sensing (in the case of SatCLIP). This is achieved by learning a shared location-to-image feature space using a CLIP-style contrastive learning objective.

While recent models like GeoCLIP and SatCLIP have advanced the integration of geographic context information into visual representation learning they primarily rely on location-only metadata. GeoCLIP uses Random Fourier Features to encode ground-level GPS coordinates, while SatCLIP employs spherical harmonics to embed satellite locations from Sentinel-2 data. In contrast, more recent studies \citep{diffusionsat} have sought to broaden the context variable scope by incorporating not only geographical but also temporal and atmospheric metadata leveraged through sensors, such as capture time, ground sampling distance (GSD), and cloud coverage, to enhance cross-domain alignment and build richer, more generalizable representations. This marks a significant step toward multimodal, context-aware satellite image modeling, expanding beyond traditional image-location pairings. Instead of naively incorporating each numerical metadata item $\bm{k}_{j}, j \in \{1, ..., M\}$, into the text caption -- a strategy that has shown to fare rather poorly when used in conjunction with continuous values \citep{radford2021learningtransferablevisualmodels} -- \cite{diffusionsat} opt to encode the metadata using the same sinusoidal timestep embedding employed by diffusion models:
\begin{equation}
\begin{aligned}
\text{Project}(\bm{k}, 2i) = \sin(\bm{k}\Omega^{-\frac{2i}{d}}), \\
\text{Project}(\bm{k}, 2i + 1) = \cos(\bm{k}\Omega^{-\frac{2i}{d}})
\end{aligned}
\end{equation}

where $\bm{k}$ is the metadata or timestep value, $i$ is the index of feature dimension in the encoding, $d$ is the dimension, and $\Omega$ is a large constant. Prior to being encoded via sinusoidal projection each metadata value $\bm{k}_{j}$ is first normalized to a value between 0 and 1000 (since the diffusion timestep $t \in \{0, ..., 1000\})$. Then, a different MLP for each metadatum encodes the projected metadata value identically to the diffusion timestep $t$ \citep{diffusion1} as follows:

\begin{equation}
f_{\theta_{j}}(\bm{k}_{j}) = MLP([Project(\bm{k}_{j},0), ..., Project(\bm{k}_{j}, d)])
\end{equation}

where $f_{\theta_{j}}$ represents the learned MLP embedding for metadata value $\bm{k}_{j}$, corresponding to metadata type $j$ (e.g. longitude). The embedding is then $f_{\theta_{j}}(\bm{k}_{j}) \in \mathbb{R}^{D}$, where $D$ is the embedding dimension. Finally, \cite{diffusionsat} proceed to add the $M$ metadata vectors together $m = f_{\theta_{1}}(\bm{k}_{1}) + ... + f_{\theta_{M}}(\bm{k}_{M})$, where $m \in \mathbb{R}^{D}$, which is then also added with the embedded timestep $t = f_{\theta}(t) \in \mathbb{R}^{D}$, so that the final conditioning vector is $c = m + t$. During training they also opt to randomly zero out the metadata vector $m$ with a probability of 0.1 so that the model remains capable of properly generating an image even when metadata are unavailable.

\section{Methodology}
\label{sec:method}

In this section we describe our methodology for a) metadata-and-text-to-satellite-image dataset curation (Section \ref{metadata_acquisition_handling}), b) diffusion-based satellite imagery generation with conditioning and (Section \ref{contextual_variable_control}), c) control signal conditional generation  (Section \ref{temporal_variable_control}).

\subsection{Dataset Acquisition}

\label{metadata_acquisition_handling}

Since no equivalent of a large, image-text-metadata dataset for satellite imagery data is readily available, we seek to address this gap by contributing a large, high-resolution generative dataset for satellite images based on text and metadata annotations that we compile for the \textbf{fMoW} dataset. fMoW: Function Map of the World \citep{fmow} is an extensive (> 1,000,000 images), global (> 200 countries), high-resolution (GSD 0.3m-1.5m) satellite image dataset. Each image includes location- and time-specific metadata, along with a label indicating one of 62 distinct object categories depicted. We centrally crop each image to 512x512 pixels. Along with the location- and time-specific metadata provided in the fMoW dataset, our approach incorporates 7 more climate-specific contextual variables. In line with previous studies \citep{diffusionsat}, we generate a caption for each image based on the semantic class and country code, formatted as follows: {\fontfamily{qcr}\selectfont
"a satellite image [of a <object>] [in <country>]"}. 

We follow by detailing our metadata acquisition strategy, aimed at providing greater control over the generation process and, consequently, enhancing the accuracy of our model's outputs. Established metadata sets up to this study typically encapsulate (1) static (i.e. non-evolving over time) satellite image information (e.g., \textit{latitude} / \textit{longitude} coordinates); (2) first-level information affecting overall image attributes (e.g., \textit{ground sampling distance} and \textit{total cloud cover}); and (3) data that only indirectly influences second-level attributes, such as flora state and sunlight levels (e.g., \textit{year}, \textit{month}, and \textit{day}). To address these limitations, we introduce 6 novel climate-specific metadata fields (\textit{2m temperature}, \textit{total precipitation}, \textit{10m U wind component}, \textit{10m V wind component}, \textit{surface net solar radiation}, and \textit{2m dewpoint temperature}) that aim to enhance metadata-specific control signals, allowing for more precise generation guidance not only on image- but also on a pixel-level scale. Temperature and precipitation are important climatic controls on vegetation growth, phenology, and land surface conditions, while wind patterns affect atmospheric clarity and moisture transport, which can adversely affect image quality and cloud formation. Solar radiation is also important for photosynthetic activity and vegetation growth and, lastly, dewpoint temperature provides atmospheric moisture information that can affect surface conditions and vegetation stress~\citep{Currier2022}. Any novel metadata as well as the already established \textit{total cloud cover} metadatum were acquired from the ERA-5 \citep{era5} dataset. ERA5 is the fifth generation ECMWF reanalysis for the global climate and weather for the past 8 decades. It provides hourly estimates for a large number of atmospheric, ocean-wave, and land-surface quantities in a gridded format, each subgrid of which covers a $\sim$25 x 25~km area (at the equator) and has hourly temporal resolution. Data were retrieved from the Copernicus Data Store (CDS) using the python \textit{cdsapi} in NetCDF format. The high spatial resolution of ERA5 (0.25$^{\circ}$ x 0.25$^{\circ}$) was a key reason for its selection, as well as its proven track record as the current state-of-the-art global climate reanalysis. The ERA5-Land dataset (spatial resolution of $\sim$9~km x 9~km) was not utilized, as it is a land-only dataset that explicitly masks out all oceanic areas, including coastlines and inland waters~\citep{Hassler2021}. Overall, our method aims to bring more control into the generation process by expanding its metadata-specific control input capabilities beyond that of previous studies.

\begin{figure*}[t!]
    \centering
    \begin{subfigure}[t]{0.45\textwidth}
        \centering
        \includegraphics[width=\linewidth]{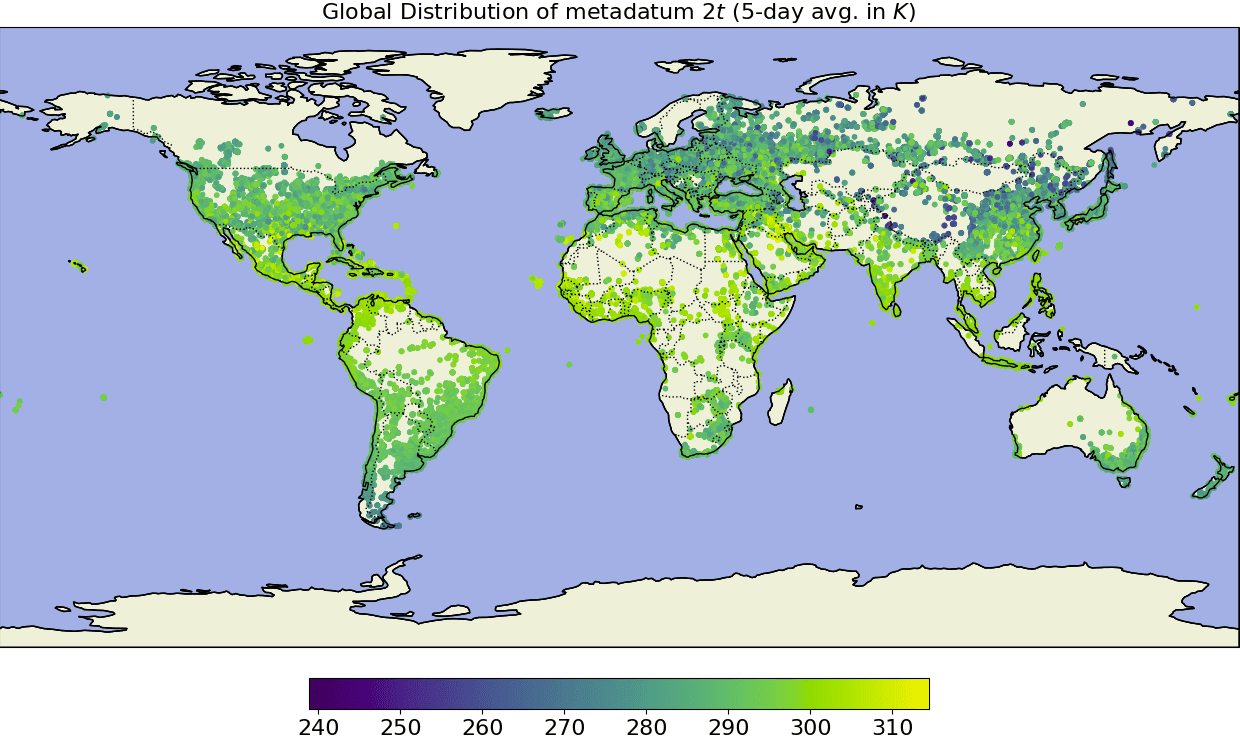}
        \label{fig:2t_global}
    \end{subfigure}%
    ~ 
    \begin{subfigure}[t]{0.45\textwidth}
       \centering
        \includegraphics[width=\linewidth]{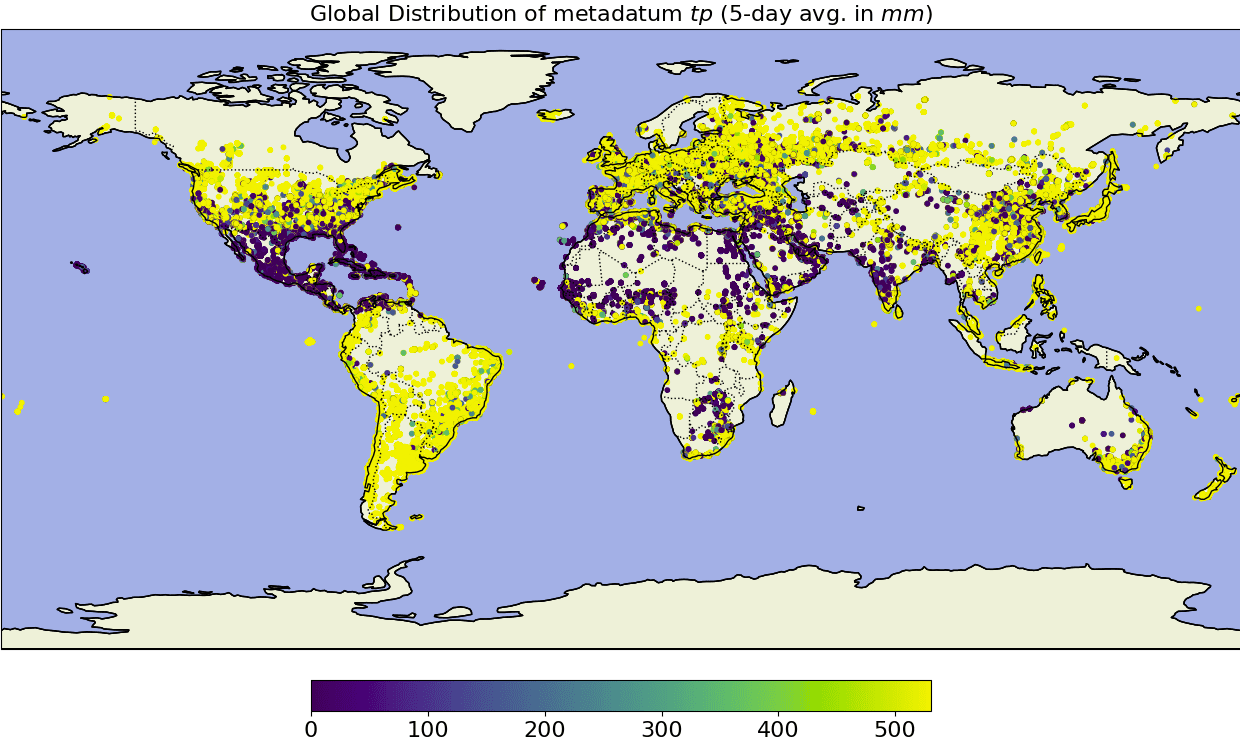}
        \label{fig:tp_global}
    \end{subfigure}\\
    \begin{subfigure}[t]{0.45\textwidth}
        \centering
        \includegraphics[width=\linewidth]{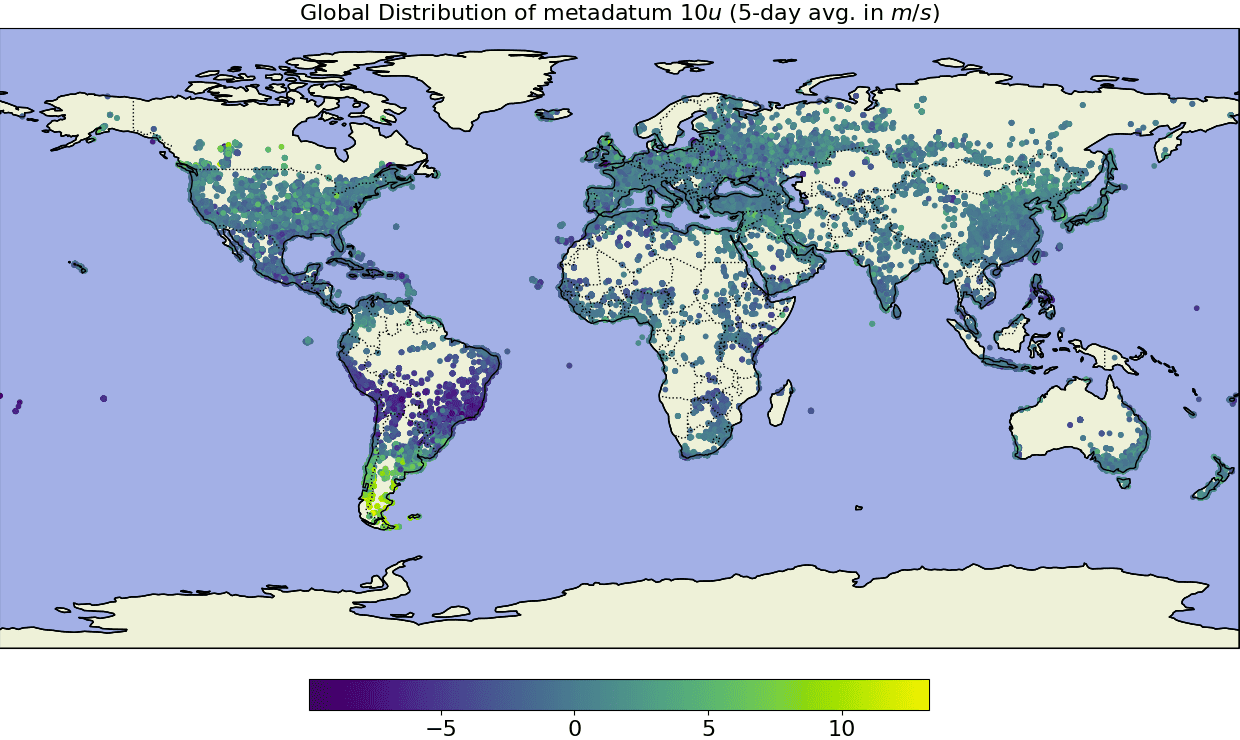}
        \label{fig:u10_global}
    \end{subfigure}
    ~
    \begin{subfigure}[t]{0.45\textwidth}
       \centering
        \includegraphics[width=\linewidth]{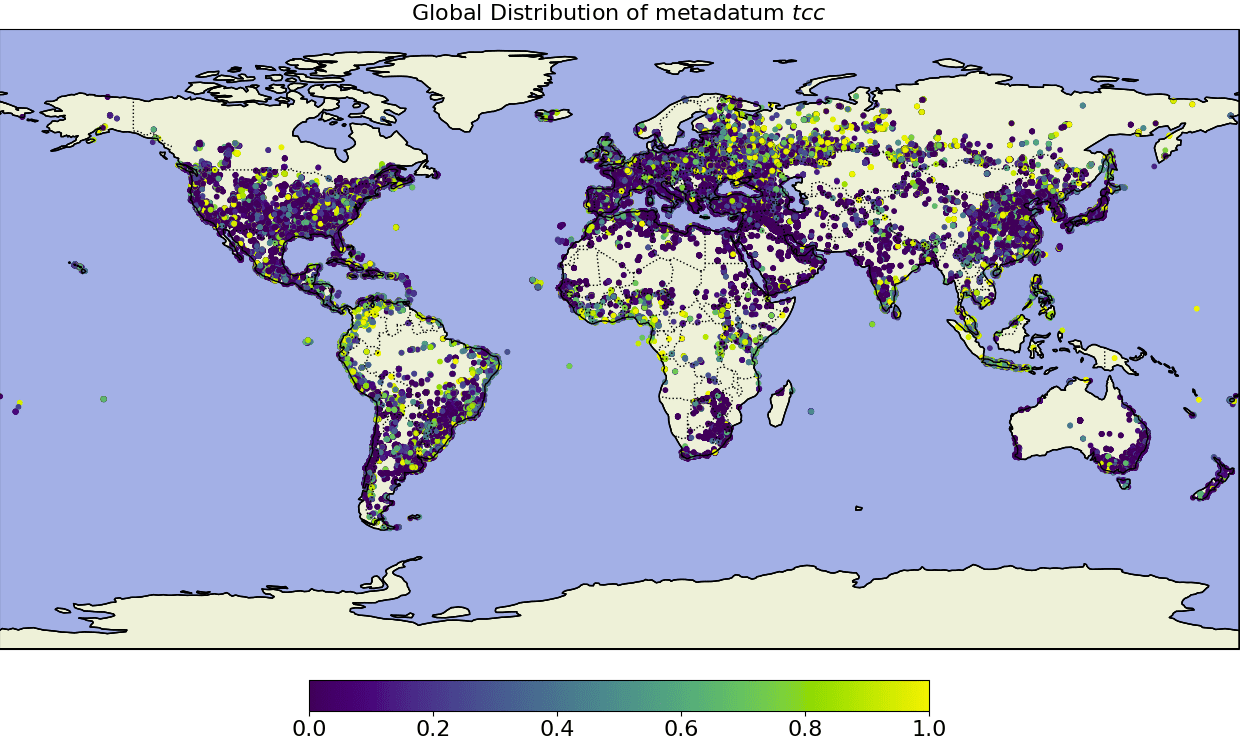}
        \label{fig:tcc_global}
    \end{subfigure}
    \caption{Global distributions of values for the metadata \{$2t$, $tp$, $10u$, $tcc$\} used in this study. Statistics for the rest metadata can be found in Figures \ref{fig:metadata_2}, \ref{fig:metadata_3} of the Appendix.}
    \label{fig:metadata_1}
\end{figure*}

To extract the corresponding metadata for a given satellite image, we first need to identify the ERA-5 grid that corresponds to the image’s location. To do this, we begin by selecting the most recent global hourly estimate relative to the image capture time. Next, we choose the grid cell from the global dataset closest to the object’s center point (i.e., the centroid coordinates) depicted in the image. This task is simplified by the fact that both the ERA-5 grids and the objects in the fMoW satellite image dataset are available in geometry format, making it straightforward to assign satellite image locations to ERA-5 grids using a simple geo-database query. However, since landscape appearance tends to be influenced by environmental changes over time rather than by single snapshots, instantaneous measurements alone may not provide optimal conditions for predicting satellite imagery. To address this, we label each image with 5-day aggregated values for each of our dynamic (i.e. non-constant) metadata variables, with the exception of \textit{total cloud cover} —a variable that directly (i.e. in a non-cumulative way) affects satellite image appearance, thus justifying using it unaggregated. More information regarding the fMoW image - ERA-5 grid matching process can be found in Fig. \ref{fig:metadata_acquisition} and in the Algorithm \ref{algo:1} of the Appendix. Information regarding a) the types of metadata we use, b) their availability in similar previous studies, and c) the aggregation strategies we used, when applicable can be found Tbl. \ref{tab:metadata}. Lastly, distributions of the metadata used in this study with respect to their global coordinates can be browsed in Figures \ref{fig:metadata_1}, \ref{fig:metadata_2}, and \ref{fig:metadata_3}.

\begin{figure*}
    \centering    \includegraphics[width = 0.8\textwidth]{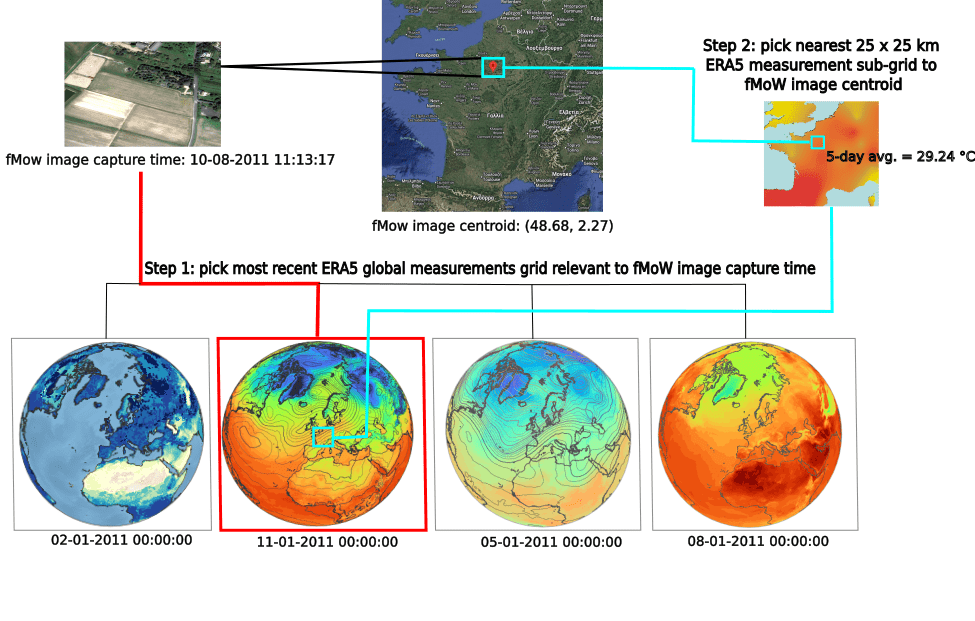}
    \caption{Details of our metadata acquisition strategy. We begin by matching the fMoW satellite image of interest to its most recent ERA5 global hourly estimate grid. We follow that by picking the nearest ERA5 subgrid to our image centroid's coordinates. Lastly, depending on the metadata, we perform any necessary aggregations -- namely a 5-day averaging as in the case of acquiring 2m temperature measurements, depicted here. The same process translates to every metadata type without loss of generality}
    \label{fig:metadata_acquisition}
\end{figure*}

\begin{table}[]
\centering
\begin{tabular}{c|c|c|c|c}
\textbf{Metadata Name}      & \textbf{\cite{diffusionsat}} & \textbf{\cite{metaearth}} & \textbf{ERA-5 Code} & \textbf{Aggregation} \\
\hline
longitude                   &   \checkmark & \checkmark         & N/A                 & N/A                       \\
latitude                    & \checkmark &   \checkmark         & N/A                 & N/A                       \\
year                        & \xmark &  \checkmark         & N/A                 & N/A                       \\
month                       &  \xmark & \checkmark         & N/A                 & N/A                       \\
day                         & \xmark &  \checkmark         & N/A                 & N/A                       \\
gsd                         & \checkmark &  \checkmark         & N/A                 & N/A                       \\
2m temperature              & \xmark &  \xmark             & 2t                  & 5-day avg.                \\
Total precipitation         & \xmark &  \xmark             & tp                  & 5-day avg.                \\
10m U wind component        & \xmark &  \xmark             & 10u                 & 5-day avg.                \\
10m V wind component        & \xmark &  \xmark             & 10v                 & 5-day avg.                \\
Surface net solar radiation & \xmark &  \xmark             & ssr                 & 5-day sum                 \\
Total cloud cover           &  \xmark & \checkmark         & tcc                 & N/A                       \\
2m dewpoint temperature     & \xmark &  \xmark             & 2d                  & 5-day avg.
\end{tabular}
\caption{Details on control input metadata employed by our method including a) their names, b) their inclusion in prior studies, c) their ERA-5 codes, and d) the type of aggregation that we perform to bring causality into play, when applicable. Metadata that are either not leveraged from the ERA-5 dataset or not aggregated are given N/A labels in the corresponding columns.}
\label{tab:metadata}
\end{table}

\subsection{Satellite image generation with Contextual Variable Control}
\label{contextual_variable_control}
Our first goal is to train a LDM to be able to generate \textit{single} satellite images of great accuracy and fidelity given an input text prompt and/or a selection of contextual variables in the form of geographical, temporal, and environmental metadata. Following the approach of prior studies we, therefore, consider datasets where each image $\bm{x} \in \mathbb{R}^{C\times H \times W}$ is paired with a text caption $\tau$, as well as a selection of numerical metadata $\bm{k} \in \mathbb{R}^M$, where $M$ is the number of distinct metadata items. We thus want to learn the data distribution $p(\bm{x}|\tau,\bm{k})$. Furthermore, our model should be capable of generating a satellite image sample of high quality regardless of whether either $\tau$ is poor or missing and/or any combination of metadata in $\bm{k}$ are missing or corrupted. To this end, we follow already established metadata handling approaches while providing extensions when necessary.


While intuitive, fusing metadata embeddings by addition precludes modeling interactions between attributes thus the resulting models have limited robustness to missing and/or corrupted data (for more information about this, refer to Section \ref{sec:experiments}), therefore arguably rendering the model useful only in cases in which either a) the metadata are available in their entirety, or b) all the metadata are unavailable (thus having the conditioning of the model default to the original Stable Diffusion conditioning of $c = t$, since the $m$ vector is zeroed out). To support partial metadata availability, a more expressive fusion strategy is required.
In this light, we firstly concatenate the $M$ metadata vectors as follows:
\[m' = concat(f_{\theta_{1}}(\bm{k}_{1}), ...,  f_{\theta_{M}}\bm{k}_{M}))\]
Clearly, the dimensionality of our $m$ vector increases linearly with respect to the number of different metadata types, $j$. Thus, in our case, $m' \in \mathbb{R}^{M  \times D}$. To match dimensionality and further capture interactions between the embeddings we train an MLP $g$ that projects the concatenated vector $m' \in \mathbb{R}^{M \times D}$ to the $\mathbb{R}^{D}$ space: 
\[g_{\phi}(m') = g_{\phi}(\text{concat}(f_{\theta_{1}}(\bm{k}_{1}), ...,  f_{\theta_{M}}(\bm{k}_{M}))) \in \mathbb{R}^{D}\]
To incorporate the conditioning vector, we firstly encode the corresponding image $\bm{x} \in \mathbb{R}^{C\times H \times W}$ using the SD variational autoencoder (VAE) \citep{ldm1,esser2021tamingtransformershighresolutionimage} to the latent representation $\bm{z} = \mathcal{E}(\bm{x}) \in \mathbb{R}^{C' \times H' \times W'}$. Gaussian noise is then added to the latent image features, thus giving the latent representation the following form: $\bm{z}_{t} = \alpha_{t}\bm{z} + \sigma_{t}\epsilon$. The conditioning vector $\bm{c}$, created from the metadata embedding and the diffusion timestep, paired with the CLIP-embedded text caption $\tau' = \mathcal{T}_{\theta}(\tau)$ are passed through a DM $\epsilon_{\theta}(\bm{z}_{t};\tau',\bm{c})$ to predict the added noise. Lastly, the denoised latents are upsampled to full resolution by the VAE decoder $\mathcal{D}$.

\begin{figure*}
    \centering
    \includegraphics[trim={0 2.5cm 0 0},clip, width=\textwidth]{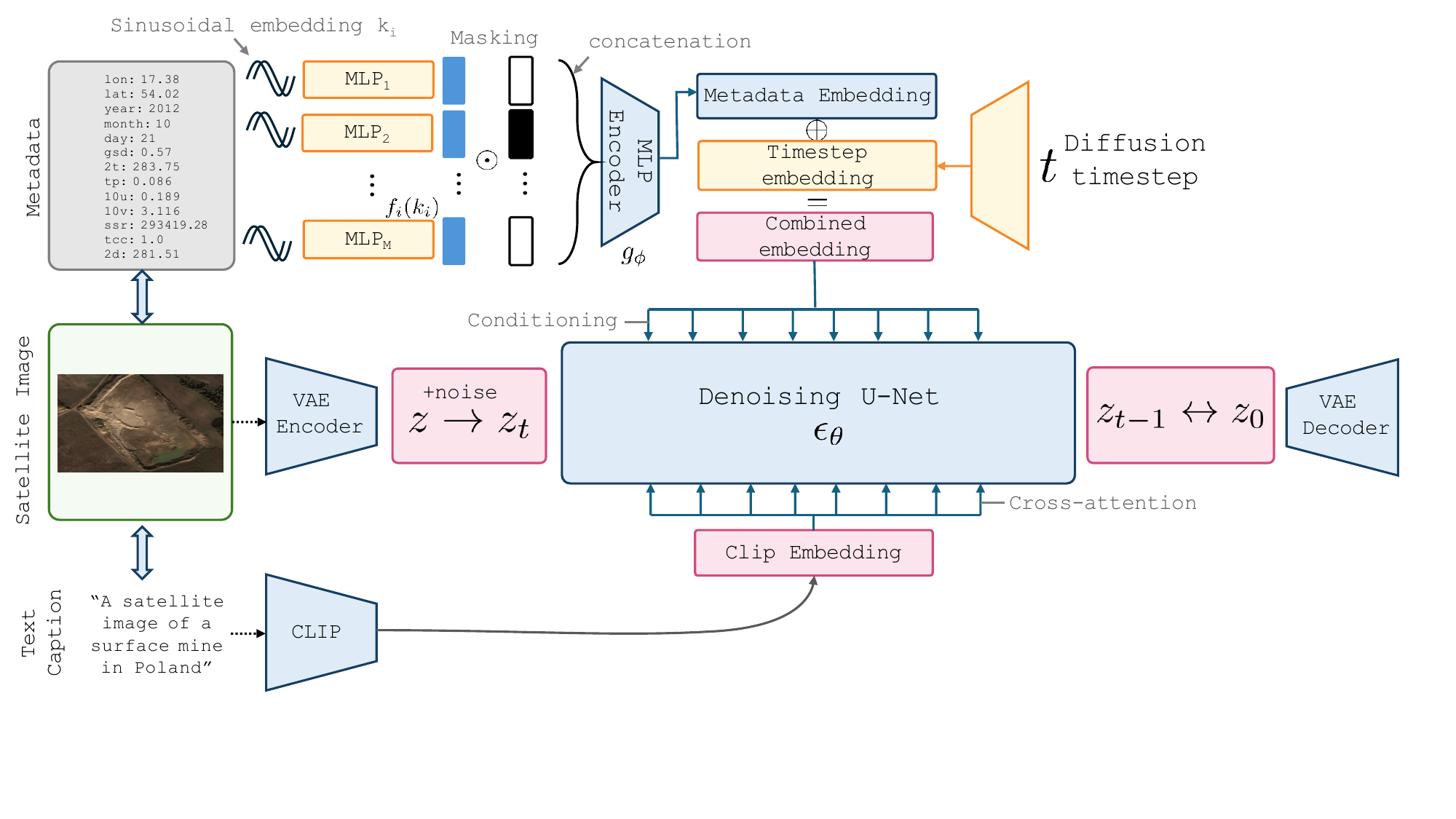}
    \caption{Overview of our model's architecture}
    \label{fig:overall}
\end{figure*}

During training we only update a) the denoising UNet $\epsilon_{\theta}$ and b) the metadata and the timestep embeddings $f_{\theta_{j}}$. Every other component - the encoder $\mathcal{E}$, the decoder $\mathcal{D}$, the CLIP text encoder $\mathcal{T}_{\theta}$ - remain frozen. Each component is initialized using SD 2.1's weights. A drop-out mechanism is also implemented for metadata embeddings by zeroing-out the complete, concatenated metadata vector $\bm{m}$ or a random subset of the metadata vectors prior to concatenation, each with a probability of 0.1. In more detail, for each metadata type $j$, we sample a binary mask $m_j \sim \text{Bernoulli}(0.9)$ and compute the masked embedding as:
$\tilde{f}_{\theta_j}(k_j) = m_j \cdot f_{\theta_j}(k_j)$.
The final concatenated metadata vector becomes:
$m' = \text{concat}(\tilde{f}_{\theta_1}(k_1), ..., \tilde{f}_{\theta_M}(k_M))$.


Moreover, to ensure that our model is capable of generating quality images even when conditioned on generic textual descriptions we also randomly and independently drop parts of the caption, again at a 10\% rate (see Section \ref{metadata_acquisition_handling}, where dropped words of the caption are  denoted by square brackets).

\subsection{Satellite image generation with Temporal Contextual Variable Control}
\label{temporal_variable_control}
Next, we extend our approach to temporal generation tasks. While our single-image model can generate high-quality satellite images from text prompts and metadata, we can leverage its pretrained weights as a foundation for more complex conditional generation. Specifically, we now consider temporal prediction tasks where we condition on a sequence of satellite images $\bm{s}\in \mathbb{R}^{T\times C' \times H' \times W'}$ with their corresponding metadata $\bm{k}_{s} \in \mathbb{R}^{T\times M}$, along with a target caption $\tau$ and target metadata $\bm{k}\in\mathbb{R}^M$. Here, $T$ represents the sequence length, while $C'$, $H'$, and $W'$ denote the dimensions of conditioning images, which may differ from the target image dimensions. The objective is to sample $\tilde{x} \sim p(\cdot|\bm{s};\bm{k}_{s};\tau;\bm{k})$, generating a target image $\tilde{x}$ that is consistent with the temporal sequence $\bm{s}$ and specified target conditions associated with caption $\tau$ and metadata $\bm{k}$.


To tackle frame-by-frame conditional temporal prediction, we adopt a ControlNet based approach. Unlike 2D ControlNet, this study employs 3D zero-convolutions between each StableDiffusion block \citep{controllable2}. Similarly to VideoLDM \citep{videoldm}, conditioning on temporal control signals is further encouraged by the use of temporal attention layers. Lastly, in order to discourage noise from early training stages from affecting our pre-trained weights, an inferred parameter $\alpha_{i}$ is introduced for each block $i$ that "mixes" in the output of the temporal attention layer.

One key advantage of this approach is the ability to assign specific {\it metadata} to each item in the control sequence
$s$. In line with the single-image strategy, each target metadata value is individually projected and embedded using an MLP. Every image-specific metadata vectors are then concatenated and mapped from the $\mathbb{R}^{M \times D}$ space to the lower $\mathbb{R}^{D}$. The embedded metadata for each image is then concatenated with the image and fed into the 2D layers of the ControlNet. This design makes the resulting model invariant to the order of images in the control sequence $s$, as each image’s timestamp in its metadata independently determines its temporal placement. Consequently, a single Stable Diffusion model can be trained to predict images both in the past and future, or to interpolate within the sequence's temporal range. 

\subsection{Implementation \& Training Details}
All models are trained using half-precision and gradient checkpointing, borrowing from the Diffusers \citep{diffusers} library. In the following,  we describe implementation details for the models used for the purposes of this study.

\subsubsection{Contextual Variable Control Model}
In the single-image setting, our codebase builds on the Hugging Face Diffusers\footnote{https://github.com/huggingface/diffusers.git} library and DiffusionSAT \footnote{https://github.com/samar-khanna/DiffusionSat}, introducing improvements regarding the metadata fusion strategy. We also re-train the model introduced in \citep{diffusionsat} with extended metadata while still leveraging the image-caption data to provide fair comparisons and ablation studies.
Since \citep{diffusionsat}'s model offers variants for two image resolutions (512x512 and 256x256 pixels), we trained our model at both resolutions as well, resulting in four distinct models. Every text-and-metadata-to-image model is trained with a batch size of 32 for 100,000 iterations, which we qualitatively determined to suffice for convergence. We choose a constant learning rate of 2e-6 with the AdamW optimizer. For sampling, we use the DDIM \citep{diffusion2} sampler with 100 steps and a guidance scale of 1.0. We generate 10,000 samples on the validation set of fMoW-RGB.

\subsubsection{Temporal Contextual Variable Control Model}
We implement a custom version of the 3D ControlNet for the temporal model from scratch based on the Hugging Face Diffusers  library. We use the 256 single-image models as our priors. We train our 3D ControlNet on sequences of at most 3 distinct conditioning images on a custom, temporal version of the fMoW dataset that we have created for this purpose. Should less than 3 distinct images be available for a given location, we pad the sequence to length with copies of random picks among the already available conditioning image sequence. Samples where there is only 1 image available per location are avoided. Our two temporal models are trained for 40,000 iterations with a learning rate of 4e-4 using the AdamW optimizer. Our sampling configuration matches the one used for the single-image model.

\section{Experiments}
\label{sec:experiments}

In this section, we describe the experiments for the tasks described in section 3. Our experimental results include a) an uncurated collection of single-image generations from our model, b) several visualizations that exhibit our model's improved coverage in terms of prompt and metadata combinations, and c) a series of conditional (temporal) generation results in which we compare our method's predictive capabilities both forwards and backwards in time to those of earlier studies. Quantitative comparisons between our concatenation-and-mapping metadata fusion method and the more straightforward, additive approach followed by earlier studies \citep{diffusionsat} are also provided. For single image generation (Subsection \ref{single_image_experiment_subsection}), we report standard visual-quality metrics such as FID \citep{fid}, Inception Score (IS), and CLIP-score \citep{radford2021learningtransferablevisualmodels}. For temporal generation (Subsection \ref{temporal_image_experiment_subsection}), we report pixel-quality metrics including SSIM \citep{ssim}, PSNR, LPIPS \citep{lpips} with VGG features, given a reference ground-truth image. LPIPS is a more relevant perceptual quality metric used in evaluating satellite images. The metrics reported are evaluated on a sample size of 10,000 images. We further back our approach by compiling a series of qualitative experiments that demonstrate our method's improved metadata fusion strategy compared to earlier approaches, as well as its broader coverage and enhanced input-output consistency in terms of control signals. 

\subsection{Satellite image generation with Contextual Variable Control}
\label{single_image_experiment_subsection}
\begin{figure*}
\renewcommand\thesubfigure{\roman{subfigure}}
    \centering
    \begin{subfigure}[t]{0.12\textwidth}
       \centering
        \includegraphics[width=\textwidth]{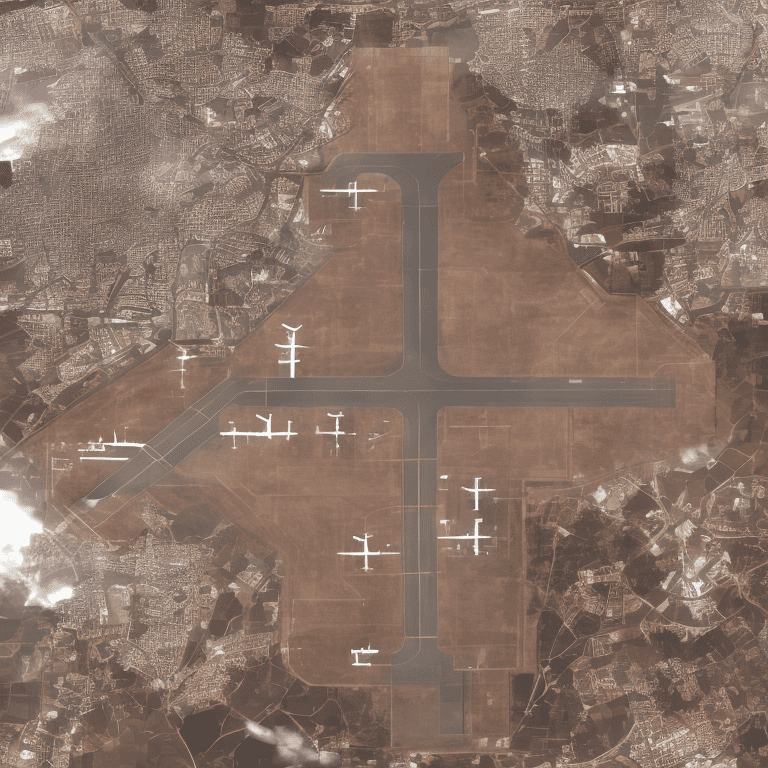}
        \caption{Airport}
    \end{subfigure}
    \begin{subfigure}[t]{0.12\textwidth}
       \centering
        \includegraphics[width=\textwidth]{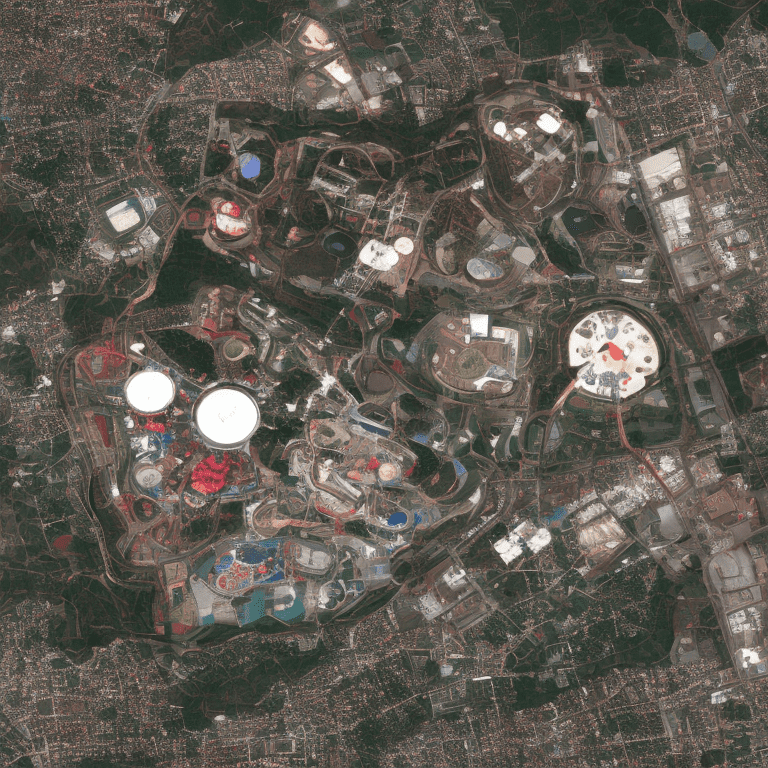}
        \caption{Amusement Park}
    \end{subfigure}
    \begin{subfigure}[t]{0.12\textwidth}
       \centering
        \includegraphics[width=\textwidth]{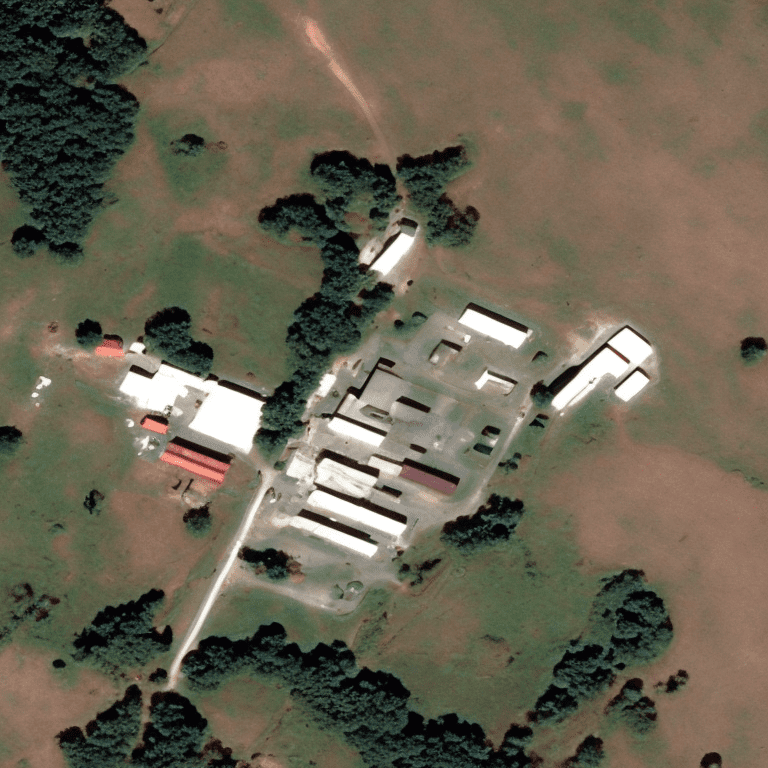}
        \caption{Barn}
    \end{subfigure}
    \begin{subfigure}[t]{0.12\textwidth}
       \centering
        \includegraphics[width=\textwidth]{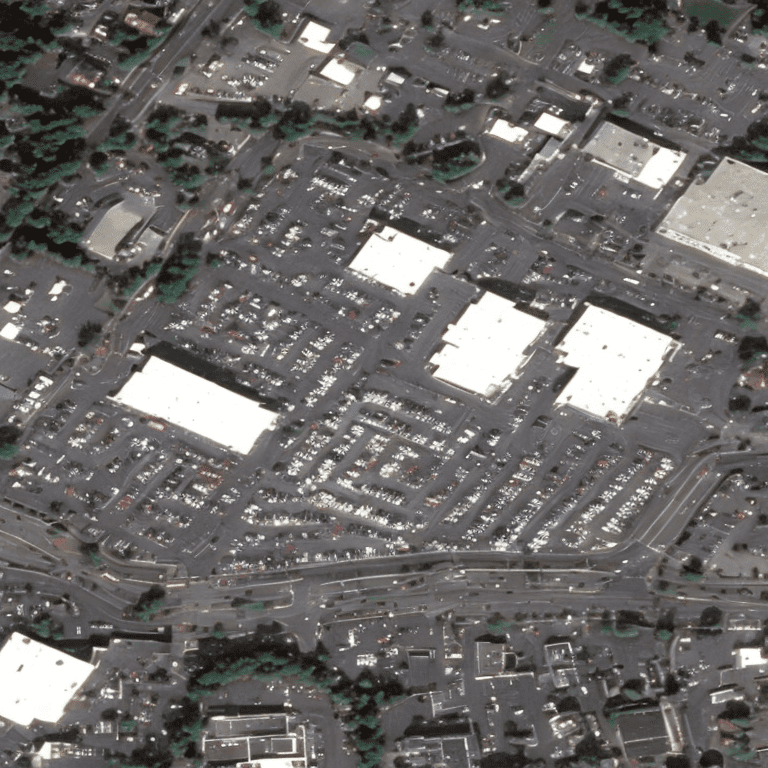}
        \caption{Car Dealership}
    \end{subfigure}
    \begin{subfigure}[t]{0.12\textwidth}
       \centering
        \includegraphics[width=\textwidth]{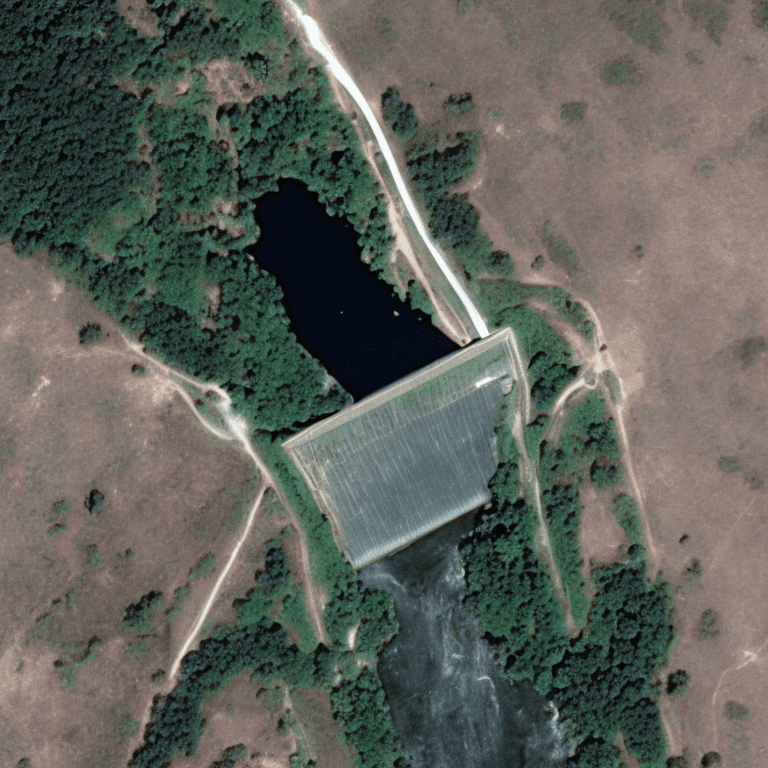}
        \caption{Dam}
    \end{subfigure}
    \begin{subfigure}[t]{0.12\textwidth}
       \centering
        \includegraphics[width=\textwidth]{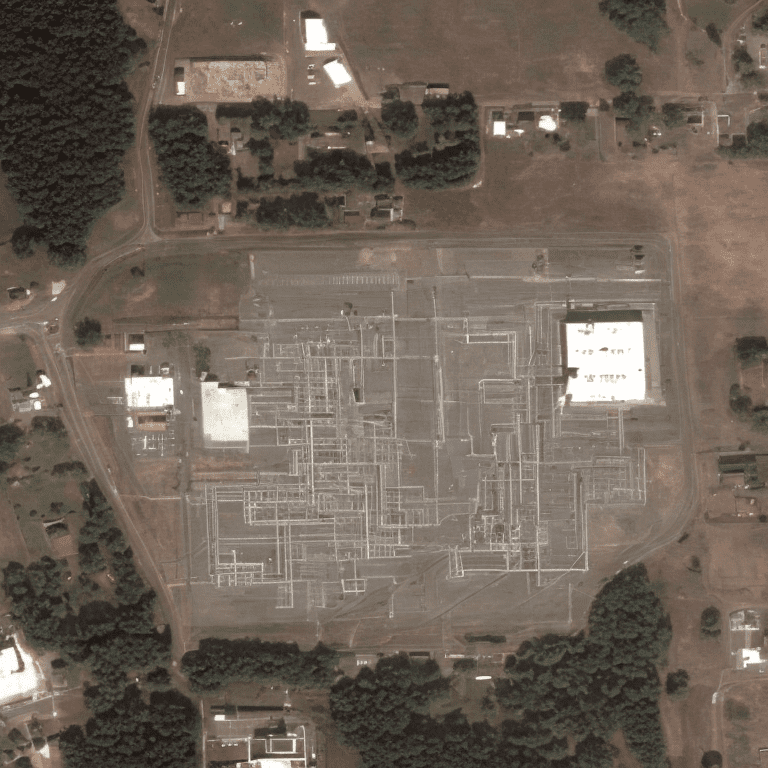}
        \caption{Elec. Substation}
    \end{subfigure}
    \begin{subfigure}[t]{0.12\textwidth}
       \centering
        \includegraphics[width=\textwidth]{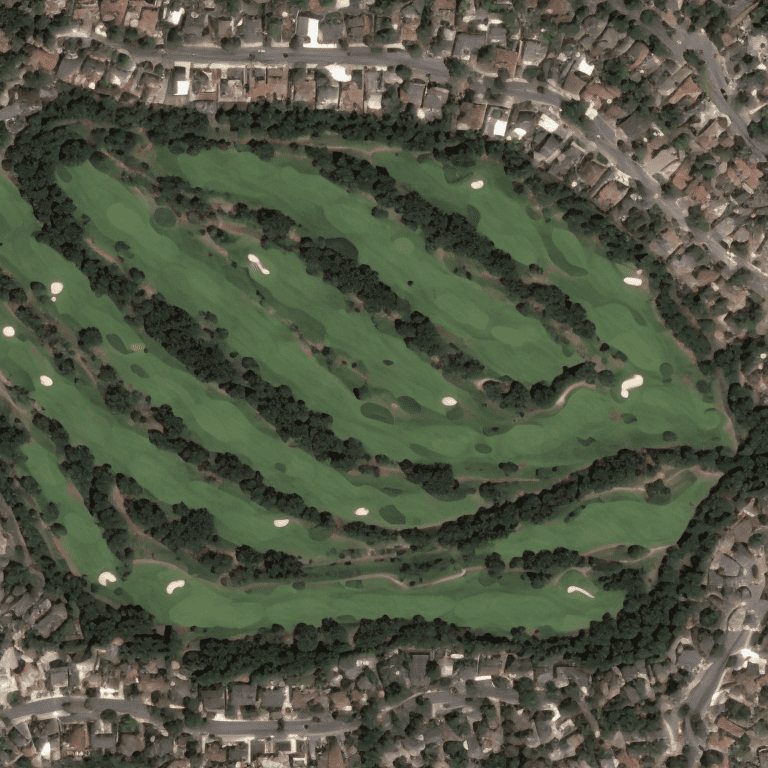}
        \caption{Golf Course}
    \end{subfigure}\\
    \begin{subfigure}[t]{0.12\textwidth}
       \centering
        \includegraphics[width=\textwidth]{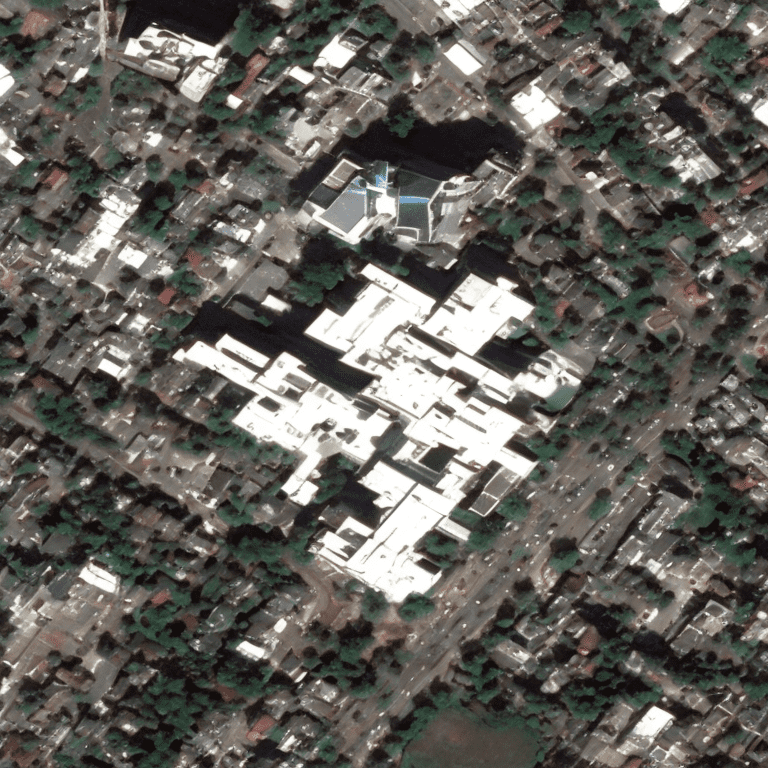}
        \caption{Hospital}
    \end{subfigure}
    \begin{subfigure}[t]{0.12\textwidth}
       \centering
        \includegraphics[width=\textwidth]{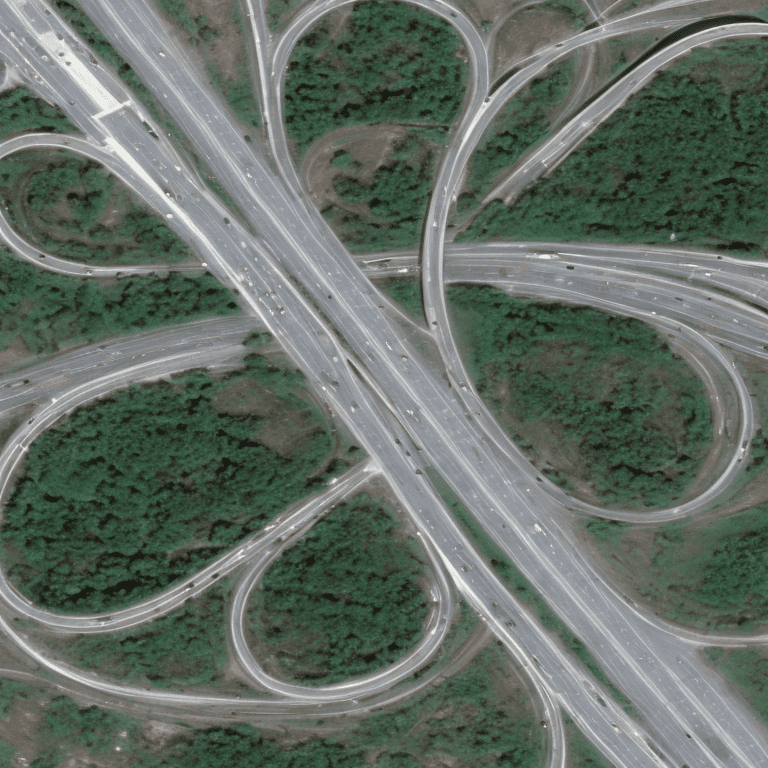}
        \caption{Interchange}
    \end{subfigure}
    \begin{subfigure}[t]{0.12\textwidth}
       \centering
        \includegraphics[width=\textwidth]{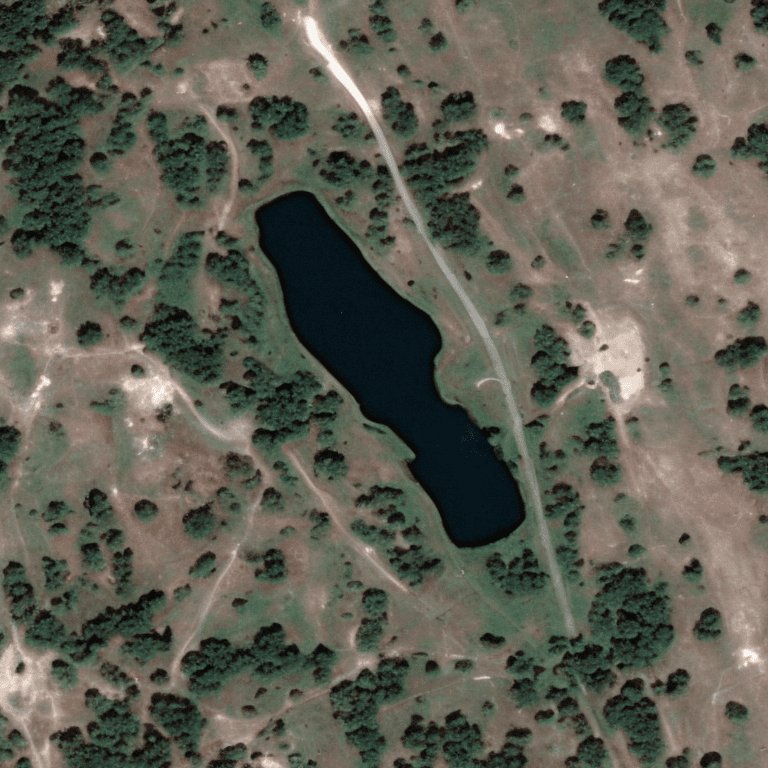}
        \caption{Lake or Pond}
    \end{subfigure}
    \begin{subfigure}[t]{0.12\textwidth}
       \centering
        \includegraphics[width=\textwidth]{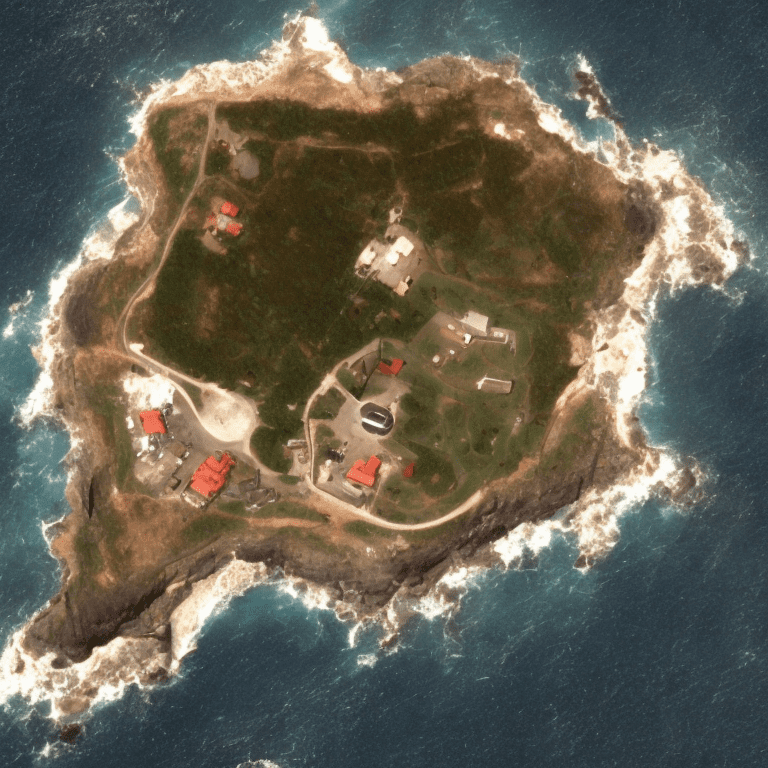}
        \caption{Lighthouse}
    \end{subfigure}
    \begin{subfigure}[t]{0.12\textwidth}
       \centering
        \includegraphics[width=\textwidth]{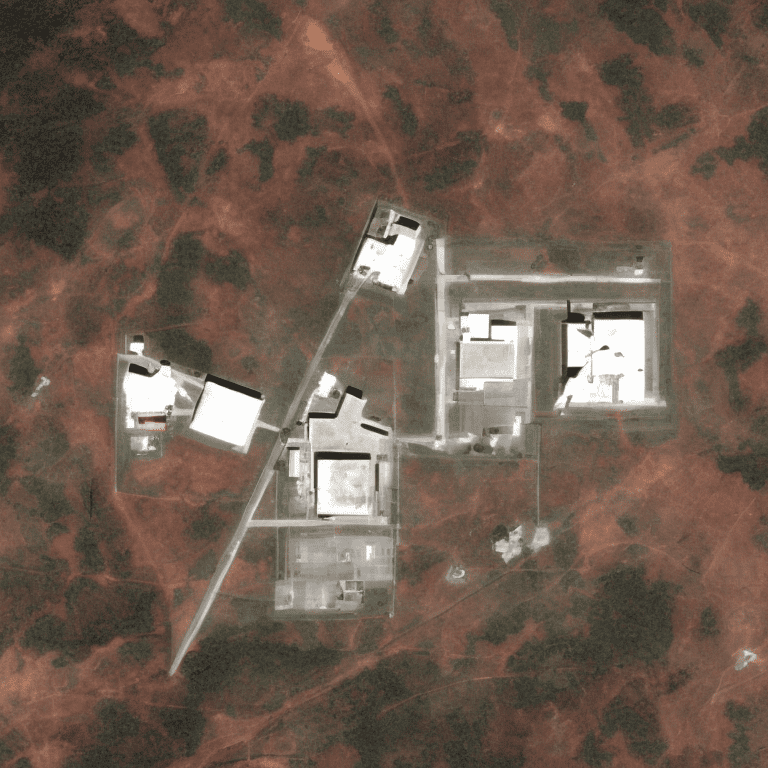}
        \caption{Mil. Facility}
    \end{subfigure}
    \begin{subfigure}[t]{0.12\textwidth}
       \centering
        \includegraphics[width=\textwidth]{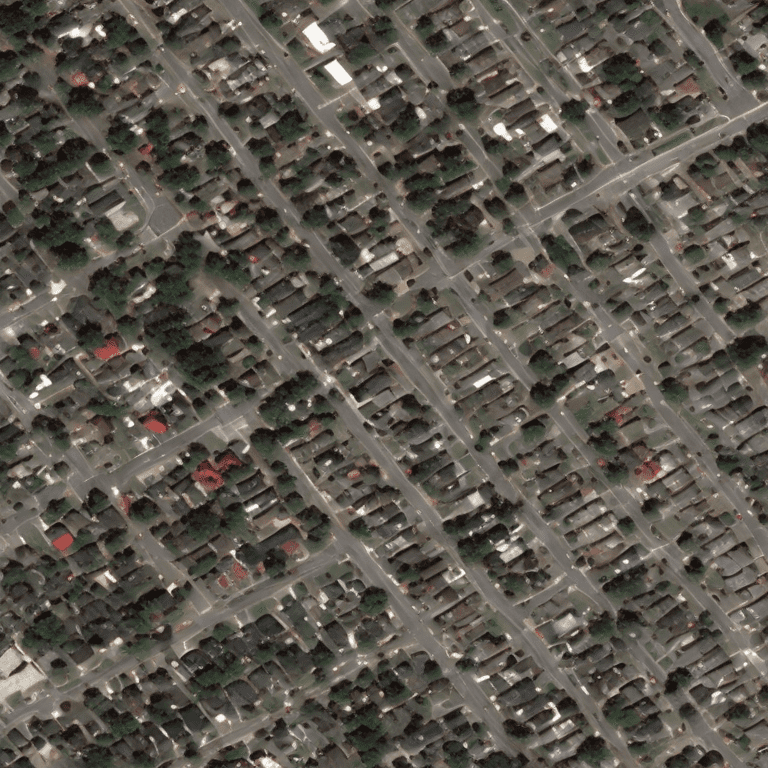}
        \caption{Multi-unit Res.}
    \end{subfigure}
    \begin{subfigure}[t]{0.12\textwidth}
       \centering
        \includegraphics[width=\textwidth]{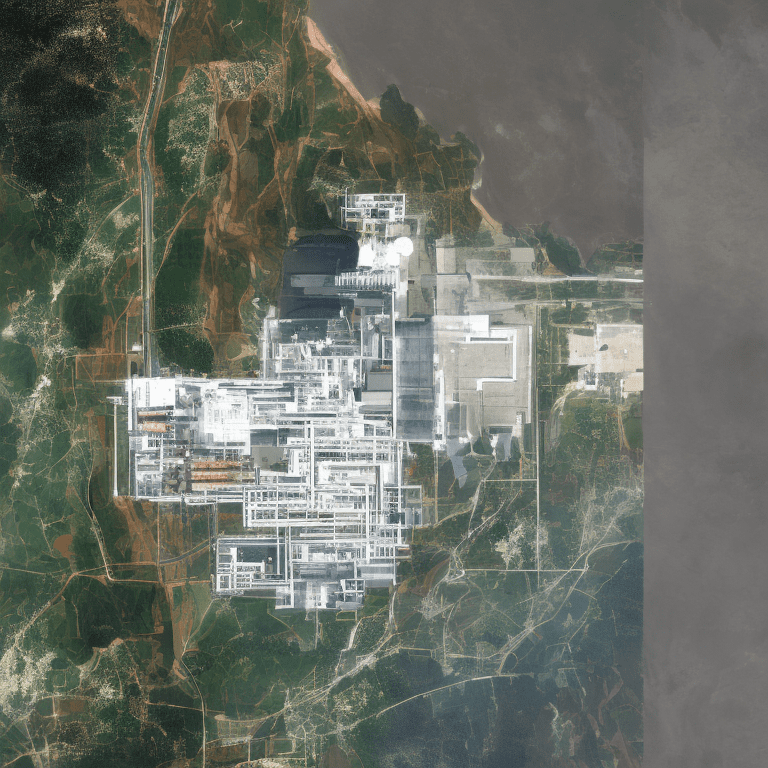}
        \caption{Nucl. Powerplant}
    \end{subfigure}\\
    \begin{subfigure}[t]{0.12\textwidth}
       \centering
        \includegraphics[width=\textwidth]{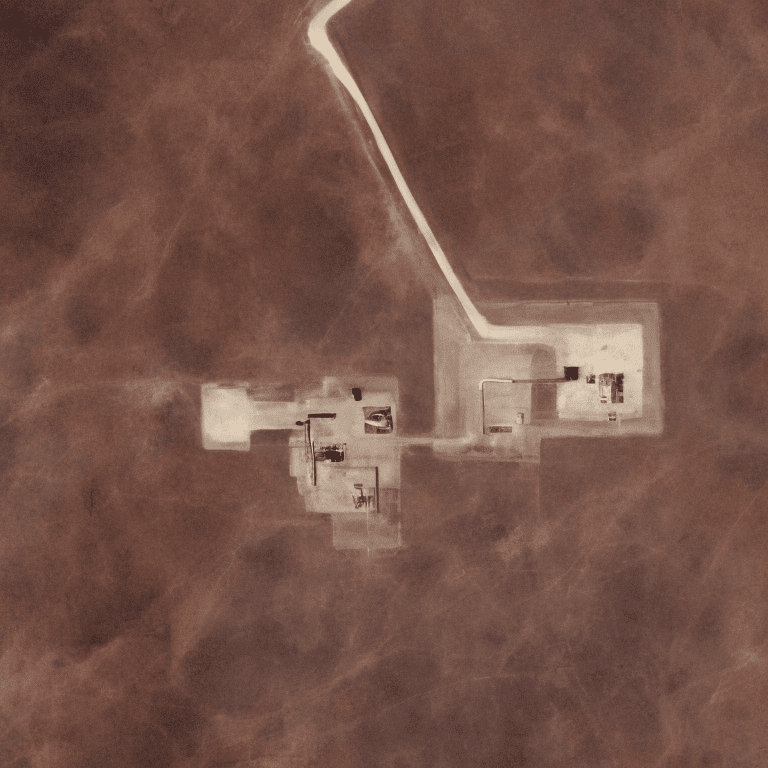}
        \caption{Oil/Gas Facility}
    \end{subfigure}
    \begin{subfigure}[t]{0.12\textwidth}
       \centering
        \includegraphics[width=\textwidth]{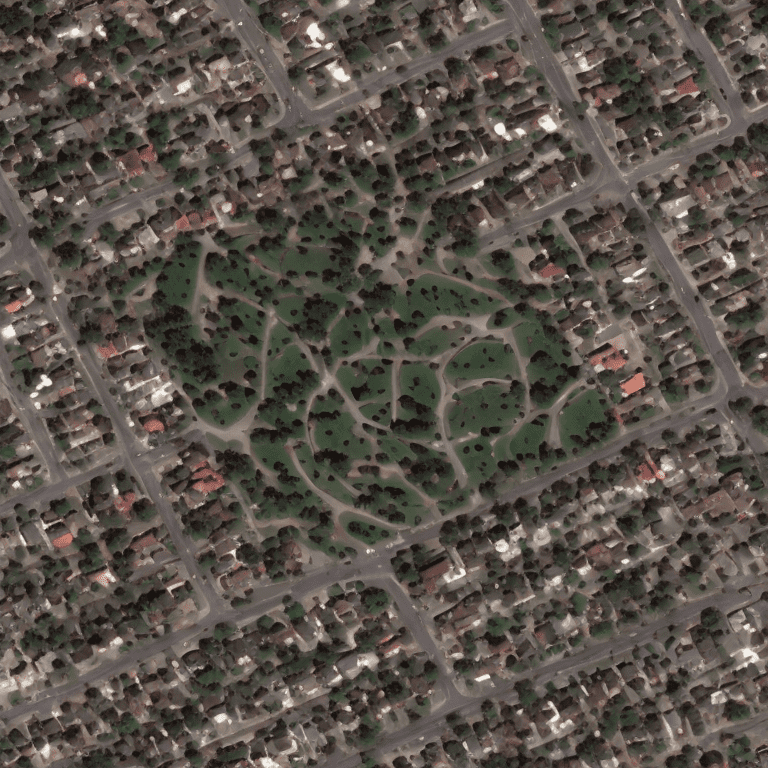}
        \caption{Park}
    \end{subfigure}
    \begin{subfigure}[t]{0.12\textwidth}
       \centering
        \includegraphics[width=\textwidth]{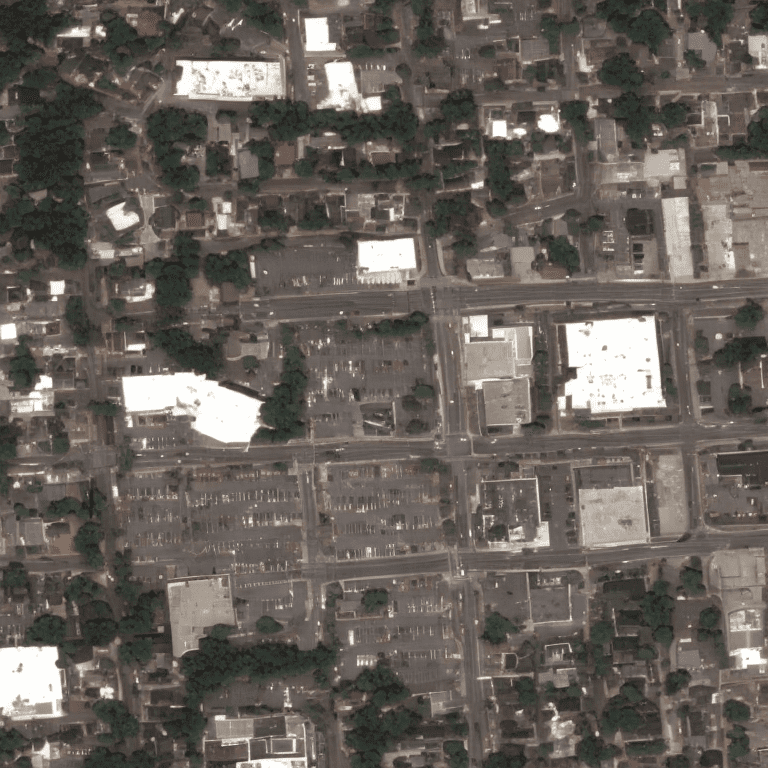}
        \caption{Park. Lot/Garage}
    \end{subfigure}
    \begin{subfigure}[t]{0.12\textwidth}
       \centering
        \includegraphics[width=\textwidth]{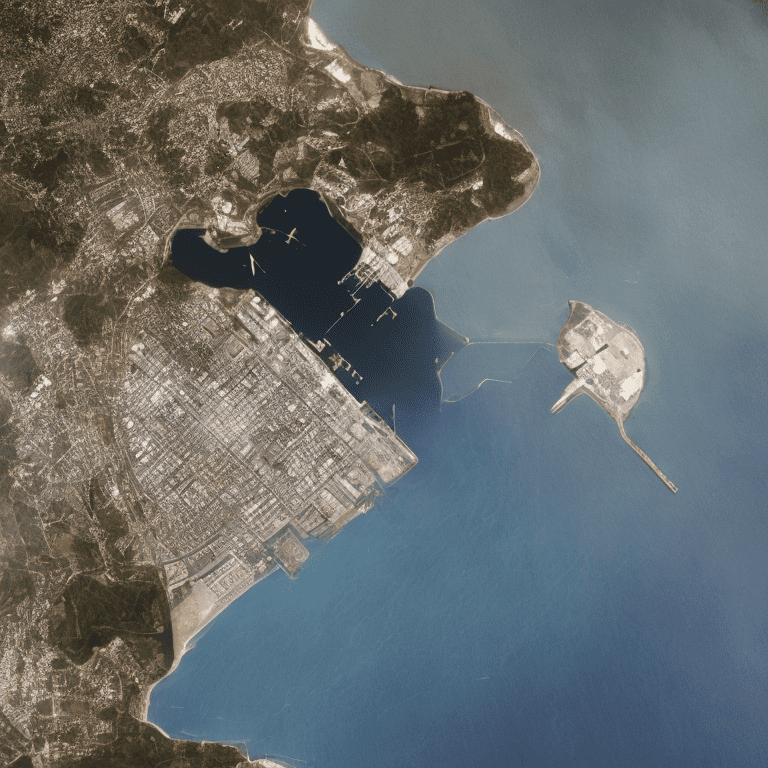}
        \caption{Port}
    \end{subfigure}
    \begin{subfigure}[t]{0.12\textwidth}
       \centering
        \includegraphics[width=\textwidth]{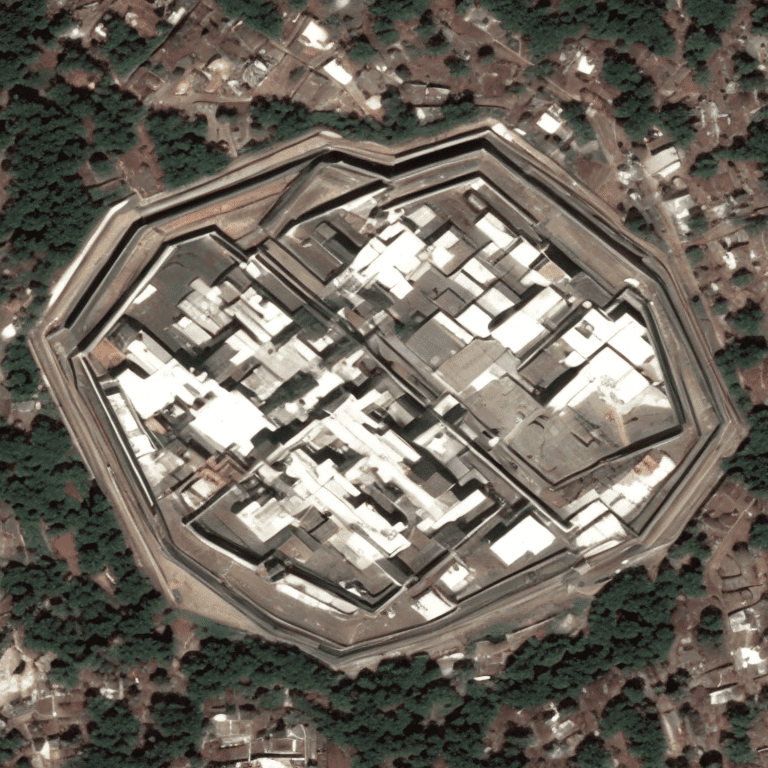}
        \caption{Prison}
    \end{subfigure}
    \begin{subfigure}[t]{0.12\textwidth}
       \centering
        \includegraphics[width=\textwidth]{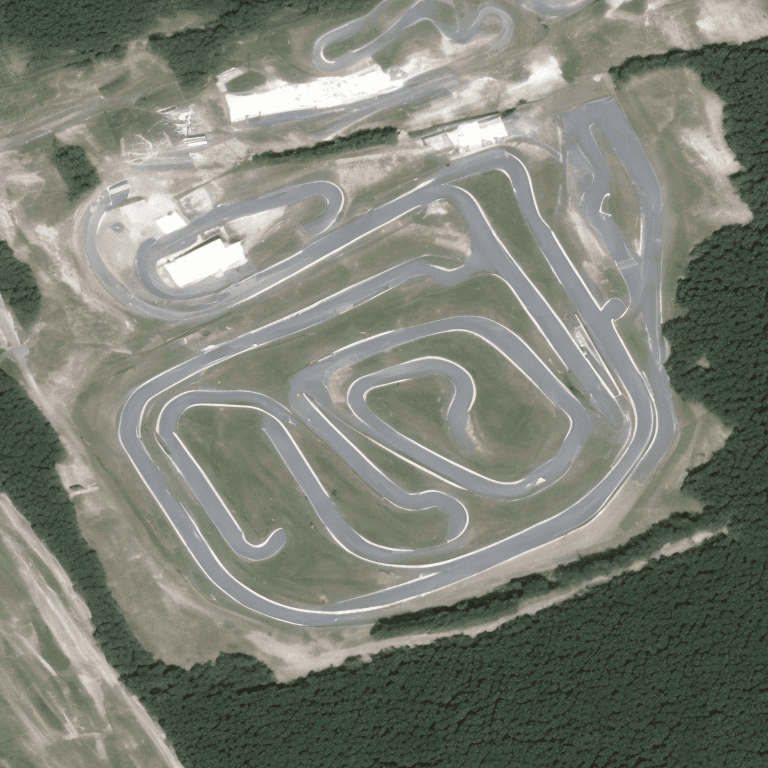}
        \caption{Race Track}
    \end{subfigure}
    \begin{subfigure}[t]{0.12\textwidth}
       \centering
        \includegraphics[width=\textwidth]{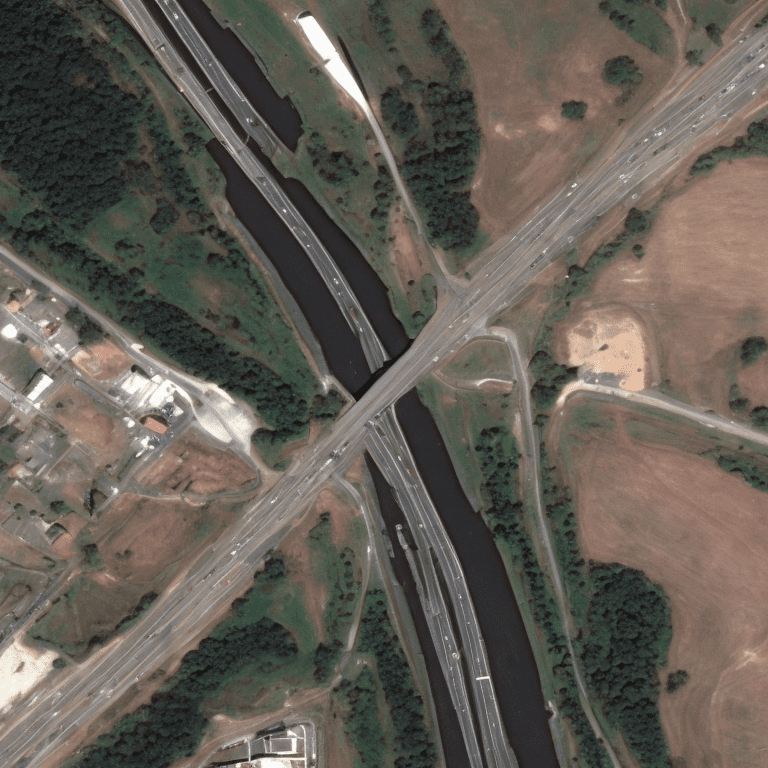}
        \caption{Road Bridge}
    \end{subfigure}\\
    \begin{subfigure}[t]{0.12\textwidth}
       \centering
        \includegraphics[width=\textwidth]{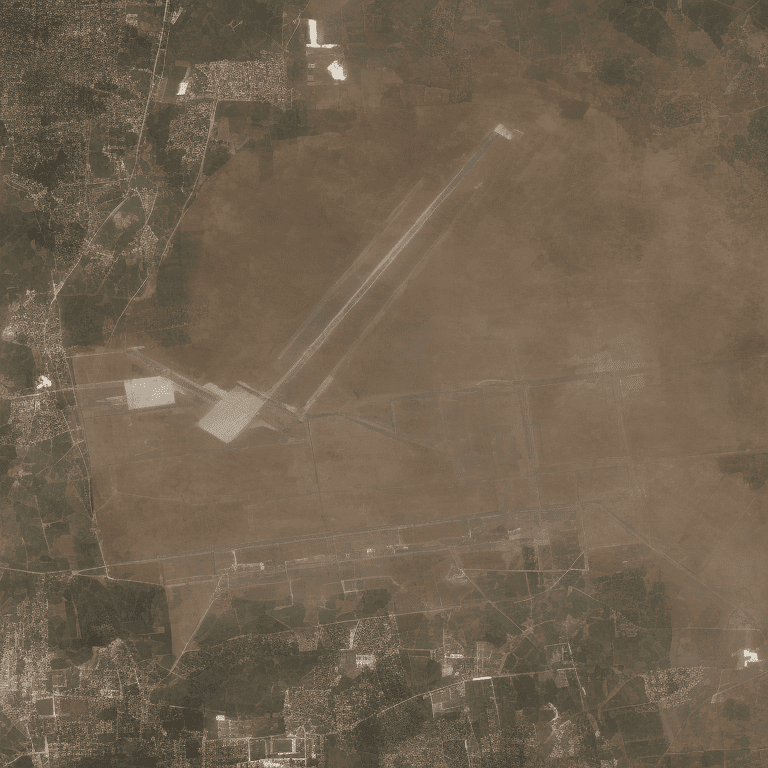}
        \caption{Runway}
    \end{subfigure}
    \begin{subfigure}[t]{0.12\textwidth}
       \centering
        \includegraphics[width=\textwidth]{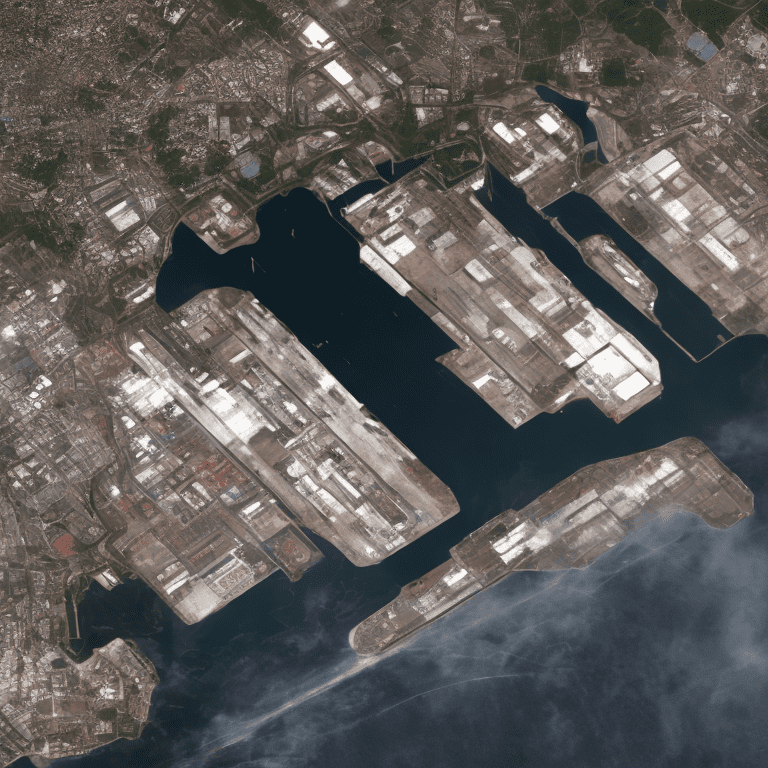}
        \caption{Shipyard}
    \end{subfigure}
    \begin{subfigure}[t]{0.12\textwidth}
       \centering
        \includegraphics[width=\textwidth]{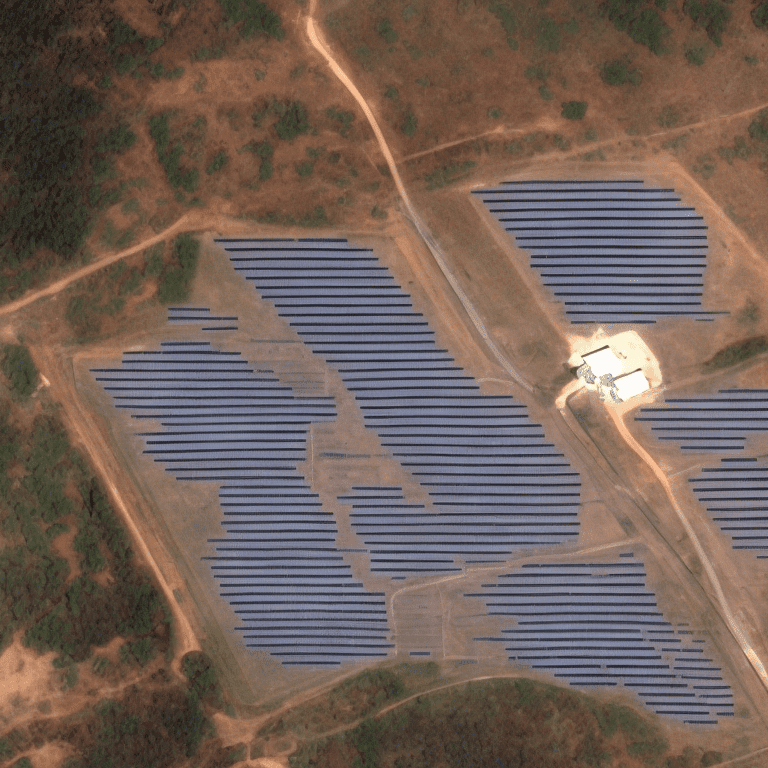}
        \caption{Solar Farm}
    \end{subfigure}
    \begin{subfigure}[t]{0.12\textwidth}
       \centering
        \includegraphics[width=\textwidth]{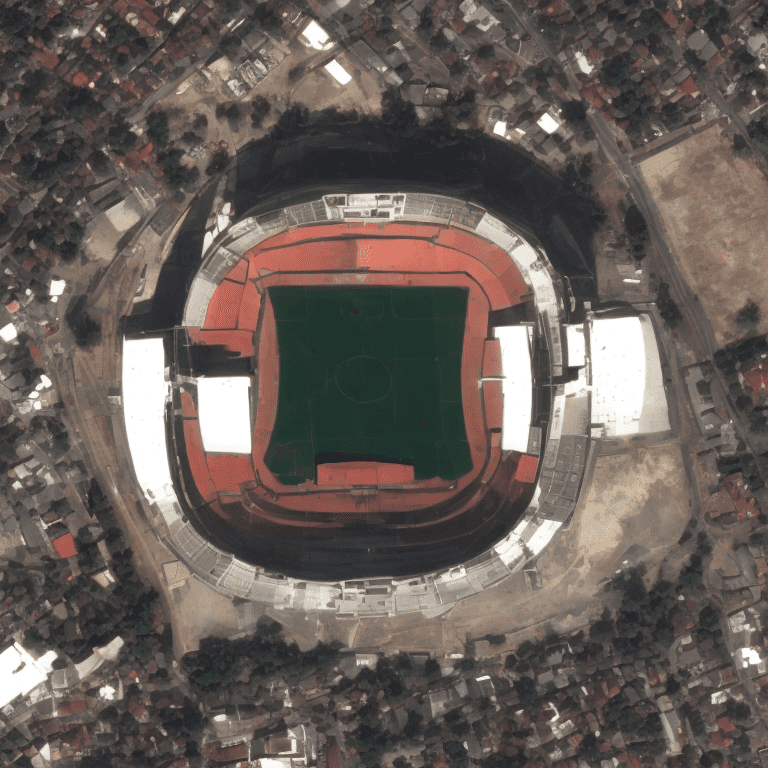}
        \caption{Stadium}
    \end{subfigure}
    \begin{subfigure}[t]{0.12\textwidth}
       \centering
        \includegraphics[width=\textwidth]{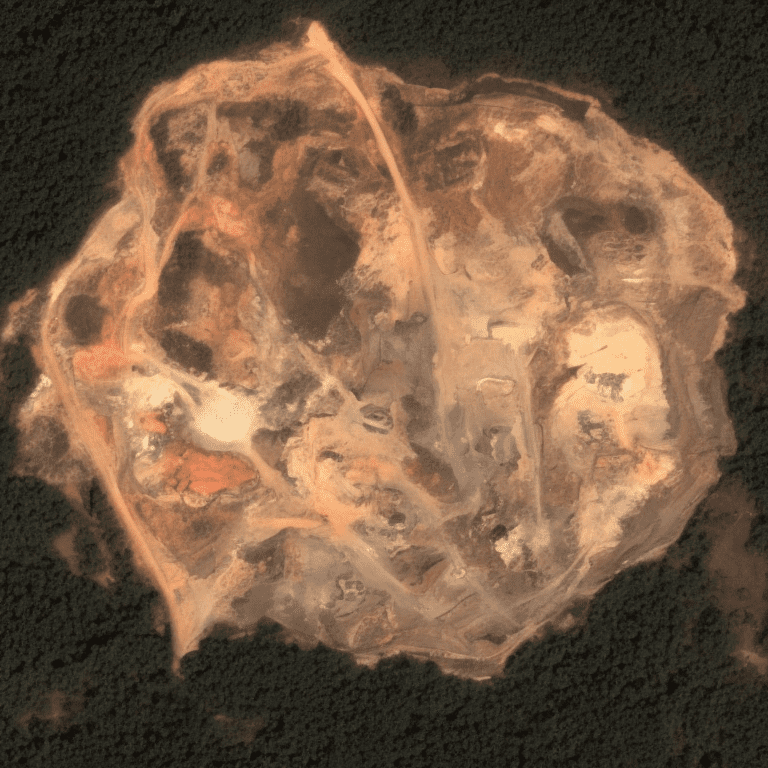}
        \caption{Surface Mine}
    \end{subfigure}
    \begin{subfigure}[t]{0.12\textwidth}
       \centering
        \includegraphics[width=\textwidth]{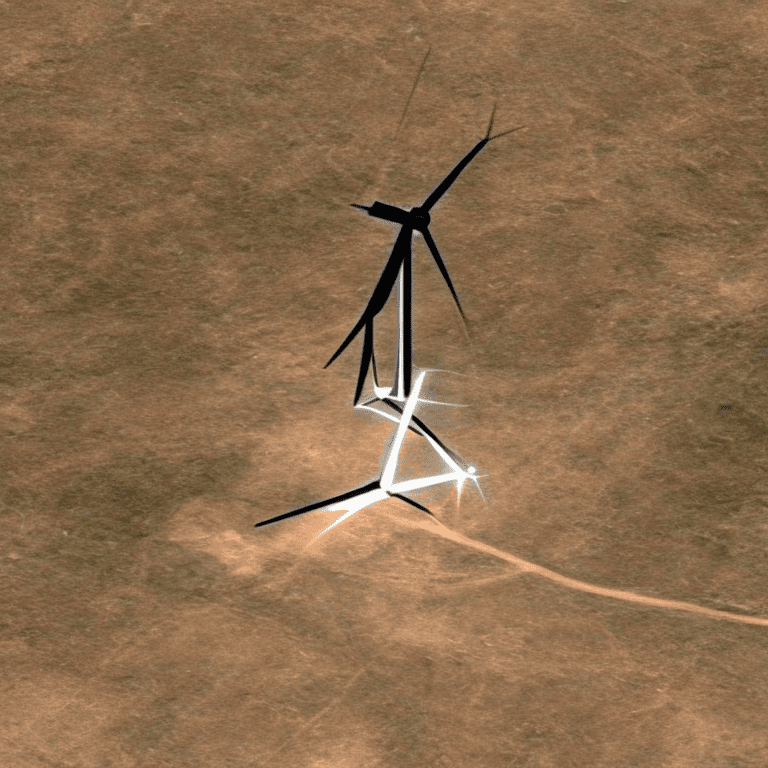}
        \caption{Wind Farm}
    \end{subfigure}
    \begin{subfigure}[t]{0.12\textwidth}
       \centering
        \includegraphics[width=\textwidth]{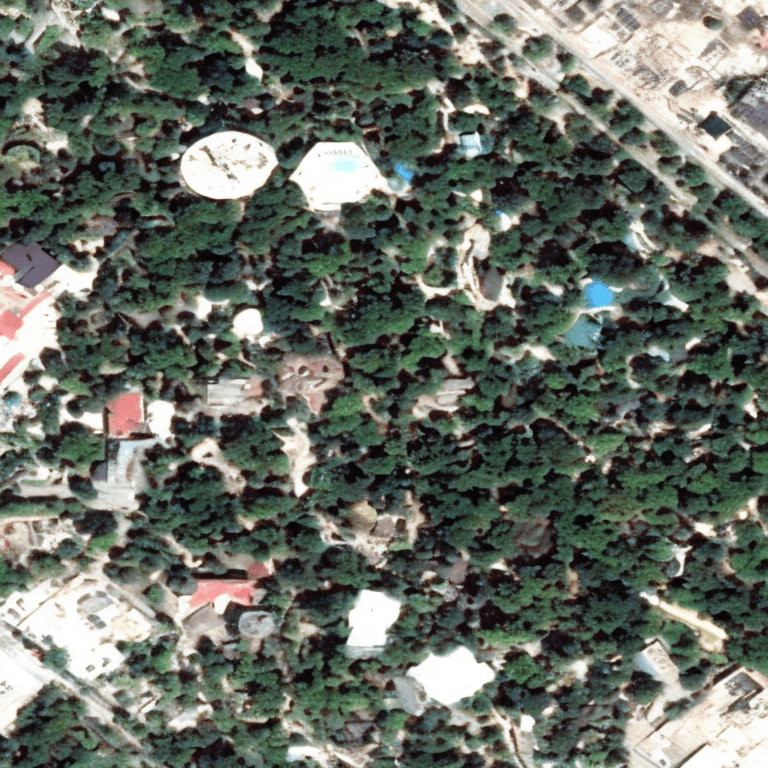}
        \caption{Zoo}
    \end{subfigure}
    \caption{Single-image generations sampled from our model given a prompt of the form "a satellite image of a $<$object class$>$" (prompt-only). Further results can be viewed in Fig. \ref{fig:single_image_overall2}.}
    \label{fig:single_image_overall1}
\end{figure*}

We first consider single-image generation -- i.e. the task on which our model is pre-trained by fine-tuning SD 2.1 on our 3-modal (i.e. combining text, metadata, and image) version of the fMoW dataset. We compare our model to the one proposed by \citep{diffusionsat} where, apart from the fMoW dataset, SD 2.1 was fine-tuned using data from two additional dataset sources -- i.e. Satlas \citep{satlas} and SpaceNet \citep{spacenet1,spacenet2}. The two architectures differ in two key aspects: (a) their approach to metadata handling/fusion -- our model produces a unified metadata vector $\mathbb{R}^{D}$ by first concatenating all $M$ post-projection (Eq. 2-3, Section 3) metadata-specific vectors, each $\in \mathbb{R}^{D}$. This results in a concatenated vector $\in \mathbb{R}^{M * D}$ which is then projected to a lower-dimensional $\mathbb{R}^{D}$ space. In contrast, the model proposed by \citep{diffusionsat} simply adds the metadata projections together; and (b) the quantity of metadata used -- our model incorporates six additional environment-specific metadata types. Uncurated single-image generation results produced by our model can be browsed in Figure \ref{fig:single_image_overall1}. 

\begin{figure}[h]
    \centering
    \begin{tabular}{>{\centering\arraybackslash}m{0.010\textwidth}
    >{\centering\arraybackslash}m{0.14\textwidth}
    @{\hspace{1pt}}|@{\hspace{1pt}}
    >{\centering\arraybackslash}m{0.14\textwidth}
    @{\hspace{1pt}}|@{\hspace{1pt}}
    >{\centering\arraybackslash}m{0.14\textwidth}}
    & \textbf{prompt-only} & \textbf{prompt + missing metadata} & \textbf{prompt + full metadata}\\
   \rotatebox[origin=l]{90}{\parbox{15mm}{\centering \textbf{DiffusionSAT \citep{diffusionsat}}}} & \includegraphics[width=0.14\textwidth]{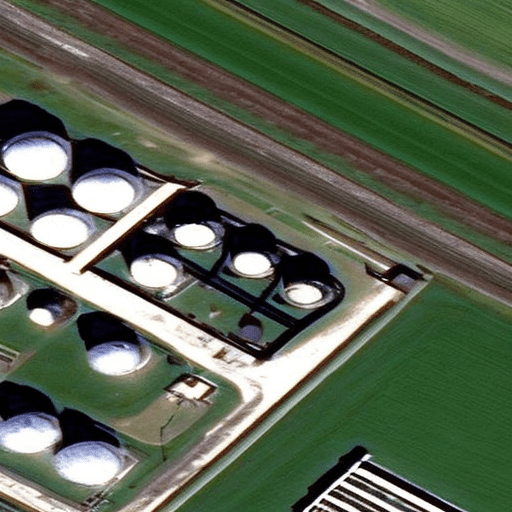} & \includegraphics[width=0.14\textwidth]{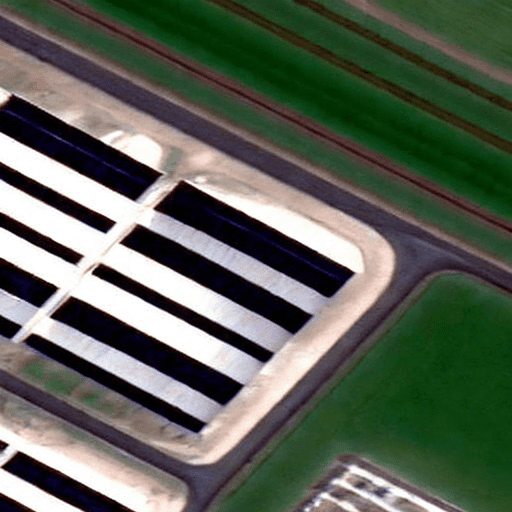} & \includegraphics[width=0.14\textwidth]{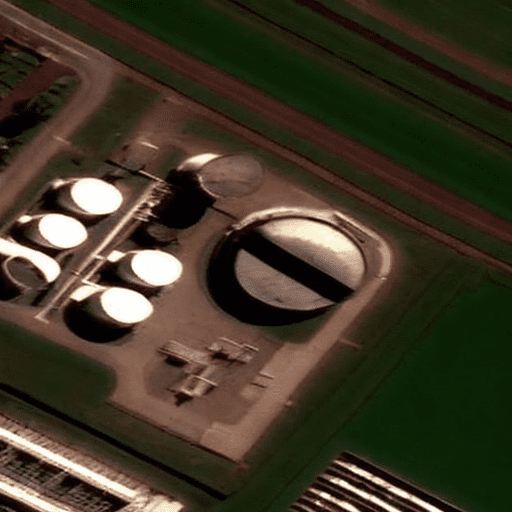}\\\cdashline{1-4}\\
   \rotatebox[origin=l]{90}{\centering \textbf{Ours}} & \includegraphics[width=0.14\textwidth]{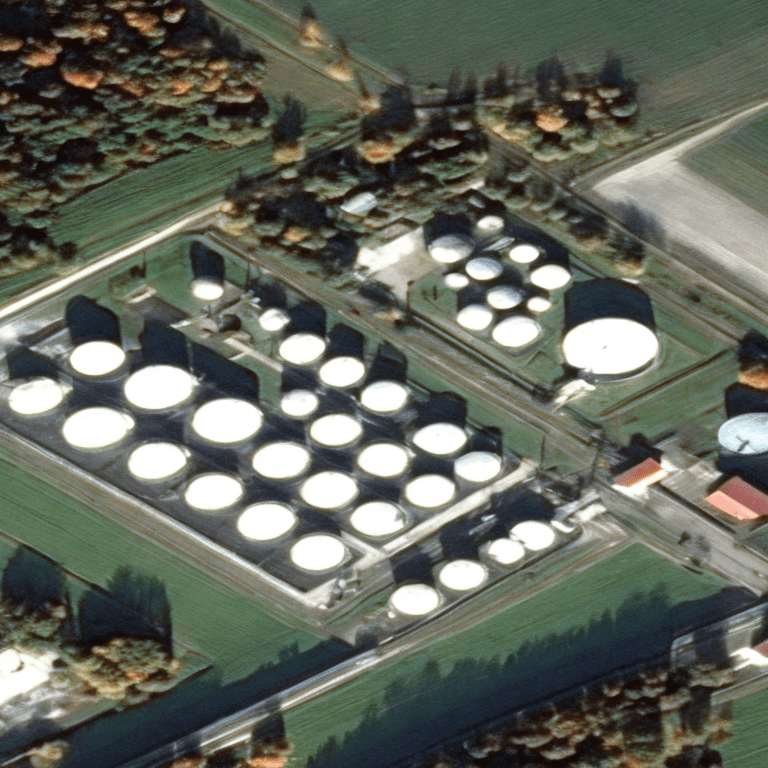} & \includegraphics[width=0.14\textwidth]{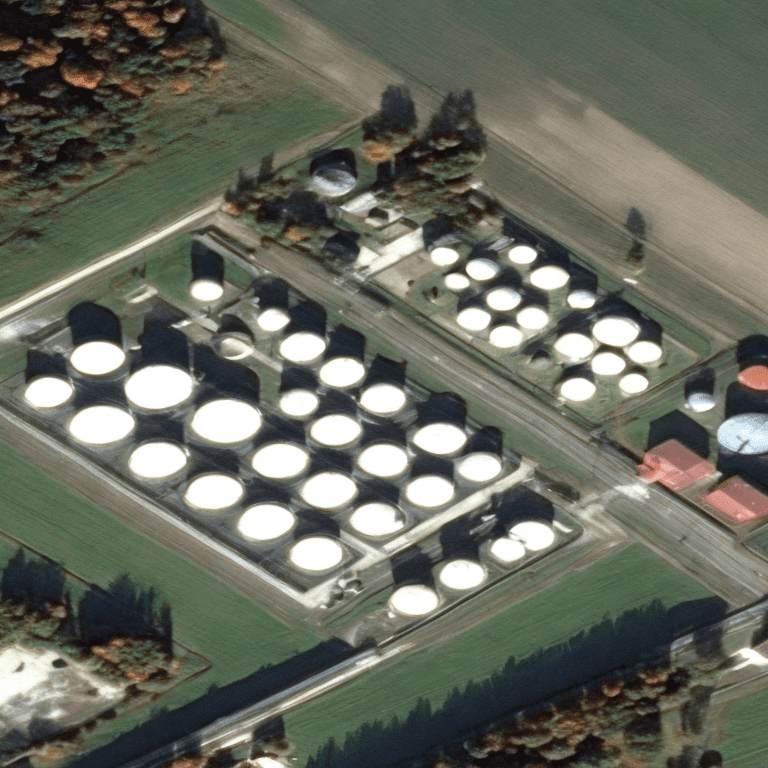}
    & \includegraphics[width=0.14\textwidth]{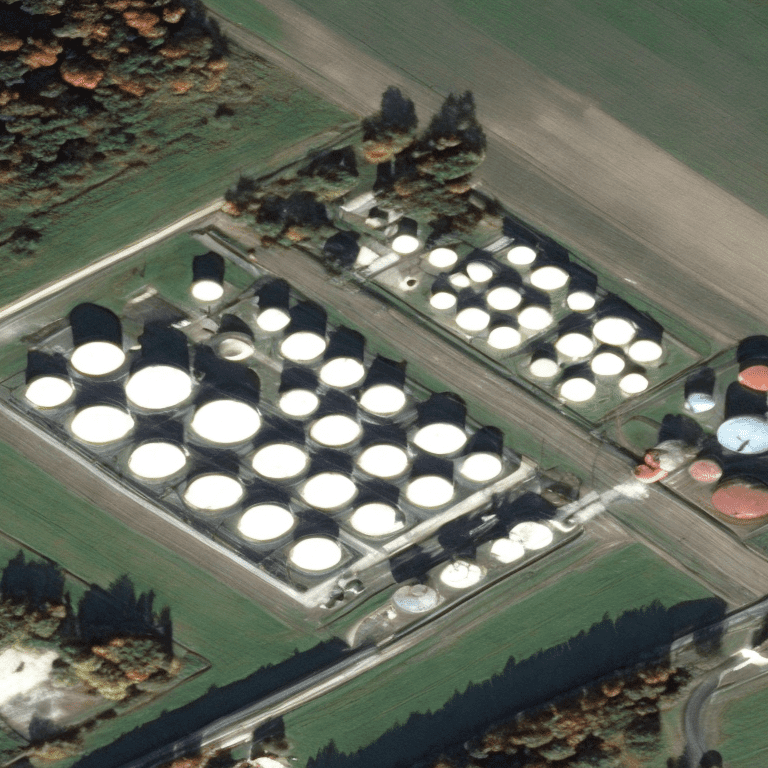}
    \end{tabular}
    \caption{Single-image generations for the "storage tank" class sampled from \citep{diffusionsat}'s (top row) and our model (bottom row) by varying metadata presence in the control input. As can be seen in the middle row, DiffusionSAT's model seems to encounter class fidelity problems when faced with partial metadata inclusion. In contrast, our model is robust to any configuration.}
    \label{tab:ref_label_example_no_metadata}
\end{figure}

By favoring the concatenation-and-mapping approach over the additive one proposed by \citep{diffusionsat} not only we allow our model to account for control inputs (i.e. metadata and/or captions) that are missing or corrupted in their entirety but also for partially missing ones. This is due to the ability of the proposed approach to maintain the order of the different metadata-specific vectors post-concatenation up to the encoder level -- a quality that is lost by having them added together as per \cite{diffusionsat}'s approach due to addition being non-invertible. Our model's ability to smoothly generate outputs under any level of input corruption is exhibited in Fig. \ref{tab:ref_label_example_no_metadata}, bottom row. Moreover, despite being trained on only one dataset (i.e. relying on less data and, thus, a more limited distribution) our architecture is shown to be able to quantitatively outperform the one introduced in \citep{diffusionsat} as reflected by the better FID, LPIPS, and CLIP-scores reported in Table \ref{tab:table_single_image_quantitative}. This indicates that enriching the model with environment-specific information in the form of metadata results in better capturing the intricacies within satellite image data, thus leading to generated samples of greater accuracy and fidelity. 

\begin{figure}[h!]
    \centering
    \begin{tabular}{
        >{\centering\arraybackslash}m{0.115\textwidth}
        @{\hspace{1pt}}
        >{\centering\arraybackslash}m{0.115\textwidth}
        @{\hspace{1pt}}|@{\hspace{1pt}}
        >{\centering\arraybackslash}m{0.115\textwidth}
        @{\hspace{1pt}}
        >{\centering\arraybackslash}m{0.115\textwidth}
    }
    \multicolumn{2}{c|}{\textbf{DiffusionSAT \cite{diffusionsat}}} & \multicolumn{2}{c}{\textbf{Ours}}\\\hline
    \textbf{minimum gsd} & \textbf{maximum gsd} & \textbf{minimum gsd} & \textbf{maximum gsd}\\\hline
    \includegraphics[width=\linewidth]{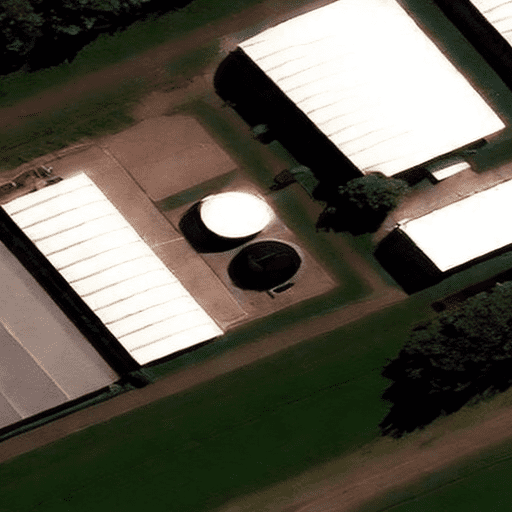} &
    \includegraphics[width=\linewidth]{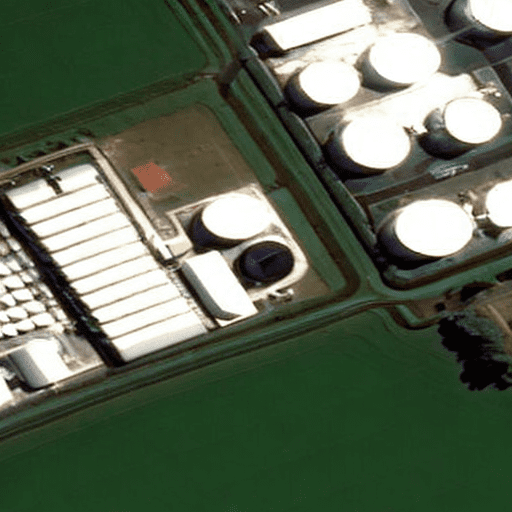} &
    \includegraphics[width=\linewidth]{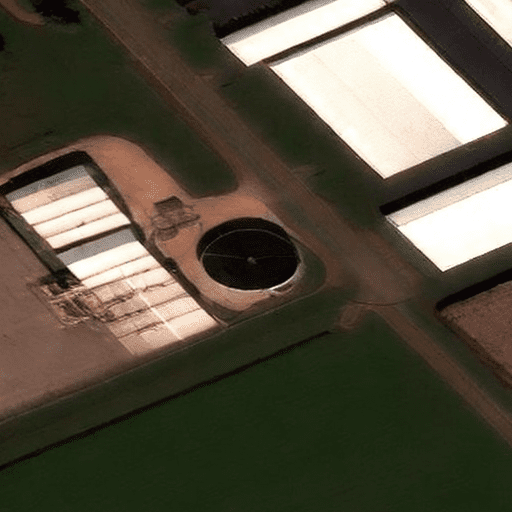} &
    \includegraphics[width=\linewidth]{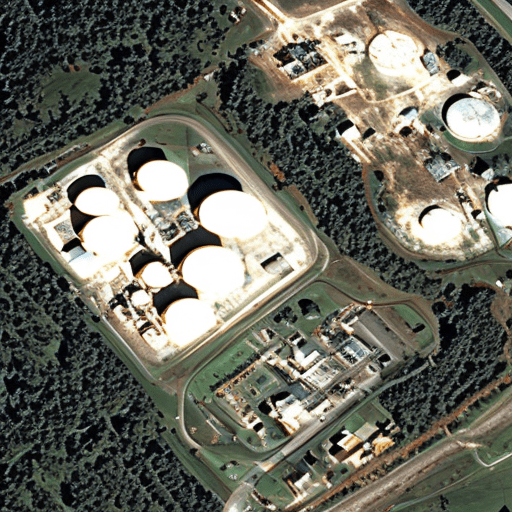}\\
    \multicolumn{4}{c}{\textbf{class: }\textit{storage tank}}\\\hline
    \includegraphics[width=\linewidth]{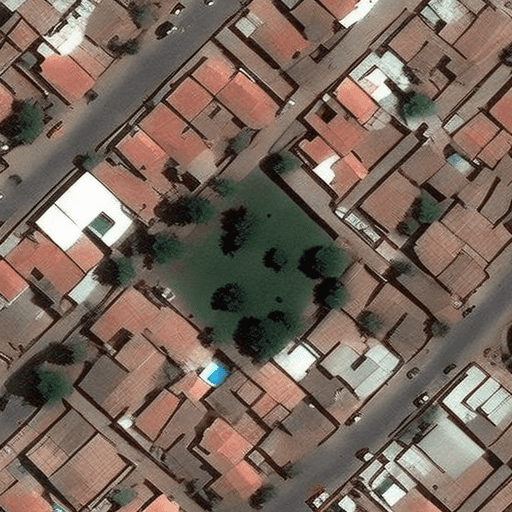} &
    \includegraphics[width=\linewidth]{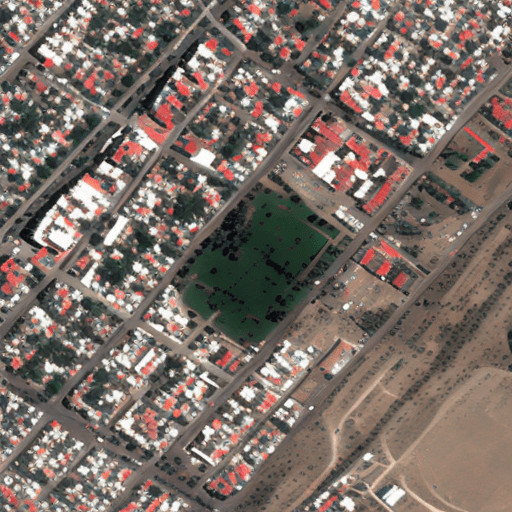} &
    \includegraphics[width=\linewidth]{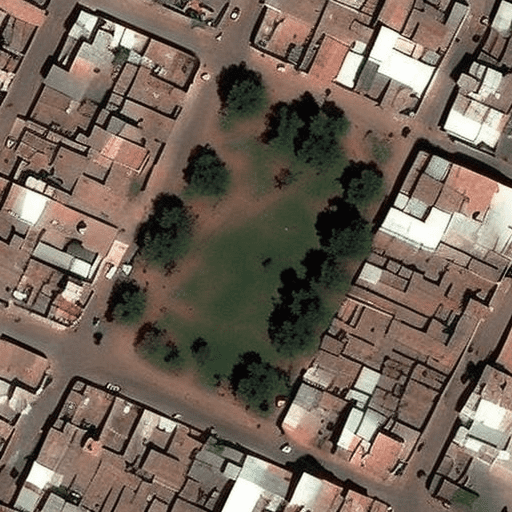} &
    \includegraphics[width=\linewidth]{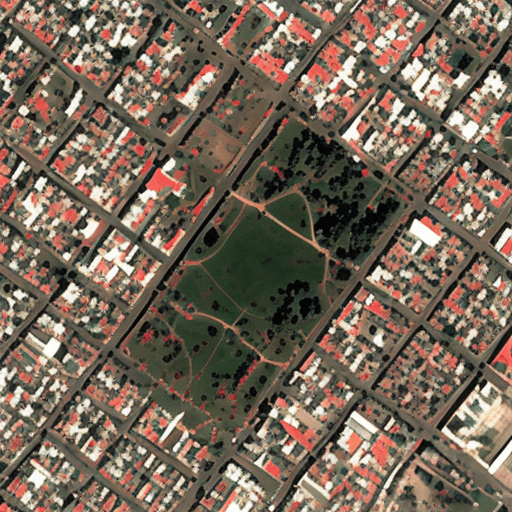}\\
    \multicolumn{4}{c}{\textbf{class: }\textit{park}}\\\hline
    \includegraphics[width=\linewidth]{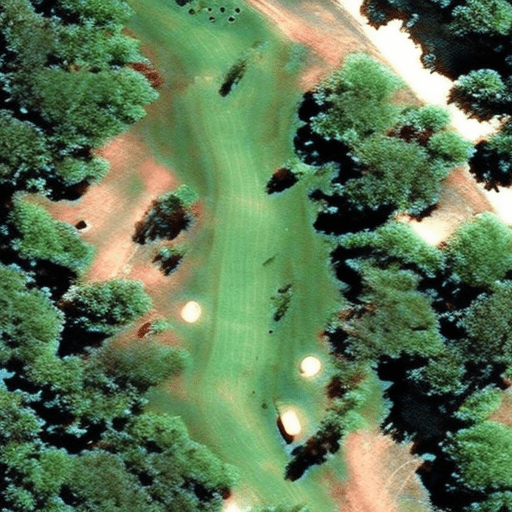} &
    \includegraphics[width=\linewidth]{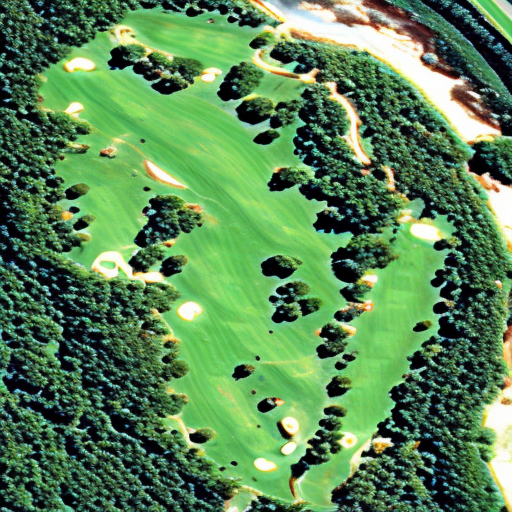} &
    \includegraphics[width=\linewidth]{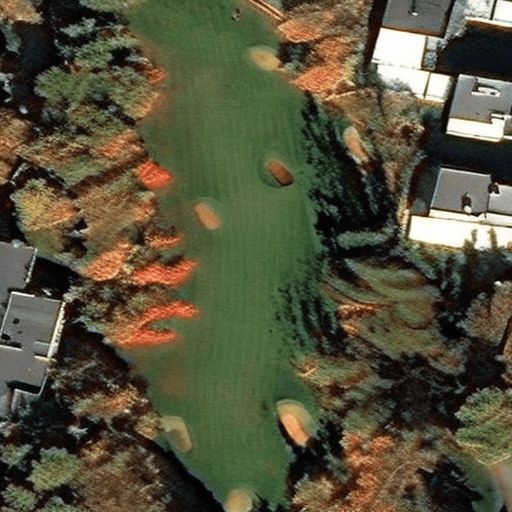} &
    \includegraphics[width=\linewidth]{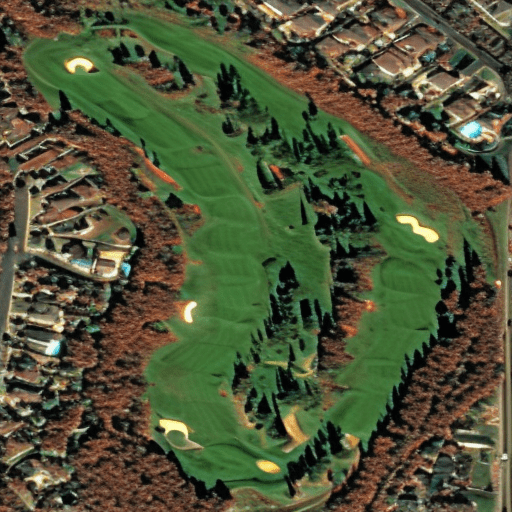}\\
    \multicolumn{4}{c}{\textbf{class: }\textit{golf course}}
    \end{tabular}
    \caption{Sample generations sampled from \citep{diffusionsat}'s and our model with varying ground-sampling distance (GSD). Lowering GSD input values (i.e. having each pixel represent shorter ground distances) results in a "zooming-in" effect in both models.}
    \label{tbl:ref_label_gsd}
\end{figure}
\begin{figure}[h]
    \centering
    \begin{tabular}{
        >{\centering\arraybackslash}m{0.115\textwidth}
        @{\hspace{1pt}}
        >{\centering\arraybackslash}m{0.115\textwidth}
        @{\hspace{1pt}}|@{\hspace{1pt}}
        >{\centering\arraybackslash}m{0.115\textwidth}
        @{\hspace{1pt}}
        >{\centering\arraybackslash}m{0.115\textwidth}
    }
    \multicolumn{2}{c|}{\textbf{DiffusionSAT \cite{diffusionsat}}}   & \multicolumn{2}{c}{\textbf{Ours}}\\\hline
    \textbf{minimum tcc} & \textbf{maximum tcc} & \textbf{minimum tcc} & \textbf{maximum tcc}\\\hline
    \includegraphics[width=\linewidth]{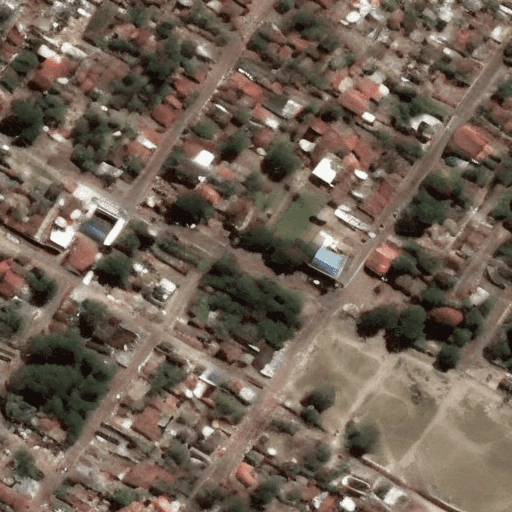} &
    \includegraphics[width=\linewidth]{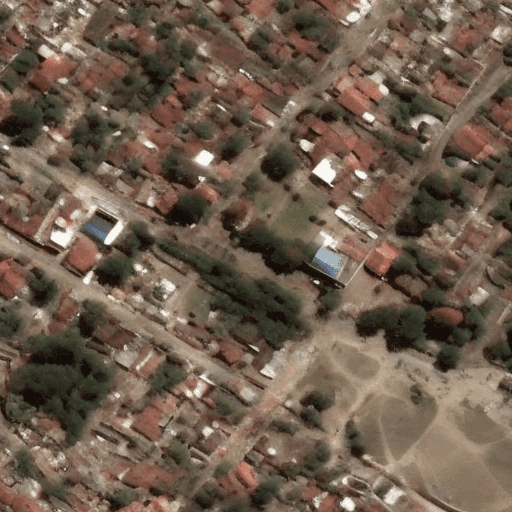} &
    \includegraphics[width=\linewidth]{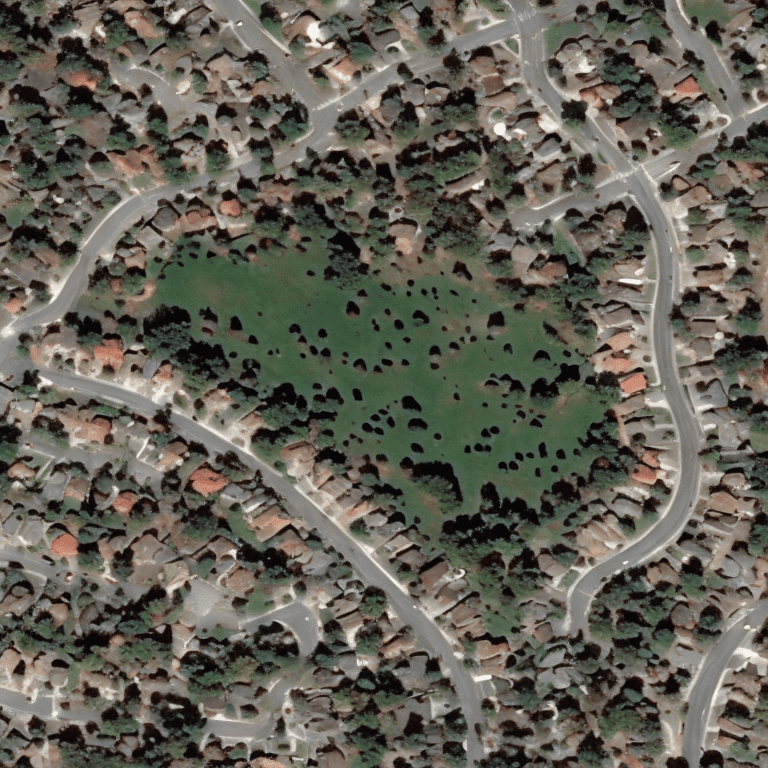} &
    \includegraphics[width=\linewidth]{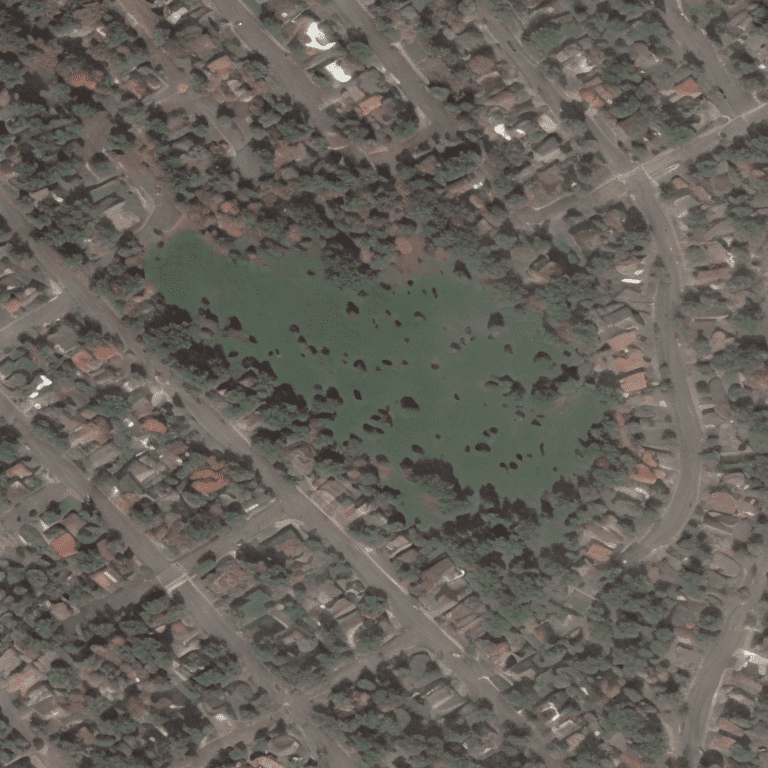}\\
    \multicolumn{4}{c}{\textbf{class: }\textit{storage tank}}\\\hline
    \includegraphics[width=\linewidth]{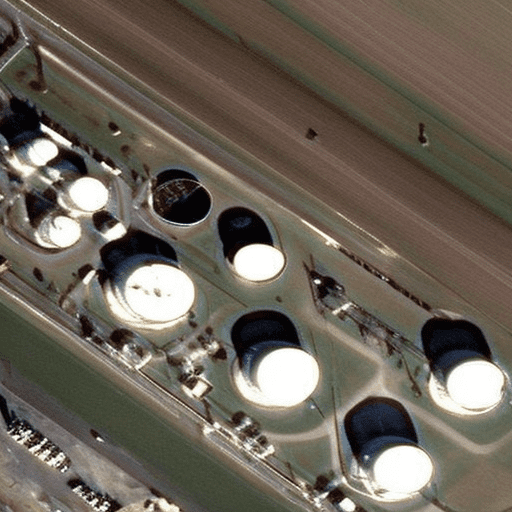} &
    \includegraphics[width=\linewidth]{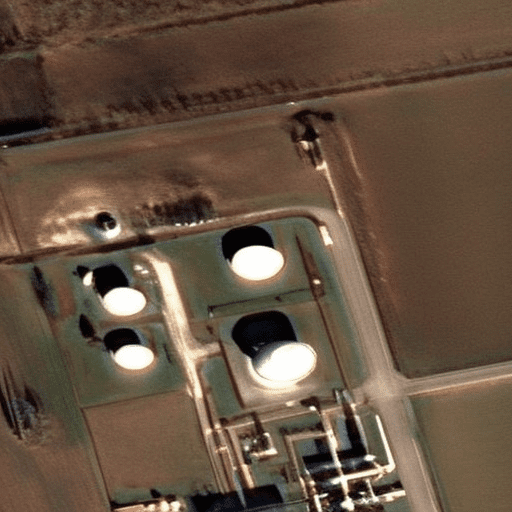} &
    \includegraphics[width=\linewidth]{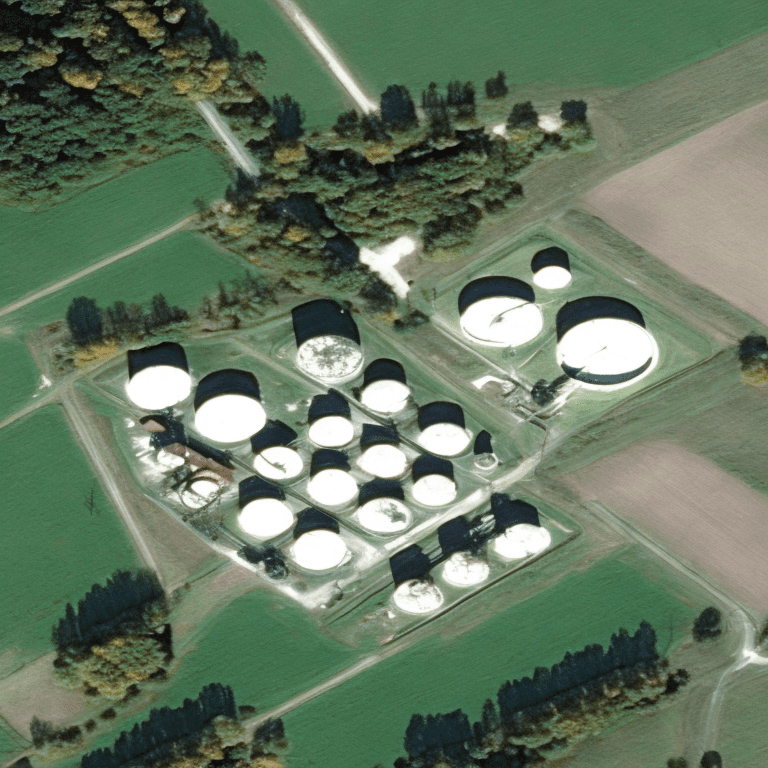} &
    \includegraphics[width=\linewidth]{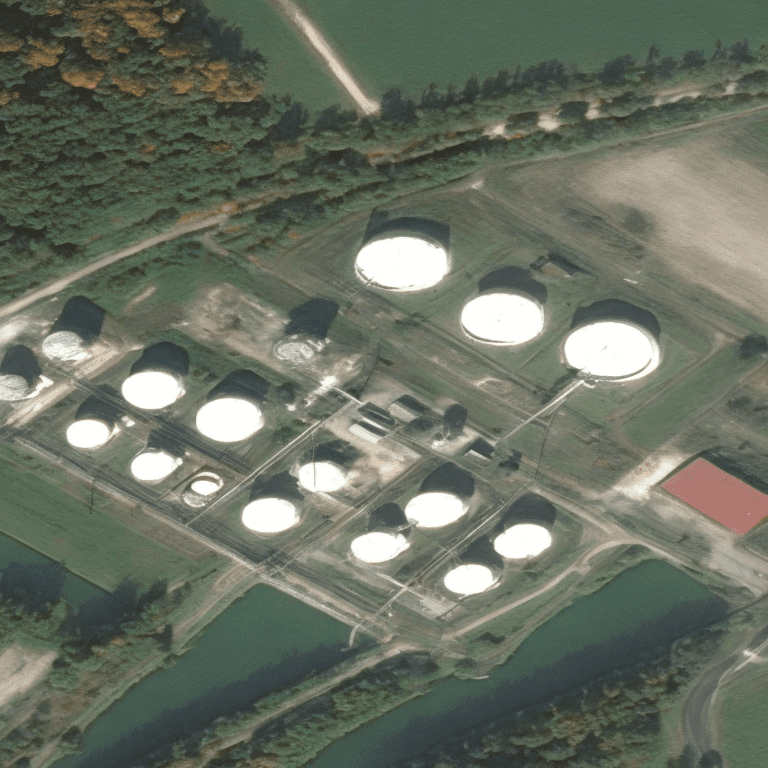}\\
    \multicolumn{4}{c}{\textbf{class: }\textit{park}}\\\hline
    \includegraphics[width=\linewidth]{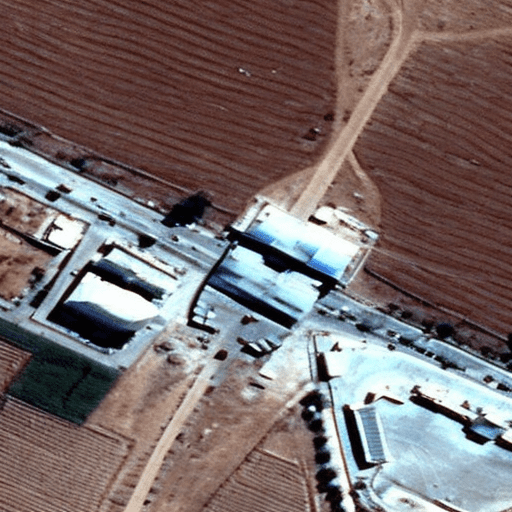} &
    \includegraphics[width=\linewidth]{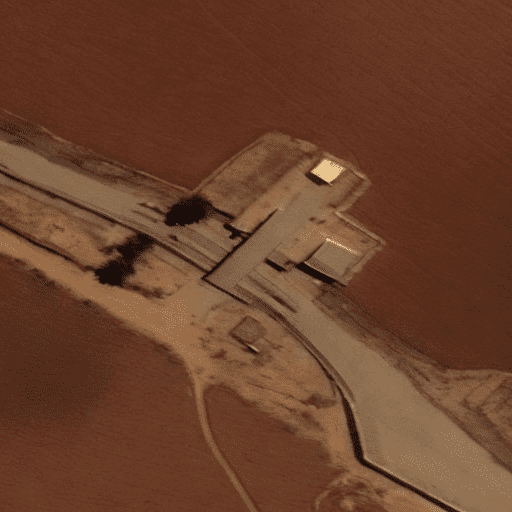} &
    \includegraphics[width=\linewidth]{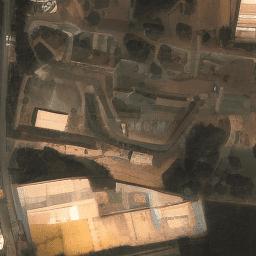} &
    \includegraphics[width=\linewidth]{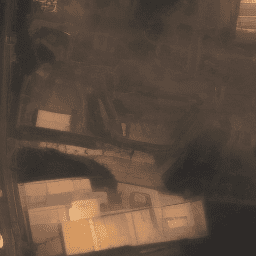}\\
    \multicolumn{4}{c}{\textbf{class: }\textit{border checkpoint}}
    \end{tabular}
    \caption{Sample generations sampled from \citep{diffusionsat}'s and our model with varying total cloud cover (TCC). While maximizing TCC input values results in foggier images and vice-versa for our model, this seems to not be the case for DiffusionSAT's generations which, albeit exhibiting slightly diminished lighting, tend to be as clear as their minimum-tcc counterparts.}
    \label{tbl:ref_label_tcc}
\end{figure}
\begin{figure}[h]
    \centering
    \captionsetup[subfigure]{justification=centering}
    \subfloat[maximum ssr (class: crop field)]{
        \label{example_1_max_ssr}
        \begin{tikzpicture}
        \node[inner sep=0pt] () at (0,0) {\includegraphics[trim = 0cm 0cm 0cm 0cm ,clip,width=0.20\textwidth]{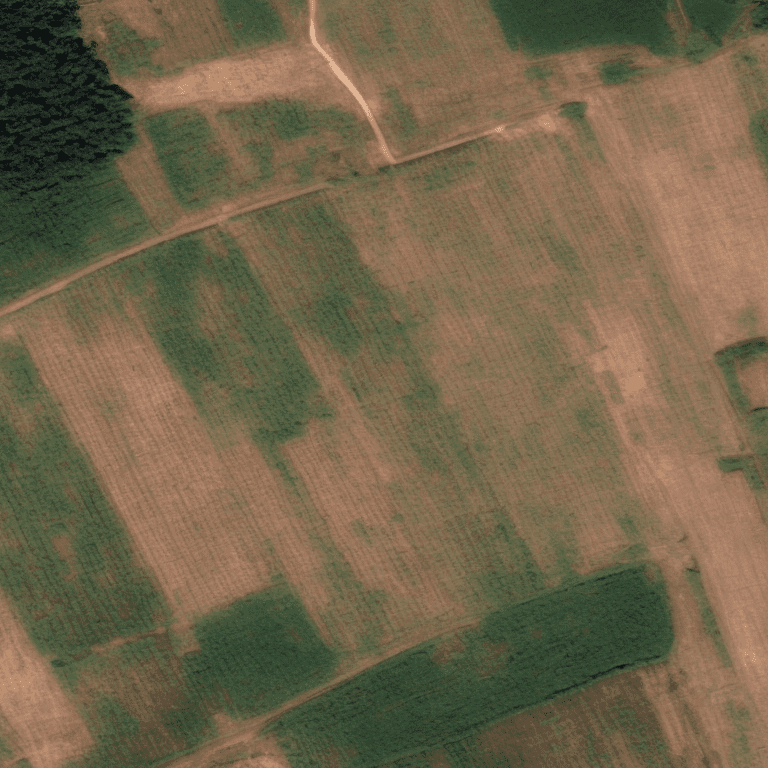}};
        
        \draw[red, very thick] (-1.8,-0.4) rectangle (0.1,1.7);
        \end{tikzpicture}%
    }
    \subfloat[minimum ssr (class: crop field)]{
        \label{example_1_min_ssr}
        \begin{tikzpicture}
        \node[inner sep=0pt] () at (0,0) {\includegraphics[trim = 0cm 0cm 0cm 0cm ,clip,width=0.20\textwidth]{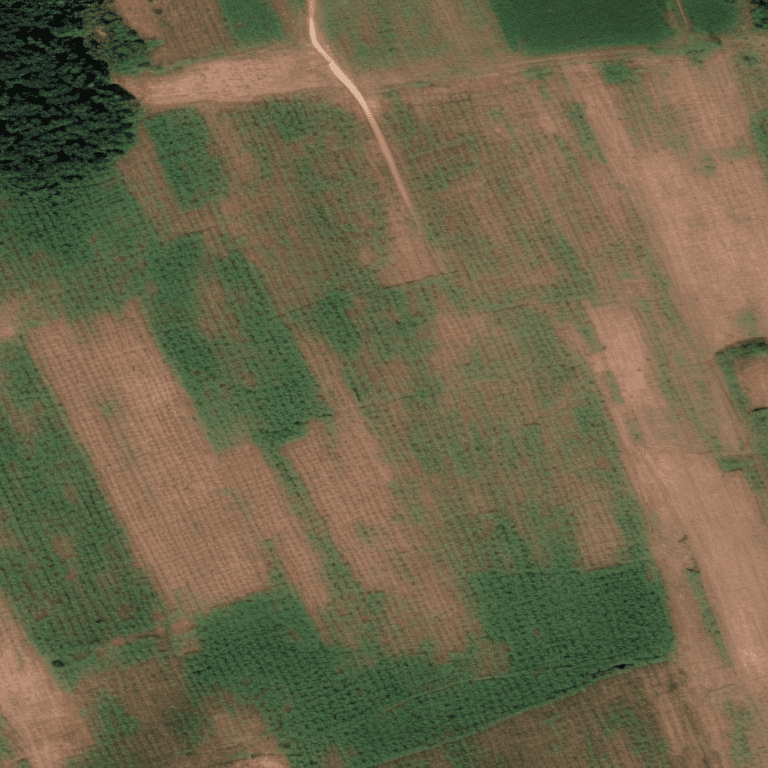}};
        
        \draw[red, very thick] (-1.8,-0.4) rectangle (0.1,1.7);
        \end{tikzpicture}%
    }\\
    \subfloat[maximum ssr (class: crop field)]{
        \label{example_2_max_ssr}
        \begin{tikzpicture}
        \node[inner sep=0pt] () at (0,0) {\includegraphics[trim = 0cm 0cm 0cm 0cm ,clip,width=0.20\textwidth]{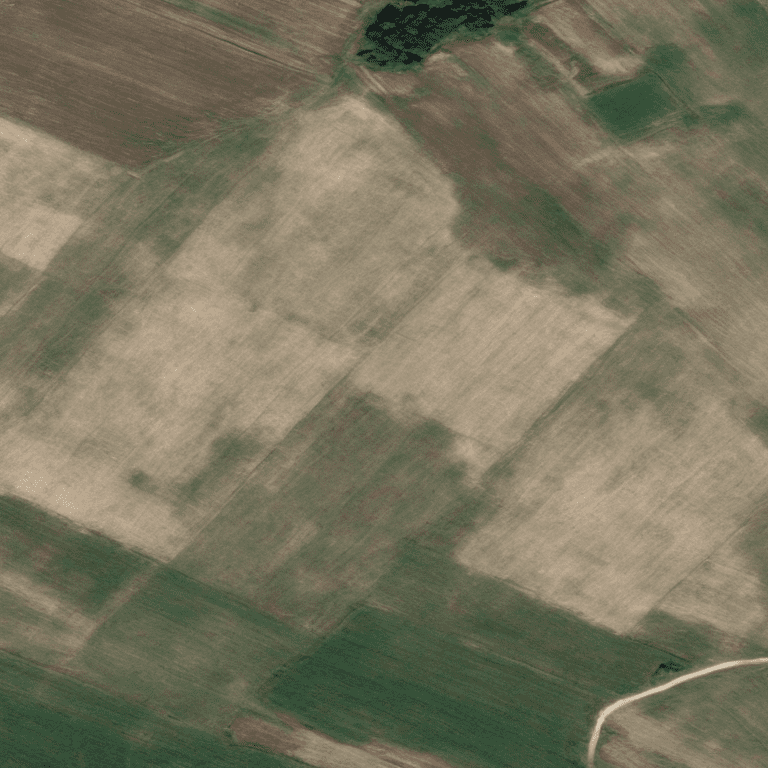}};
        
        \draw[red, very thick] (-1.8,-0.7) rectangle (1.8,-1.8);
        \end{tikzpicture}%
    }
    \subfloat[minimum ssr (class: crop field)]{
        \label{example_2_min_ssr}
        \begin{tikzpicture}
        \node[inner sep=0pt] () at (0,0) {\includegraphics[trim = 0cm 0cm 0cm 0cm ,clip,width=0.20\textwidth]{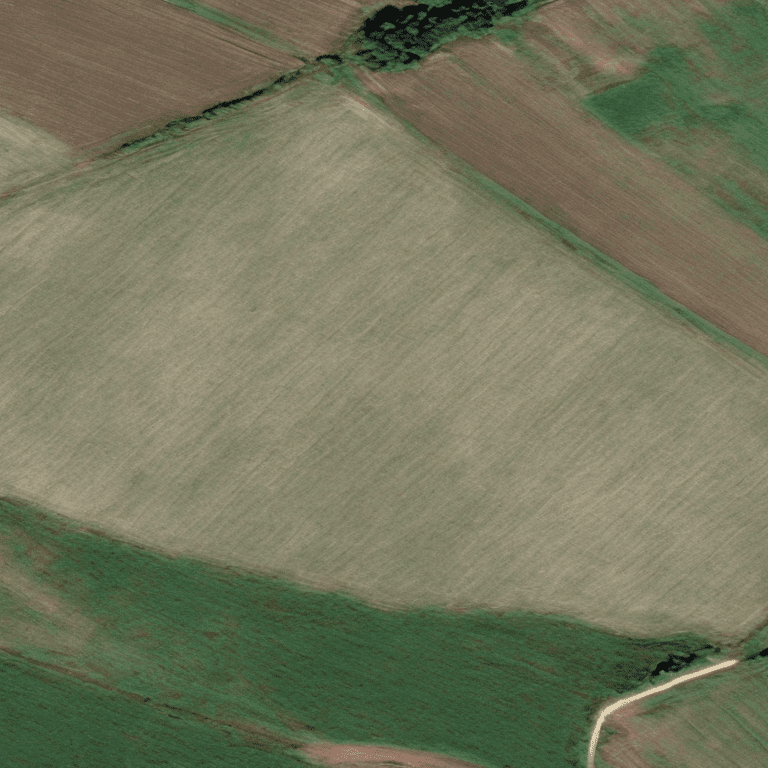}};
        
        \draw[red, very thick] (-1.8,-0.7) rectangle (1.8,-1.8);
        \end{tikzpicture}%
    }\\
    \subfloat[maximum ssr (class: crop field)]{
        \label{example_3_max_ssr}
        \begin{tikzpicture}
        \node[inner sep=0pt] () at (0,0) {\includegraphics[trim = 0cm 0cm 0cm 0cm ,clip,width=0.20\textwidth]{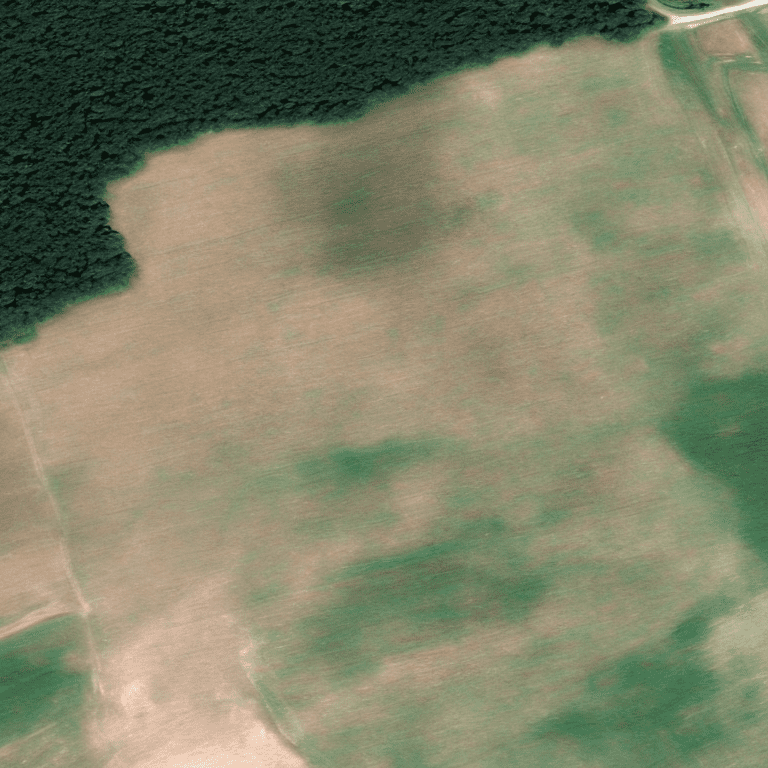}};
        
        \draw[red, very thick] (-0.7,-0.3) rectangle (1.8,-1.8);
        \end{tikzpicture}%
    }
    \subfloat[minimum ssr (class: crop field)]{
        \label{example_3_min_ssr}
        \begin{tikzpicture}
        \node[inner sep=0pt] () at (0,0) {\includegraphics[trim = 0cm 0cm 0cm 0cm ,clip,width=0.20\textwidth]{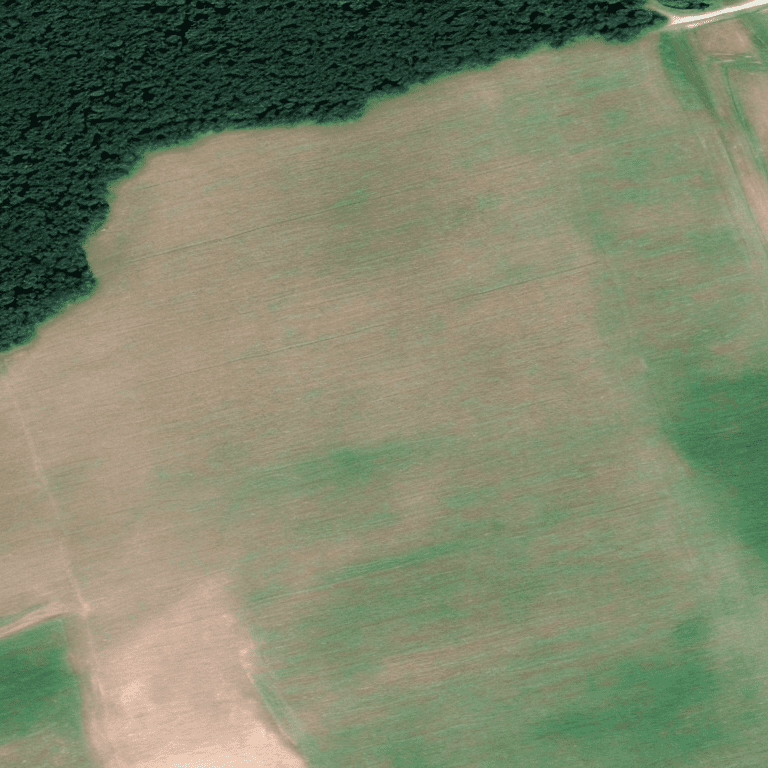}};
        
        \draw[red, very thick] (-0.7,-0.3) rectangle (1.8,-1.8);
        \end{tikzpicture}%
    }
    \caption{Sample generations with varying Surface net short-wave (solar) radiation (SSR). Higher SSR input values result in more saturated greens due to heavy sun exposure and vice-versa.}
    \label{fig:ref_label_ssr}
\end{figure}
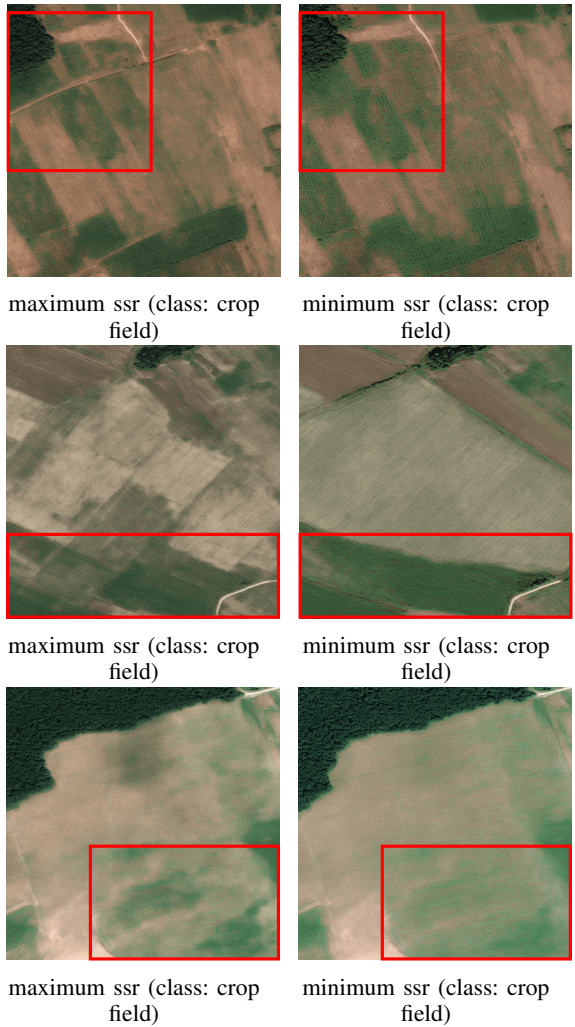
\begin{figure}[h]
    \centering
    \captionsetup[subfigure]{justification=centering}
    \subfloat[maximum tp (class: crop field)]{
        \label{example_1_max_tp}
        \begin{tikzpicture}
        \node[inner sep=0pt] () at (0,0) {\includegraphics[trim = 0cm 0cm 0cm 0cm ,clip,width=0.20\textwidth]{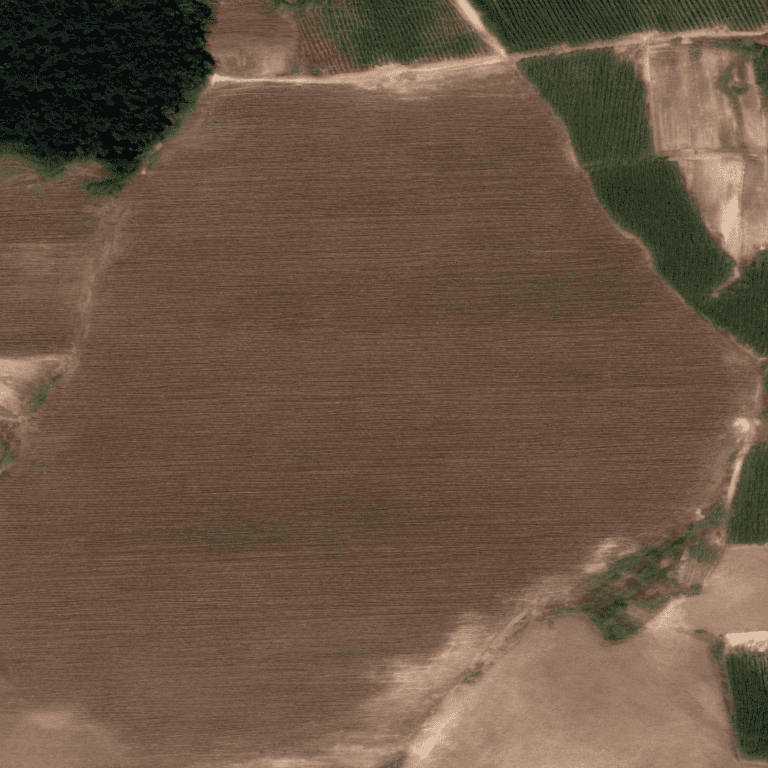}};
        \draw[red, very thick] (-0.5,1.8) rectangle (1.8,1.1);
        \draw[red, very thick] (1.0,0) rectangle (1.8,-1.78);
        \end{tikzpicture}%
    }
    \subfloat[minimum tp (class: crop field)]{
        \label{example_1_min_tp}
        \begin{tikzpicture}
        \node[inner sep=0pt] () at (0,0) {\includegraphics[trim = 0cm 0cm 0cm 0cm ,clip,width=0.20\textwidth]{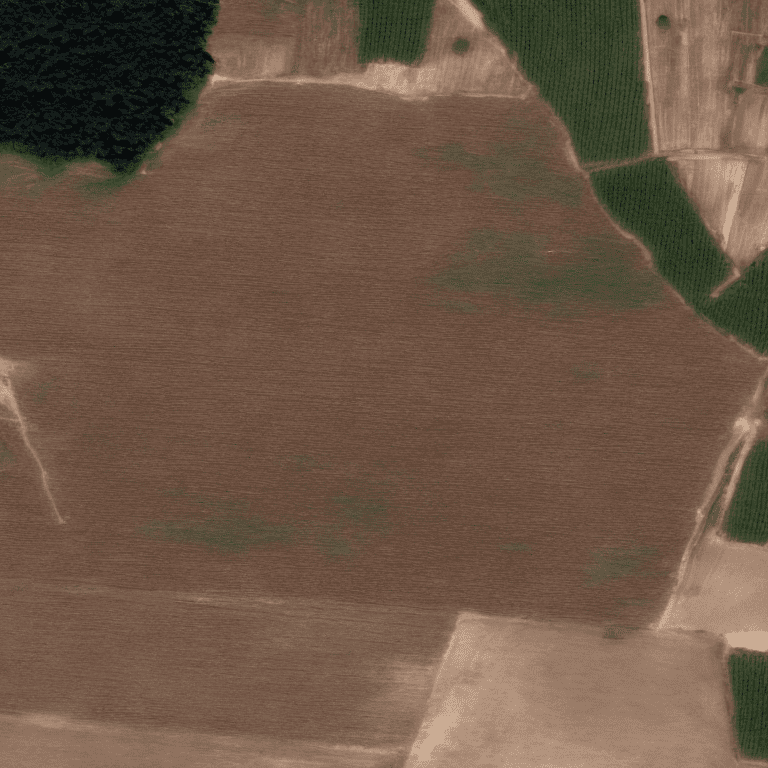}};
        \draw[red, very thick] (-0.5,1.8) rectangle (1.8,1.1);
        \draw[red, very thick] (1.0,0) rectangle (1.8,-1.78);
        \end{tikzpicture}%
    }\\
    \subfloat[maximum tp (class: dam)]{
        \label{example_2_max_tp}
        \begin{tikzpicture}
        \node[inner sep=0pt] () at (0,0) {\includegraphics[trim = 0cm 0cm 0cm 0cm ,clip,width=0.20\textwidth]{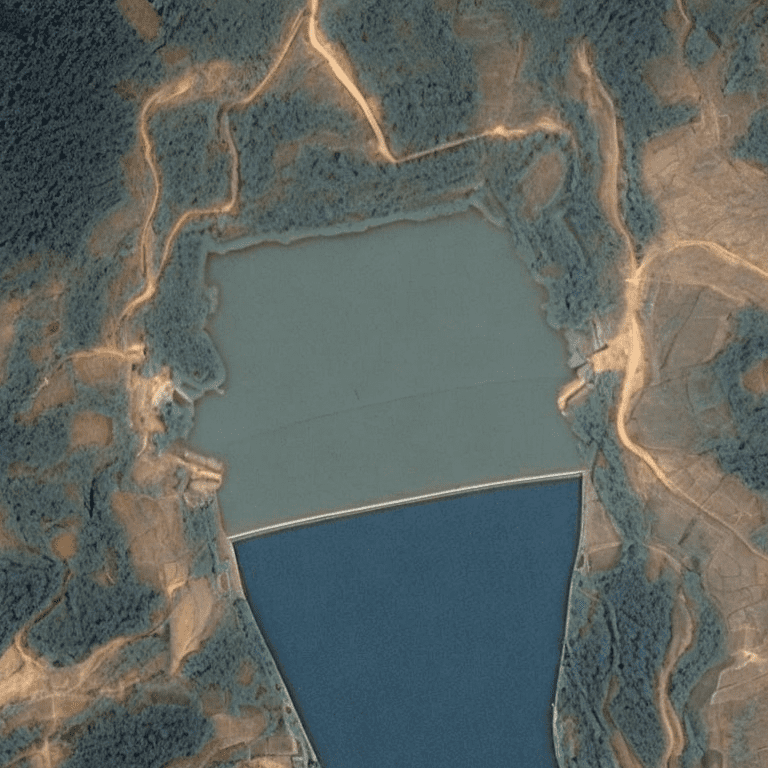}};
        \draw[red, very thick] (-0.9,1.2) rectangle (1,-0.2);
        \end{tikzpicture}%
    }
    \subfloat[minimum tp (class: dam)]{
        \label{example_2_min_tp}
        \begin{tikzpicture}
        \node[inner sep=0pt] () at (0,0) {\includegraphics[trim = 0cm 0cm 0cm 0cm ,clip,width=0.20\textwidth]{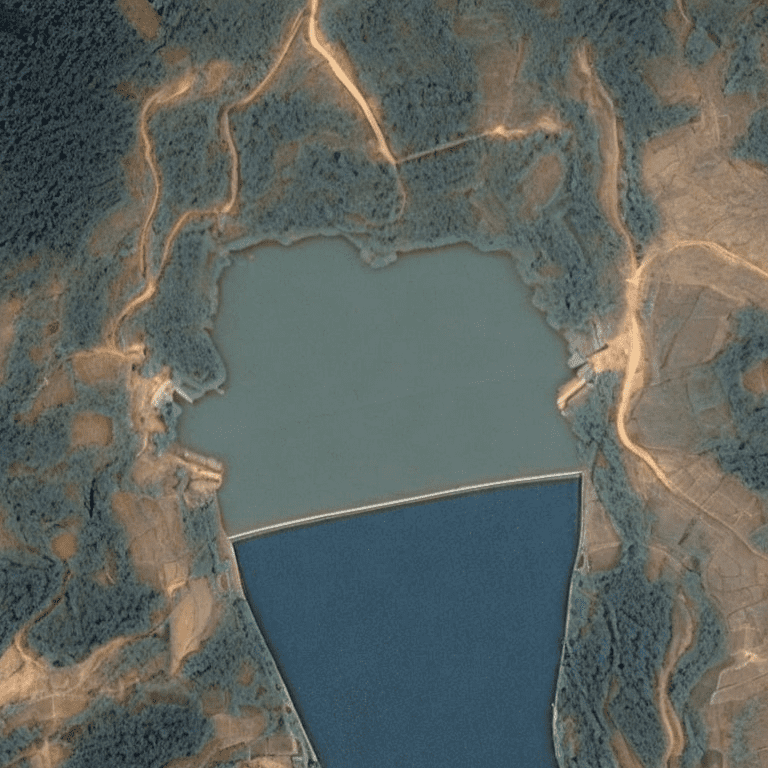}};
        \draw[red, very thick] (-0.9,1.2) rectangle (1,-0.2);
        \end{tikzpicture}%
    }\\
    \subfloat[maximum tp (class: dam)]{
        \label{example_3_max_tp}
        \begin{tikzpicture}
        \node[inner sep=0pt] () at (0,0) {\includegraphics[trim = 0cm 0cm 0cm 0cm ,clip,width=0.20\textwidth]{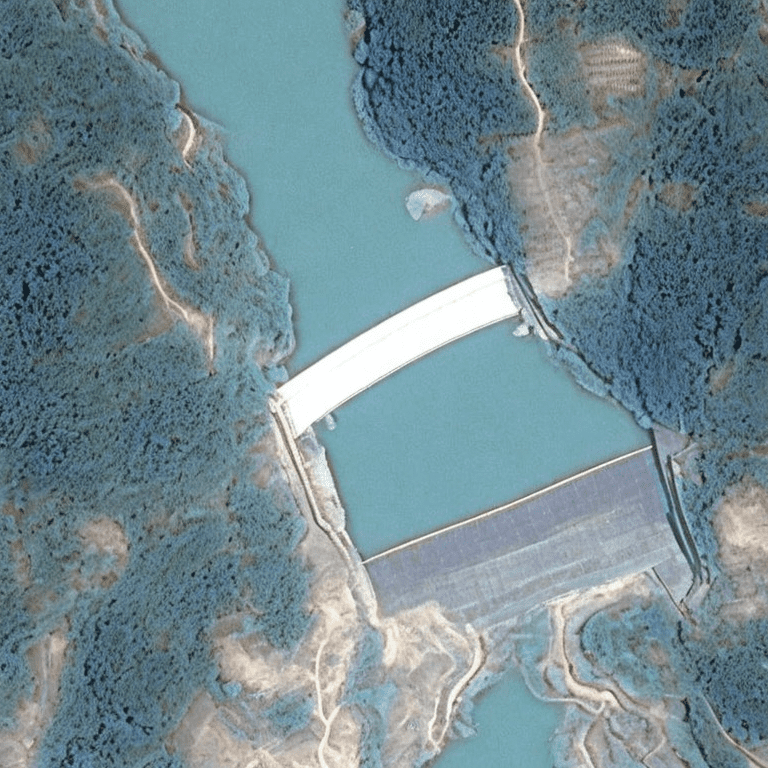}};
        \draw[red, very thick] (-0.3,1.8) rectangle (0.5,0.3);
        \draw[red, very thick] (0.7,0.3) rectangle (1.7,-0.7);
        \end{tikzpicture}%
    }
    \subfloat[minimum tp (class: dam)]{
        \label{example_3_min_tp}
        \begin{tikzpicture}
        \node[inner sep=0pt] () at (0,0) {\includegraphics[trim = 0cm 0cm 0cm 0cm ,clip,width=0.20\textwidth]{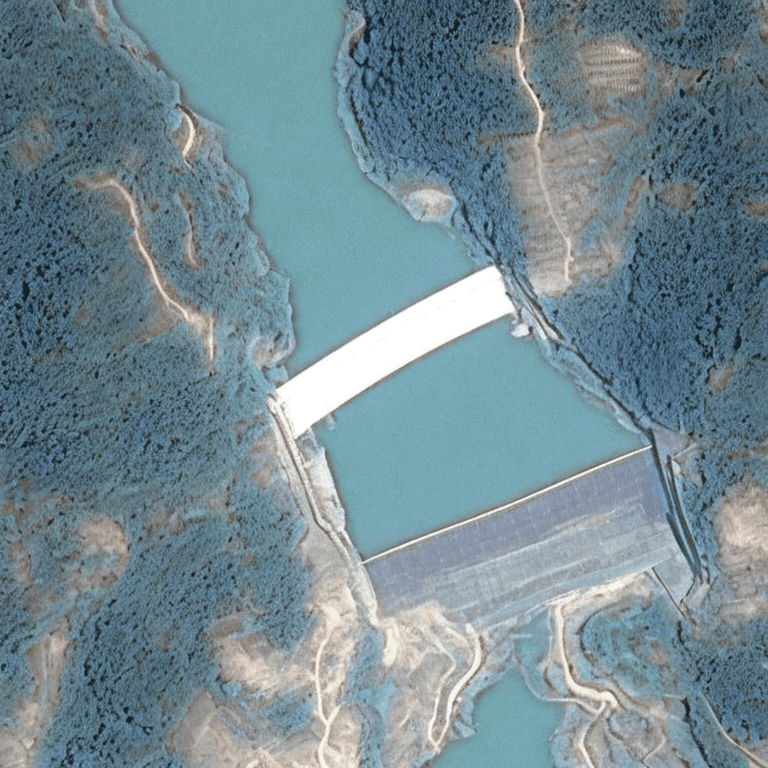}};
        \draw[red, very thick] (-0.3,1.8) rectangle (0.5,0.3);
        \draw[red, very thick] (0.7,0.3) rectangle (1.7,-0.7);
        \end{tikzpicture}%
    }
    
    \caption{Sample generations with varying Total Precipitation (TP). Higher TP input values result in either more defined greens or increased water presence.}
    \label{fig:ref_label_tp}
\end{figure}
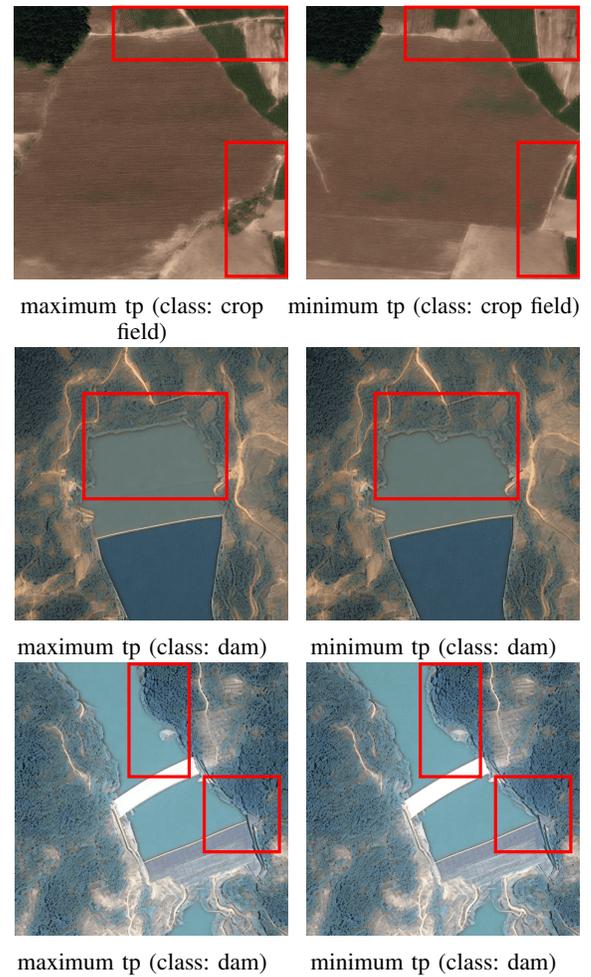

Further evidence that our model offers a more reliable alternative to the one proposed by \citep{diffusionsat} is also provided via a series of qualitative experiments (Figures \ref{tbl:ref_label_gsd}-\ref{tbl:ref_class_variation}) in which we investigate the effect different control inputs (i.e. metadata and/or captions) have on an image generated either by our or \cite{diffusionsat}'s models. In these experiments, we found that, while altering the value for the $gsd$ metadatum seems to indeed yield the expected results for both models (Table \ref{tbl:ref_label_gsd}), this appears to not be the case for the rest of the control inputs examined. More specifically, despite DiffusionSAT being adequately equipped to handle different values for the $tcc$, $month$, and the two ($lat$, $lon$) coordinate metadata, any alterations in these input values fail to be reflected in most of this model's sample generations (Figures \ref{tbl:ref_label_tcc}, \ref{tbl:ref_label_epoch}, and \ref{tbl:ref_class_variation}, respectively). This comes in contrast to our model's capabilities of reliably adapting the cloud coverage (Fig. \ref{tbl:ref_label_tcc}) the seasonal elements (Fig. \ref{tbl:ref_label_epoch}), and the locality-specific elements of a target class (Fig. \ref{tbl:ref_class_variation}) within a picture to the input data provided. Furthermore, by equipping our model with further environment-specific metadata such as \textit{surface net solar radiation} and  \textit{total precipitation} we are able to introduce more control over the generation process as evidenced by metadata-driven changes in flora colour saturation and water presence in Figures \ref{fig:ref_label_ssr} and \ref{fig:ref_label_tp}, respectively. Overall, it is evident that conditioning on multiple types of context (i.e., incorporating both location-, time-, along with a great variety of climate-based context variables to our control signals) leads to generations of greater quality and fidelity, thus rendering our model able to outperform prior studies that tend to leverage a more limited context selection for their generations.

Last but not least, we sought to further assess our model's capabilities at providing quality generations (i.e. generations of high fidelity and accuracy) given varying levels of missing metadata. Given its concatenation-and-mapping metadata approach when handling metadata -- a feature that enables modeling possible interactions between them -- it is expected that our model will exhibit great robustness in such a setting, thus differentiating itself from more direct methods used in prior works \citep{diffusionsat} that opt to form their conditioning embeddings by simply adding any attribute embeddings together. To that end, we incorporate all novel metadata fields introduced in this study into its metadata selection while keeping the rest of its architecture unchanged. This model differs from our proposed version solely in its metadata handling. As can be seen in Fig. \ref{tab:ref_label_addition_missing}, despite having access to more metadata fields, our custom augmented version - albeit still addition-based - continues to suffer from robustness issues when metadata is missing. Problems related to class fidelity and accuracy become even more pronounced as the available metadata decrease.


\renewcommand{\arraystretch}{0.95} 
\renewcommand{\arraystretch}{1.0} 

\begin{figure}[h]
    \centering
    \renewcommand{\arraystretch}{0.9} 
    \setlength{\tabcolsep}{1pt} 
    \begin{tabular}{@{}ll|ll@{}}
    \multicolumn{2}{c|}{\textbf{DiffusionSAT \cite{diffusionsat}}} & \multicolumn{2}{c}{\textbf{Ours}}\\\hline
    \textbf{month: }\textit{Jan.} & \textbf{month: }\textit{Aug.} & \textbf{month: }\textit{Jan.} & \textbf{month: }\textit{Aug.}\\\hline
    \includegraphics[width=0.12\textwidth]{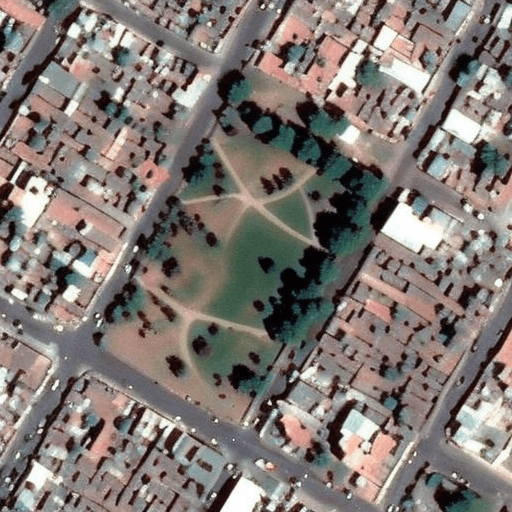} &
    \includegraphics[width=0.12\textwidth]{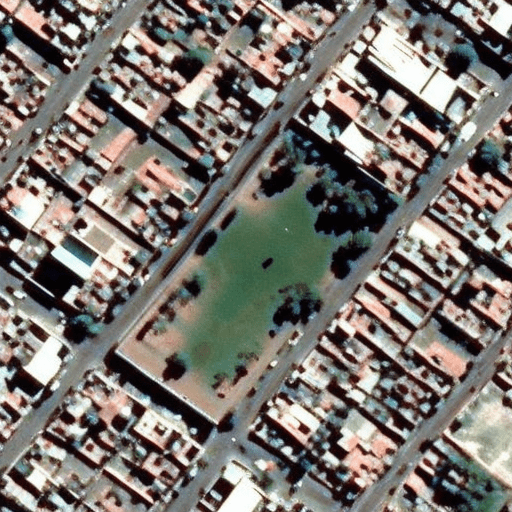} &
    \includegraphics[width=0.12\textwidth]{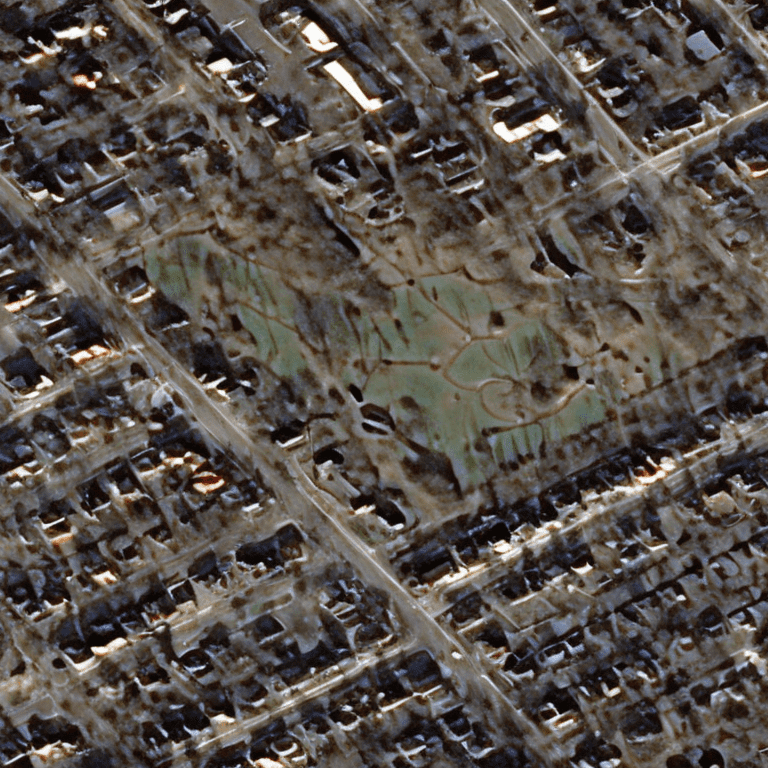} &
    \includegraphics[width=0.12\textwidth]{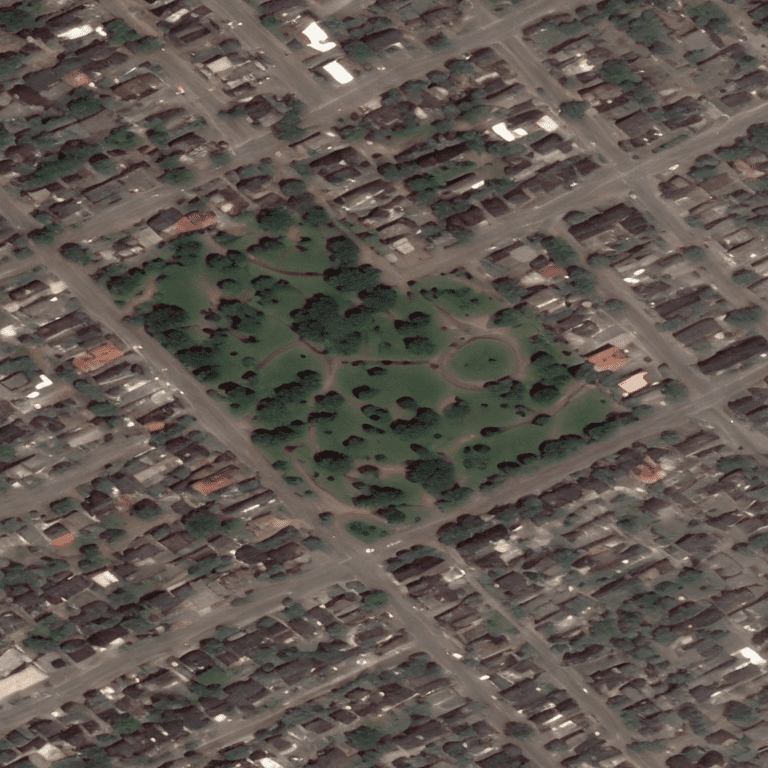}\\
    \multicolumn{4}{c}{\textbf{class: }\textit{park}}\\\hline
    \textbf{month: }\textit{Jan.} & \textbf{month: }\textit{Aug.} & \textbf{month: }\textit{Jan.} & \textbf{month: }\textit{Aug.}\\\hline
    \includegraphics[width=0.12\textwidth]{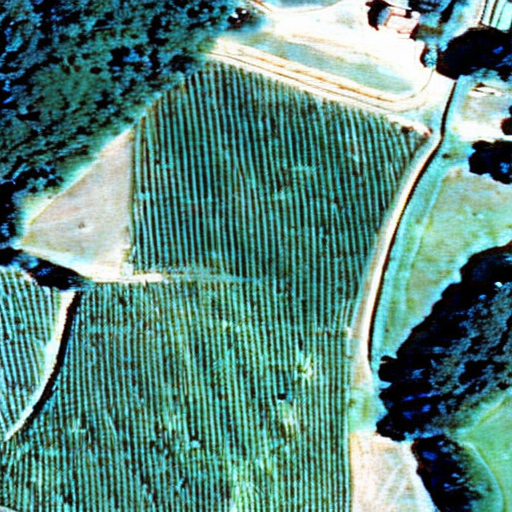} &
    \includegraphics[width=0.12\textwidth]{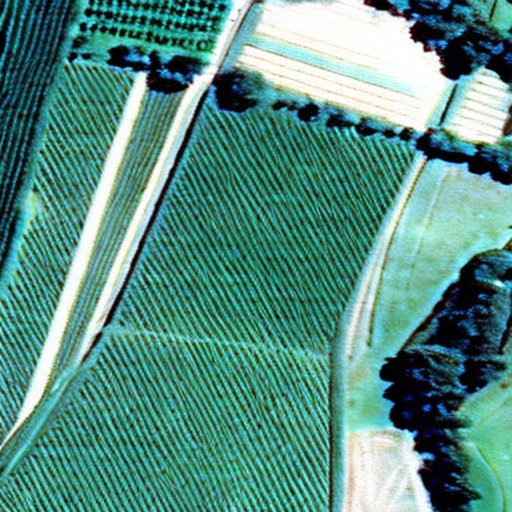} &
    \includegraphics[width=0.12\textwidth]{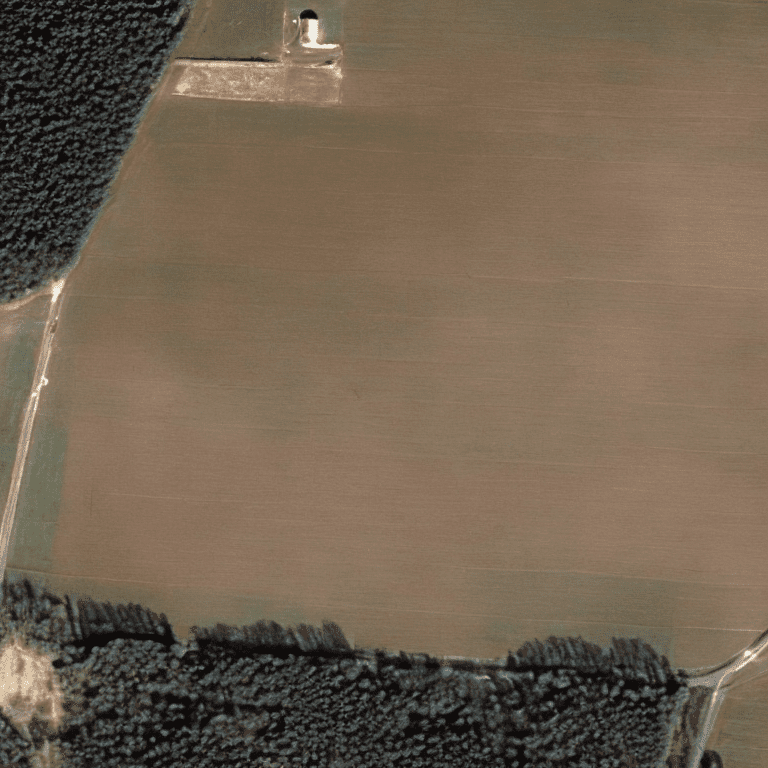} &
    \includegraphics[width=0.12\textwidth]{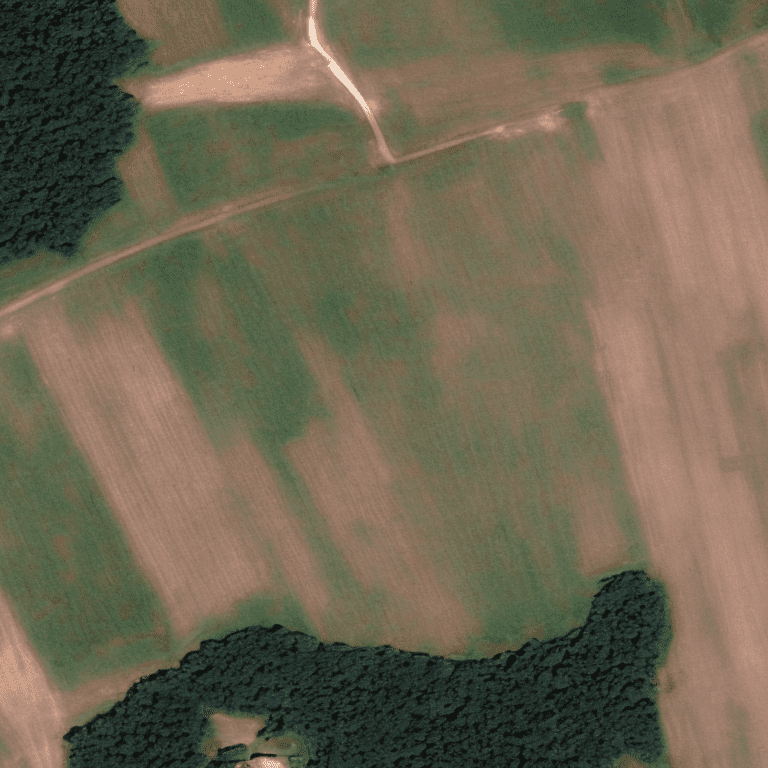}\\
    \multicolumn{4}{c}{\textbf{class: }\textit{crop field}}\\\hline
    \textbf{month: }\textit{Nov.} & \textbf{month: }\textit{Feb.} & \textbf{month: }\textit{Nov.} & \textbf{month: }\textit{Feb.}\\\hline
    \includegraphics[width=0.12\textwidth]{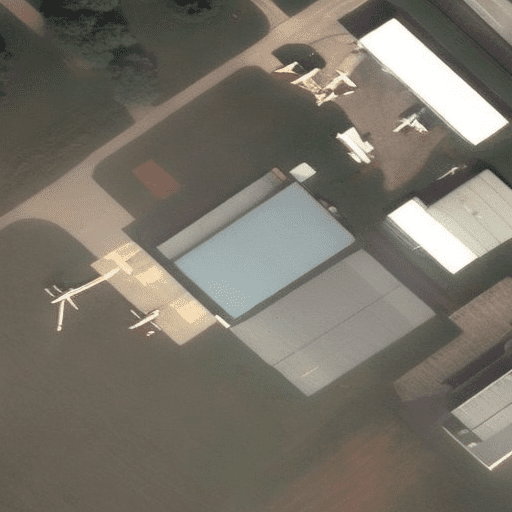} &
    \includegraphics[width=0.12\textwidth]{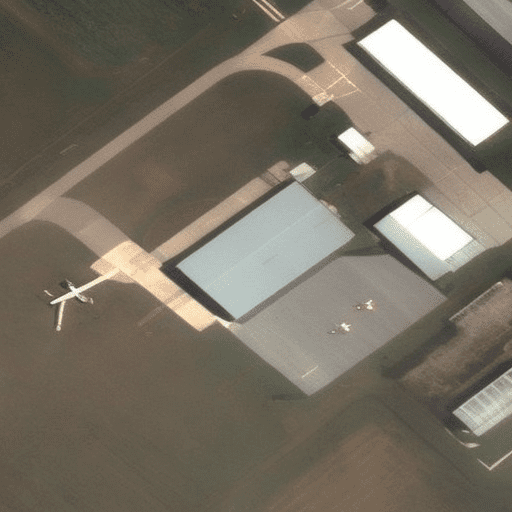} &
    \includegraphics[width=0.12\textwidth]{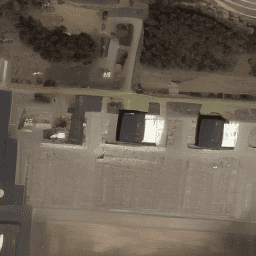} &
    \includegraphics[width=0.12\textwidth]{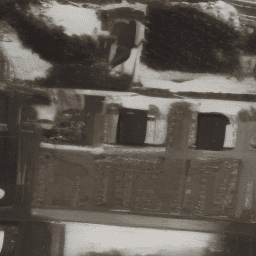}\\
    \multicolumn{4}{c}{\textbf{class: }\textit{airport hangar}}
    \end{tabular}
    \caption{Sample generations from \citep{diffusionsat}'s and our model with varying the month metadatum. Flora colour and growth change with season in our model's outputs. DiffusionSAT's outputs seem less affected.}
    \label{tbl:ref_label_epoch}
\end{figure}

\captionsetup[table]{skip=2pt} 


\begin{figure}[h]
    \centering
    \renewcommand{\arraystretch}{0.9} 
    \setlength{\tabcolsep}{1pt} 
    \begin{tabular}{lll|ll}
    & \multicolumn{2}{c|}{\textbf{DiffusionSAT \cite{diffusionsat}}}   & \multicolumn{2}{c}{\textbf{Ours}}\\\hline
    & \textbf{place: } \textit{France} & \textbf{place: } \textit{USA} &  \textbf{place: } \textit{France} & \textbf{place: } \textit{USA}\\
     \rotatebox[origin=l]{90}{\parbox{10mm}{\centering prompt}} & \includegraphics[width=0.115\textwidth]{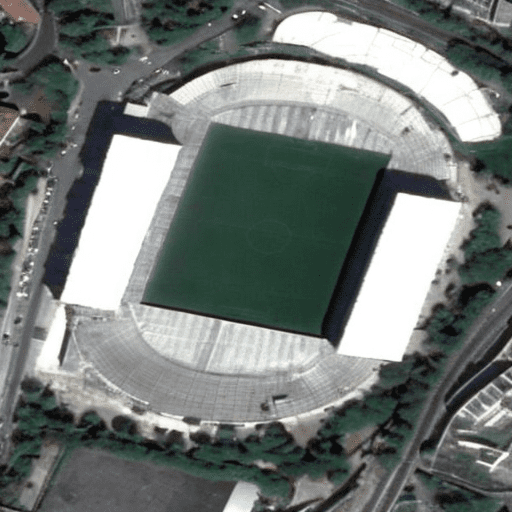} &
    \includegraphics[width=0.115\textwidth]{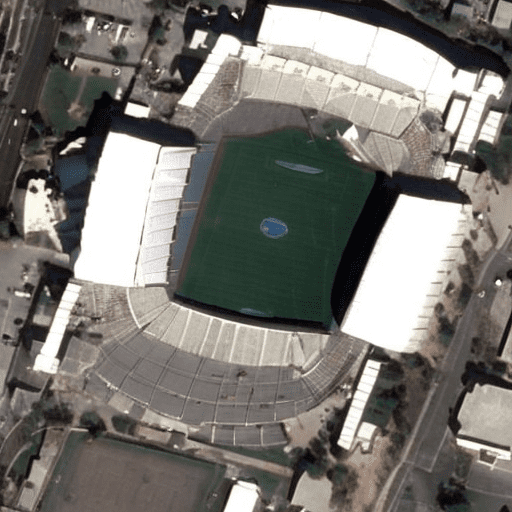} &
    \includegraphics[width=0.115\textwidth]{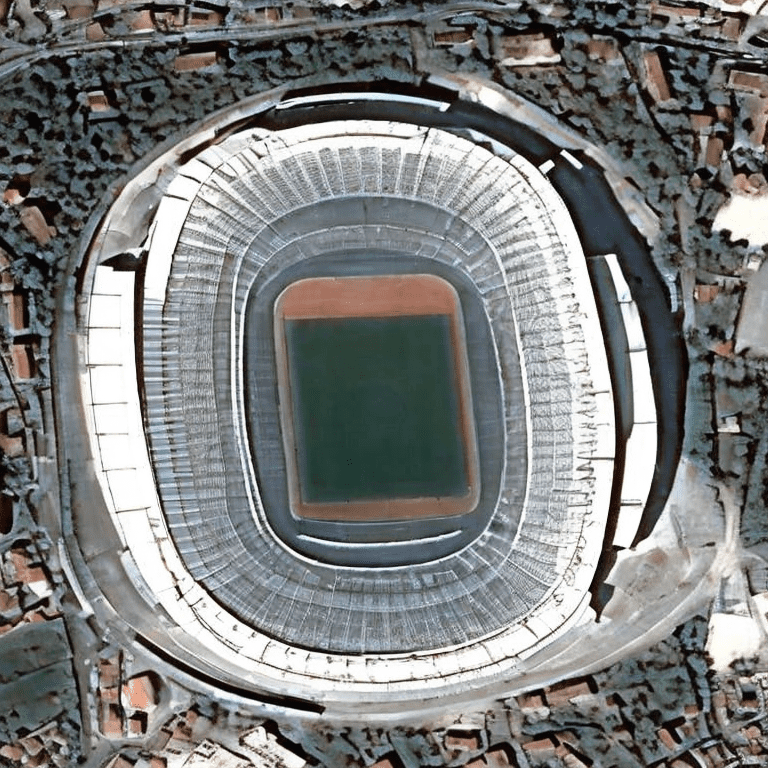} & \includegraphics[width=0.115\textwidth]{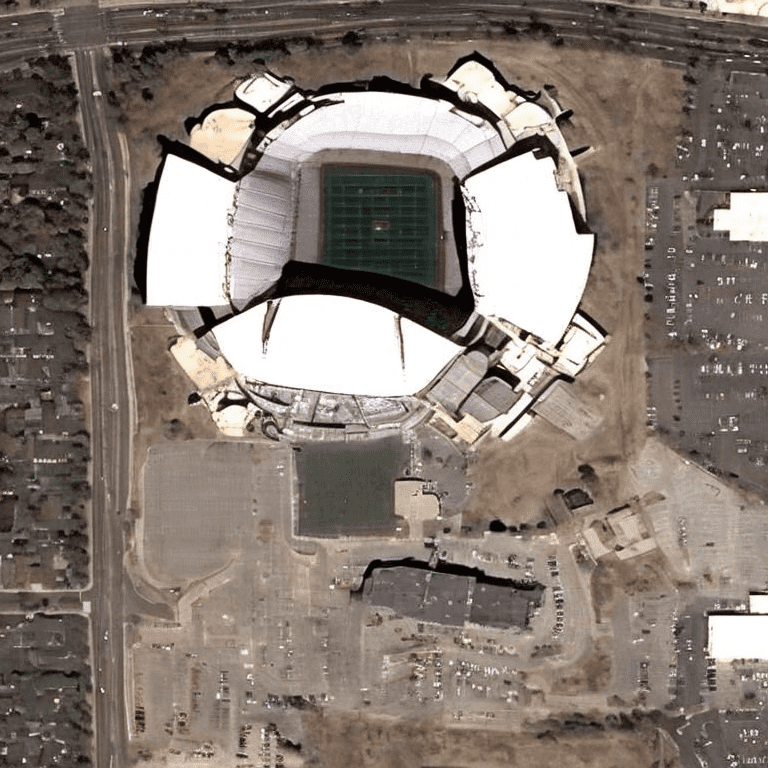}\\
     \rotatebox[origin=l]{90}{\parbox{10mm}{\centering metadata}} & \includegraphics[width=0.115\textwidth]{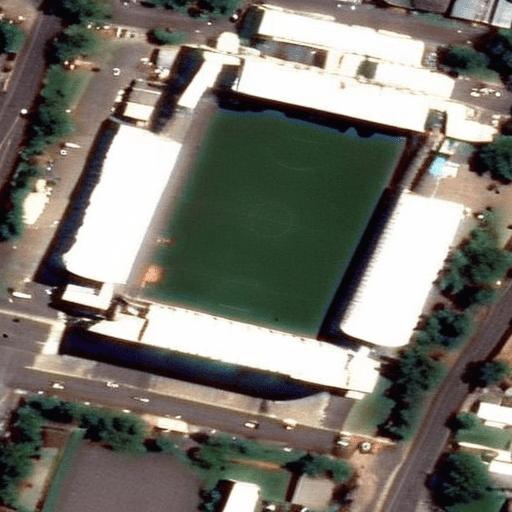} &
    \includegraphics[width=0.115\textwidth]{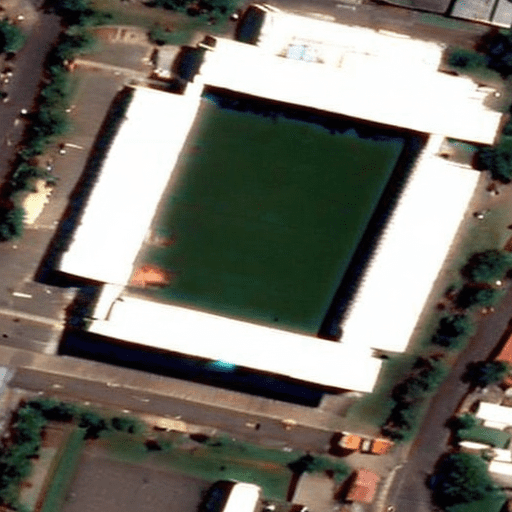} &
    \includegraphics[width=0.115\textwidth]{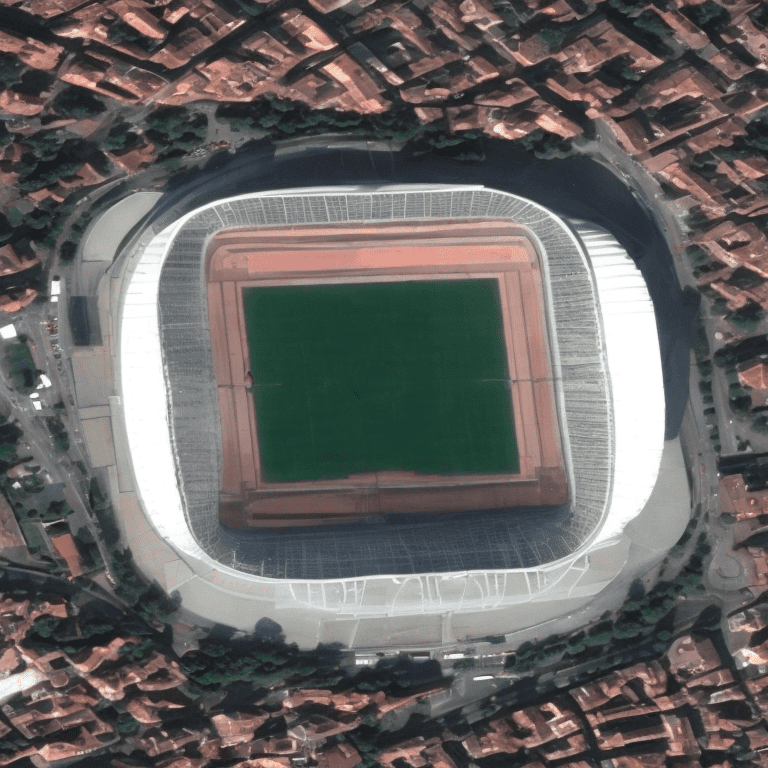} & \includegraphics[width=0.115\textwidth]{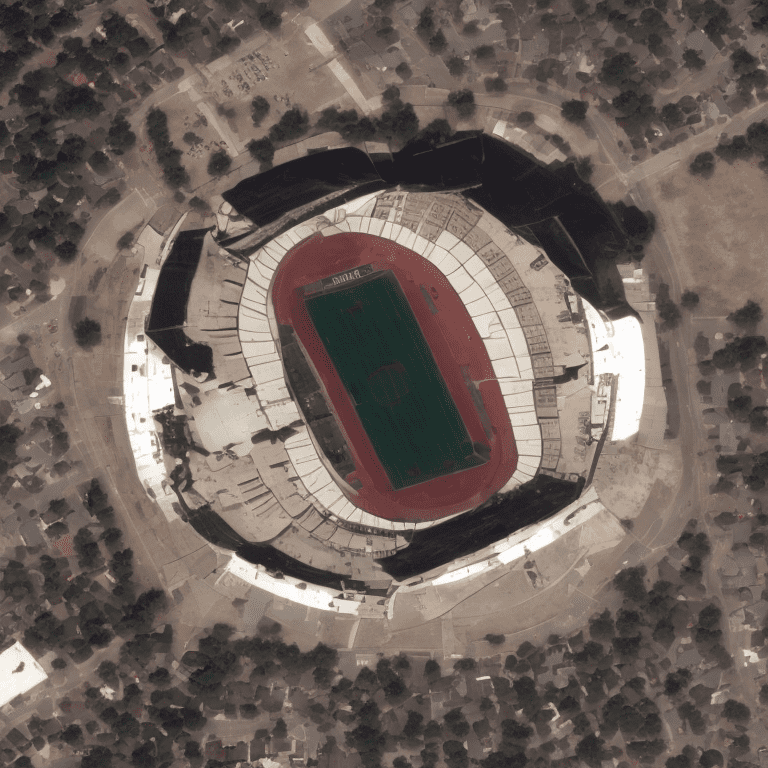}\\
    \rotatebox[origin=l]{90}{\parbox{10mm}{\centering both}} & \includegraphics[width=0.115\textwidth]{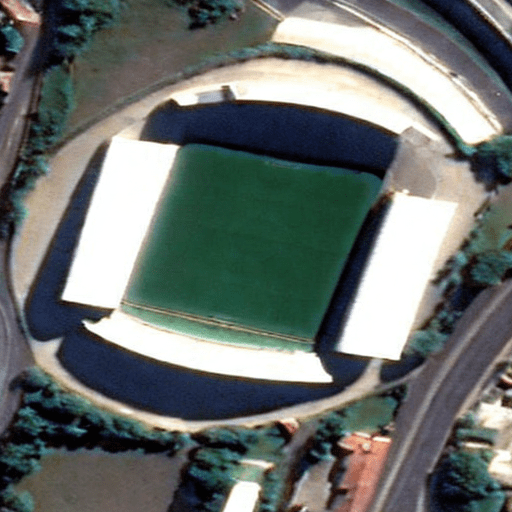} &
    \includegraphics[width=0.115\textwidth]{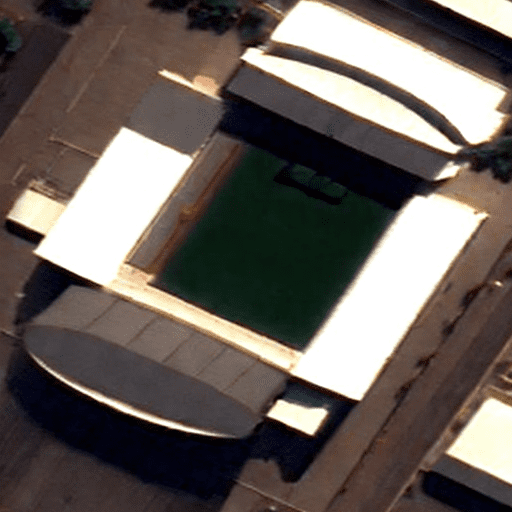} &
    \includegraphics[width=0.115\textwidth]{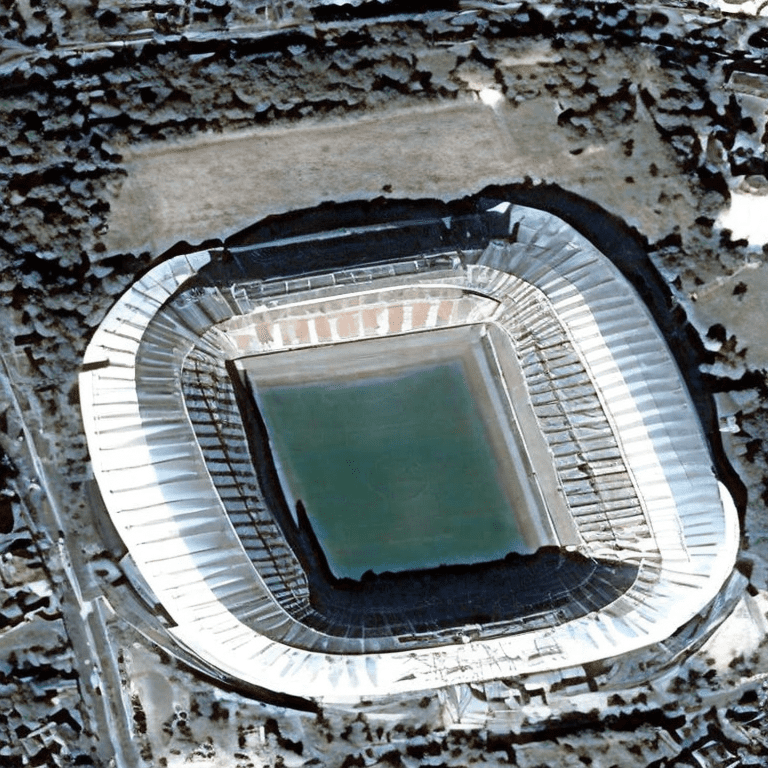} & \includegraphics[width=0.115\textwidth]{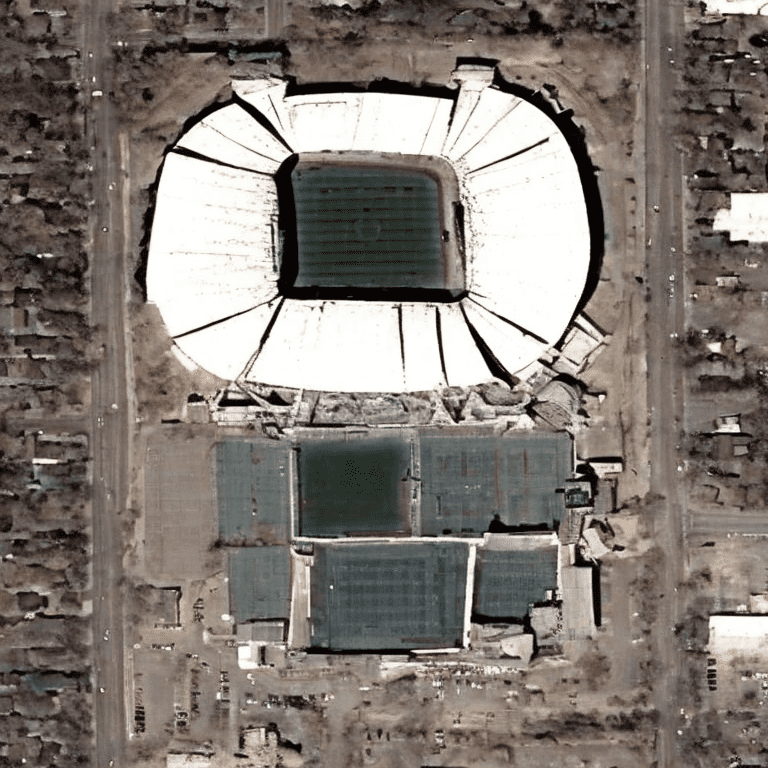}\\
    & \multicolumn{4}{c}{\textbf{class: }\textit{stadium}}\\\hline
    & \textbf{place: } \textit{Greece} & \textbf{place: } \textit{Japan} &  \textbf{place: } \textit{Greece} & \textbf{place: } \textit{Japan}\\
    \rotatebox[origin=l]{90}{\parbox{10mm}{\centering prompt}} & \includegraphics[width=0.115\textwidth]{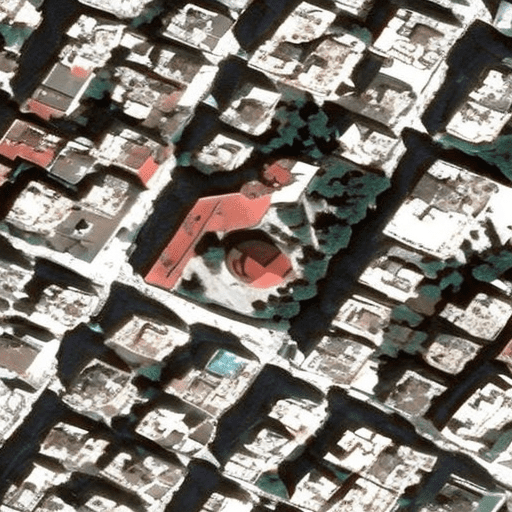} &
    \includegraphics[width=0.115\textwidth]{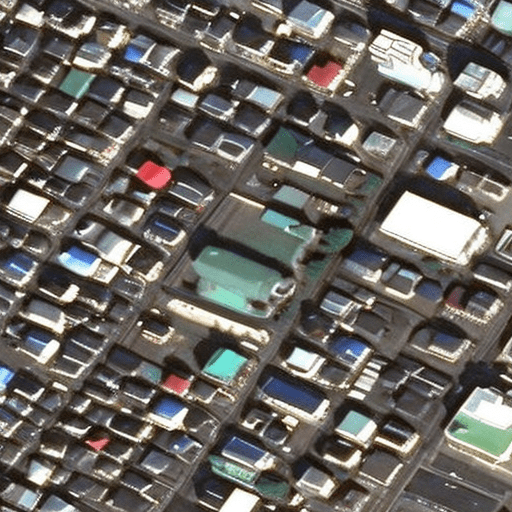} &
    \includegraphics[width=0.115\textwidth]{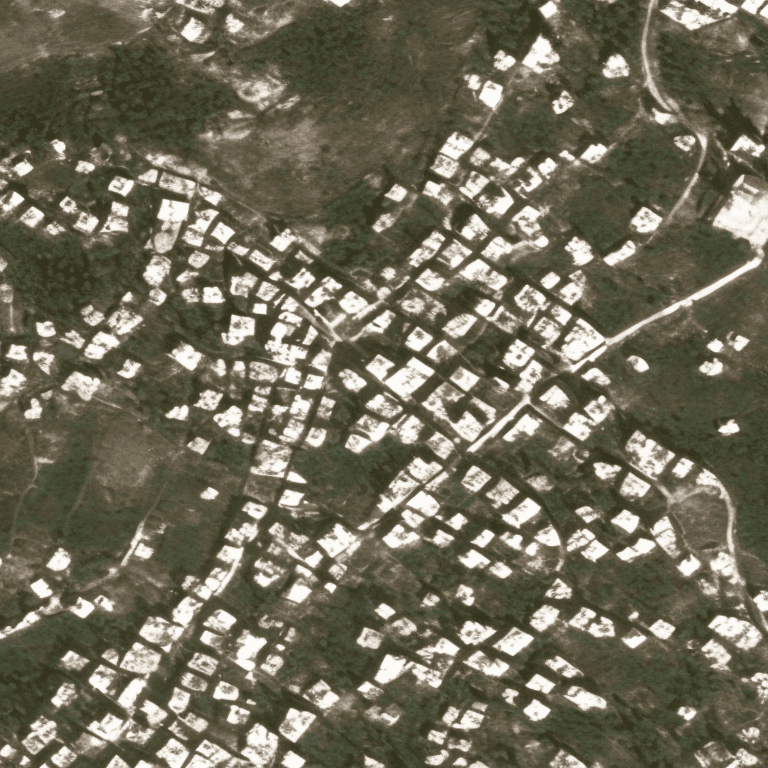} & \includegraphics[width=0.115\textwidth]{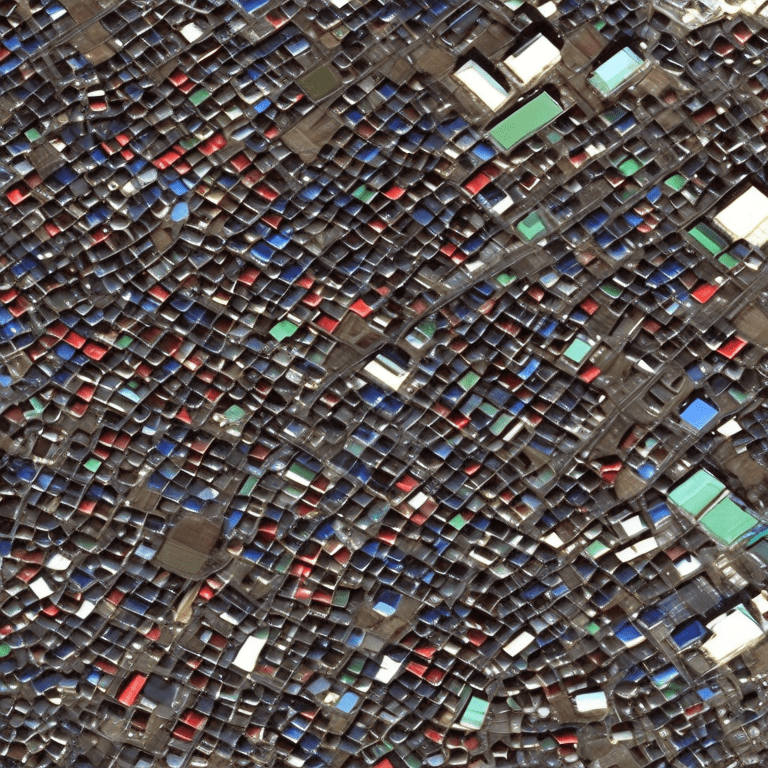}\\
    \rotatebox[origin=l]{90}{\parbox{10mm}{\centering metadata}} & \includegraphics[width=0.115\textwidth]{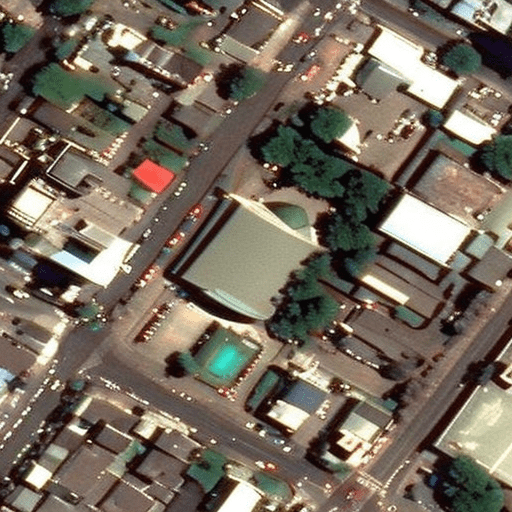} &
    \includegraphics[width=0.115\textwidth]{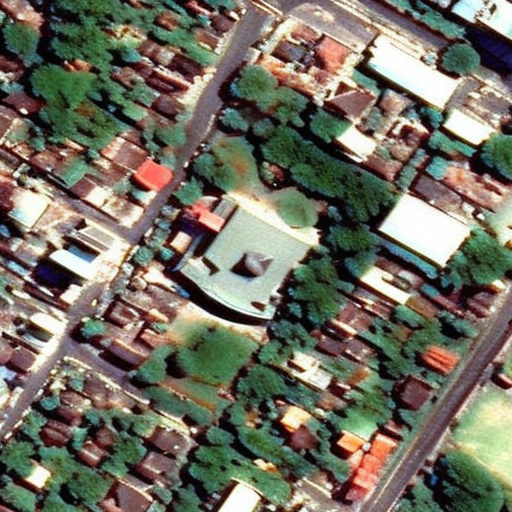} &
    \includegraphics[width=0.115\textwidth]{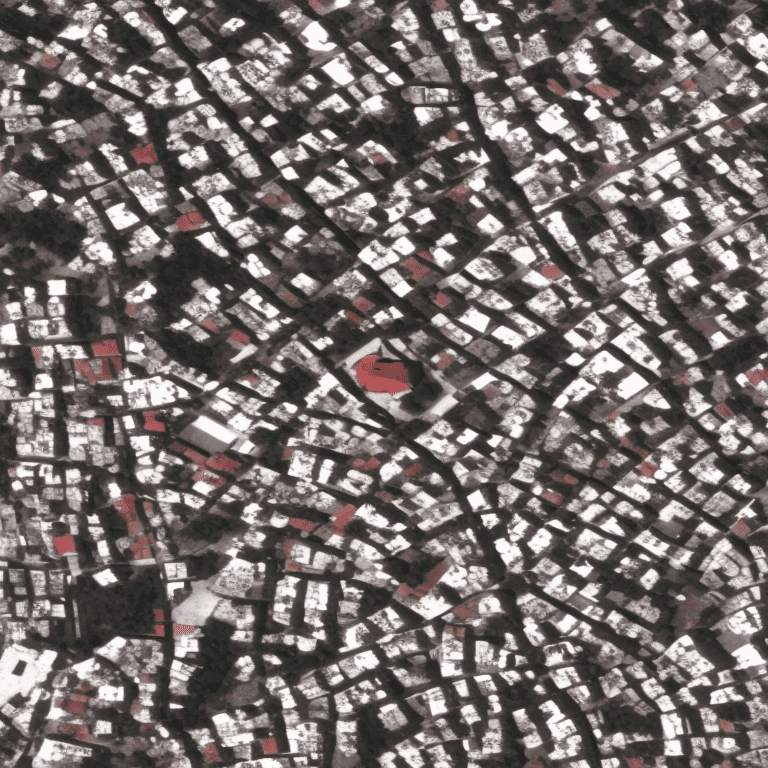} & \includegraphics[width=0.115\textwidth]{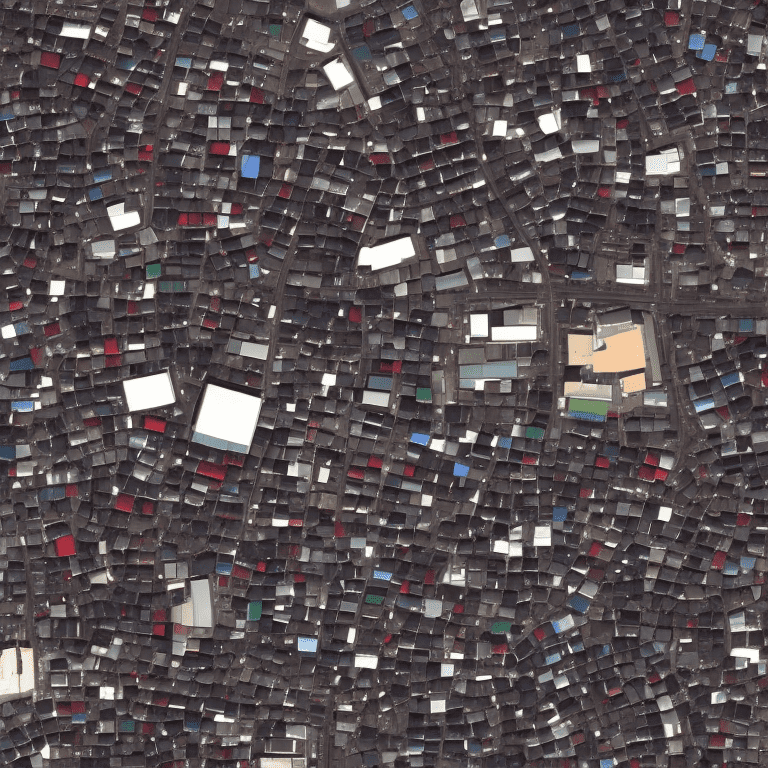}\\
    \rotatebox[origin=l]{90}{\parbox{10mm}{\centering both}} & \includegraphics[width=0.115\textwidth]{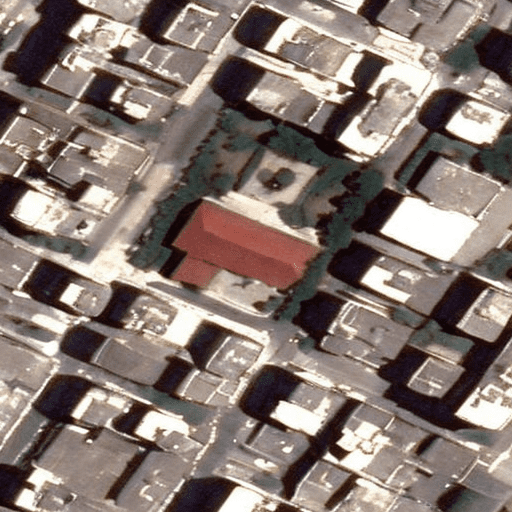} &
    \includegraphics[width=0.115\textwidth]{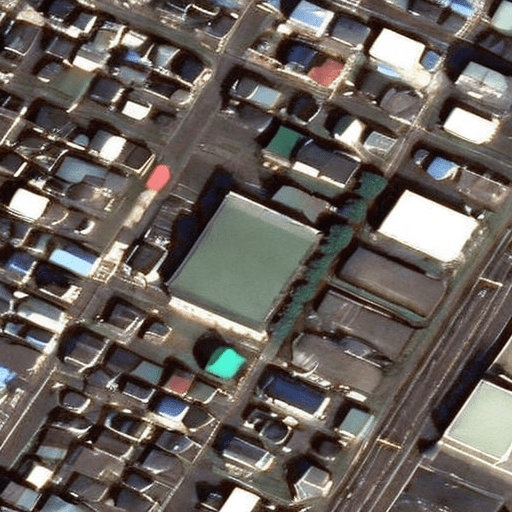} &
    \includegraphics[width=0.115\textwidth]{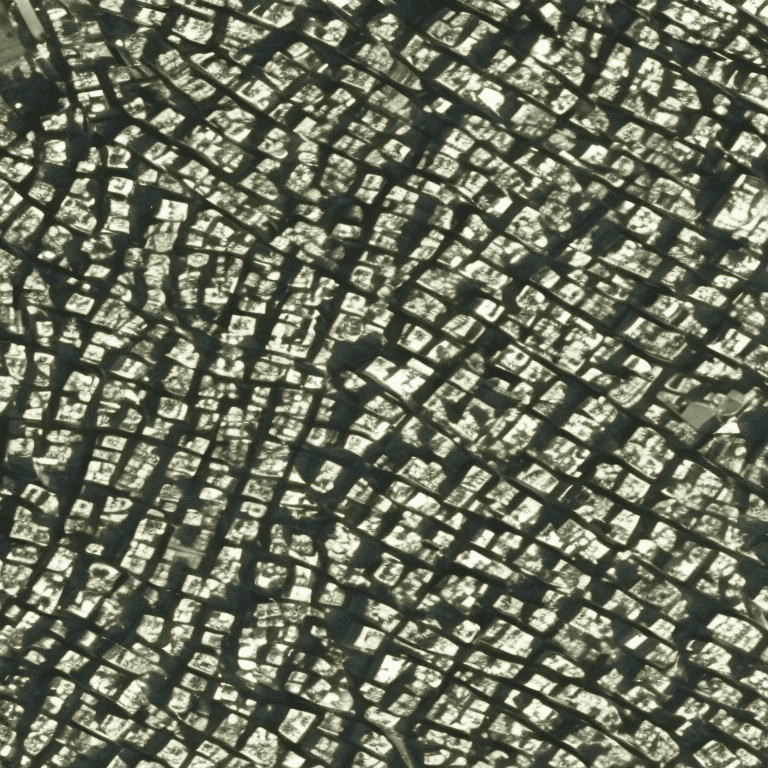} & \includegraphics[width=0.115\textwidth]{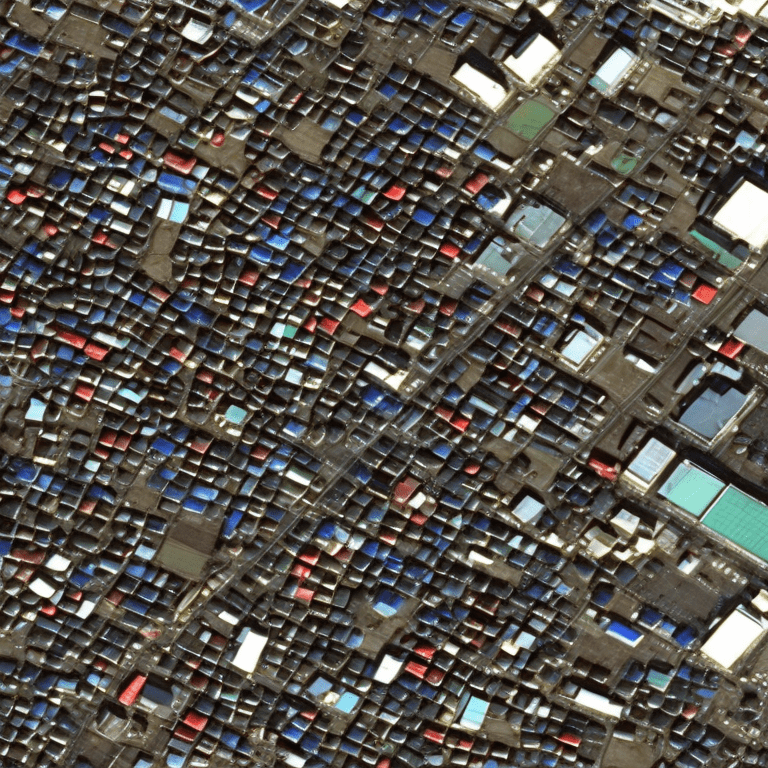}\\
    & \multicolumn{4}{c}{\textbf{class: }\textit{place of worship}}\\\hline
    \end{tabular}
    \caption{Sample generations sampled from \citep{diffusionsat}'s and our model depicting class variation based on topography. As far as our model's generations are concerned, stadium type varies between that of american football and continental football given corresponding control input values either on a) prompt-level, b) metadata-level, or both. Place of worship type also varies, depending on location. DiffusionSAT, on the other hand, most of the time fails to adapt its generations to the control input and, for cases when it does (here, in the \textit{place\_of\_worship} class examples), explicit location information presence within the prompt (i.e. "in Japan" and/or "in Greece") seems essential.}
    \label{tbl:ref_class_variation}
\end{figure}

\begin{figure}[h]
    \centering
    \setlength{\tabcolsep}{1pt} 
    \renewcommand{\arraystretch}{0.85} 
    \begin{tabular}{@{}>{\centering\arraybackslash}p{0.03\textwidth}
                      >{\centering\arraybackslash}p{0.11\textwidth}|
                      >{\centering\arraybackslash}p{0.11\textwidth}|
                      >{\centering\arraybackslash}p{0.11\textwidth}|
                      >{\centering\arraybackslash}p{0.11\textwidth}@{}}
    & \textbf{missing metadata count: 0} 
    & \textbf{missing metadata count: 3} 
    & \textbf{missing metadata count: 5} 
    & \textbf{missing metadata count: 7}\\
   \rotatebox[origin=l]{90}{\parbox{14mm}{\centering \textbf{gsd: \textit{min}}}} 
   & \includegraphics[width=0.11\textwidth]{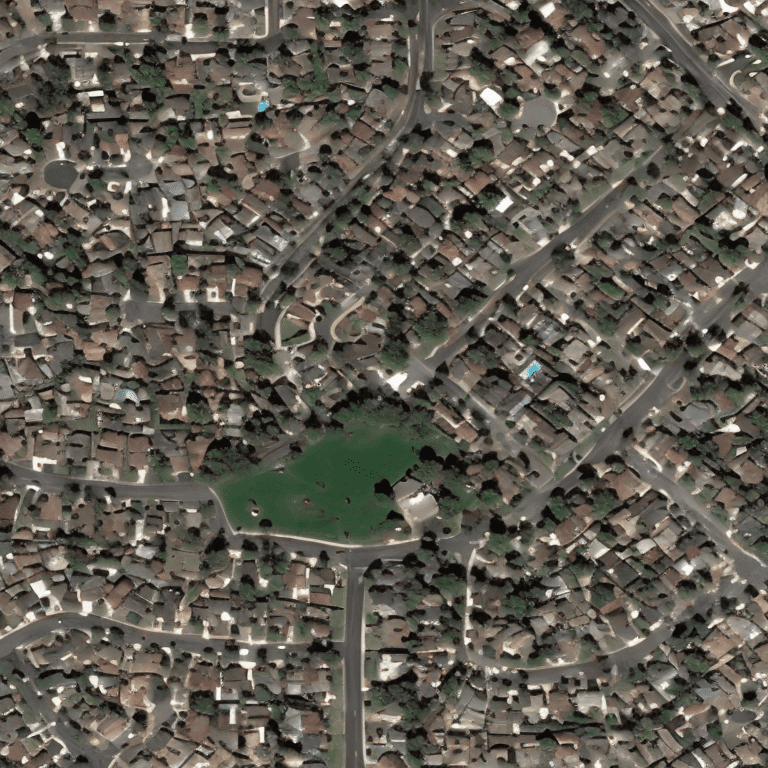} 
   & \includegraphics[width=0.11\textwidth]{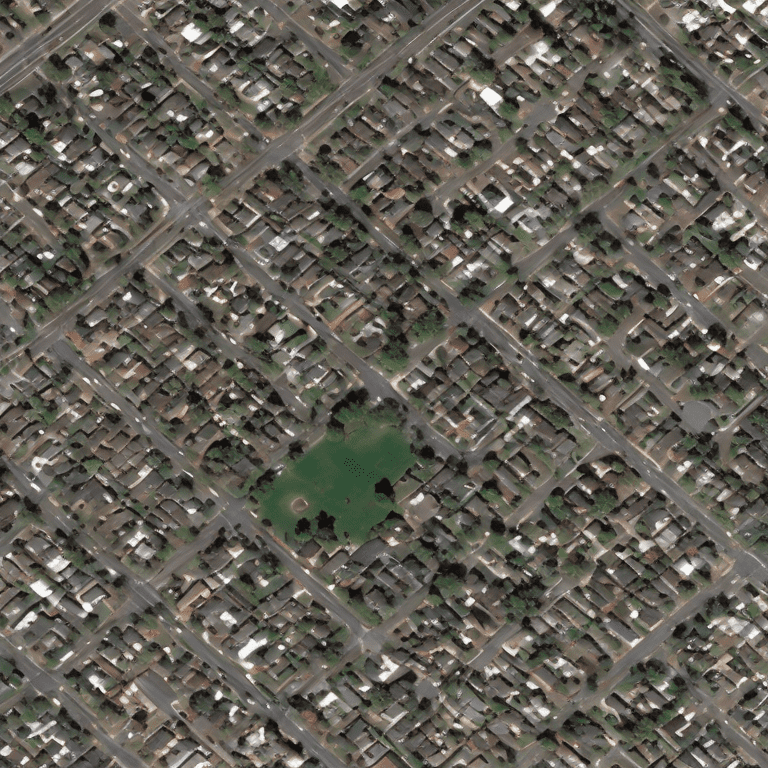} 
   & \includegraphics[width=0.11\textwidth]{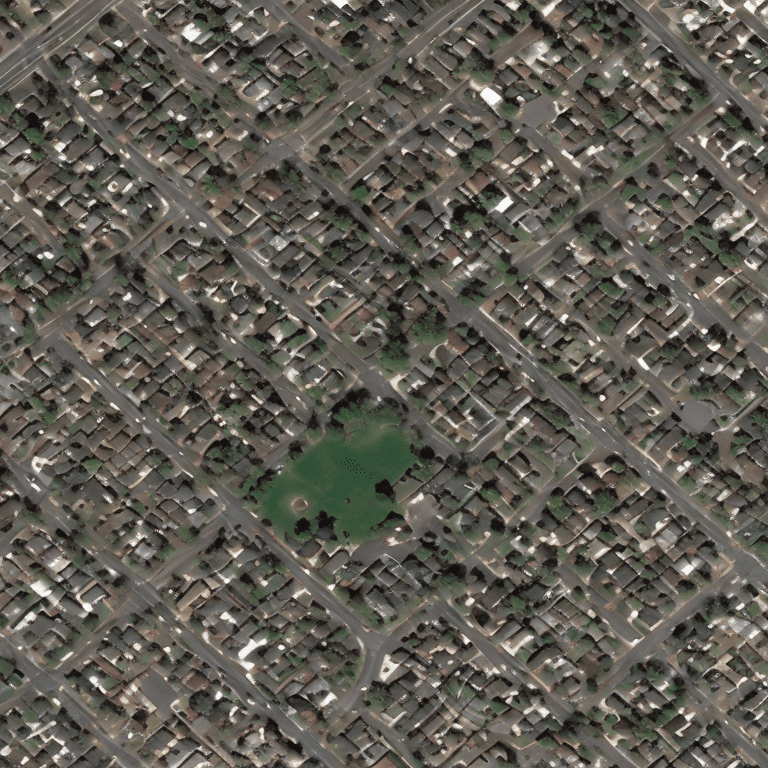} 
   & \includegraphics[width=0.11\textwidth]{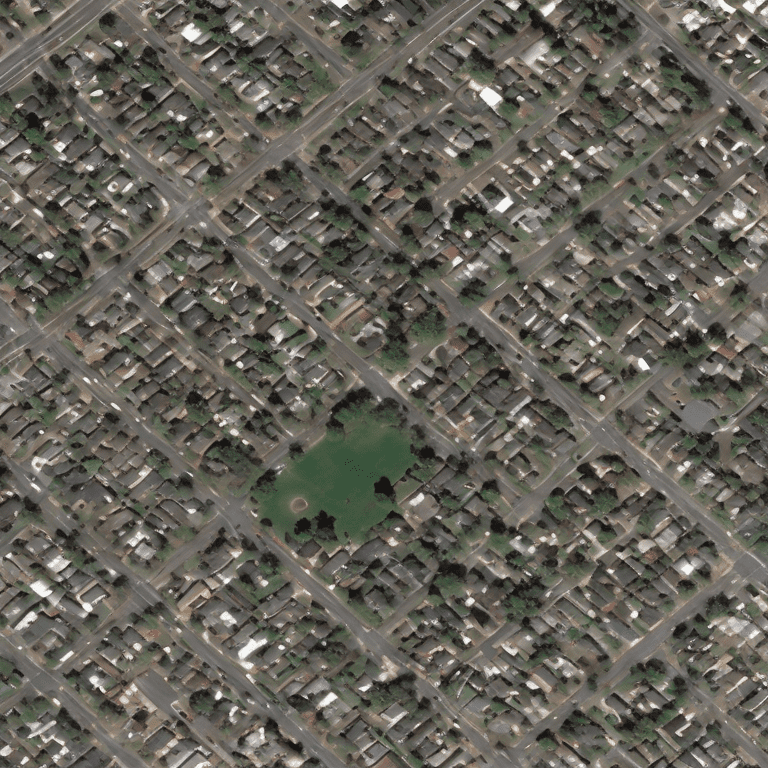} \\
   
   \rotatebox[origin=l]{90}{\parbox{14mm}{\centering \textbf{gsd: \textit{max}}}} 
   & \includegraphics[width=0.11\textwidth]{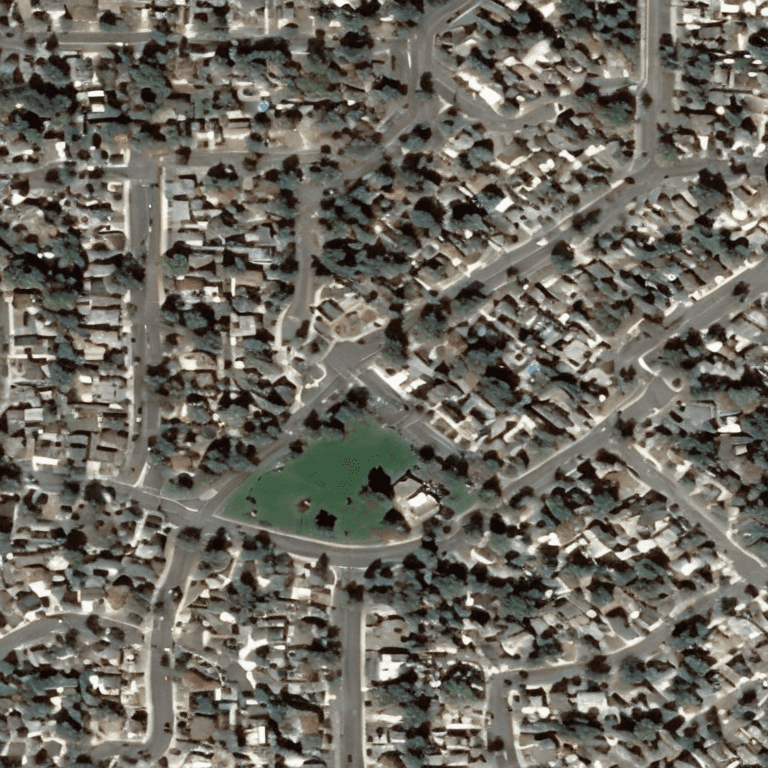} 
   & \includegraphics[width=0.11\textwidth]{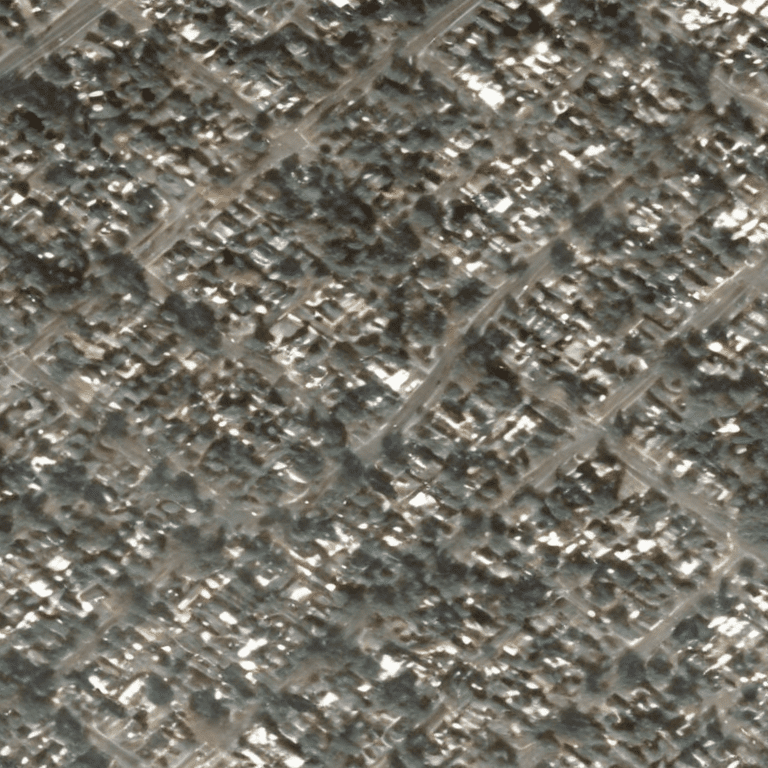} 
   & \includegraphics[width=0.11\textwidth]{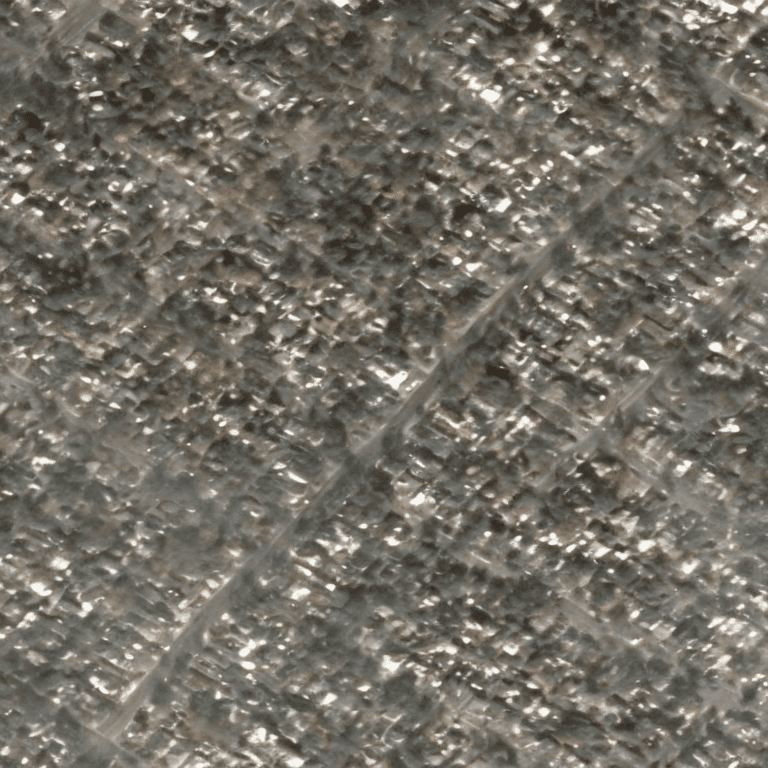} 
   & \includegraphics[width=0.11\textwidth]{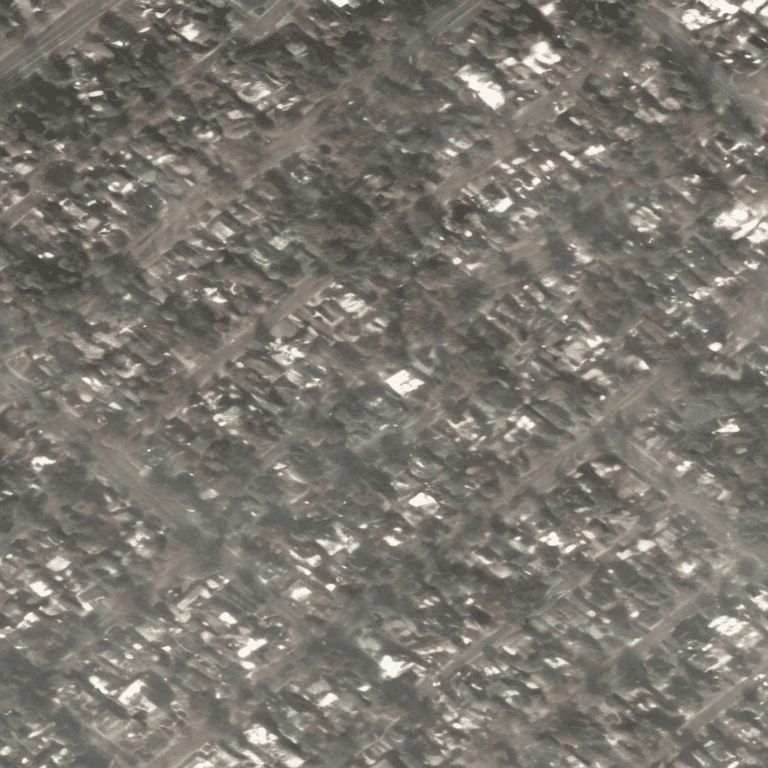} \\
   \cdashline{1-5}
   
   \rotatebox[origin=l]{90}{\parbox{14mm}{\centering \textbf{gsd: \textit{min}}}} 
   & \includegraphics[width=0.11\textwidth]{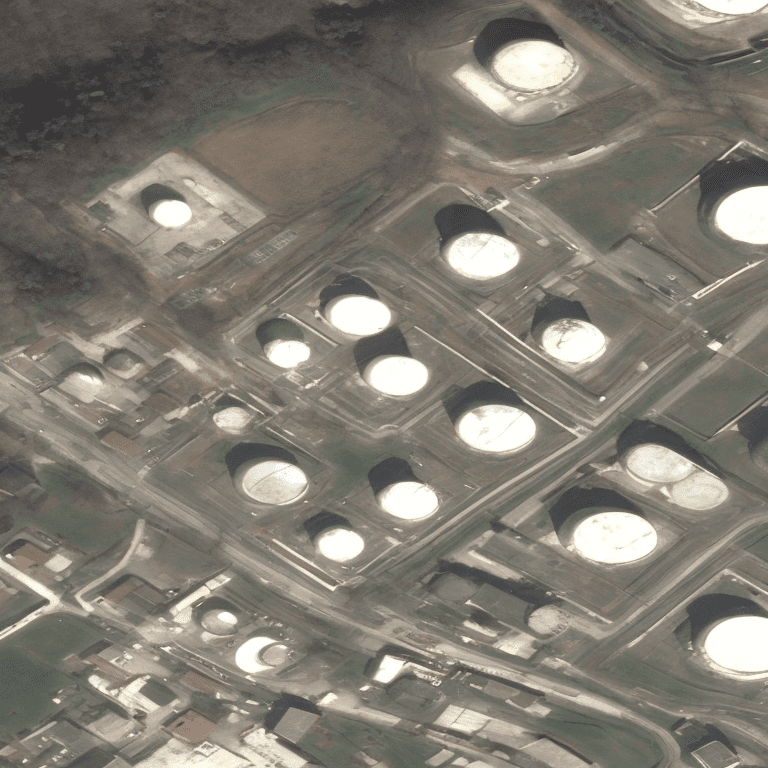} 
   & \includegraphics[width=0.11\textwidth]{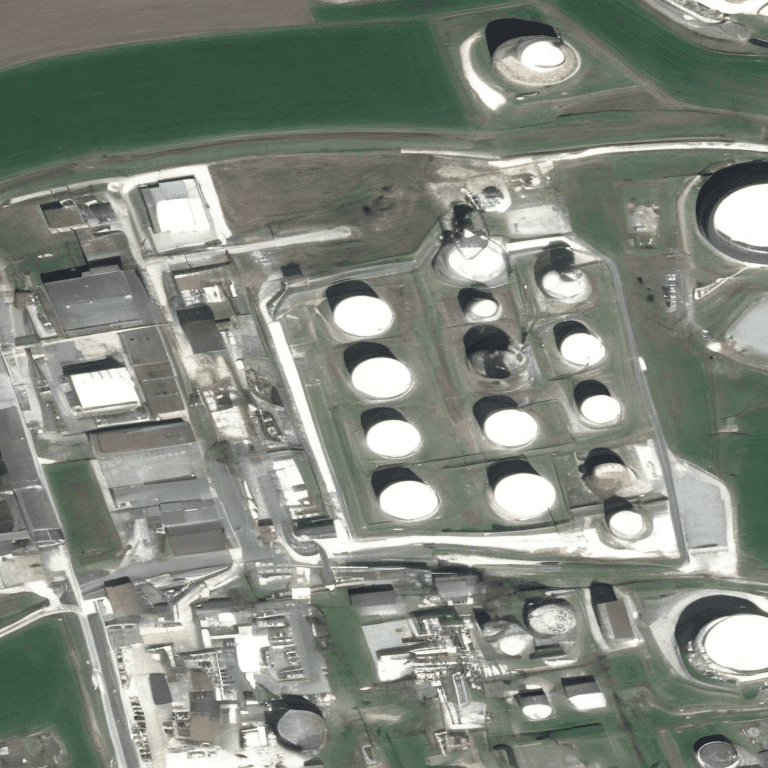}
   & \includegraphics[width=0.11\textwidth]{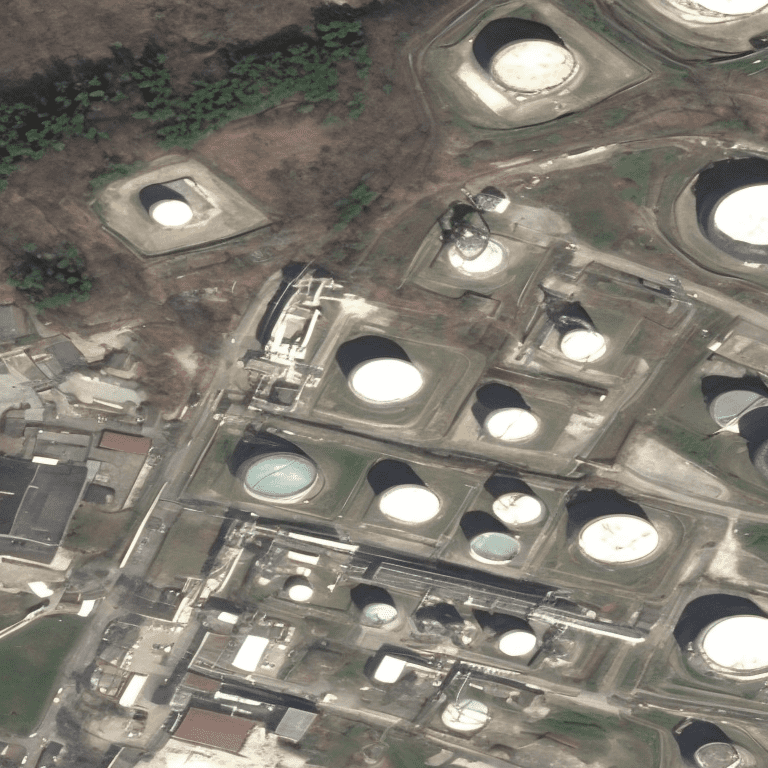} 
   & \includegraphics[width=0.11\textwidth]{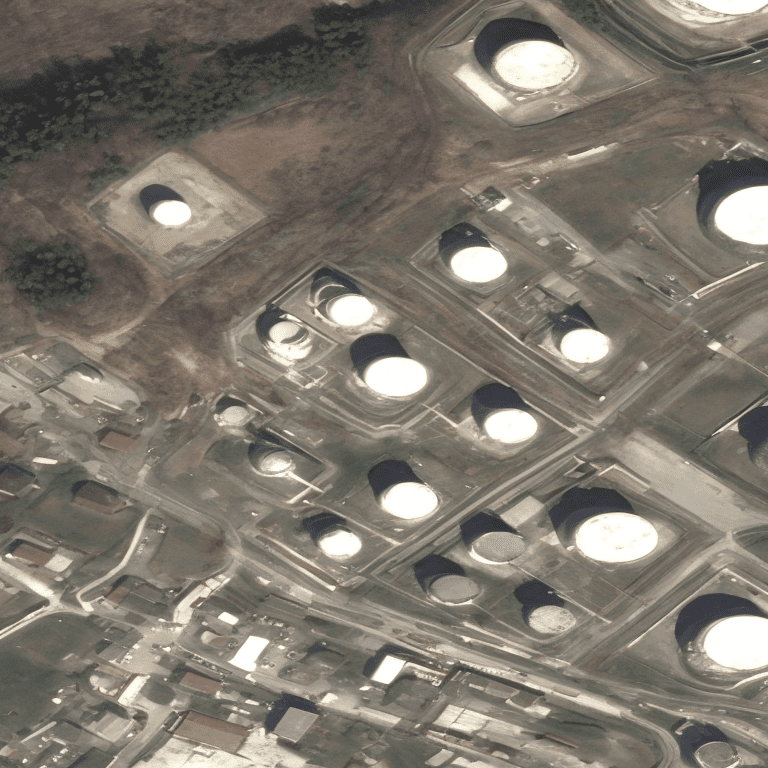} \\
   
   \rotatebox[origin=l]{90}{\parbox{14mm}{\centering \textbf{gsd: \textit{max}}}} 
   & \includegraphics[width=0.11\textwidth]{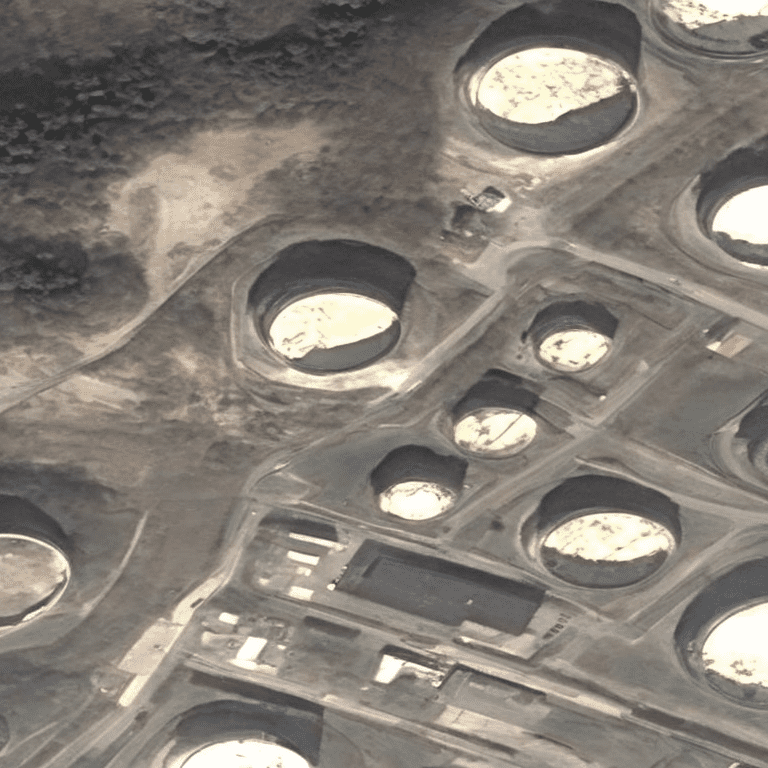} 
   & \includegraphics[width=0.11\textwidth]{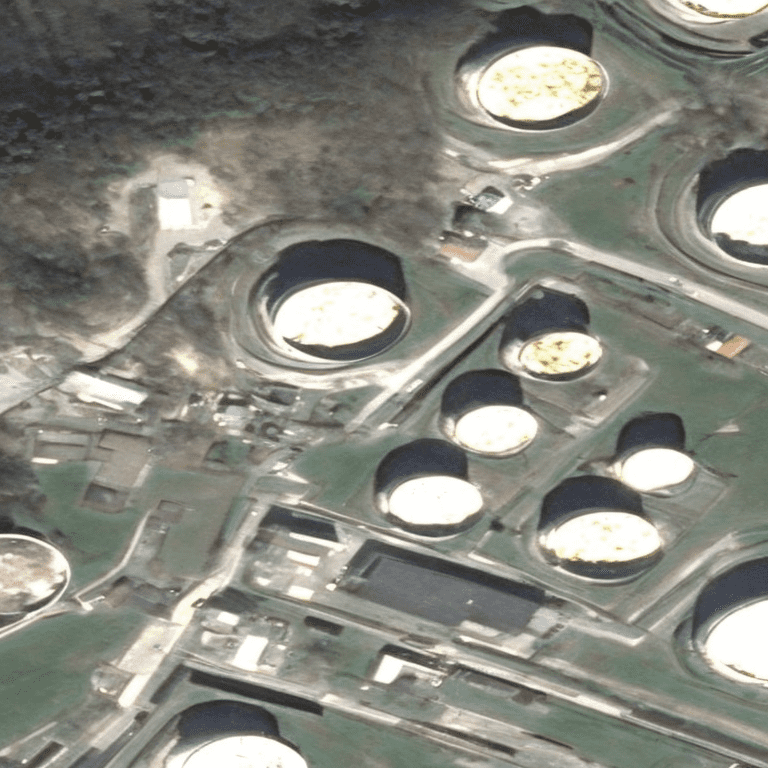}
   & \includegraphics[width=0.11\textwidth]{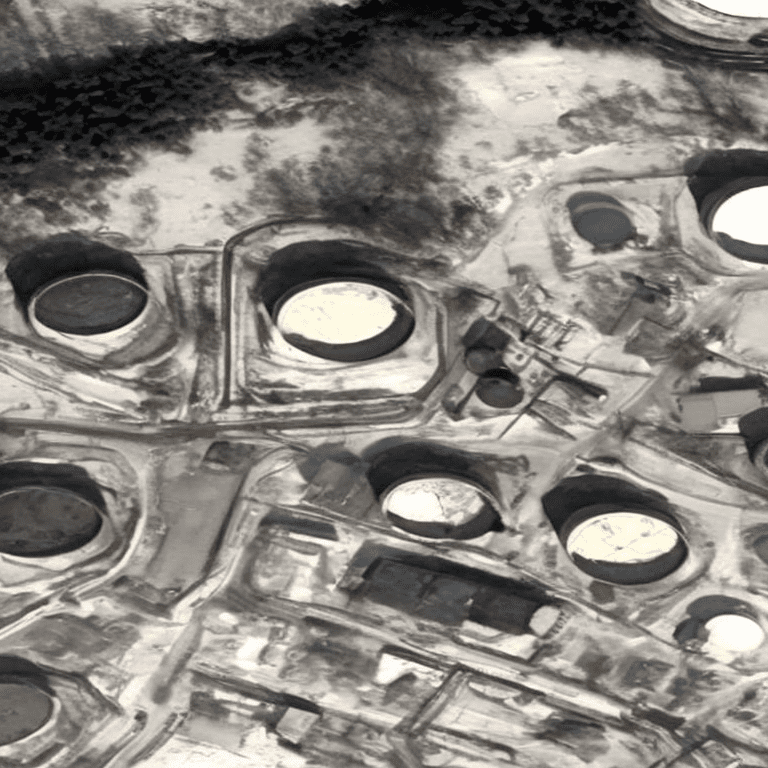}
   & \includegraphics[width=0.11\textwidth]{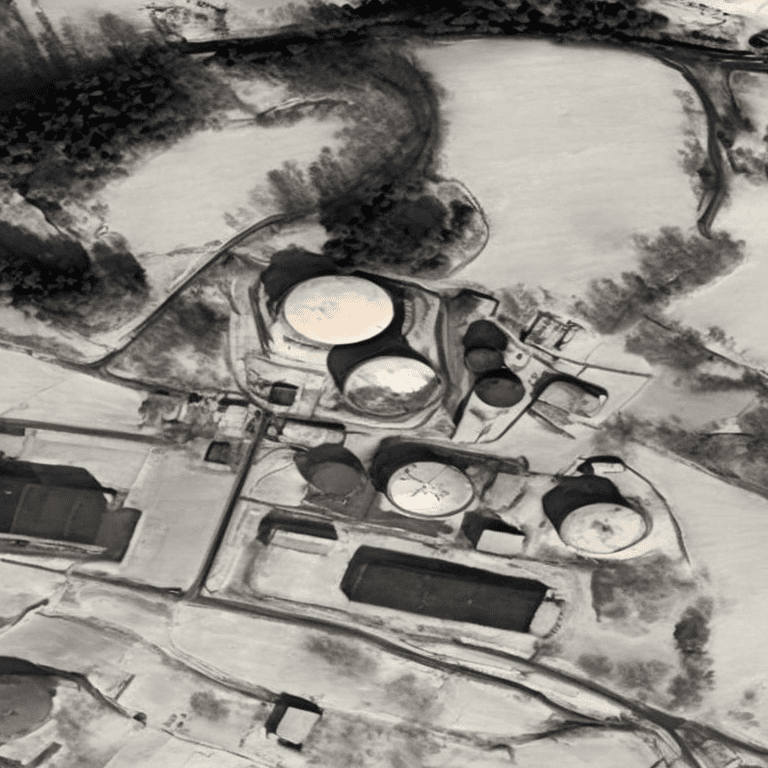}
    \end{tabular}
    \caption{Single-image generations for the "park" and "storage tank" classes sampled from our custom, metadata-augmented version of DiffusionSAT \citep{diffusionsat}, with varying gsd and metadata presence (rest of the metadata values remain fixed). As can be seen from the results, the addition-based metadata handling approach adopted by DiffusionSAT encounters problems relevant to both class fidelity (any evidence of a park-like structure disappears in the second row) and accuracy (cloud cover and epoch seem to change in the "park" and "storage tank" examples, respectively, despite no such change in the corresponding metadata) that become more and more pronounced as the amount of missing metadata increases.}
    \label{tab:ref_label_addition_missing}
\end{figure}

\begin{table}[h]
\centering
\begin{tabular}{ccccc}
\hline
\multicolumn{1}{l}{} & \multicolumn{1}{l}{\textit{Metric}} & \textbf{FID} $\downarrow$ & \textbf{IS} $\uparrow$  & \textbf{CLIP} $\uparrow$ \\ \hline
\textit{Model}      & \textit{}                           & \textbf{}             & \textbf{}     & \textbf{}      \\ \hline
\textbf{DiffusionSAT}       &                                     & 67.36                 & 1.33          & 26.42          \\ \hline
\textbf{Ours}        &                                     & \textbf{56.70}        & \textbf{1.39} & \textbf{28.11} \\ \hline
\end{tabular}
\caption{Sample quality results on single-image 512x512 generation on the fMoW validation data (quantitative).}
\label{tab:table_single_image_quantitative}
\end{table}

\subsection{Satellite image generation with Temporal Contextual Variable Control}
\label{temporal_image_experiment_subsection}
\begin{figure*}[h!]
    \centering
    \begin{tabular}{ll|l|l|ll}
    \multicolumn{3}{c|}{\textit{t' \textless t}}   & \multicolumn{3}{c}{\textit{t' \textgreater t}}\\
    \textbf{DiffusionSAT} & \textbf{Ours} & \textbf{G.T.} & \textbf{G.T} & \textbf{Ours} & \textbf{DiffusionSAT}\\
       \includegraphics[width=0.13\textwidth, clip=true, trim={9cm 0 0 0}]{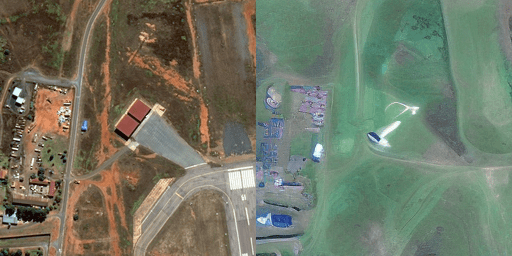} & 
       \includegraphics[width=0.13\textwidth, clip=true, trim={9cm 0 0 0}]{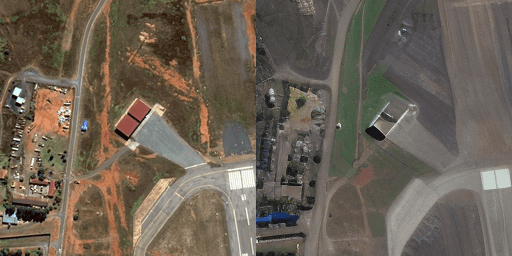} & 
       \includegraphics[width=0.13\textwidth, clip=true, trim={0 0 9cm 0}]{temporal_qualitative/ours/img2img_airport_hangar_1_1,2,3,0_gs=1.0_iters=20_backward_0.png} & 
       \includegraphics[width=0.13\textwidth, clip=true, trim={0 0 9cm 0}]{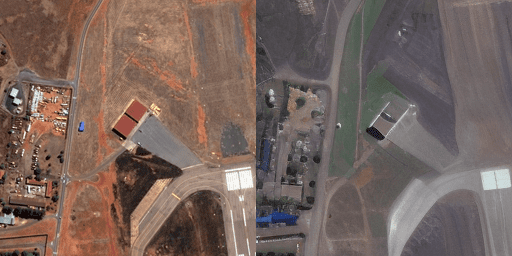} & 
       \includegraphics[width=0.13\textwidth, clip=true, trim={9cm 0 0 0}]{temporal_qualitative/ours/img2img_airport_hangar_1_1,2,3,5_gs=1.0_iters=20_forward_0.png} &
       \includegraphics[width=0.13\textwidth, clip=true, trim={9cm 0 0 0}]{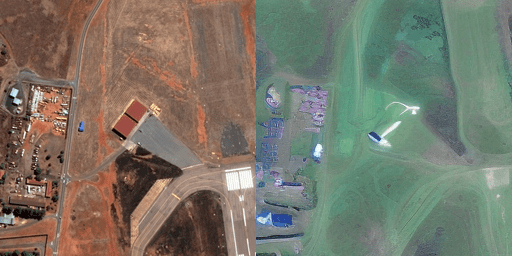}\\
       \includegraphics[width=0.13\textwidth, clip=true, trim={9cm 0 0 0}]{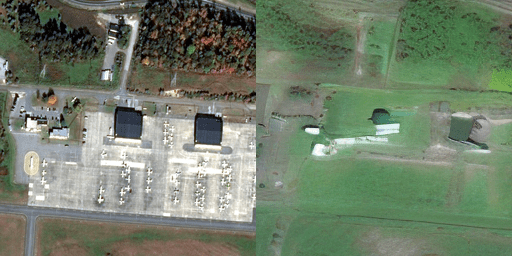} & 
       \includegraphics[width=0.13\textwidth, clip=true, trim={9cm 0 0 0}]{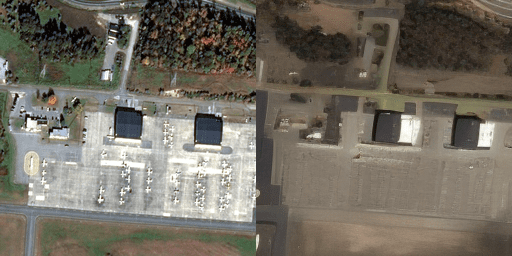} & 
       \includegraphics[width=0.13\textwidth, clip=true, trim={0 0 9cm 0}]{temporal_qualitative/ours/img2img_airport_hangar_4_1,3,7,0_gs=1.0_iters=20_backward_0.png} & 
       \includegraphics[width=0.13\textwidth, clip=true, trim={0 0 9cm 0}]{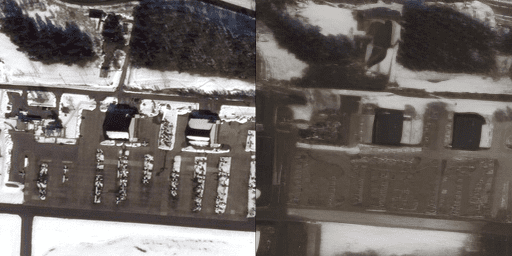} & 
       \includegraphics[width=0.13\textwidth, clip=true, trim={9cm 0 0 0}]{temporal_qualitative/ours/img2img_airport_hangar_4_1,3,7,8_gs=1.0_iters=20_forward_0.png} & 
       \includegraphics[width=0.13\textwidth, clip=true, trim={9cm 0 0 0}]{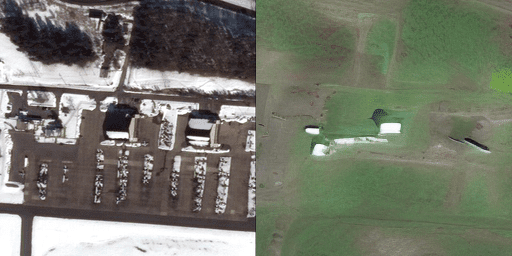}\\
       \includegraphics[width=0.13\textwidth, clip=true, trim={9cm 0 0 0}]{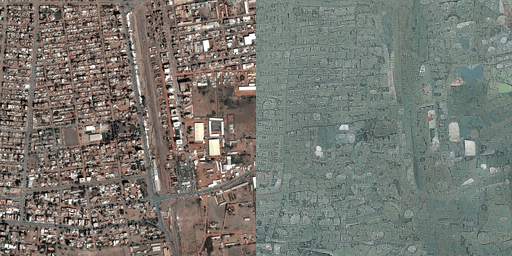} & 
       \includegraphics[width=0.13\textwidth, clip=true, trim={9cm 0 0 0}]{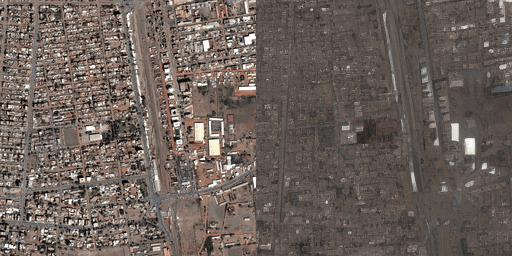} & 
       \includegraphics[width=0.13\textwidth, clip=true, trim={0 0 9cm 0}]{temporal_qualitative/ours/img2img_amusement_park_0_3,4,5,0_gs=1.0_iters=20_backward_0.png} & 
       \includegraphics[width=0.13\textwidth, clip=true, trim={0 0 9cm 0}]{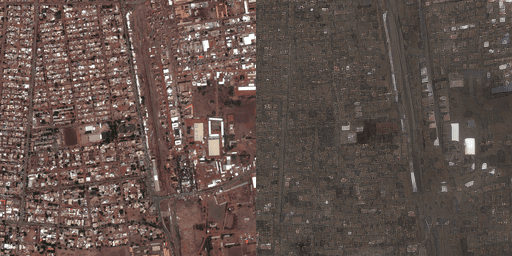} & 
       \includegraphics[width=0.13\textwidth, clip=true, trim={9cm 0 0 0}]{temporal_qualitative/ours/img2img_amusement_park_0_3,4,5,9_gs=1.0_iters=20_forward_0.png} & 
       \includegraphics[width=0.13\textwidth, clip=true, trim={9cm 0 0 0}]{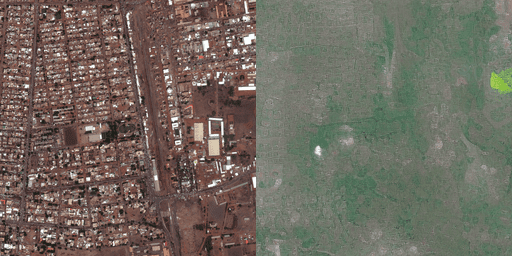}\\
       \includegraphics[width=0.13\textwidth, clip=true, trim={9cm 0 0 0}]{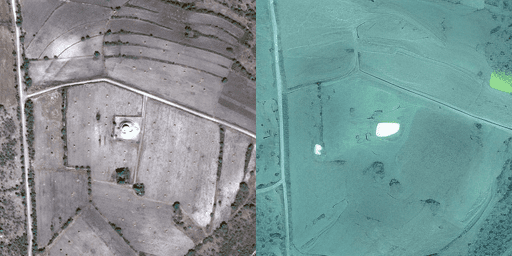} & 
       \includegraphics[width=0.13\textwidth, clip=true, trim={9cm 0 0 0}]{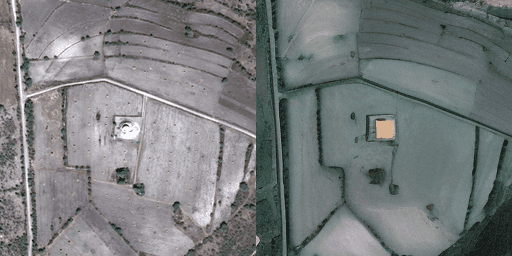} & 
       \includegraphics[width=0.13\textwidth, clip=true, trim={0 0 9cm 0}]{temporal_qualitative/ours/img2img_archaeological_site_2_1,2,3,0_gs=1.0_iters=20_backward_0.png} & 
       \includegraphics[width=0.13\textwidth, clip=true, trim={0 0 9cm 0}]{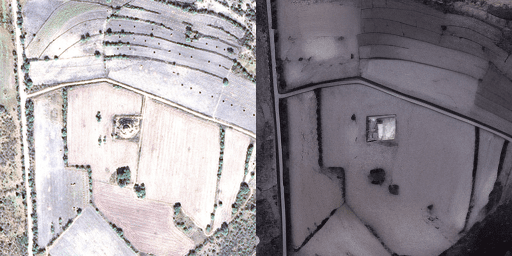} & 
       \includegraphics[width=0.13\textwidth, clip=true, trim={9cm 0 0 0}]{temporal_qualitative/ours/img2img_archaeological_site_2_0,1,2,3_gs=1.0_iters=20_forward_0.png} & 
       \includegraphics[width=0.13\textwidth, clip=true, trim={9cm 0 0 0}]{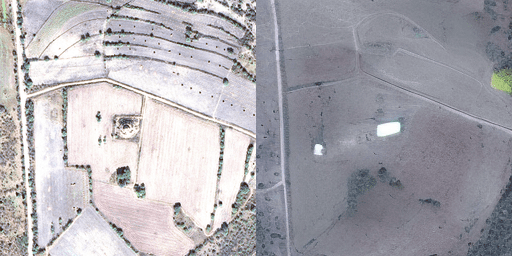}\\
       \includegraphics[width=0.13\textwidth, clip=true, trim={9cm 0 0 0}]{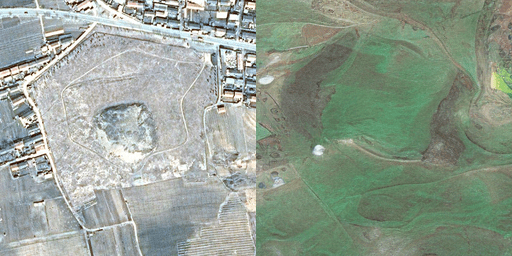} & 
       \includegraphics[width=0.13\textwidth, clip=true, trim={9cm 0 0 0}]{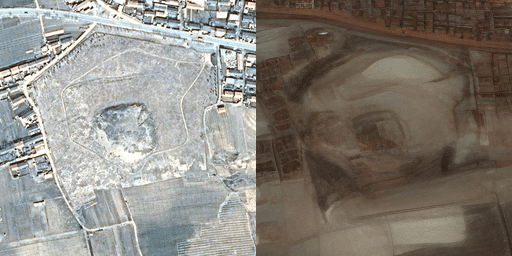} & 
       \includegraphics[width=0.13\textwidth, clip=true, trim={0 0 9cm 0}]{temporal_qualitative/ours/img2img_archaeological_site_3_2,3,4,0_gs=1.0_iters=20_backward_0.png} & 
       \includegraphics[width=0.13\textwidth, clip=true, trim={0 0 9cm 0}]{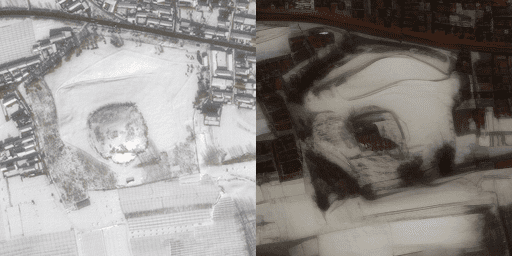} & 
       \includegraphics[width=0.13\textwidth, clip=true, trim={9cm 0 0 0}]{temporal_qualitative/ours/img2img_archaeological_site_3_2,3,4,5_gs=1.0_iters=20_forward_0.png} & 
       \includegraphics[width=0.13\textwidth, clip=true, trim={9cm 0 0 0}]{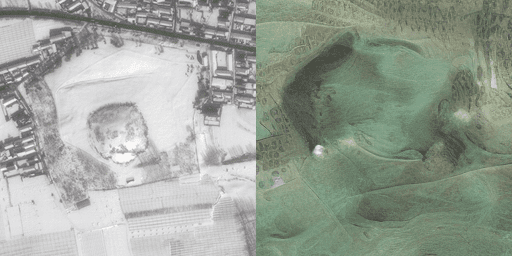}\\
       \includegraphics[width=0.13\textwidth, clip=true, trim={9cm 0 0 0}]{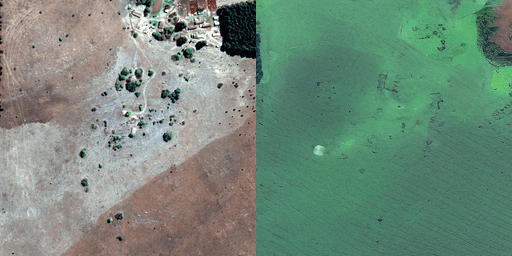} & 
       \includegraphics[width=0.13\textwidth, clip=true, trim={9cm 0 0 0}]{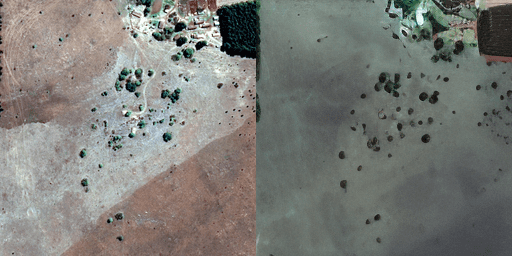} & 
       \includegraphics[width=0.13\textwidth, clip=true, trim={0 0 9cm 0}]{temporal_qualitative/ours/img2img_archaeological_site_4_4,7,9,0_gs=1.0_iters=20_backward_0.png} & 
       \includegraphics[width=0.13\textwidth, clip=true, trim={0 0 9cm 0}]{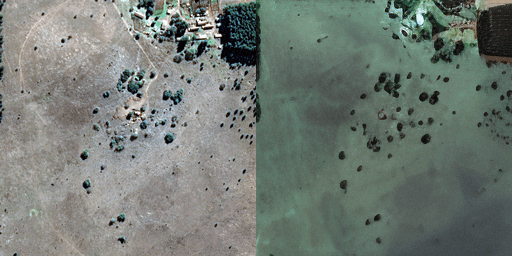} & 
       \includegraphics[width=0.13\textwidth, clip=true, trim={9cm 0 0 0}]{temporal_qualitative/ours/img2img_archaeological_site_4_4,7,9,10_gs=1.0_iters=20_forward_0.png} & 
       \includegraphics[width=0.13\textwidth, clip=true, trim={9cm 0 0 0}]{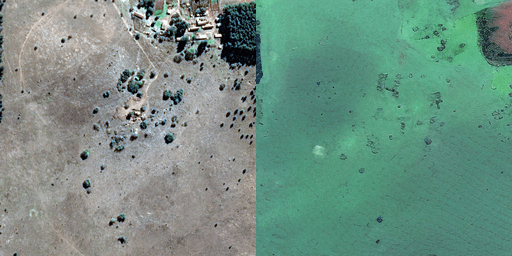}\\
       \includegraphics[width=0.13\textwidth, clip=true, trim={9cm 0 0 0}]{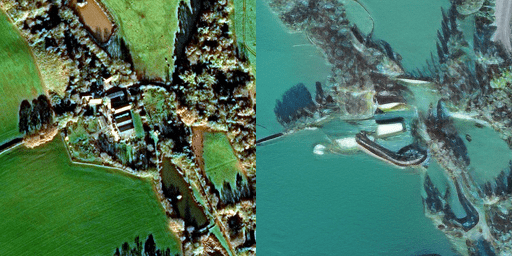} & 
       \includegraphics[width=0.13\textwidth, clip=true, trim={9cm 0 0 0}]{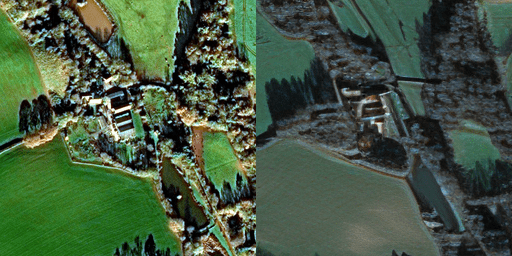} & 
       \includegraphics[width=0.13\textwidth, clip=true, trim={0 0 9cm 0}]{temporal_qualitative/ours/img2img_barn_2_2,3,4,1_gs=1.0_iters=20_backward_0.png} & 
       \includegraphics[width=0.13\textwidth, clip=true, trim={0 0 9cm 0}]{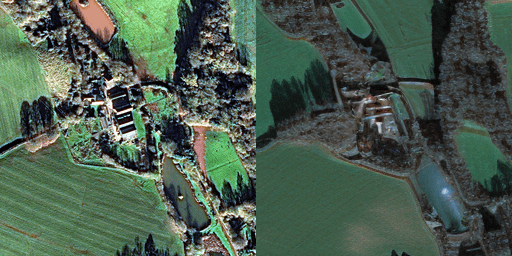} & 
       \includegraphics[width=0.13\textwidth, clip=true, trim={9cm 0 0 0}]{temporal_qualitative/ours/img2img_barn_2_2,3,4,5_gs=1.0_iters=20_forward_0.png} & 
       \includegraphics[width=0.13\textwidth, clip=true, trim={9cm 0 0 0}]{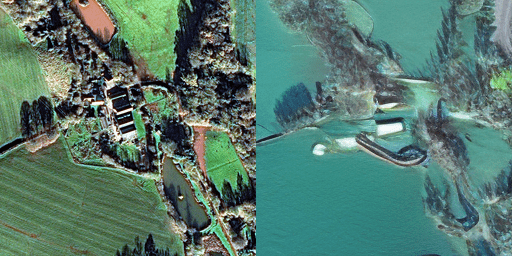}\\
       \includegraphics[width=0.13\textwidth, clip=true, trim={9cm 0 0 0}]{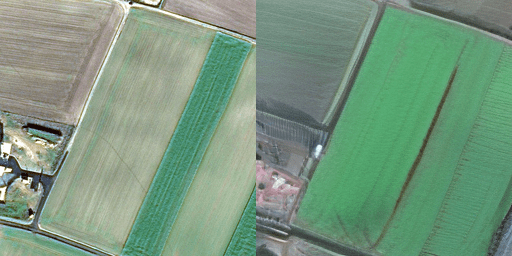} & 
       \includegraphics[width=0.13\textwidth, clip=true, trim={9cm 0 0 0}]{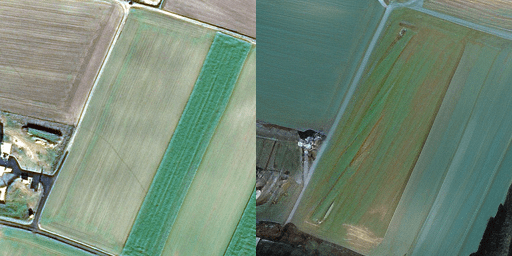} & 
       \includegraphics[width=0.13\textwidth, clip=true, trim={0 0 9cm 0}]{temporal_qualitative/ours/img2img_crop_field_13_1,4,6,0_gs=1.0_iters=20_backward_0.png} & 
       \includegraphics[width=0.13\textwidth, clip=true, trim={0 0 9cm 0}]{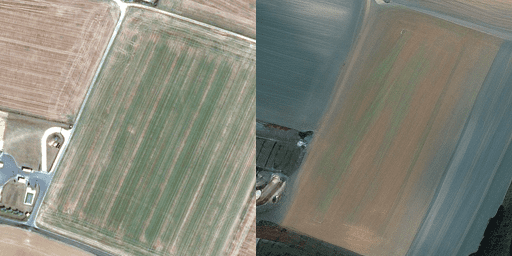} & 
       \includegraphics[width=0.13\textwidth, clip=true, trim={9cm 0 0 0}]{temporal_qualitative/ours/img2img_crop_field_13_1,4,6,7_gs=1.0_iters=20_forward_0.png} & 
       \includegraphics[width=0.13\textwidth, clip=true, trim={9cm 0 0 0}]{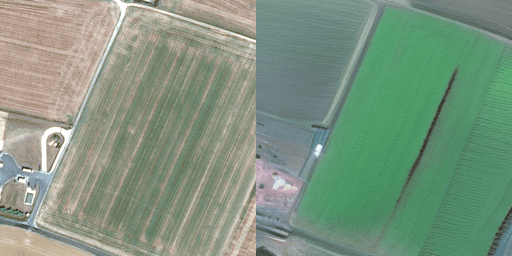}\\
       \includegraphics[width=0.13\textwidth, clip=true, trim={9cm 0 0 0}]{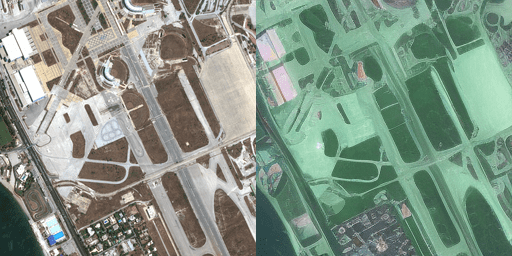} & 
       \includegraphics[width=0.13\textwidth, clip=true, trim={9cm 0 0 0}]{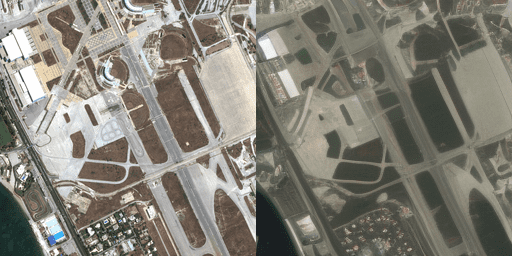} & 
       \includegraphics[width=0.13\textwidth, clip=true, trim={0 0 9cm 0}]{temporal_qualitative/ours/img2img_runway_3_1,4,6,0_gs=1.0_iters=20_backward_0.png} & 
       \includegraphics[width=0.13\textwidth, clip=true, trim={0 0 9cm 0}]{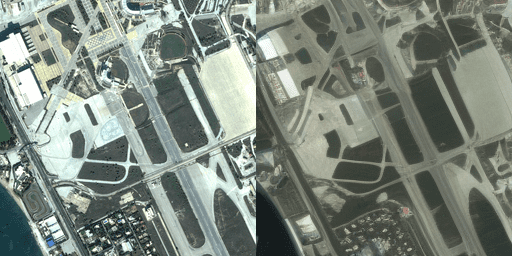} & 
       \includegraphics[width=0.13\textwidth, clip=true, trim={9cm 0 0 0}]{temporal_qualitative/ours/img2img_runway_3_1,4,6,9_gs=1.0_iters=20_forward_0.png} & 
       \includegraphics[width=0.13\textwidth, clip=true, trim={9cm 0 0 0}]{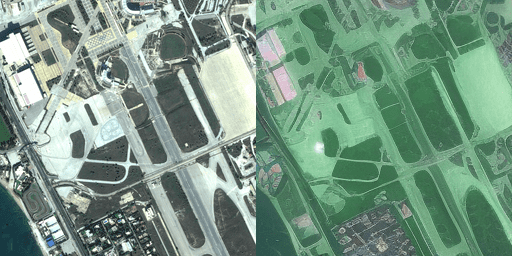}
\end{tabular}
\caption{}
\label{tab:temporal_1}
\end{figure*}

In this step, we employ our single-image model as an effective prior for the task of temporal generation/prediction and demonstrate results using our 3D conditioning approach on our custom, temporal version of the fMoW dataset. Given a sequence $s$ of conditioning images along with their corresponding captions and/or metadata, we want our model to reliably predict an image at any desired target time when provided with relative control inputs (i.e. a target caption and/or target metadata). Here we consider tasks where the target image predates or postdates the first image in $s$. Quantitative and qualitative results can be found in Table \ref{tab:3d_controlnet_quantitative} and Figure \ref{tab:temporal_1}, respectively (more results regarding the latter can be viewed in Tables \ref{tab:appendix_temporal}-\ref{tab:appendix_temporal_b}). As demonstrated in both cases, our environment-aware model is able to ourperform the model proposed in DiffusionSAT across all three reported metrics in both prediction settings (i.e. backwards or forwards in time) while also producing visual results of greater accuracy and fidelity.

\begin{table}[h]
\centering
\setlength{\tabcolsep}{2pt}
\renewcommand{\arraystretch}{0.8}
\begin{tabular}{@{}c ccc ccc@{}}
\toprule
\multirow{2}{*}{\textit{Model}} & \multicolumn{3}{c}{\textit{t' \textgreater t}} & \multicolumn{3}{c}{\textit{t' \textless t}} \\
\cmidrule(lr){2-4}\cmidrule(lr){5-7}
& \textbf{SSIM} $\uparrow$ & \textbf{PSNR} $\uparrow$ & \textbf{LPIPS} $\downarrow$ 
& \textbf{SSIM} $\uparrow$ & \textbf{PSNR} $\uparrow$ & \textbf{LPIPS} $\downarrow$ \\
\midrule
\textbf{DiffusionSAT} & 0.1704 & 10.897 & 0.7242 & 0.1573 & 10.7413 & 0.7329 \\
\textbf{Ours}         & \textbf{0.1777} & \textbf{11.094} & \textbf{0.7201} 
                      & \textbf{0.1761} & \textbf{11.0936} & \textbf{0.7199} \\
\bottomrule
\end{tabular}
\caption{Sample quality results on temporal generation on the fMoW validation data. t' $>$ t: generating an image in the past given three random future reference images; t' $<$ t: generating a future image given three random past reference images.}
\label{tab:3d_controlnet_quantitative}
\end{table}

\section{Conclusion}
In this work, we presented a novel, environment-aware diffusion framework for satellite image generation, advancing the state of the art in multimodal conditioning for remote sensing applications. Our approach introduces several key technical innovations over existing methods, most notably a concatenation-based metadata fusion strategy that preserves attribute interactions while remaining robust to partially missing metadata, thus overcoming the limitations of previous additive fusion approaches. Furthermore, our model is the first, to our knowledge, to integrate dynamic environmental metadata --namely wind, precipitation, temperature, and solar radiation-- aggregated over multi-day windows, thereby moving beyond static metadata to enable contextually grounded image generation. This enriched conditioning mechanism translates into enhanced robustness: our model can handle any combination of missing or partial control signals while maintaining high generation quality. Empirical results confirm significant performance gains, including improved qualitative metrics (e.g., FID improved from 67.36 to 56.70, and CLIP-score from 26.42 to 28.11), higher responsiveness to user-defined inputs, and superior temporal generation fidelity. Our model also demonstrates enhanced realism in the presence of complex environmental phenomena such as cloud cover, seasonal transitions, and geographically distinct features. These contributions, supported by the release of the first publicly available tri-modal dataset for satellite image generation, position our method as a promising foundation for future work in both scientific and operational downstream tasks in remote sensing.
\clearpage
\bibliographystyle{abbrvnat}
\bibliography{paper}

\begin{thebibliography}{92}
\providecommand{\natexlab}[1]{#1}
\providecommand{\url}[1]{\texttt{#1}}
\expandafter\ifx\csname urlstyle\endcsname\relax
  \providecommand{\doi}[1]{doi: #1}\else
  \providecommand{\doi}{doi: \begingroup \urlstyle{rm}\Url}\fi

\bibitem[Bastani et~al.(2023{\natexlab{a}})Bastani, Wolters, Gupta, Ferdinando, and Kembhavi]{satlas}
F.~Bastani, P.~Wolters, R.~Gupta, J.~Ferdinando, and A.~Kembhavi.
\newblock Satlaspretrain: A large-scale dataset for remote sensing image understanding, 2023{\natexlab{a}}.
\newblock URL \url{https://arxiv.org/abs/2211.15660}.

\bibitem[Bastani et~al.(2023{\natexlab{b}})Bastani, Wolters, Gupta, Ferdinando, and Kembhavi]{sisr4}
F.~Bastani, P.~Wolters, R.~Gupta, J.~Ferdinando, and A.~Kembhavi.
\newblock Satlaspretrain: A large-scale dataset for remote sensing image understanding, 2023{\natexlab{b}}.
\newblock URL \url{https://arxiv.org/abs/2211.15660}.

\bibitem[Blattmann et~al.(2023)Blattmann, Rombach, Ling, Dockhorn, Kim, Fidler, and Kreis]{videoldm}
A.~Blattmann, R.~Rombach, H.~Ling, T.~Dockhorn, S.~W. Kim, S.~Fidler, and K.~Kreis.
\newblock Align your latents: High-resolution video synthesis with latent diffusion models, 2023.
\newblock URL \url{https://arxiv.org/abs/2304.08818}.

\bibitem[Brooks et~al.(2023)Brooks, Holynski, and Efros]{controllable1}
T.~Brooks, A.~Holynski, and A.~A. Efros.
\newblock Instructpix2pix: Learning to follow image editing instructions, 2023.
\newblock URL \url{https://arxiv.org/abs/2211.09800}.

\bibitem[Brown et~al.(2020)Brown, Mann, Ryder, Subbiah, Kaplan, Dhariwal, Neelakantan, Shyam, Sastry, Askell, Agarwal, Herbert-Voss, Krueger, Henighan, Child, Ramesh, Ziegler, Wu, Winter, Hesse, Chen, Sigler, Litwin, Gray, Chess, Clark, Berner, McCandlish, Radford, Sutskever, and Amodei]{gpt}
T.~B. Brown, B.~Mann, N.~Ryder, M.~Subbiah, J.~Kaplan, P.~Dhariwal, A.~Neelakantan, P.~Shyam, G.~Sastry, A.~Askell, S.~Agarwal, A.~Herbert-Voss, G.~Krueger, T.~Henighan, R.~Child, A.~Ramesh, D.~M. Ziegler, J.~Wu, C.~Winter, C.~Hesse, M.~Chen, E.~Sigler, M.~Litwin, S.~Gray, B.~Chess, J.~Clark, C.~Berner, S.~McCandlish, A.~Radford, I.~Sutskever, and D.~Amodei.
\newblock Language models are few-shot learners, 2020.
\newblock URL \url{https://arxiv.org/abs/2005.14165}.

\bibitem[Cepeda et~al.(2023)Cepeda, Nayak, and Shah]{geoclip}
V.~V. Cepeda, G.~K. Nayak, and M.~Shah.
\newblock Geoclip: Clip-inspired alignment between locations and images for effective worldwide geo-localization, 2023.
\newblock URL \url{https://arxiv.org/abs/2309.16020}.

\bibitem[Chan et~al.(2023)Chan, Nagano, Chan, Bergman, Park, Levy, Aittala, Mello, Karras, and Wetzstein]{graphics3}
E.~R. Chan, K.~Nagano, M.~A. Chan, A.~W. Bergman, J.~J. Park, A.~Levy, M.~Aittala, S.~D. Mello, T.~Karras, and G.~Wetzstein.
\newblock Generative novel view synthesis with 3d-aware diffusion models, 2023.
\newblock URL \url{https://arxiv.org/abs/2304.02602}.

\bibitem[Chen et~al.(2020)Chen, Kornblith, Norouzi, and Hinton]{chen2020simpleframeworkcontrastivelearning}
T.~Chen, S.~Kornblith, M.~Norouzi, and G.~Hinton.
\newblock A simple framework for contrastive learning of visual representations, 2020.
\newblock URL \url{https://arxiv.org/abs/2002.05709}.

\bibitem[Cheng et~al.(2017)Cheng, Han, and Lu]{resisc}
G.~Cheng, J.~Han, and X.~Lu.
\newblock Remote sensing image scene classification: Benchmark and state of the art.
\newblock \emph{Proceedings of the IEEE}, 105\penalty0 (10):\penalty0 1865--1883, 2017.

\bibitem[Christie et~al.(2018)Christie, Fendley, Wilson, and Mukherjee]{fmow}
G.~Christie, N.~Fendley, J.~Wilson, and R.~Mukherjee.
\newblock Functional map of the world, 2018.
\newblock URL \url{https://arxiv.org/abs/1711.07846}.

\bibitem[Cong et~al.(2023)Cong, Khanna, Meng, Liu, Rozi, He, Burke, Lobell, and Ermon]{satmae}
Y.~Cong, S.~Khanna, C.~Meng, P.~Liu, E.~Rozi, Y.~He, M.~Burke, D.~B. Lobell, and S.~Ermon.
\newblock Satmae: Pre-training transformers for temporal and multi-spectral satellite imagery, 2023.
\newblock URL \url{https://arxiv.org/abs/2207.08051}.

\bibitem[Cornebise et~al.(2022)Cornebise, Oršolić, and Kalaitzis]{sisr5}
J.~Cornebise, I.~Oršolić, and F.~Kalaitzis.
\newblock Open high-resolution satellite imagery: The worldstrat dataset -- with application to super-resolution, 2022.
\newblock URL \url{https://arxiv.org/abs/2207.06418}.

\bibitem[Currier and Sala(2022)]{Currier2022}
C.~M. Currier and O.~E. Sala.
\newblock Precipitation versus temperature as phenology controls in drylands.
\newblock \emph{Ecology}, 103\penalty0 (11), July 2022.
\newblock ISSN 1939-9170.
\newblock \doi{10.1002/ecy.3793}.
\newblock URL \url{http://dx.doi.org/10.1002/ecy.3793}.

\bibitem[Dado et~al.(2020)Dado, Deines, Patel, Liang, and Lobell]{crop2}
W.~Dado, J.~Deines, R.~Patel, S.-Z. Liang, and D.~Lobell.
\newblock High-resolution soybean yield mapping across the us midwest using subfield harvester data.
\newblock \emph{Remote Sensing}, 12:\penalty0 3471, 10 2020.
\newblock \doi{10.3390/rs12213471}.

\bibitem[Devlin et~al.(2019)Devlin, Chang, Lee, and Toutanova]{bert}
J.~Devlin, M.-W. Chang, K.~Lee, and K.~Toutanova.
\newblock {BERT}: Pre-training of deep bidirectional transformers for language understanding.
\newblock In J.~Burstein, C.~Doran, and T.~Solorio, editors, \emph{Proceedings of the 2019 Conference of the North {A}merican Chapter of the Association for Computational Linguistics: Human Language Technologies, Volume 1 (Long and Short Papers)}, pages 4171--4186, Minneapolis, Minnesota, June 2019. Association for Computational Linguistics.
\newblock \doi{10.18653/v1/N19-1423}.
\newblock URL \url{https://aclanthology.org/N19-1423/}.

\bibitem[DILEEP et~al.(2020)DILEEP, Zimmerle, Beveridge, and Vaughn]{oil}
S.~DILEEP, D.~Zimmerle, R.~Beveridge, and T.~Vaughn.
\newblock Automated identification of oil field features using cnns.
\newblock In \emph{NeurIPS 2020 Workshop on Tackling Climate Change with Machine Learning}, 2020.
\newblock URL \url{https://www.climatechange.ai/papers/neurips2020/32}.

\bibitem[Esser et~al.(2021)Esser, Rombach, and Ommer]{esser2021tamingtransformershighresolutionimage}
P.~Esser, R.~Rombach, and B.~Ommer.
\newblock Taming transformers for high-resolution image synthesis, 2021.
\newblock URL \url{https://arxiv.org/abs/2012.09841}.

\bibitem[Etten et~al.(2019)Etten, Lindenbaum, and Bacastow]{spacenet1}
A.~V. Etten, D.~Lindenbaum, and T.~M. Bacastow.
\newblock Spacenet: A remote sensing dataset and challenge series, 2019.
\newblock URL \url{https://arxiv.org/abs/1807.01232}.

\bibitem[Feng et~al.(2022)Feng, Zhang, Yu, Fang, Li, Chen, Lu, Liu, Yin, Feng, Sun, Tian, Wu, and Wang]{unknown}
Z.~Feng, Z.~Zhang, X.~Yu, Y.~Fang, L.~Li, X.~Chen, Y.~Lu, J.~Liu, W.~Yin, S.~Feng, Y.~Sun, H.~Tian, H.~Wu, and H.~Wang.
\newblock Ernie-vilg 2.0: Improving text-to-image diffusion model with knowledge-enhanced mixture-of-denoising-experts, 10 2022.

\bibitem[Gong et~al.(2021)Gong, Liao, Zhang, Zhang, Chen, Zhu, Tan, and Lv]{sisr2}
Y.~Gong, P.~Liao, X.~Zhang, L.~Zhang, G.~Chen, K.~Zhu, X.~Tan, and Z.~Lv.
\newblock Enlighten-gan for super resolution reconstruction in mid-resolution remote sensing images.
\newblock \emph{Remote Sensing}, 13\penalty0 (6), 2021.
\newblock ISSN 2072-4292.
\newblock \doi{10.3390/rs13061104}.
\newblock URL \url{https://www.mdpi.com/2072-4292/13/6/1104}.

\bibitem[Gupta et~al.(2019)Gupta, Hosfelt, Sajeev, Patel, Goodman, Doshi, Heim, Choset, and Gaston]{xview2}
R.~Gupta, R.~Hosfelt, S.~Sajeev, N.~Patel, B.~Goodman, J.~Doshi, E.~Heim, H.~Choset, and M.~Gaston.
\newblock xbd: A dataset for assessing building damage from satellite imagery, 2019.
\newblock URL \url{https://arxiv.org/abs/1911.09296}.

\bibitem[Hassler and Lauer(2021)]{Hassler2021}
B.~Hassler and A.~Lauer.
\newblock Comparison of reanalysis and observational precipitation datasets including era5 and wfde5.
\newblock \emph{Atmosphere}, 12\penalty0 (11):\penalty0 1462, Nov. 2021.
\newblock ISSN 2073-4433.
\newblock \doi{10.3390/atmos12111462}.
\newblock URL \url{http://dx.doi.org/10.3390/atmos12111462}.

\bibitem[Hendrycks et~al.(2019)Hendrycks, Mazeika, Kadavath, and Song]{hendrycks2019usingselfsupervisedlearningimprove}
D.~Hendrycks, M.~Mazeika, S.~Kadavath, and D.~Song.
\newblock Using self-supervised learning can improve model robustness and uncertainty, 2019.
\newblock URL \url{https://arxiv.org/abs/1906.12340}.

\bibitem[Hersbach et~al.(2020)Hersbach, Bell, Berrisford, Hirahara, Horányi, Muñoz-Sabater, Nicolas, Peubey, Radu, Schepers, Simmons, Soci, Abdalla, Abellan, Balsamo, Bechtold, Biavati, Bidlot, Bonavita, De~Chiara, Dahlgren, Dee, Diamantakis, Dragani, Flemming, Forbes, Fuentes, Geer, Haimberger, Healy, Hogan, Hólm, Janisková, Keeley, Laloyaux, Lopez, Lupu, Radnoti, de~Rosnay, Rozum, Vamborg, Villaume, and Thépaut]{era5}
H.~Hersbach, B.~Bell, P.~Berrisford, S.~Hirahara, A.~Horányi, J.~Muñoz-Sabater, J.~Nicolas, C.~Peubey, R.~Radu, D.~Schepers, A.~Simmons, C.~Soci, S.~Abdalla, X.~Abellan, G.~Balsamo, P.~Bechtold, G.~Biavati, J.~Bidlot, M.~Bonavita, G.~De~Chiara, P.~Dahlgren, D.~Dee, M.~Diamantakis, R.~Dragani, J.~Flemming, R.~Forbes, M.~Fuentes, A.~Geer, L.~Haimberger, S.~Healy, R.~J. Hogan, E.~Hólm, M.~Janisková, S.~Keeley, P.~Laloyaux, P.~Lopez, C.~Lupu, G.~Radnoti, P.~de~Rosnay, I.~Rozum, F.~Vamborg, S.~Villaume, and J.-N. Thépaut.
\newblock The era5 global reanalysis.
\newblock \emph{Quarterly Journal of the Royal Meteorological Society}, 146\penalty0 (730):\penalty0 1999--2049, 2020.
\newblock \doi{https://doi.org/10.1002/qj.3803}.
\newblock URL \url{https://rmets.onlinelibrary.wiley.com/doi/abs/10.1002/qj.3803}.

\bibitem[Heusel et~al.(2018)Heusel, Ramsauer, Unterthiner, Nessler, and Hochreiter]{fid}
M.~Heusel, H.~Ramsauer, T.~Unterthiner, B.~Nessler, and S.~Hochreiter.
\newblock Gans trained by a two time-scale update rule converge to a local nash equilibrium, 2018.
\newblock URL \url{https://arxiv.org/abs/1706.08500}.

\bibitem[Ho et~al.(2020)Ho, Jain, and Abbeel]{diffusion1}
J.~Ho, A.~Jain, and P.~Abbeel.
\newblock Denoising diffusion probabilistic models, 2020.
\newblock URL \url{https://arxiv.org/abs/2006.11239}.

\bibitem[Ho et~al.(2022)Ho, Chan, Saharia, Whang, Gao, Gritsenko, Kingma, Poole, Norouzi, Fleet, and Salimans]{ldm3}
J.~Ho, W.~Chan, C.~Saharia, J.~Whang, R.~Gao, A.~Gritsenko, D.~P. Kingma, B.~Poole, M.~Norouzi, D.~J. Fleet, and T.~Salimans.
\newblock Imagen video: High definition video generation with diffusion models, 2022.
\newblock URL \url{https://arxiv.org/abs/2210.02303}.

\bibitem[Isola et~al.(2018)Isola, Zhu, Zhou, and Efros]{pix2pix}
P.~Isola, J.-Y. Zhu, T.~Zhou, and A.~A. Efros.
\newblock Image-to-image translation with conditional adversarial networks, 2018.
\newblock URL \url{https://arxiv.org/abs/1611.07004}.

\bibitem[Kaplan et~al.(2020)Kaplan, McCandlish, Henighan, Brown, Chess, Child, Gray, Radford, Wu, and Amodei]{kaplan2020scalinglawsneurallanguage}
J.~Kaplan, S.~McCandlish, T.~Henighan, T.~B. Brown, B.~Chess, R.~Child, S.~Gray, A.~Radford, J.~Wu, and D.~Amodei.
\newblock Scaling laws for neural language models, 2020.
\newblock URL \url{https://arxiv.org/abs/2001.08361}.

\bibitem[Kerner et~al.(2020)Kerner, Tseng, Becker-Reshef, Nakalembe, Barker, Munshell, Paliyam, and Hosseini]{crop1}
H.~Kerner, G.~Tseng, I.~Becker-Reshef, C.~Nakalembe, B.~Barker, B.~Munshell, M.~Paliyam, and M.~Hosseini.
\newblock Rapid response crop maps in data sparse regions, 2020.
\newblock URL \url{https://arxiv.org/abs/2006.16866}.

\bibitem[Khanna et~al.(2024)Khanna, Liu, Zhou, Meng, Rombach, Burke, Lobell, and Ermon]{diffusionsat}
S.~Khanna, P.~Liu, L.~Zhou, C.~Meng, R.~Rombach, M.~Burke, D.~Lobell, and S.~Ermon.
\newblock Diffusionsat: A generative foundation model for satellite imagery, 2024.
\newblock URL \url{https://arxiv.org/abs/2312.03606}.

\bibitem[Kingma et~al.(2021)Kingma, Salimans, Poole, and Ho]{diffusion3}
D.~P. Kingma, T.~Salimans, B.~Poole, and J.~Ho.
\newblock Variational diffusion models, 2021.
\newblock URL \url{https://arxiv.org/abs/2107.00630}.

\bibitem[Klemmer et~al.(2024)Klemmer, Rolf, Robinson, Mackey, and Rußwurm]{satclip}
K.~Klemmer, E.~Rolf, C.~Robinson, L.~Mackey, and M.~Rußwurm.
\newblock Satclip: Global, general-purpose location embeddings with satellite imagery, 2024.
\newblock URL \url{https://arxiv.org/abs/2311.17179}.

\bibitem[Kong et~al.(2021)Kong, Ping, Huang, Zhao, and Catanzaro]{speech1}
Z.~Kong, W.~Ping, J.~Huang, K.~Zhao, and B.~Catanzaro.
\newblock Diffwave: A versatile diffusion model for audio synthesis, 2021.
\newblock URL \url{https://arxiv.org/abs/2009.09761}.

\bibitem[Kumari et~al.(2023)Kumari, Zhang, Zhang, Shechtman, and Zhu]{subject1}
N.~Kumari, B.~Zhang, R.~Zhang, E.~Shechtman, and J.-Y. Zhu.
\newblock Multi-concept customization of text-to-image diffusion, 2023.
\newblock URL \url{https://arxiv.org/abs/2212.04488}.

\bibitem[Ledig et~al.(2017)Ledig, Theis, Huszar, Caballero, Cunningham, Acosta, Aitken, Tejani, Totz, Wang, and Shi]{superresolution2}
C.~Ledig, L.~Theis, F.~Huszar, J.~Caballero, A.~Cunningham, A.~Acosta, A.~Aitken, A.~Tejani, J.~Totz, Z.~Wang, and W.~Shi.
\newblock Photo-realistic single image super-resolution using a generative adversarial network, 2017.
\newblock URL \url{https://arxiv.org/abs/1609.04802}.

\bibitem[Li et~al.(2021)Li, Yang, Chang, Feng, Xu, Li, and Chen]{li2021srdiffsingleimagesuperresolution}
H.~Li, Y.~Yang, M.~Chang, H.~Feng, Z.~Xu, Q.~Li, and Y.~Chen.
\newblock Srdiff: Single image super-resolution with diffusion probabilistic models, 2021.
\newblock URL \url{https://arxiv.org/abs/2104.14951}.

\bibitem[Lin et~al.(2023)Lin, Gao, Tang, Takikawa, Zeng, Huang, Kreis, Fidler, Liu, and Lin]{text3d1}
C.-H. Lin, J.~Gao, L.~Tang, T.~Takikawa, X.~Zeng, X.~Huang, K.~Kreis, S.~Fidler, M.-Y. Liu, and T.-Y. Lin.
\newblock Magic3d: High-resolution text-to-3d content creation, 2023.
\newblock URL \url{https://arxiv.org/abs/2211.10440}.

\bibitem[Liu et~al.(2023{\natexlab{a}})Liu, Chen, Chen, Zou, and Shi]{DBLP:journals/tgrs/LiuCCZS23}
L.~Liu, B.~Chen, H.~Chen, Z.~Zou, and Z.~Shi.
\newblock Diverse hyperspectral remote sensing image synthesis with diffusion models.
\newblock \emph{{IEEE} Trans. Geosci. Remote. Sens.}, 61:\penalty0 1--16, 2023{\natexlab{a}}.
\newblock \doi{10.1109/TGRS.2023.3335975}.
\newblock URL \url{https://doi.org/10.1109/TGRS.2023.3335975}.

\bibitem[Liu et~al.(2023{\natexlab{b}})Liu, Zhang, Shen, Zheng, Zhu, Feng, Liu, Zhao, Zhou, and Cao]{subject2}
Z.~Liu, Y.~Zhang, Y.~Shen, K.~Zheng, K.~Zhu, R.~Feng, Y.~Liu, D.~Zhao, J.~Zhou, and Y.~Cao.
\newblock Cones 2: Customizable image synthesis with multiple subjects, 2023{\natexlab{b}}.
\newblock URL \url{https://arxiv.org/abs/2305.19327}.

\bibitem[Lobry et~al.(2020)Lobry, Le~Saux, Boulch, and Randrianarivo]{rsvqa}
S.~Lobry, B.~Le~Saux, A.~Boulch, and H.~Randrianarivo.
\newblock Rsvqa: Visual question answering for remote sensing data.
\newblock \emph{IEEE Transactions on Geoscience and Remote Sensing}, 58\penalty0 (12):\penalty0 8555--8566, 2020.

\bibitem[Lu et~al.(2017)Lu, Wang, Zheng, and Li]{rsicd}
X.~Lu, B.~Wang, X.~Zheng, and X.~Li.
\newblock Exploring models and data for remote sensing image caption generation.
\newblock \emph{IEEE Transactions on Geoscience and Remote Sensing}, 56\penalty0 (4):\penalty0 2183--2195, 2017.
\newblock \doi{10.1109/TGRS.2017.2776321}.

\bibitem[Luo and Hu(2021)]{3dgeometry1}
S.~Luo and W.~Hu.
\newblock Diffusion probabilistic models for 3d point cloud generation, 2021.
\newblock URL \url{https://arxiv.org/abs/2103.01458}.

\bibitem[Lütjens et~al.(2019)Lütjens, Liebenwein, and Kramer]{forest_carbon}
B.~Lütjens, L.~Liebenwein, and K.~Kramer.
\newblock Machine learning-based estimation of forest carbon stocks to increase transparency of forest preservation efforts, 2019.
\newblock URL \url{https://arxiv.org/abs/1912.07850}.

\bibitem[Ma et~al.(2019)Ma, Pan, Yuan, and Lei]{sisr3}
W.~Ma, Z.~Pan, F.~Yuan, and B.~Lei.
\newblock Super-resolution of remote sensing images via a dense residual generative adversarial network.
\newblock \emph{Remote Sensing}, 11\penalty0 (21), 2019.
\newblock ISSN 2072-4292.
\newblock \doi{10.3390/rs11212578}.
\newblock URL \url{https://www.mdpi.com/2072-4292/11/21/2578}.

\bibitem[McGovern et~al.(2017)McGovern, Elmore, Gagne, Haupt, Karstens, Lagerquist, Smith, and Williams]{weather}
A.~McGovern, K.~L. Elmore, D.~J. Gagne, S.~E. Haupt, C.~D. Karstens, R.~Lagerquist, T.~Smith, and J.~K. Williams.
\newblock Using artificial intelligence to improve real-time decision-making for high-impact weather.
\newblock \emph{Bulletin of the American Meteorological Society}, 98\penalty0 (10):\penalty0 2073 -- 2090, 2017.
\newblock \doi{10.1175/BAMS-D-16-0123.1}.
\newblock URL \url{https://journals.ametsoc.org/view/journals/bams/98/10/bams-d-16-0123.1.xml}.

\bibitem[Mendieta et~al.(2023)Mendieta, Han, Shi, Zhu, and Chen]{geopile}
M.~Mendieta, B.~Han, X.~Shi, Y.~Zhu, and C.~Chen.
\newblock Towards geospatial foundation models via continual pretraining, 2023.
\newblock URL \url{https://arxiv.org/abs/2302.04476}.

\bibitem[Mou et~al.(2023)Mou, Wang, Xie, Wu, Zhang, Qi, Shan, and Qie]{controllable3}
C.~Mou, X.~Wang, L.~Xie, Y.~Wu, J.~Zhang, Z.~Qi, Y.~Shan, and X.~Qie.
\newblock T2i-adapter: Learning adapters to dig out more controllable ability for text-to-image diffusion models, 2023.
\newblock URL \url{https://arxiv.org/abs/2302.08453}.

\bibitem[Nedungadi et~al.(2024)Nedungadi, Kariryaa, Oehmcke, Belongie, Igel, and Lang]{mmearth}
V.~Nedungadi, A.~Kariryaa, S.~Oehmcke, S.~Belongie, C.~Igel, and N.~Lang.
\newblock Mmearth: Exploring multi-modal pretext tasks for geospatial representation learning, 2024.
\newblock URL \url{https://arxiv.org/abs/2405.02771}.

\bibitem[Nichol et~al.(2022)Nichol, Dhariwal, Ramesh, Shyam, Mishkin, McGrew, Sutskever, and Chen]{glide}
A.~Nichol, P.~Dhariwal, A.~Ramesh, P.~Shyam, P.~Mishkin, B.~McGrew, I.~Sutskever, and M.~Chen.
\newblock Glide: Towards photorealistic image generation and editing with text-guided diffusion models, 2022.
\newblock URL \url{https://arxiv.org/abs/2112.10741}.

\bibitem[Patil et~al.(2024)Patil, Patil, Sitapure, Patil, and Shelke]{article}
S.~Patil, S.~Patil, S.~Sitapure, M.~Patil, and M.~Shelke.
\newblock Text-guided artistic image synthesis using diffusion model.
\newblock \emph{International Research Journal on Advanced Science Hub}, 6:\penalty0 157--166, 06 2024.
\newblock \doi{10.47392/IRJASH.2024.024}.

\bibitem[Poole et~al.(2022)Poole, Jain, Barron, and Mildenhall]{graphics2}
B.~Poole, A.~Jain, J.~T. Barron, and B.~Mildenhall.
\newblock Dreamfusion: Text-to-3d using 2d diffusion, 2022.
\newblock URL \url{https://arxiv.org/abs/2209.14988}.

\bibitem[Popov et~al.(2021)Popov, Vovk, Gogoryan, Sadekova, and Kudinov]{speech2}
V.~Popov, I.~Vovk, V.~Gogoryan, T.~Sadekova, and M.~Kudinov.
\newblock Grad-tts: A diffusion probabilistic model for text-to-speech, 2021.
\newblock URL \url{https://arxiv.org/abs/2105.06337}.

\bibitem[Rabbi et~al.(2020)Rabbi, Ray, Schubert, Chowdhury, and Chao]{sisr6}
J.~Rabbi, N.~Ray, M.~Schubert, S.~Chowdhury, and D.~Chao.
\newblock Small-object detection in remote sensing images with end-to-end edge-enhanced gan and object detector network, 2020.
\newblock URL \url{https://arxiv.org/abs/2003.09085}.

\bibitem[Radford et~al.(2021)Radford, Kim, Hallacy, Ramesh, Goh, Agarwal, Sastry, Askell, Mishkin, Clark, Krueger, and Sutskever]{radford2021learningtransferablevisualmodels}
A.~Radford, J.~W. Kim, C.~Hallacy, A.~Ramesh, G.~Goh, S.~Agarwal, G.~Sastry, A.~Askell, P.~Mishkin, J.~Clark, G.~Krueger, and I.~Sutskever.
\newblock Learning transferable visual models from natural language supervision, 2021.
\newblock URL \url{https://arxiv.org/abs/2103.00020}.

\bibitem[Ramesh et~al.(2022)Ramesh, Dhariwal, Nichol, Chu, and Chen]{hierarchical}
A.~Ramesh, P.~Dhariwal, A.~Nichol, C.~Chu, and M.~Chen.
\newblock Hierarchical text-conditional image generation with clip latents, 2022.
\newblock URL \url{https://arxiv.org/abs/2204.06125}.

\bibitem[Reed et~al.(2023)Reed, Gupta, Li, Brockman, Funk, Clipp, Keutzer, Candido, Uyttendaele, and Darrell]{scalemae}
C.~J. Reed, R.~Gupta, S.~Li, S.~Brockman, C.~Funk, B.~Clipp, K.~Keutzer, S.~Candido, M.~Uyttendaele, and T.~Darrell.
\newblock Scale-mae: A scale-aware masked autoencoder for multiscale geospatial representation learning, 2023.
\newblock URL \url{https://arxiv.org/abs/2212.14532}.

\bibitem[Rolnick et~al.(2019)Rolnick, Donti, Kaack, Kochanski, Lacoste, Sankaran, Ross, Milojevic-Dupont, Jaques, Waldman-Brown, Luccioni, Maharaj, Sherwin, Mukkavilli, Kording, Gomes, Ng, Hassabis, Platt, Creutzig, Chayes, and Bengio]{climatechange}
D.~Rolnick, P.~L. Donti, L.~H. Kaack, K.~Kochanski, A.~Lacoste, K.~Sankaran, A.~S. Ross, N.~Milojevic-Dupont, N.~Jaques, A.~Waldman-Brown, A.~Luccioni, T.~Maharaj, E.~D. Sherwin, S.~K. Mukkavilli, K.~P. Kording, C.~Gomes, A.~Y. Ng, D.~Hassabis, J.~C. Platt, F.~Creutzig, J.~Chayes, and Y.~Bengio.
\newblock Tackling climate change with machine learning, 2019.
\newblock URL \url{https://arxiv.org/abs/1906.05433}.

\bibitem[Rombach et~al.(2022)Rombach, Blattmann, Lorenz, Esser, and Ommer]{ldm1}
R.~Rombach, A.~Blattmann, D.~Lorenz, P.~Esser, and B.~Ommer.
\newblock High-resolution image synthesis with latent diffusion models, 2022.
\newblock URL \url{https://arxiv.org/abs/2112.10752}.

\bibitem[Ruiz et~al.(2023)Ruiz, Li, Jampani, Pritch, Rubinstein, and Aberman]{subject3}
N.~Ruiz, Y.~Li, V.~Jampani, Y.~Pritch, M.~Rubinstein, and K.~Aberman.
\newblock Dreambooth: Fine tuning text-to-image diffusion models for subject-driven generation, 2023.
\newblock URL \url{https://arxiv.org/abs/2208.12242}.

\bibitem[Saharia et~al.(2021)Saharia, Ho, Chan, Salimans, Fleet, and Norouzi]{saharia2021imagesuperresolutioniterativerefinement}
C.~Saharia, J.~Ho, W.~Chan, T.~Salimans, D.~J. Fleet, and M.~Norouzi.
\newblock Image super-resolution via iterative refinement, 2021.
\newblock URL \url{https://arxiv.org/abs/2104.07636}.

\bibitem[Saharia et~al.(2022)Saharia, Chan, Saxena, Li, Whang, Denton, Ghasemipour, Ayan, Mahdavi, Lopes, Salimans, Ho, Fleet, and Norouzi]{ldm2}
C.~Saharia, W.~Chan, S.~Saxena, L.~Li, J.~Whang, E.~Denton, S.~K.~S. Ghasemipour, B.~K. Ayan, S.~S. Mahdavi, R.~G. Lopes, T.~Salimans, J.~Ho, D.~J. Fleet, and M.~Norouzi.
\newblock Photorealistic text-to-image diffusion models with deep language understanding, 2022.
\newblock URL \url{https://arxiv.org/abs/2205.11487}.

\bibitem[Sebaq and ElHelw(2024)]{Sebaq_2024}
A.~Sebaq and M.~ElHelw.
\newblock Rsdiff: remote sensing image generation from text using diffusion model.
\newblock \emph{Neural Computing and Applications}, 36\penalty0 (36):\penalty0 23103–23111, Oct. 2024.
\newblock ISSN 1433-3058.
\newblock \doi{10.1007/s00521-024-10363-3}.
\newblock URL \url{http://dx.doi.org/10.1007/s00521-024-10363-3}.

\bibitem[Sheng et~al.(2020)Sheng, Irvin, Munukutla, Zhang, Cross, Story, Rustowicz, Elsworth, Yang, Omara, Gautam, Jackson, and Ng]{ognet}
H.~Sheng, J.~Irvin, S.~Munukutla, S.~Zhang, C.~Cross, K.~Story, R.~Rustowicz, C.~Elsworth, Z.~Yang, M.~Omara, R.~Gautam, R.~B. Jackson, and A.~Y. Ng.
\newblock Ognet: Towards a global oil and gas infrastructure database using deep learning on remotely sensed imagery, 2020.
\newblock URL \url{https://arxiv.org/abs/2011.07227}.

\bibitem[Shue et~al.(2022)Shue, Chan, Po, Ankner, Wu, and Wetzstein]{graphics1}
J.~R. Shue, E.~R. Chan, R.~Po, Z.~Ankner, J.~Wu, and G.~Wetzstein.
\newblock 3d neural field generation using triplane diffusion, 2022.
\newblock URL \url{https://arxiv.org/abs/2211.16677}.

\bibitem[Song et~al.(2020)Song, Meng, and Ermon]{diffusion2}
J.~Song, C.~Meng, and S.~Ermon.
\newblock Denoising diffusion implicit models, 2020.
\newblock URL \url{https://arxiv.org/abs/2010.02502}.

\bibitem[Tang et~al.(2024)Tang, Cao, Hou, Jiang, Liu, and Meng]{crsdiff}
D.~Tang, X.~Cao, X.~Hou, Z.~Jiang, J.~Liu, and D.~Meng.
\newblock Crs-diff: Controllable remote sensing image generation with diffusion model, 2024.
\newblock URL \url{https://arxiv.org/abs/2403.11614}.

\bibitem[Thomee et~al.(2016)Thomee, Shamma, Friedland, Elizalde, Ni, Poland, Borth, and Li]{yfcc}
B.~Thomee, D.~A. Shamma, G.~Friedland, B.~Elizalde, K.~Ni, D.~Poland, D.~Borth, and L.-J. Li.
\newblock Yfcc100m: The new data in multimedia research.
\newblock \emph{Communications of the ACM}, 59\penalty0 (2):\penalty0 64--73, 2016.

\bibitem[Van~Etten et~al.(2021)Van~Etten, Hogan, Manso, Shermeyer, Weir, and Lewis]{spacenet2}
A.~Van~Etten, D.~Hogan, J.~M. Manso, J.~Shermeyer, N.~Weir, and R.~Lewis.
\newblock The multi-temporal urban development spacenet dataset.
\newblock In \emph{2021 IEEE/CVF Conference on Computer Vision and Pattern Recognition (CVPR)}, pages 6394--6403, 2021.
\newblock \doi{10.1109/CVPR46437.2021.00633}.

\bibitem[Van~Horn et~al.(2018)Van~Horn, Mac~Aodha, Song, Cui, Sun, Shepard, Adam, Perona, and Belongie]{inat}
G.~Van~Horn, O.~Mac~Aodha, Y.~Song, Y.~Cui, C.~Sun, A.~Shepard, H.~Adam, P.~Perona, and S.~Belongie.
\newblock The inaturalist species classification and detection dataset.
\newblock \emph{arXiv preprint arXiv:1707.06642}, 2018.

\bibitem[von Platen et~al.(2022)von Platen, Patil, Lozhkov, Cuenca, Lambert, Rasul, Davaadorj, Nair, Paul, Berman, Xu, Liu, and Wolf]{diffusers}
P.~von Platen, S.~Patil, A.~Lozhkov, P.~Cuenca, N.~Lambert, K.~Rasul, M.~Davaadorj, D.~Nair, S.~Paul, W.~Berman, Y.~Xu, S.~Liu, and T.~Wolf.
\newblock Diffusers: State-of-the-art diffusion models.
\newblock \url{https://github.com/huggingface/diffusers}, 2022.

\bibitem[Wallace et~al.(2023)Wallace, Gokul, and Naik]{Wallace_2023_CVPR}
B.~Wallace, A.~Gokul, and N.~Naik.
\newblock Edict: Exact diffusion inversion via coupled transformations.
\newblock In \emph{Proceedings of the IEEE/CVF Conference on Computer Vision and Pattern Recognition (CVPR)}, pages 22532--22541, June 2023.

\bibitem[Wang et~al.(2018)Wang, Yu, Wu, Gu, Liu, Dong, Loy, Qiao, and Tang]{sisr7}
X.~Wang, K.~Yu, S.~Wu, J.~Gu, Y.~Liu, C.~Dong, C.~C. Loy, Y.~Qiao, and X.~Tang.
\newblock Esrgan: Enhanced super-resolution generative adversarial networks, 2018.
\newblock URL \url{https://arxiv.org/abs/1809.00219}.

\bibitem[Wang et~al.(2004)Wang, Bovik, Sheikh, and Simoncelli]{ssim}
Z.~Wang, A.~Bovik, H.~Sheikh, and E.~Simoncelli.
\newblock Image quality assessment: from error visibility to structural similarity.
\newblock \emph{IEEE Transactions on Image Processing}, 13\penalty0 (4):\penalty0 600--612, 2004.
\newblock \doi{10.1109/TIP.2003.819861}.

\bibitem[Wang et~al.(2020)Wang, Jiang, Yi, Han, and He]{sisr1}
Z.~Wang, K.~Jiang, P.~Yi, Z.~Han, and Z.~He.
\newblock Ultra-dense gan for satellite imagery super-resolution.
\newblock \emph{Neurocomputing}, 398:\penalty0 328--337, 2020.
\newblock ISSN 0925-2312.
\newblock \doi{https://doi.org/10.1016/j.neucom.2019.03.106}.
\newblock URL \url{https://www.sciencedirect.com/science/article/pii/S0925231219314602}.

\bibitem[Wang et~al.(2023)Wang, Lu, Wang, Bao, Li, Su, and Zhu]{text3d2}
Z.~Wang, C.~Lu, Y.~Wang, F.~Bao, C.~Li, H.~Su, and J.~Zhu.
\newblock Prolificdreamer: High-fidelity and diverse text-to-3d generation with variational score distillation, 2023.
\newblock URL \url{https://arxiv.org/abs/2305.16213}.

\bibitem[Wanyan et~al.(2024)Wanyan, Seneviratne, Shen, and Kirley]{dino_mc}
X.~Wanyan, S.~Seneviratne, S.~Shen, and M.~Kirley.
\newblock Extending global-local view alignment for self-supervised learning with remote sensing imagery, 2024.
\newblock URL \url{https://arxiv.org/abs/2303.06670}.

\bibitem[Wu et~al.(2023)Wu, Wang, Mao, and Li]{wu2023hsrdiffhyperspectralimagesuperresolutionconditional}
C.~Wu, D.~Wang, H.~Mao, and Y.~Li.
\newblock Hsr-diff:hyperspectral image super-resolution via conditional diffusion models, 2023.
\newblock URL \url{https://arxiv.org/abs/2306.12085}.

\bibitem[Xia et~al.(2017)Xia, Hu, Hu, Shi, Bai, Zhong, Zhang, and Lu]{aid}
G.-S. Xia, J.~Hu, F.~Hu, B.~Shi, X.~Bai, Y.~Zhong, L.~Zhang, and X.~Lu.
\newblock Aid: A benchmark dataset for performance evaluation of aerial scene classification.
\newblock \emph{IEEE Transactions on Geoscience and Remote Sensing}, 55\penalty0 (7):\penalty0 3965--3981, 2017.

\bibitem[Xu et~al.(2022)Xu, Yu, Song, Shi, Ermon, and Tang]{3dgeometry2}
M.~Xu, L.~Yu, Y.~Song, C.~Shi, S.~Ermon, and J.~Tang.
\newblock Geodiff: a geometric diffusion model for molecular conformation generation, 2022.
\newblock URL \url{https://arxiv.org/abs/2203.02923}.

\bibitem[Yang and Newsam(2010)]{ucmerced}
Y.~Yang and S.~Newsam.
\newblock Bag-of-visual-words and spatial extensions for land-use classification.
\newblock \emph{Proceedings of the 18th SIGSPATIAL International Conference on Advances in Geographic Information Systems}, pages 270--279, 2010.

\bibitem[Yu et~al.(2024)Yu, Liu, Liu, Shi, and Zou]{metaearth}
Z.~Yu, C.~Liu, L.~Liu, Z.~Shi, and Z.~Zou.
\newblock Metaearth: A generative foundation model for global-scale remote sensing image generation, 2024.
\newblock URL \url{https://arxiv.org/abs/2405.13570}.

\bibitem[Yuan et~al.(2023)Yuan, Hao, Zhou, Chen, Yu, Zhang, Wang, and Sun]{10105619}
Z.~Yuan, C.~Hao, R.~Zhou, J.~Chen, M.~Yu, W.~Zhang, H.~Wang, and X.~Sun.
\newblock Efficient and controllable remote sensing fake sample generation based on diffusion model.
\newblock \emph{IEEE Transactions on Geoscience and Remote Sensing}, 61:\penalty0 1--12, 2023.
\newblock \doi{10.1109/TGRS.2023.3268331}.

\bibitem[Zavras et~al.(2025)Zavras, Michail, Zhu, Demir, and Papoutsis]{gaia}
A.~Zavras, D.~Michail, X.~X. Zhu, B.~Demir, and I.~Papoutsis.
\newblock Gaia: A global, multi-modal, multi-scale vision-language dataset for remote sensing image analysis, 2025.
\newblock URL \url{https://arxiv.org/abs/2502.09598}.

\bibitem[Zhang et~al.(2024)Zhang, Li, Lin, Zhang, Fan, and Liu]{cyclegansar}
H.~Zhang, H.~Li, J.~Lin, Y.~Zhang, J.~Fan, and H.~Liu.
\newblock Seg-cyclegan : Sar-to-optical image translation guided by a downstream task, 2024.
\newblock URL \url{https://arxiv.org/abs/2408.05777}.

\bibitem[Zhang et~al.(2023{\natexlab{a}})Zhang, Rao, and Agrawala]{controllable2}
L.~Zhang, A.~Rao, and M.~Agrawala.
\newblock Adding conditional control to text-to-image diffusion models, 2023{\natexlab{a}}.
\newblock URL \url{https://arxiv.org/abs/2302.05543}.

\bibitem[Zhang et~al.(2018)Zhang, Isola, Efros, Shechtman, and Wang]{lpips}
R.~Zhang, P.~Isola, A.~A. Efros, E.~Shechtman, and O.~Wang.
\newblock The unreasonable effectiveness of deep features as a perceptual metric, 2018.
\newblock URL \url{https://arxiv.org/abs/1801.03924}.

\bibitem[Zhang et~al.(2023{\natexlab{b}})Zhang, Huang, Tang, Huang, Ma, Dong, and Xu]{zhang2023inversionbasedstyletransferdiffusion}
Y.~Zhang, N.~Huang, F.~Tang, H.~Huang, C.~Ma, W.~Dong, and C.~Xu.
\newblock Inversion-based style transfer with diffusion models, 2023{\natexlab{b}}.
\newblock URL \url{https://arxiv.org/abs/2211.13203}.

\bibitem[Zhao and Xiong(2022)]{sydney}
K.~Zhao and W.~Xiong.
\newblock Exploring data and models in sar ship image captioning.
\newblock \emph{IEEE Access}, PP:\penalty0 1--1, 01 2022.
\newblock \doi{10.1109/ACCESS.2022.3202193}.

\bibitem[Zhao et~al.(2024)Zhao, Wang, Ouyang, Chen, Liu, Zhang, Yu, Wang, Xie, Li, et~al.]{cell}
T.~Zhao, S.~Wang, C.~Ouyang, M.~Chen, C.~Liu, J.~Zhang, L.~Yu, F.~Wang, Y.~Xie, J.~Li, et~al.
\newblock Artificial intelligence for geoscience: Progress, challenges and perspectives.
\newblock \emph{The Innovation}, 2024.

\bibitem[Zhou et~al.(2021)Zhou, Du, and Wu]{3dgeometry3}
L.~Zhou, Y.~Du, and J.~Wu.
\newblock 3d shape generation and completion through point-voxel diffusion, 2021.
\newblock URL \url{https://arxiv.org/abs/2104.03670}.

\bibitem[Zhu et~al.(2020)Zhu, Park, Isola, and Efros]{cyclegan}
J.-Y. Zhu, T.~Park, P.~Isola, and A.~A. Efros.
\newblock Unpaired image-to-image translation using cycle-consistent adversarial networks, 2020.
\newblock URL \url{https://arxiv.org/abs/1703.10593}.

\end{thebibliography}

\appendix
\section{Additional Results}
\label{appendix:extra_results}

\begin{algorithm}
\caption{Aligning fMoW Images with ERA-5 Environmental Measurements}
\begin{algorithmic}[1]
\For{each image $x \in \mathbb{R}^{C \times H \times W}$}
    \State Extract capture time $t$ and centroid coordinates $(\text{lat}, \text{lon})$
    \State Identify the ERA-5 global grid closest to time $t$
    \State Within that grid, find the $25km^{2}$ sub-grid tile closest to $(\text{lat}, \text{lon})$
    \For{each environmental measurement field} 
        \If{that field $\in \{2t, tp, 10u, 10v, ssr, 2d\}$}
            \State Compute 5-day aggregate of that field for that tile
            \State Assign aggregated value for that field to image $x$
        \Else
            \State Directly assign non-aggregated value for that field to image $x$
        \EndIf
    \EndFor
\EndFor
\end{algorithmic}
\label{algo:1}
\end{algorithm}

\renewcommand{\thefigure}{A\arabic{figure}}
\setcounter{figure}{0}

\begin{figure*}[t]
   \centering
\begin{subfigure}[t]{0.12\textwidth}
    \includegraphics[width=\textwidth]{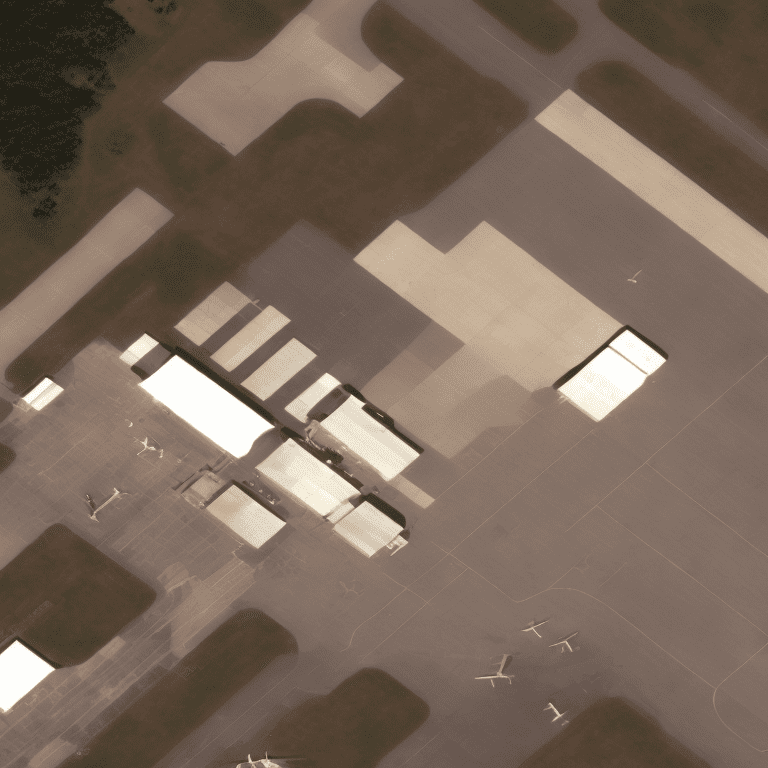}
    \caption{Airport Hangar}
\end{subfigure}
\begin{subfigure}[t]{0.12\textwidth}
   \centering
    \includegraphics[width=\textwidth]{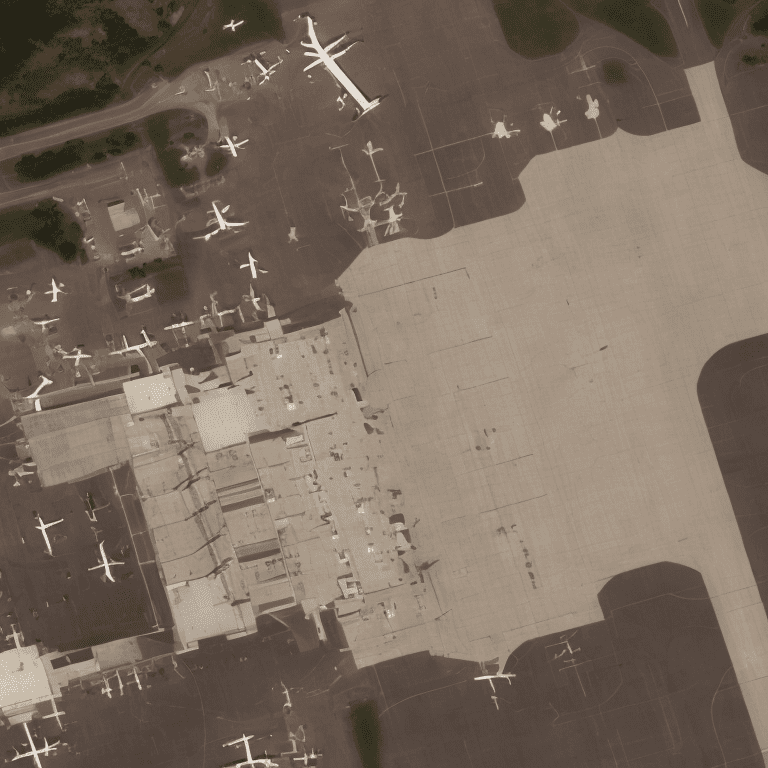}
    \caption{Airport Terminal}
\end{subfigure}
\begin{subfigure}[t]{0.12\textwidth}
   \centering
    \includegraphics[width=\textwidth]{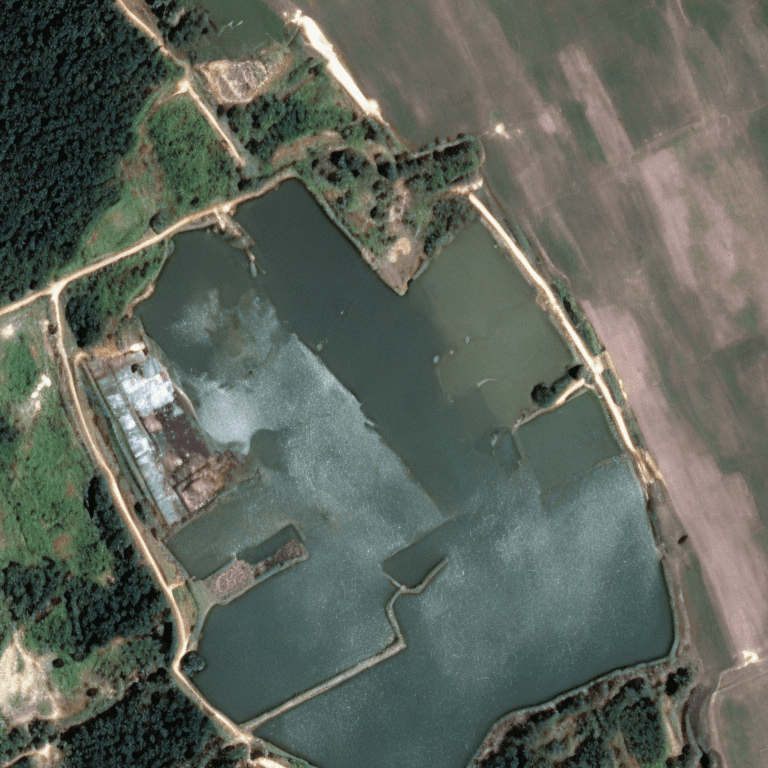}
    \caption{Aquaculture}
\end{subfigure}
\begin{subfigure}[t]{0.12\textwidth}
   \centering
    \includegraphics[width=\textwidth]{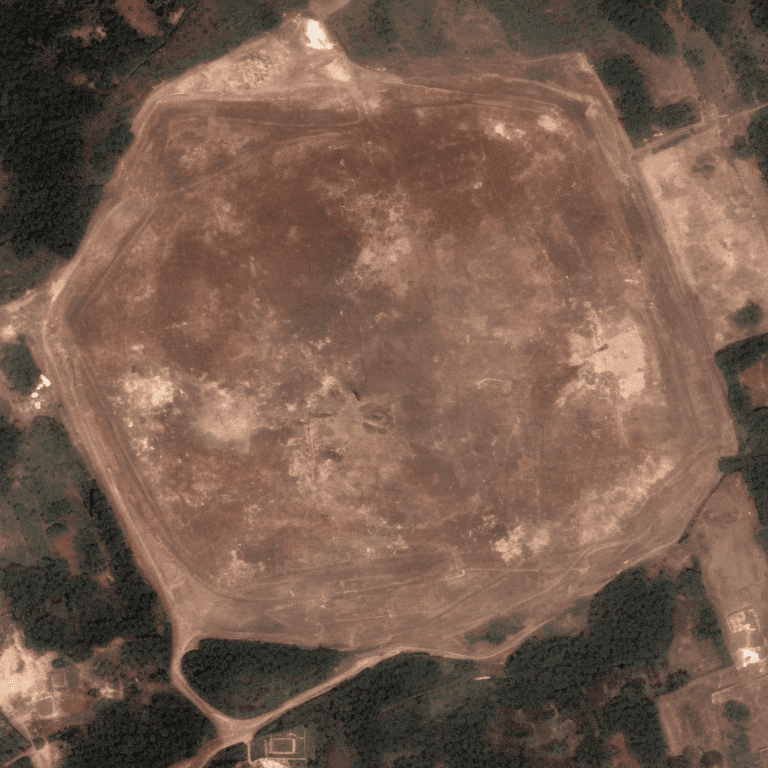}
    \caption{Arch. Site}
\end{subfigure}
\begin{subfigure}[t]{0.12\textwidth}
   \centering
    \includegraphics[width=\textwidth]{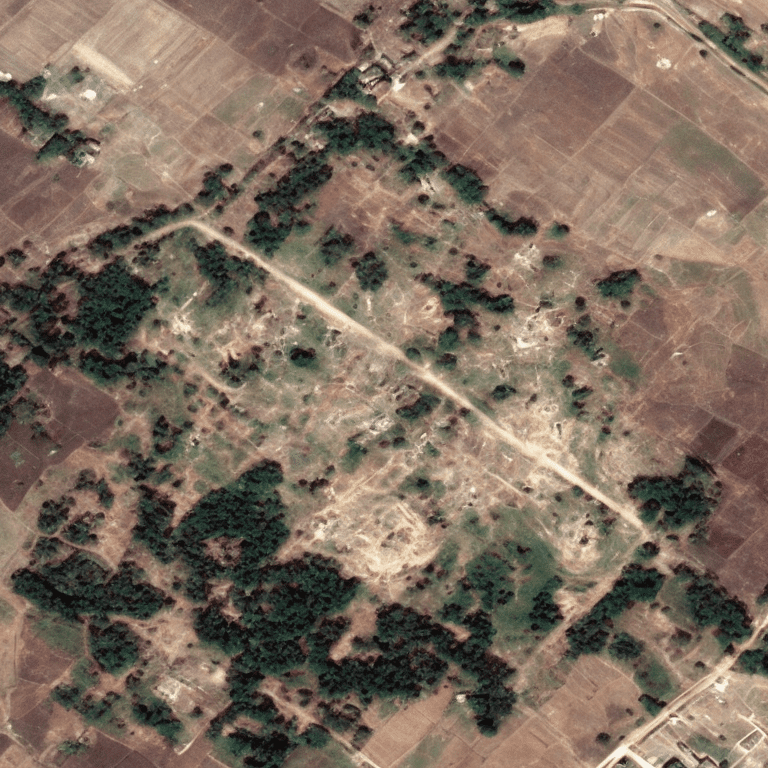}
    \caption{Burial Site}
\end{subfigure}
\begin{subfigure}[t]{0.12\textwidth}
   \centering
    \includegraphics[width=\textwidth]{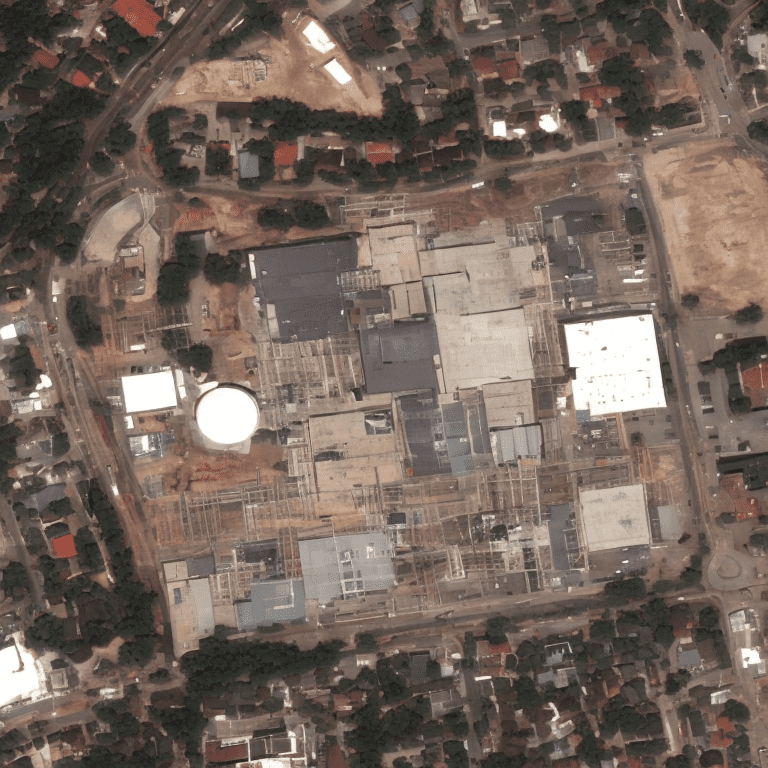}
    \caption{Construction Site}
\end{subfigure}
\begin{subfigure}[t]{0.12\textwidth}
   \centering
    \includegraphics[width=\textwidth]{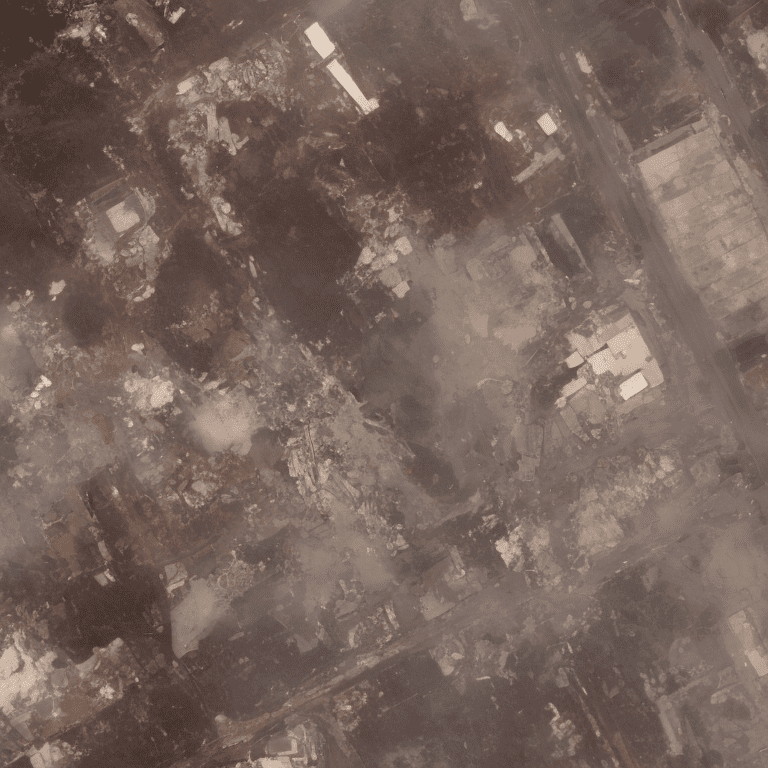}
    \caption{Debris or Rubble}
\end{subfigure}\\
\begin{subfigure}[t]{0.12\textwidth}
   \centering
    \includegraphics[width=\textwidth]{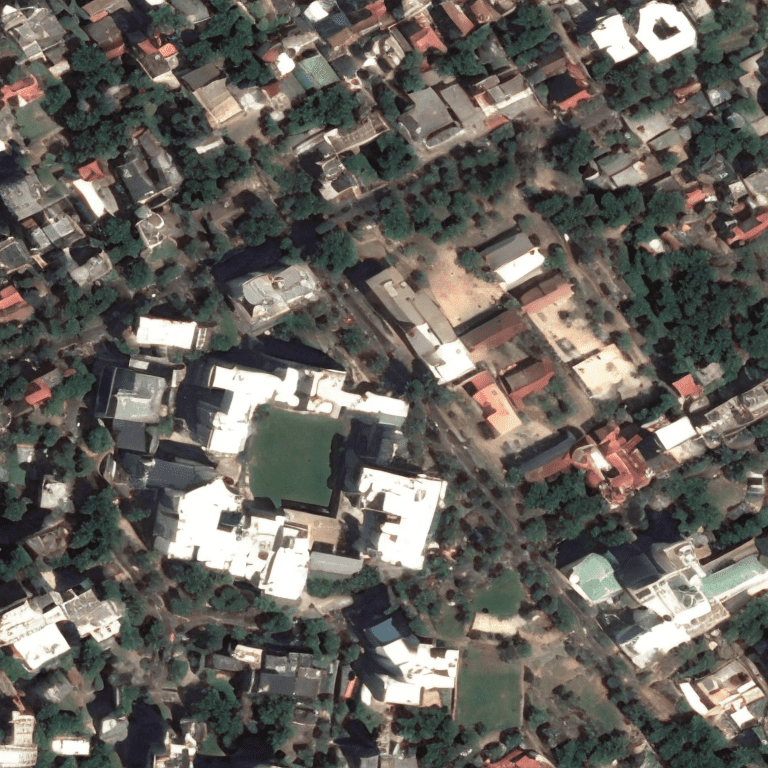}
    \caption{Edu. Institution}
\end{subfigure}
\begin{subfigure}[t]{0.12\textwidth}
   \centering
    \includegraphics[width=\textwidth]{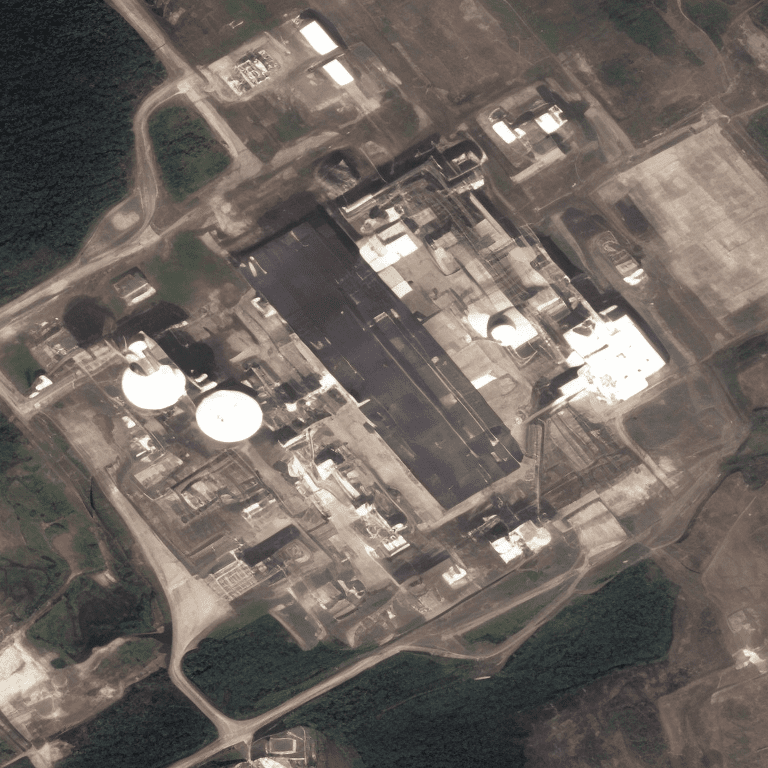}
    \caption{Factory}
\end{subfigure}
\begin{subfigure}[t]{0.12\textwidth}
   \centering
    \includegraphics[width=\textwidth]{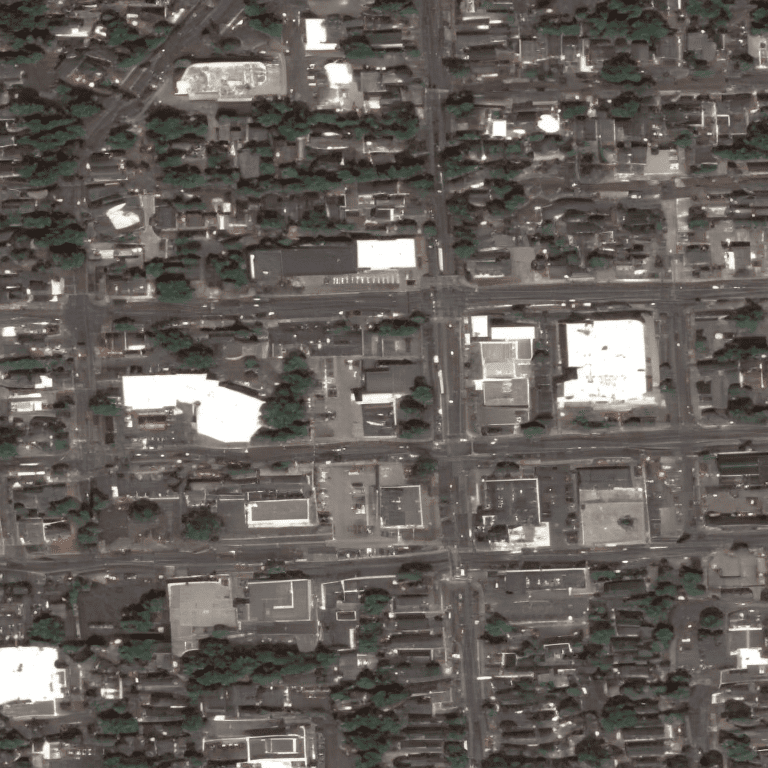}
    \caption{Fire Station}
\end{subfigure}
\begin{subfigure}[t]{0.12\textwidth}
   \centering
    \includegraphics[width=\textwidth]{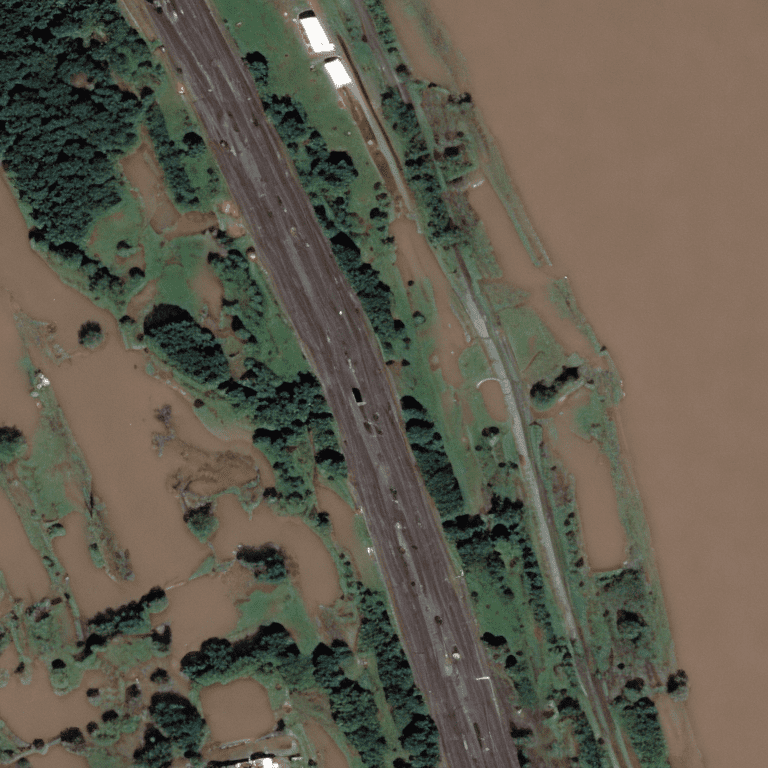}
    \caption{Flooded Road}
\end{subfigure}
\begin{subfigure}[t]{0.12\textwidth}
   \centering
    \includegraphics[width=\textwidth]{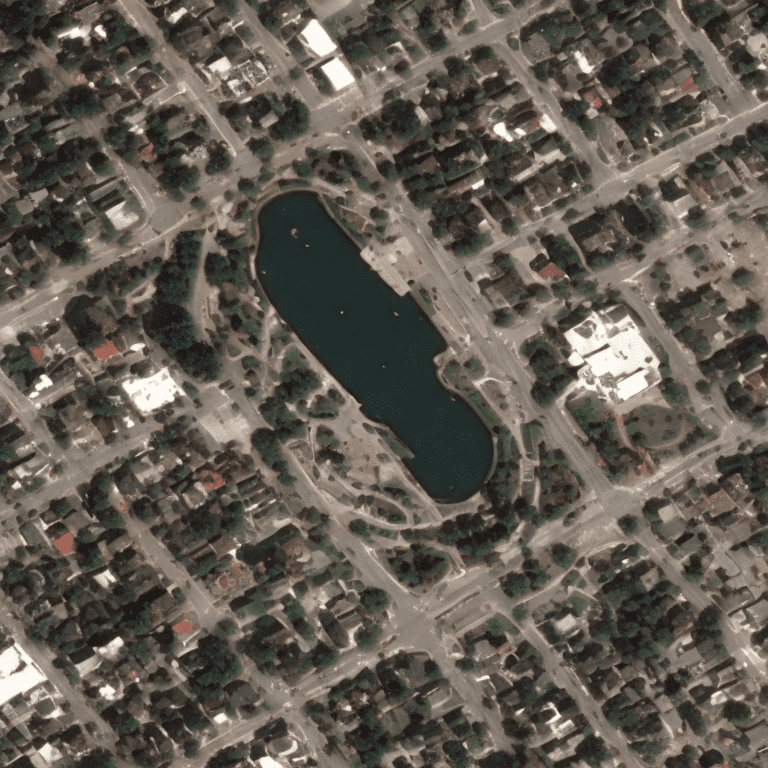}
    \caption{Fountain}
\end{subfigure}
\begin{subfigure}[t]{0.12\textwidth}
   \centering
    \includegraphics[width=\textwidth]{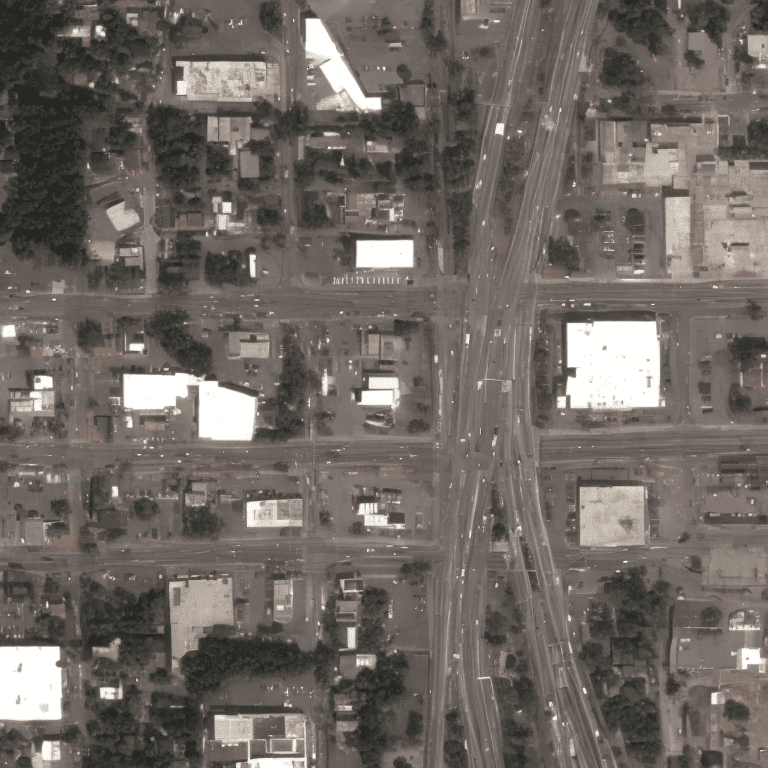}
    \caption{Gas Station}
\end{subfigure}
\begin{subfigure}[t]{0.12\textwidth}
   \centering
    \includegraphics[width=\textwidth]{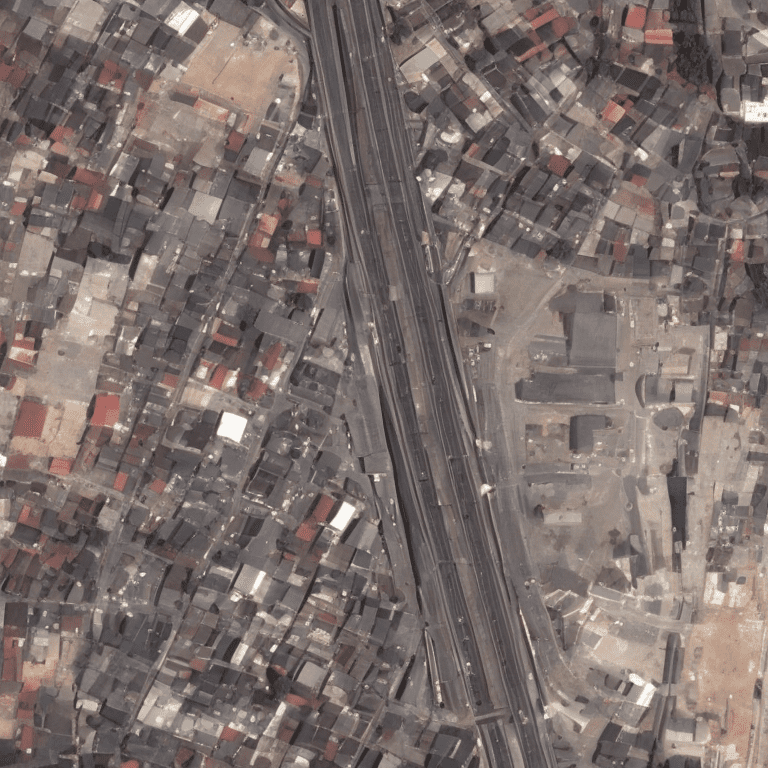}
    \caption{Gr. Transp. Station}
\end{subfigure}\\
\begin{subfigure}[t]{0.12\textwidth}
   \centering
    \includegraphics[width=\textwidth]{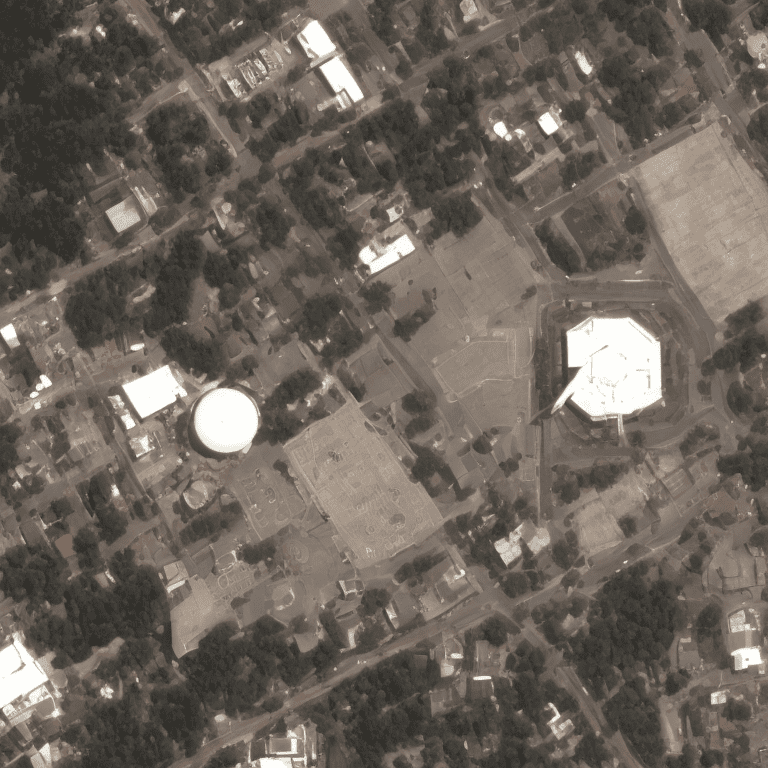}
    \caption{Helipad}
\end{subfigure}
\begin{subfigure}[t]{0.12\textwidth}
   \centering
    \includegraphics[width=\textwidth]{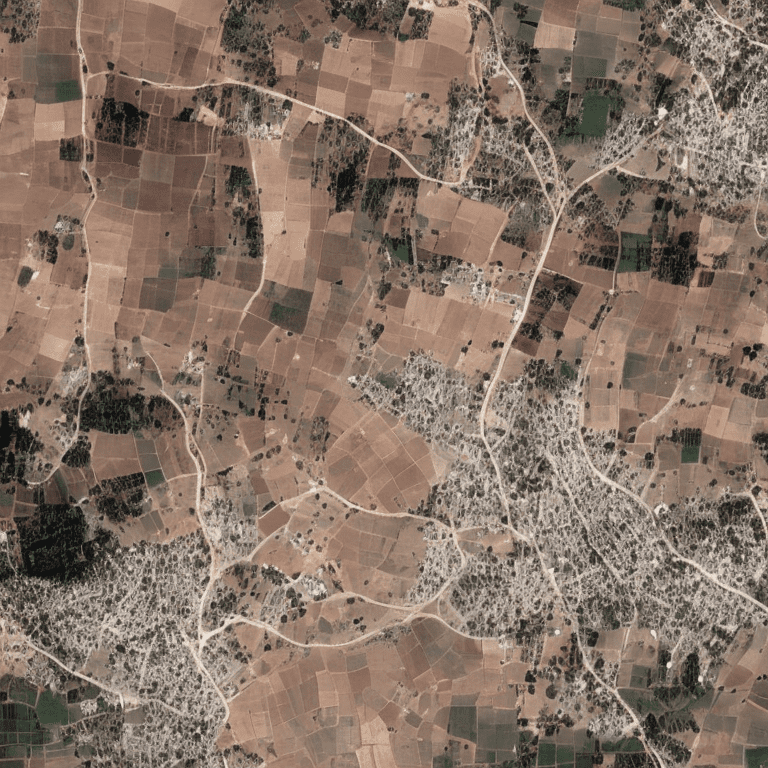}
    \caption{Impov. Settlement}
\end{subfigure}
\begin{subfigure}[t]{0.12\textwidth}
   \centering
    \includegraphics[width=\textwidth]{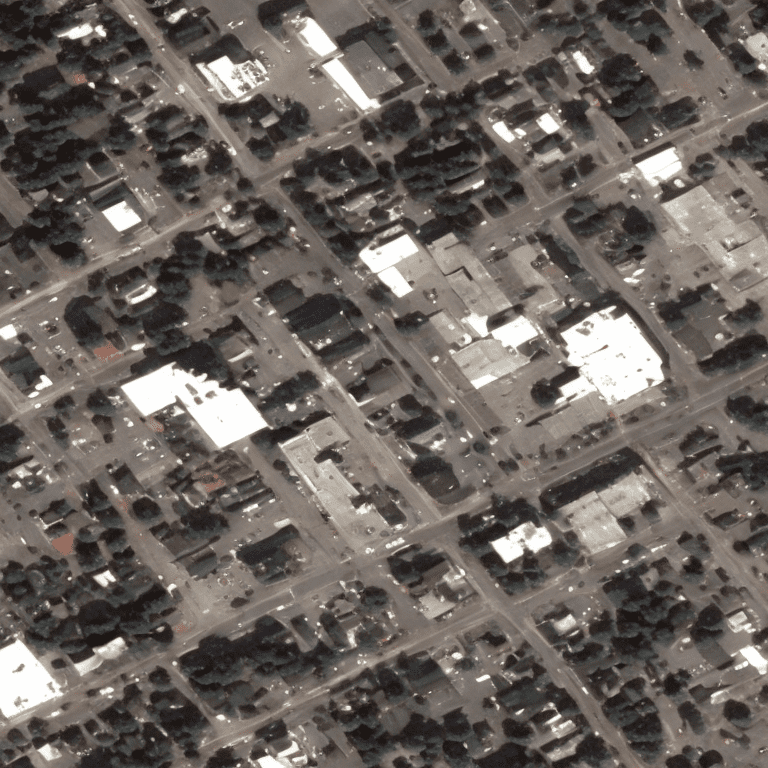}
    \caption{Office Building}
\end{subfigure}
\begin{subfigure}[t]{0.12\textwidth}
   \centering
    \includegraphics[width=\textwidth]{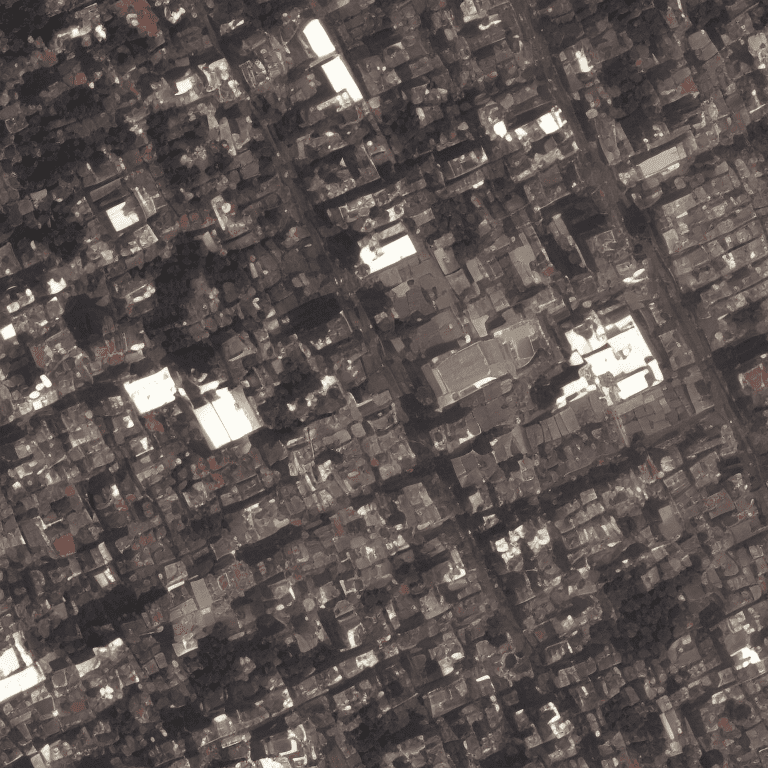}
    \caption{Place of Worship}
\end{subfigure}
 \begin{subfigure}[t]{0.12\textwidth}
   \centering
    \includegraphics[width=\textwidth]{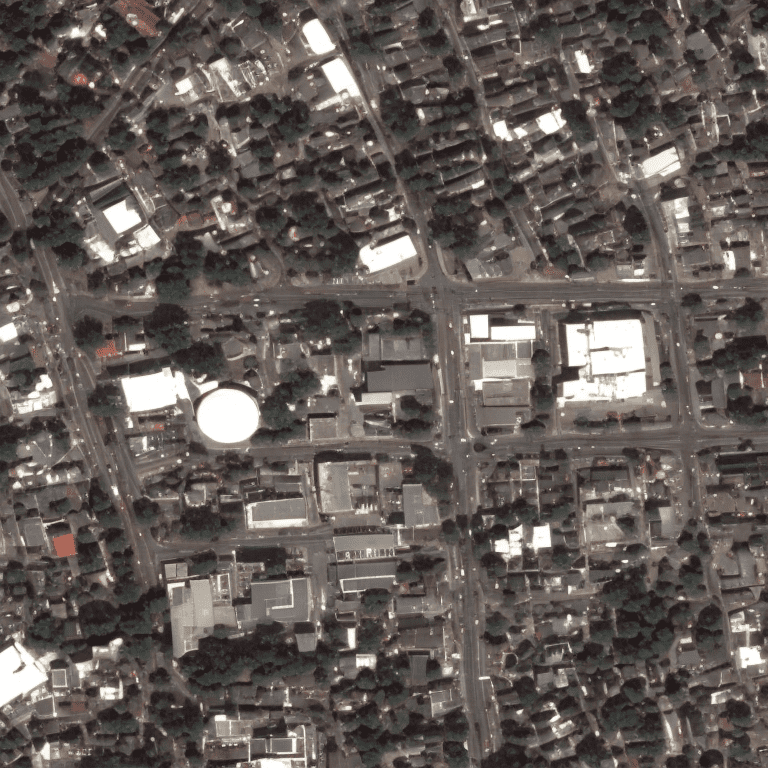}
    \caption{Police Station}
\end{subfigure}
\begin{subfigure}[t]{0.12\textwidth}
   \centering
    \includegraphics[width=\textwidth]{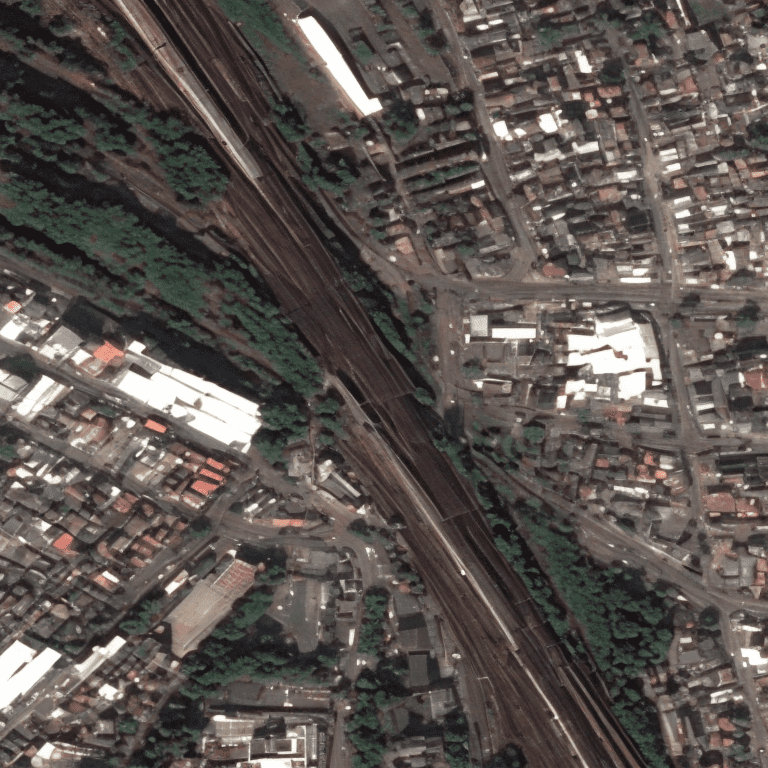}
    \caption{Railway Bridge}
\end{subfigure}
\begin{subfigure}[t]{0.12\textwidth}
   \centering
    \includegraphics[width=\textwidth]{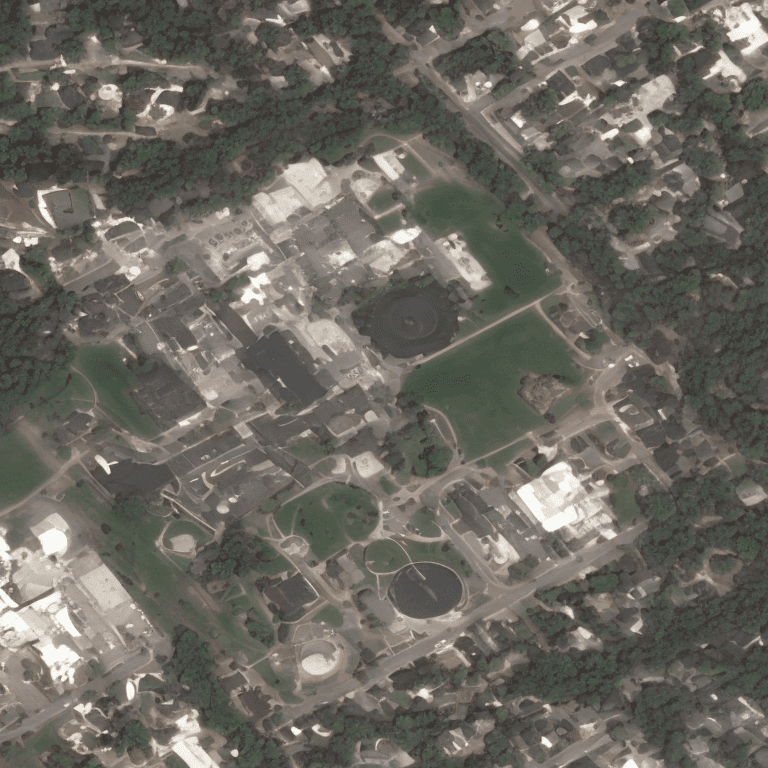}
    \caption{Recr. Facility}
\end{subfigure}\\
\begin{subfigure}[t]{0.12\textwidth}
   \centering
    \includegraphics[width=\textwidth]{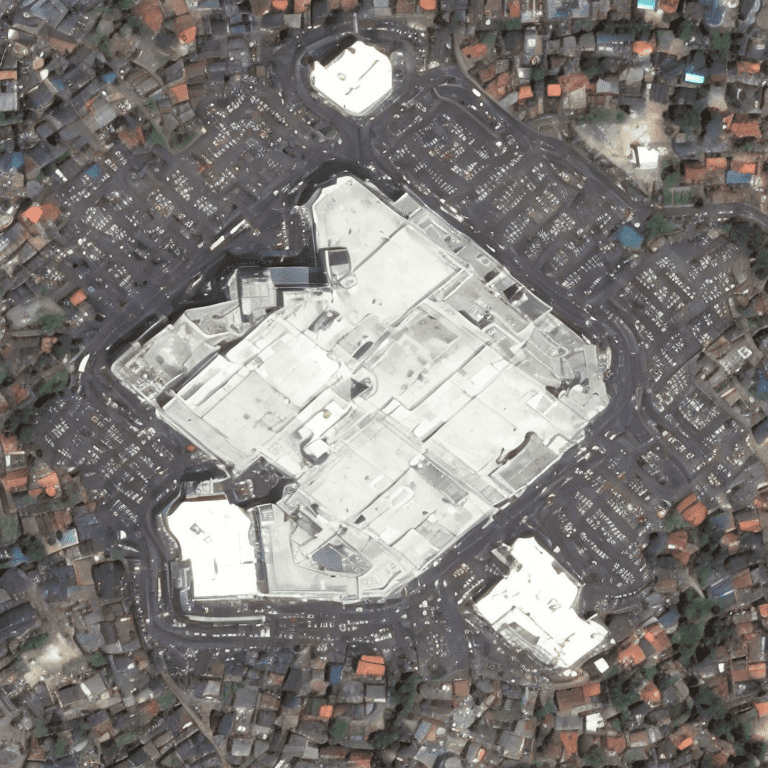}
    \caption{Shopping Mall}
\end{subfigure}
\begin{subfigure}[t]{0.12\textwidth}
   \centering
    \includegraphics[width=\textwidth]{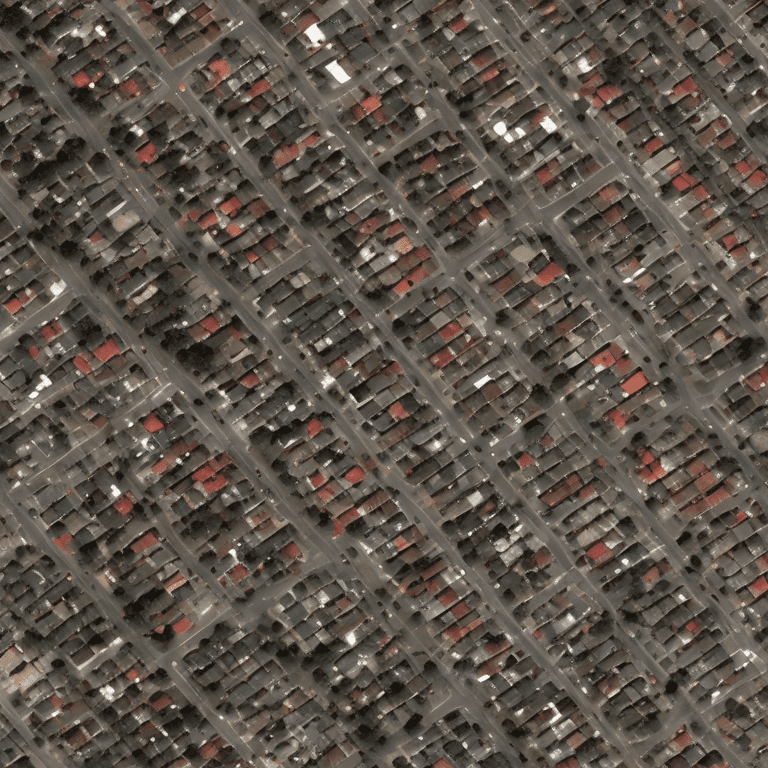}
    \caption{Single-unit Res.}
\end{subfigure}
\begin{subfigure}[t]{0.12\textwidth}
   \centering
    \includegraphics[width=\textwidth]{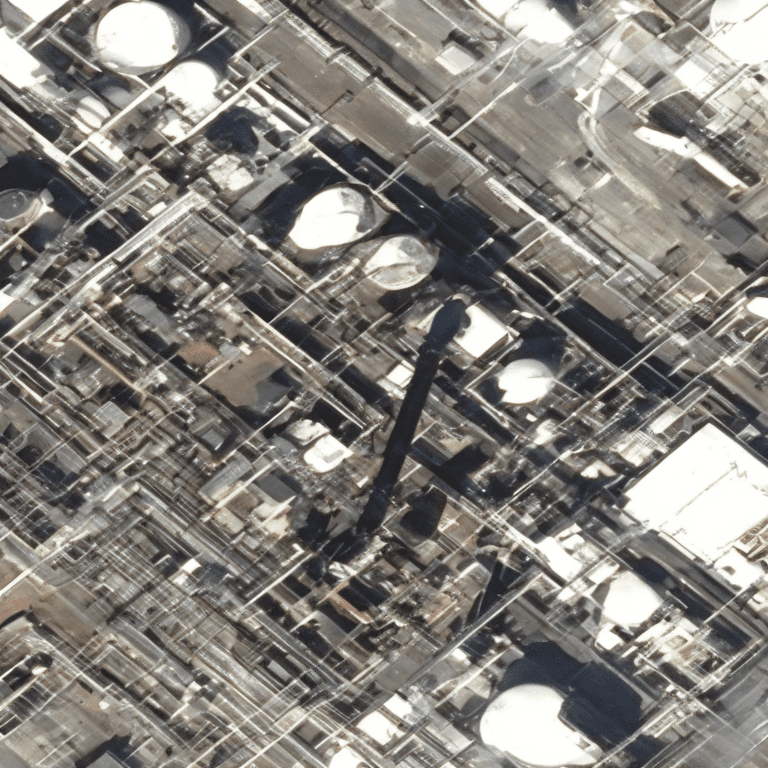}
    \caption{Smokestack}
\end{subfigure}
 \begin{subfigure}[t]{0.12\textwidth}
   \centering
    \includegraphics[width=\textwidth]{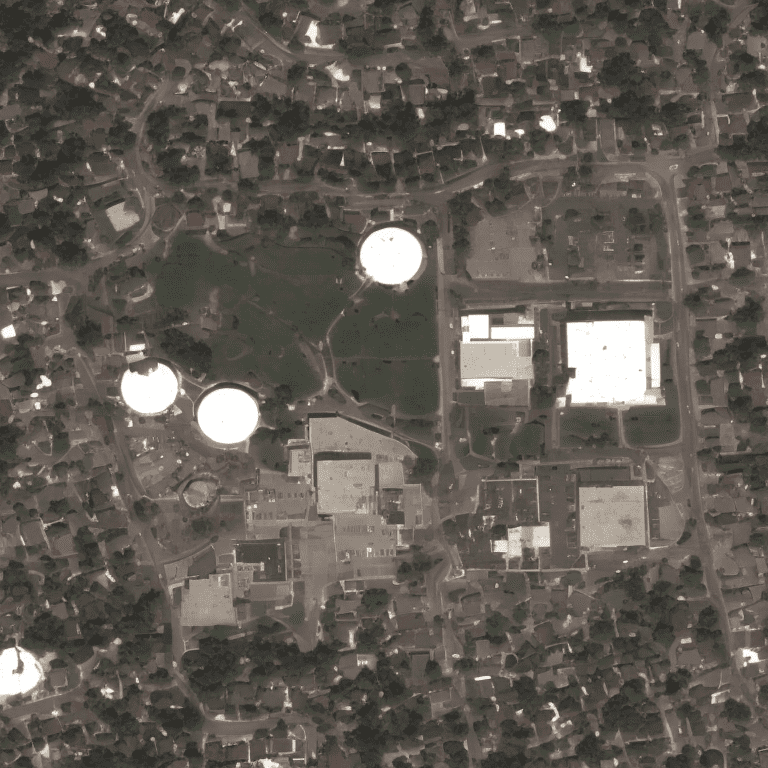}
    \caption{Space Facility}
\end{subfigure}
\begin{subfigure}[t]{0.12\textwidth}
   \centering
    \includegraphics[width=\textwidth]{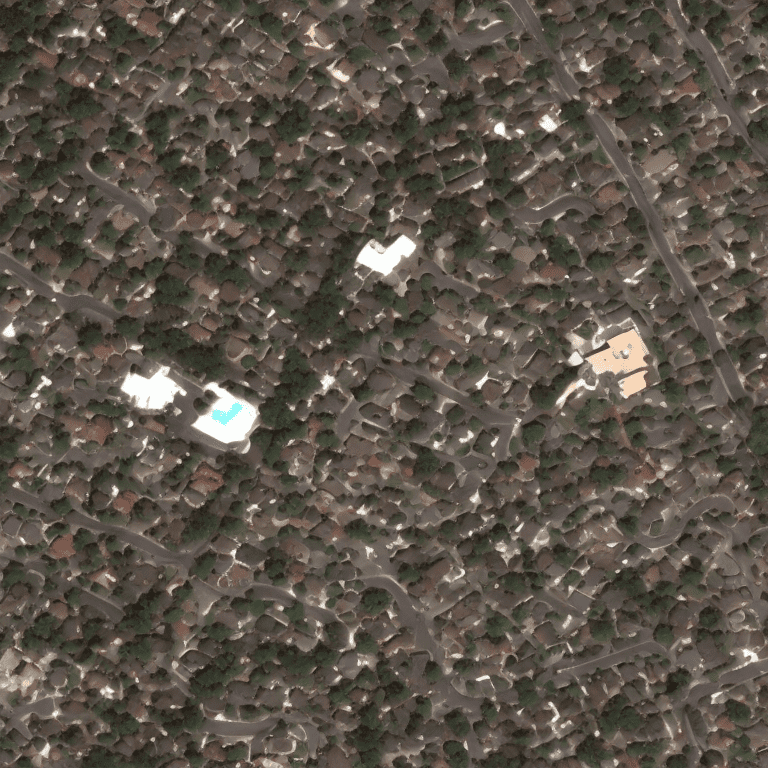}
    \caption{Swimming Pool}
\end{subfigure}
\begin{subfigure}[t]{0.12\textwidth}
   \centering
    \includegraphics[width=\textwidth]{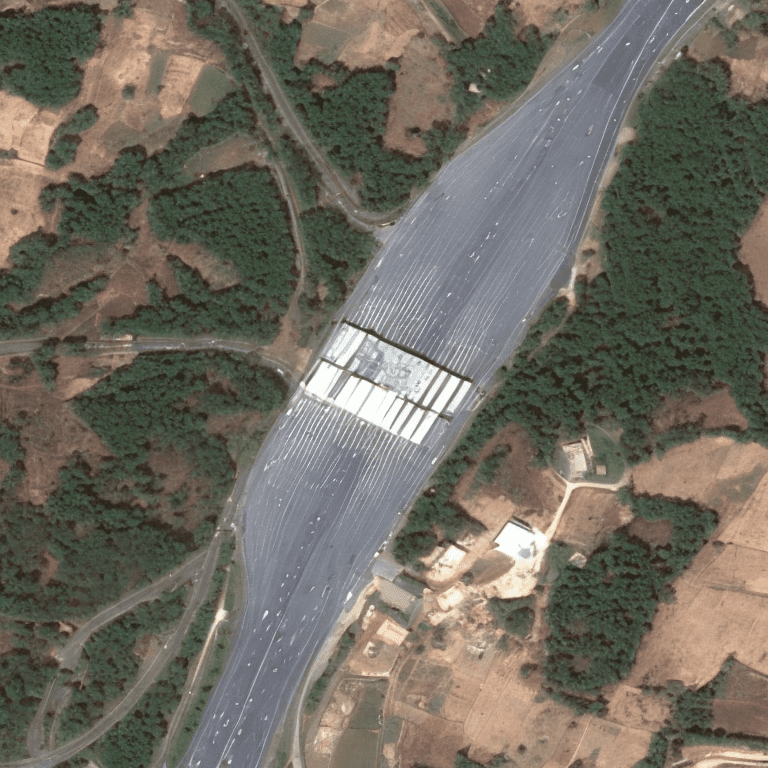}
    \caption{Toll Booth}
\end{subfigure}
\begin{subfigure}[t]{0.12\textwidth}
   \centering
    \includegraphics[width=\textwidth]{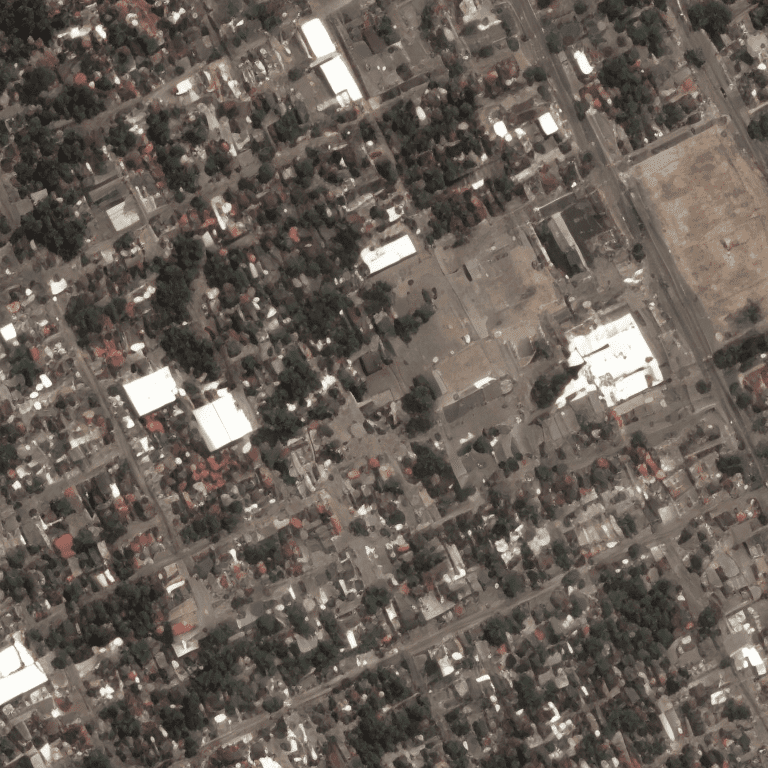}
    \caption{Tower}
\end{subfigure}\\
\begin{subfigure}[t]{0.12\textwidth}
   \centering
    \includegraphics[width=\textwidth]{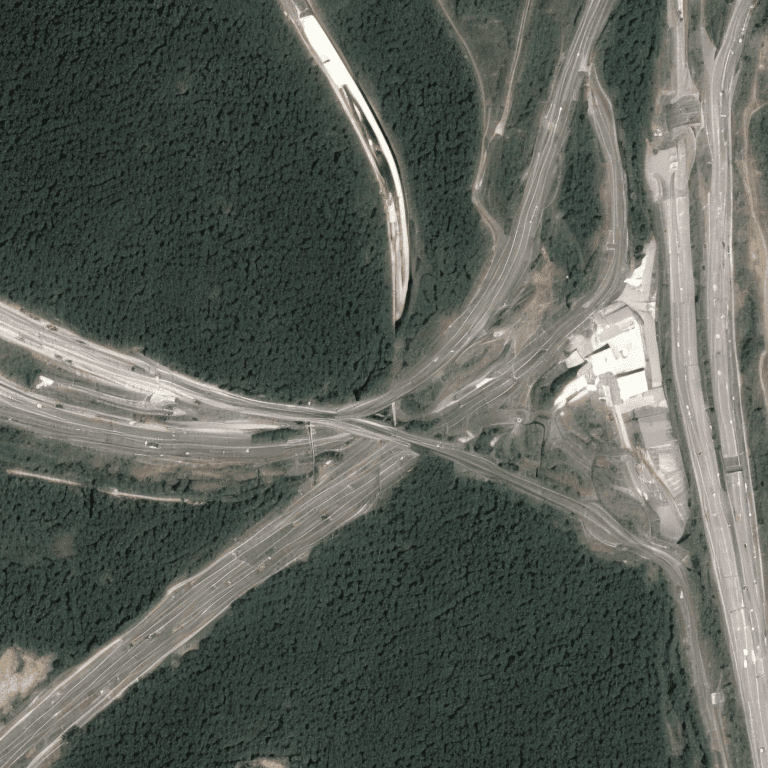}
    \caption{Tunnel Opening}
\end{subfigure}
\begin{subfigure}[t]{0.12\textwidth}
   \centering
    \includegraphics[width=\textwidth]{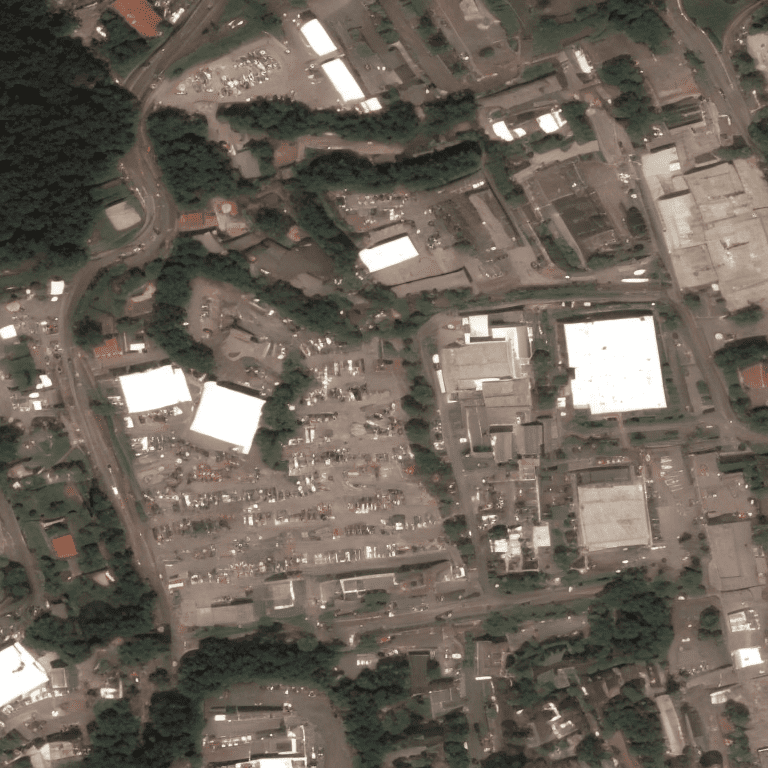}
    \caption{Waste Disposal}
\end{subfigure}
\begin{subfigure}[t]{0.12\textwidth}
   \centering
    \includegraphics[width=\textwidth]{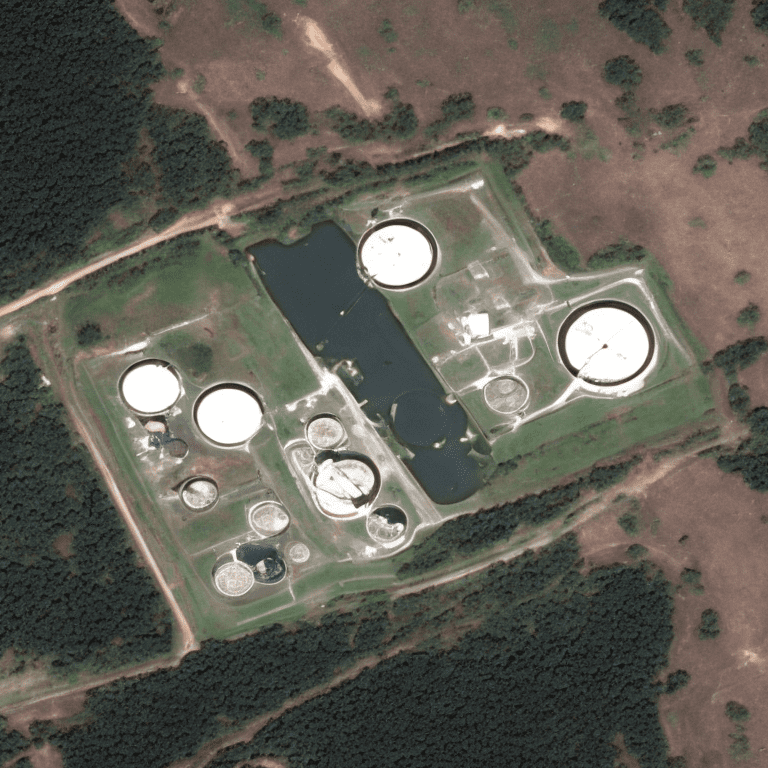}
    \caption{Water Treatment Facility}
\end{subfigure}
\caption{Single-image generations sampled from our model given a prompt of the form "a satellite image of a $<$object class$>$" (prompt-only). This figure acts as supplementary material to Fig.\ref{fig:single_image_overall1}.}
\label{fig:single_image_overall2}
\end{figure*}

\begin{figure*}[h!]
    \centering
    \begin{subfigure}[t]{0.3\textwidth}
        \centering
        \includegraphics[width=\linewidth]{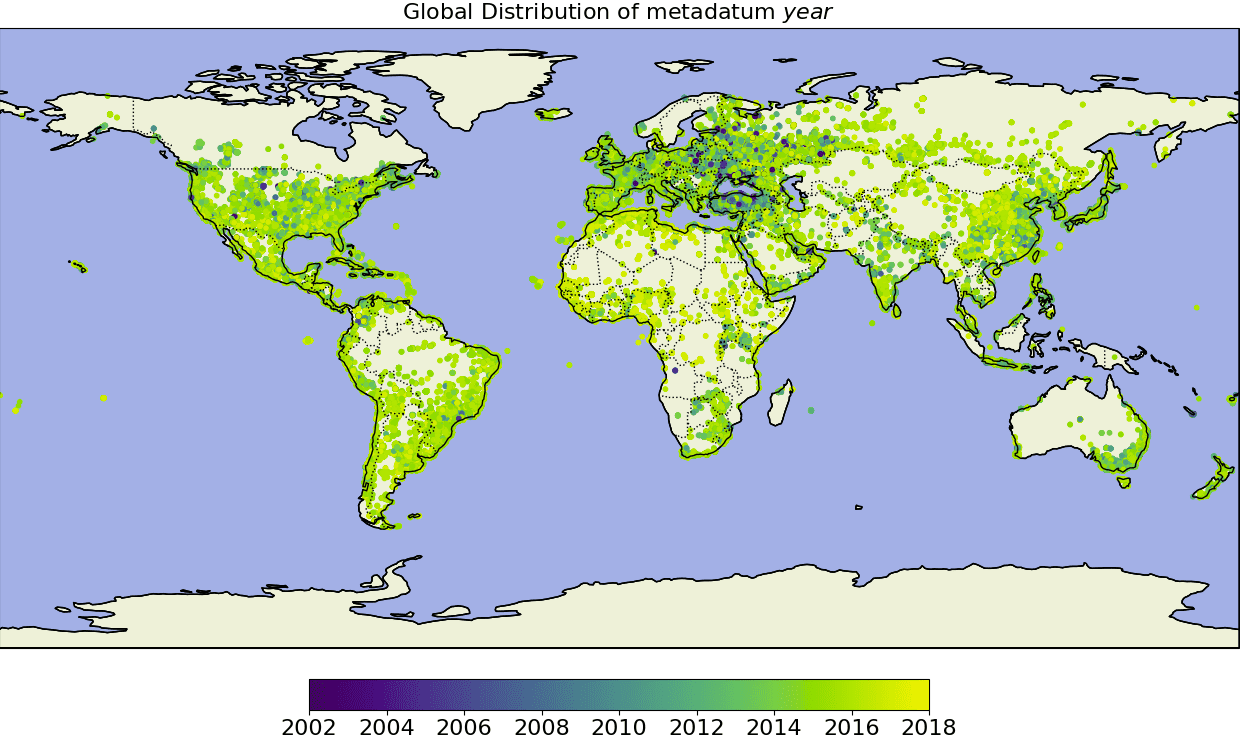}
        \label{fig:year_global}
    \end{subfigure}
    \begin{subfigure}[t]{0.3\textwidth}
        \centering
        \includegraphics[width=\linewidth]{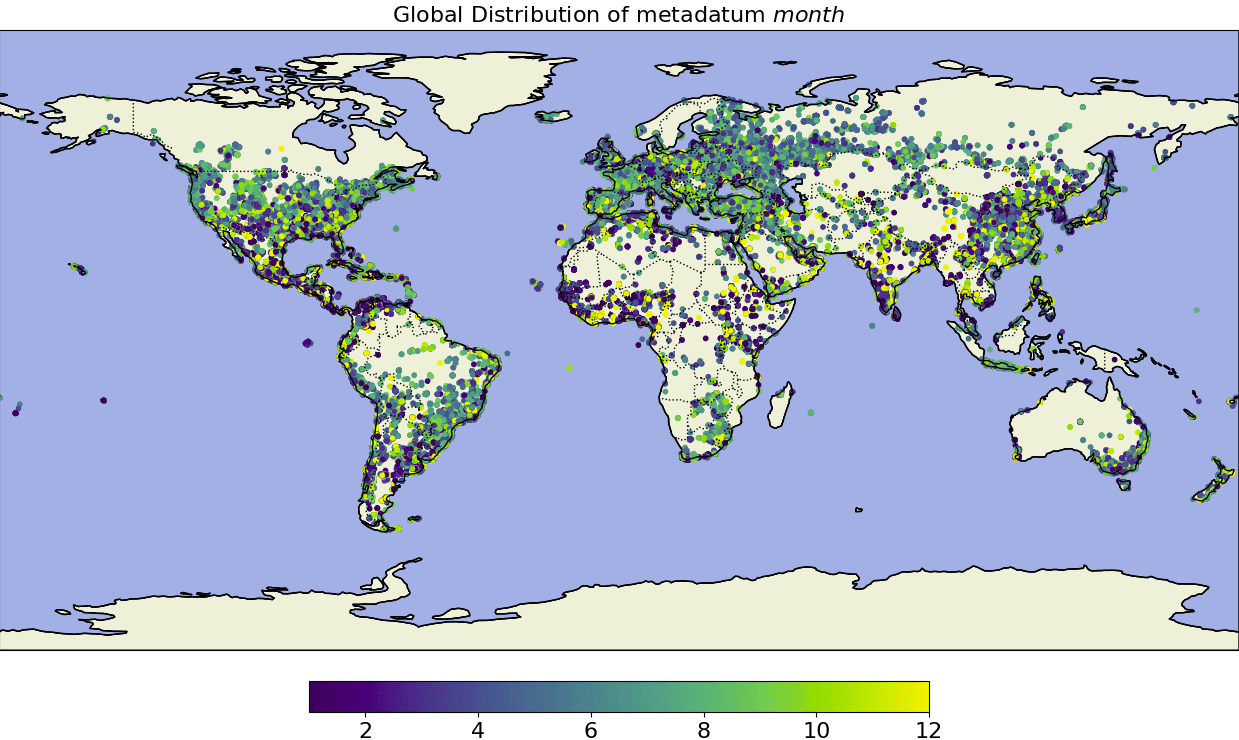}
        \label{fig:month_global}
    \end{subfigure}
    \begin{subfigure}[t]{0.3\textwidth}
        \centering
        \includegraphics[width=\linewidth]{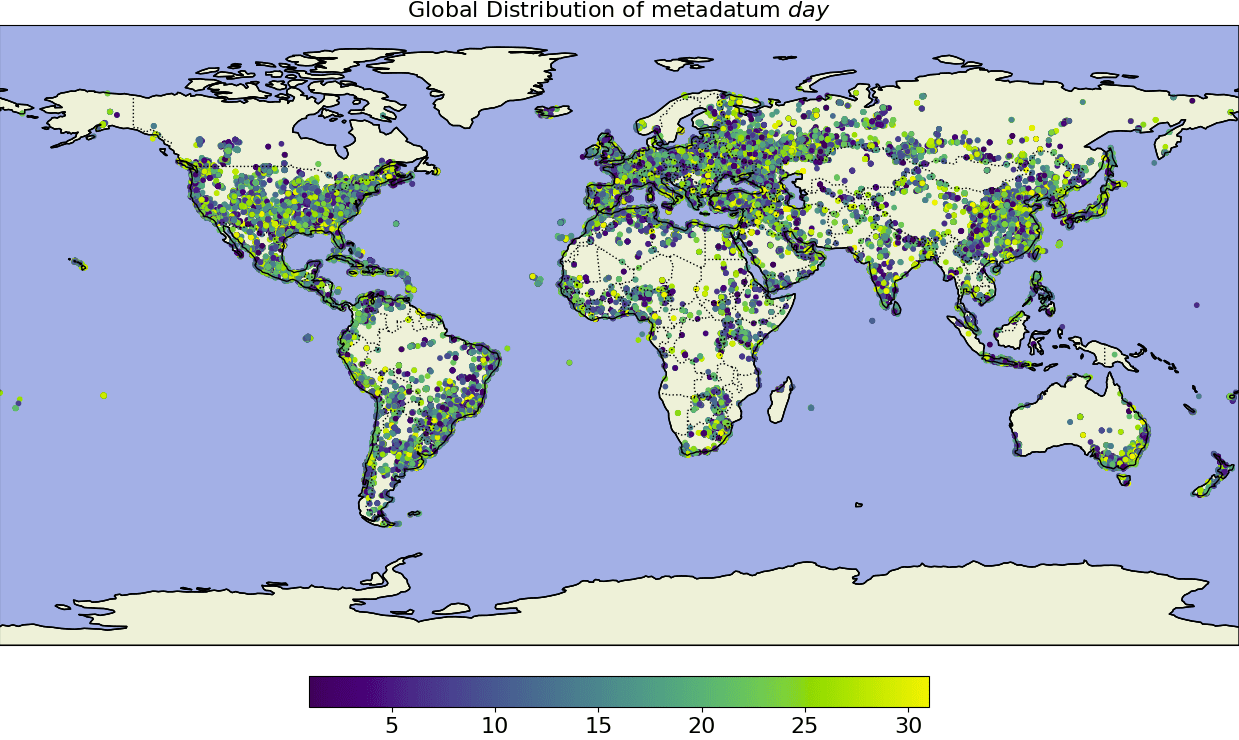}
        \label{fig:day_global}
    \end{subfigure}
    \caption{Global distributions of values for the metadata \{$year$, $month$, $day$\} used in this study. This figure acts as supplementary material to Fig.\ref{fig:metadata_1}.}
    \label{fig:metadata_2}
\end{figure*}

\begin{figure}[h!]
    \centering
    \begin{subfigure}[t]{0.45\textwidth}
        \centering
        \includegraphics[width=\linewidth]{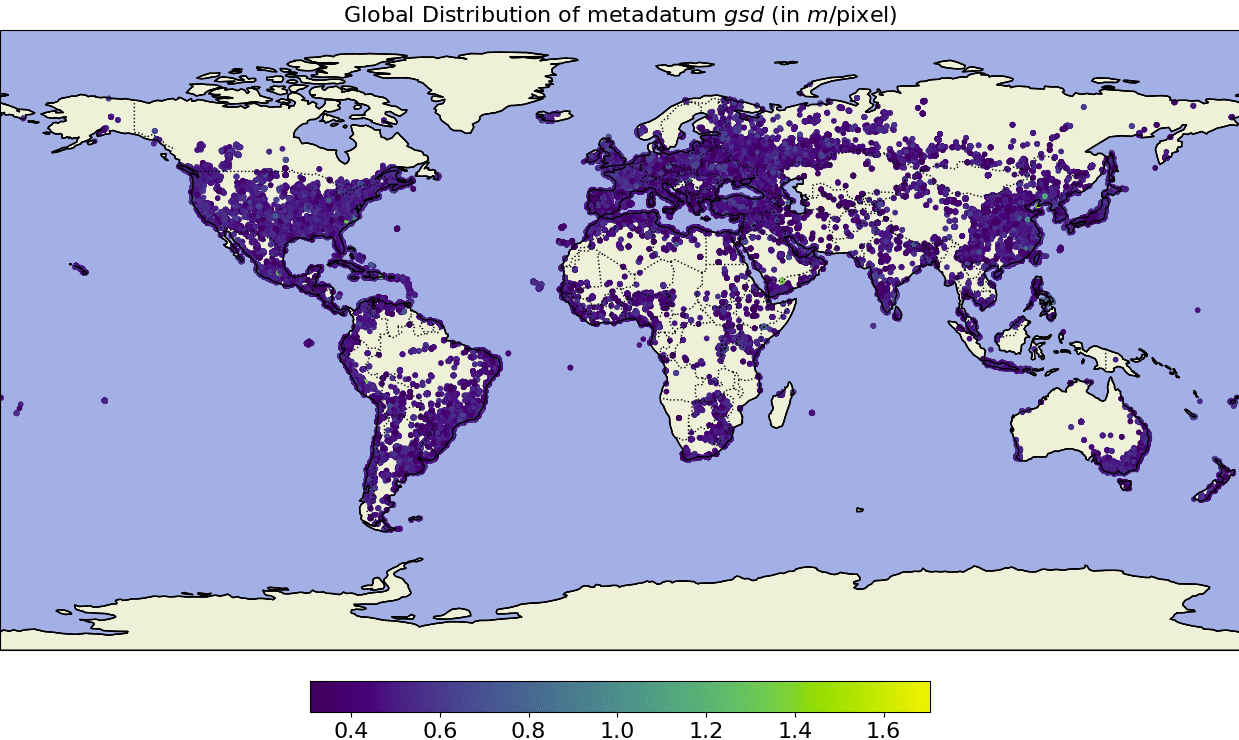}
        \label{fig:gsd_global}
    \end{subfigure}\\
    \begin{subfigure}[t]{0.45\textwidth}
       \centering
        \includegraphics[width=\linewidth]{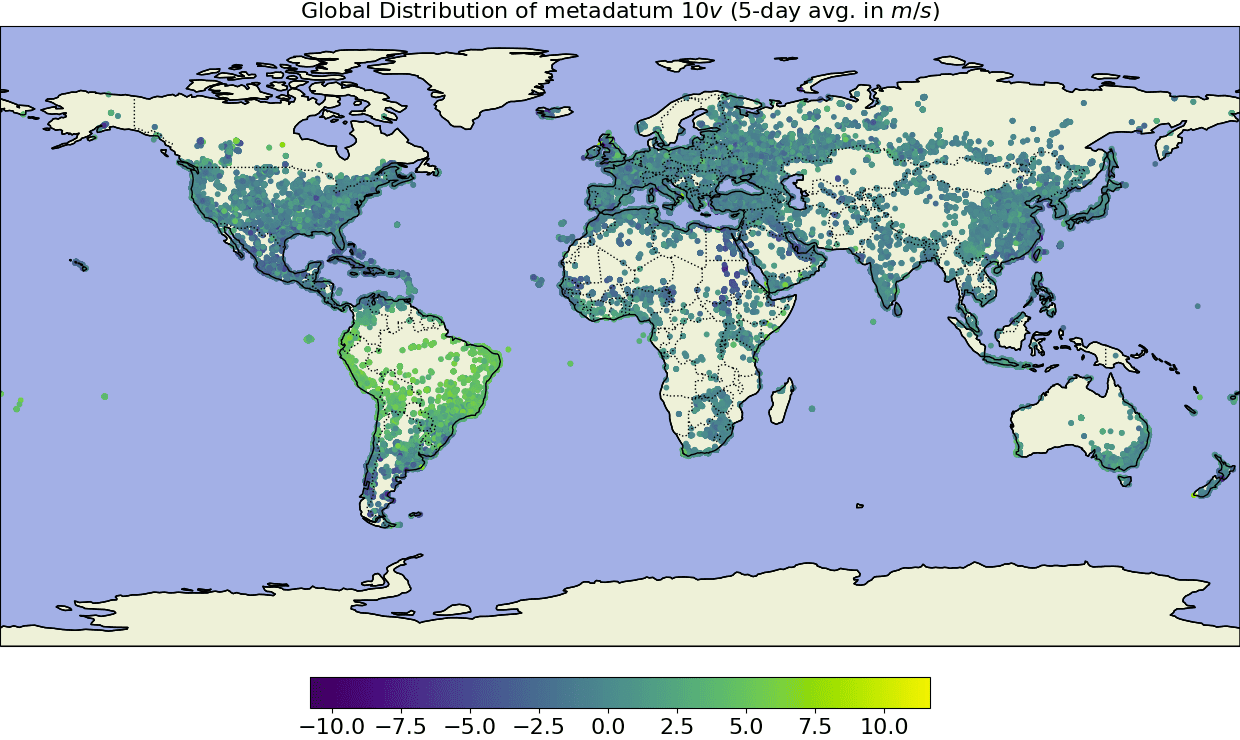}
        \label{fig:v10_global}
    \end{subfigure}\\
    \begin{subfigure}[t]{0.45\textwidth}
       \centering
        \includegraphics[width=\linewidth]{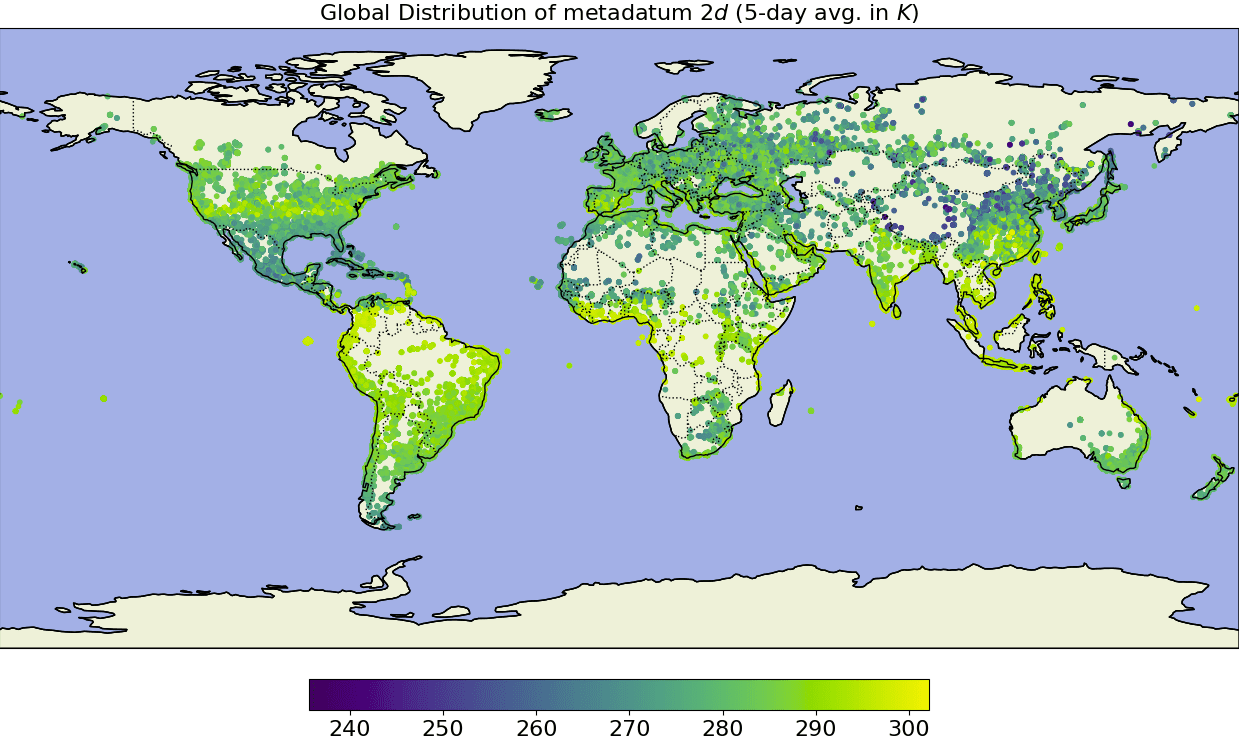}
        \label{fig:2d_global}
    \end{subfigure}\\
    \begin{subfigure}[t]{0.45\textwidth}
        \centering
        \includegraphics[width=\linewidth]{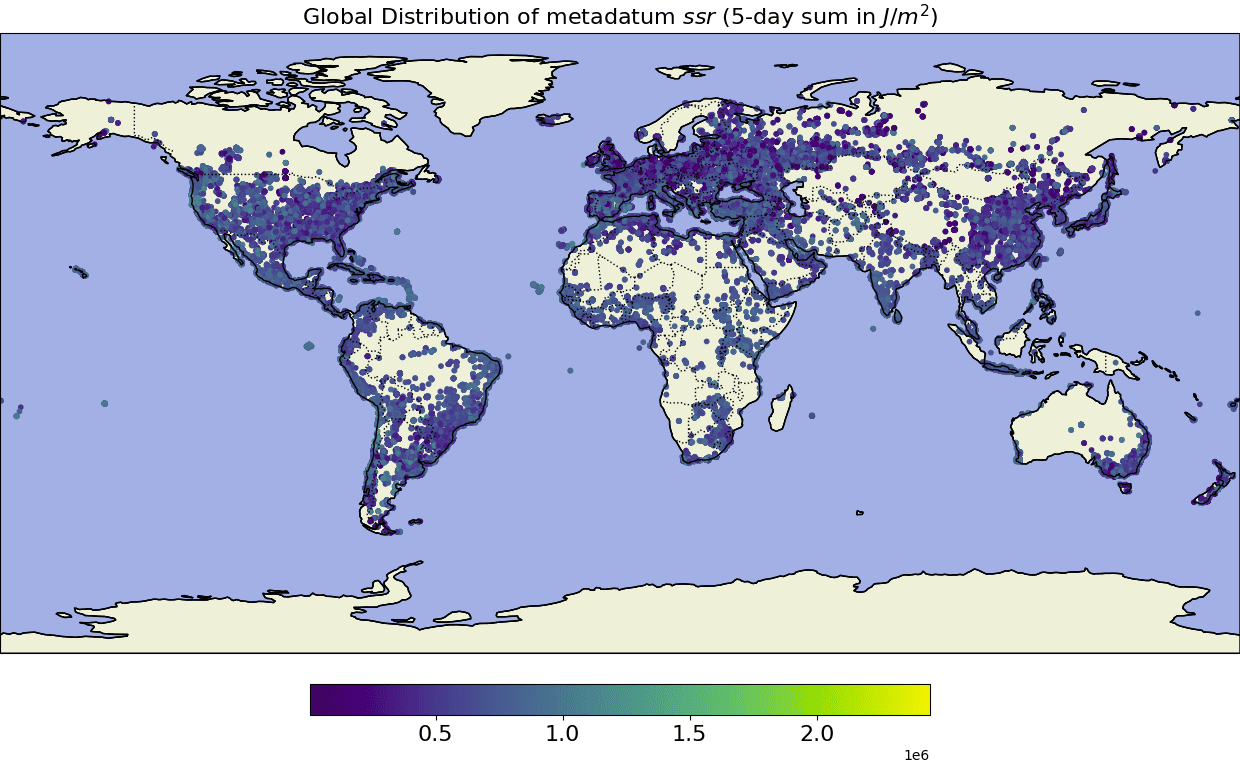}
        \label{fig:ssr_global}
    \end{subfigure}%
    \caption{Global distributions of values for the metadata \{$10v$, $2d$, and $ssr$\} used in this study. This figure acts as supplementary material to Fig.\ref{fig:metadata_1}.}
    \label{fig:metadata_3}
\end{figure}

\begin{figure}[h!]
\includegraphics[width = 0.48\textwidth]{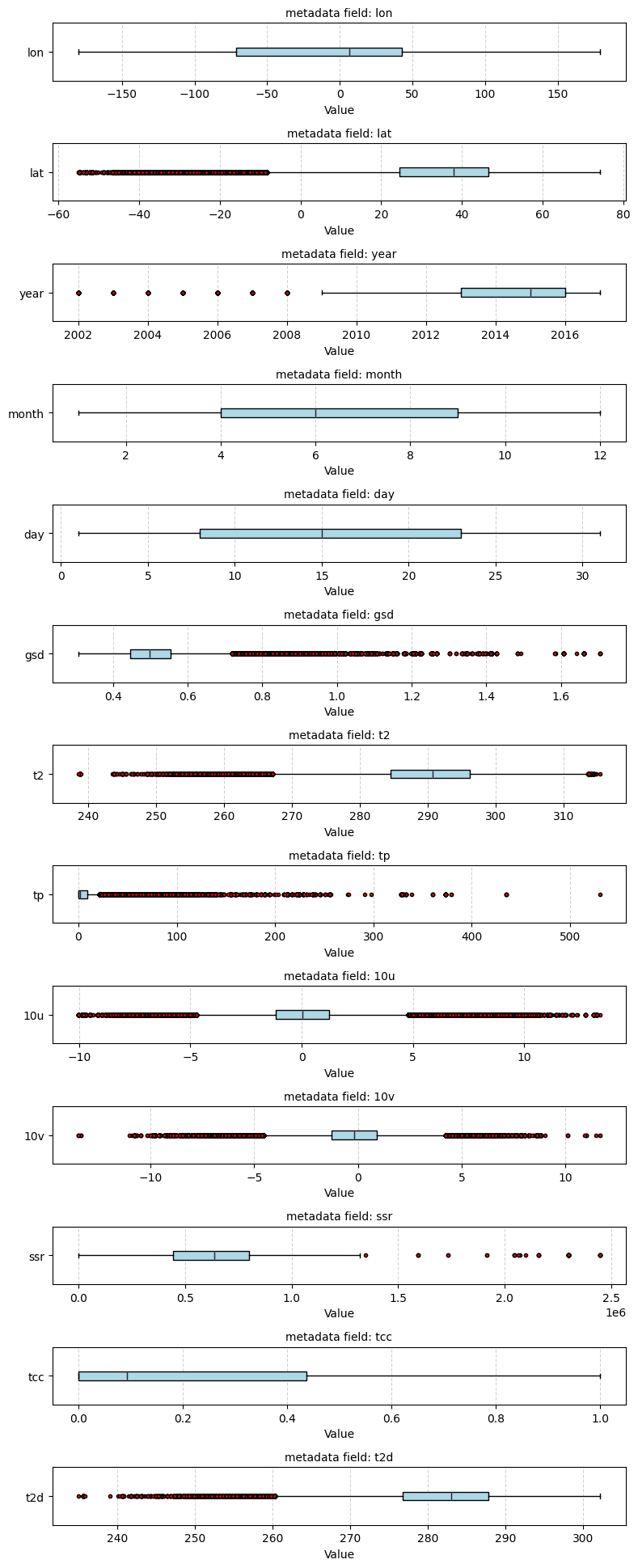}
\caption{Individual boxplots for each metadata variable in our dataset, showing the distribution, median, interquartile range, and outliers. This visualization allows for direct comparison of the value spread and skewness of each variable in its original scale, without normalization.}
\label{fig:dataset_boxplots}
\end{figure}

\begin{figure*}[ht]
    \centering
    \begin{tabular}{ll|l|l|ll}
    \multicolumn{3}{c|}{\textit{t' \textless t}}   & \multicolumn{3}{c}{\textit{t' \textgreater t}}\\
    \textbf{DiffusionSAT} & \textbf{Ours} & \textbf{G.T.} & \textbf{G.T} & \textbf{Ours} & \textbf{DiffusionSAT}\\
       \includegraphics[width=0.13\textwidth, clip=true, trim={9cm 0 0 0}]{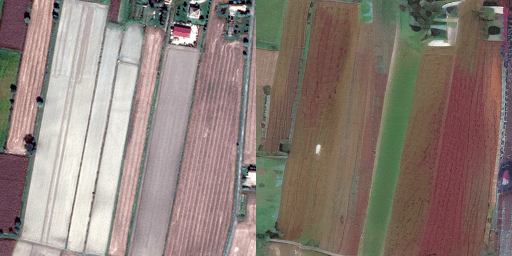} & 
       \includegraphics[width=0.13\textwidth, clip=true, trim={9cm 0 0 0}]{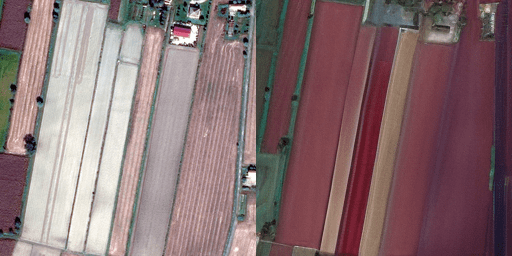} & 
       \includegraphics[width=0.13\textwidth, clip=true, trim={0 0 9cm 0}]{temporal_qualitative/ours/img2img_crop_field_17_1,2,3,0_gs=1.0_iters=20_backward_0.png} & 
       \includegraphics[width=0.13\textwidth, clip=true, trim={0 0 9cm 0}]{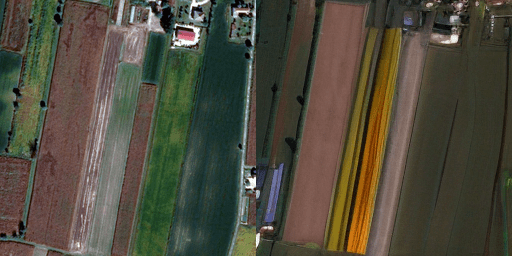} & 
       \includegraphics[width=0.13\textwidth, clip=true, trim={9cm 0 0 0}]{temporal_qualitative/ours/img2img_crop_field_17_0,1,2,3_gs=1.0_iters=20_forward_0.png} &
       \includegraphics[width=0.13\textwidth, clip=true, trim={9cm 0 0 0}]{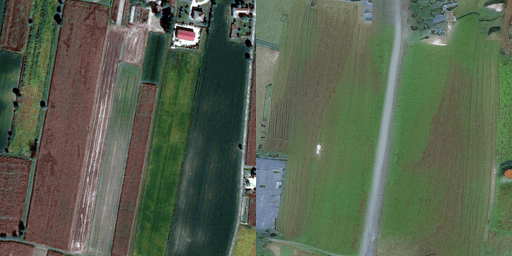}\\
       \includegraphics[width=0.13\textwidth, clip=true, trim={9cm 0 0 0}]{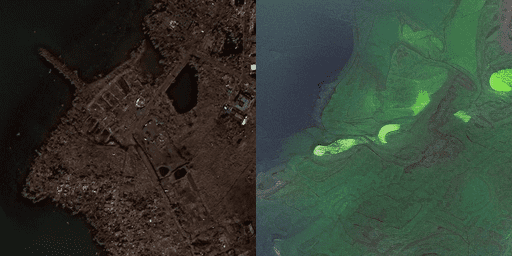} & 
       \includegraphics[width=0.13\textwidth, clip=true, trim={9cm 0 0 0}]{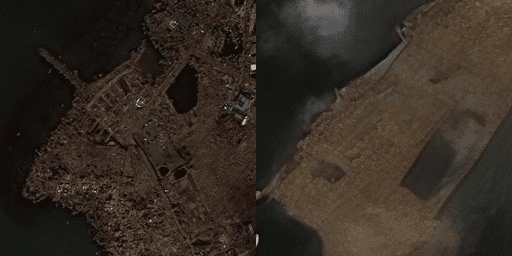} & 
       \includegraphics[width=0.13\textwidth, clip=true, trim={0 0 9cm 0}]{temporal_qualitative/ours/img2img_debris_or_rubble_0_1,2,3,0_gs=1.0_iters=20_backward_0.png} & 
       \includegraphics[width=0.13\textwidth, clip=true, trim={0 0 9cm 0}]{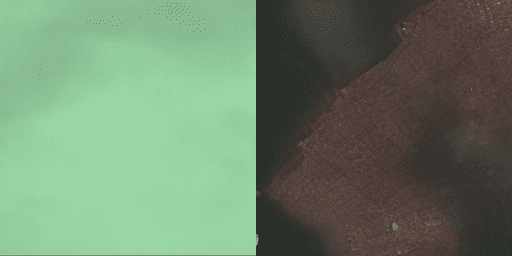} & 
       \includegraphics[width=0.13\textwidth, clip=true, trim={9cm 0 0 0}]{temporal_qualitative/ours/img2img_debris_or_rubble_0_2,3,4,5_gs=1.0_iters=20_forward_0.png} & 
       \includegraphics[width=0.13\textwidth, clip=true, trim={9cm 0 0 0}]{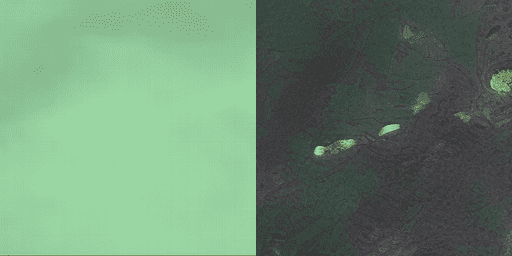}\\
       \includegraphics[width=0.13\textwidth, clip=true, trim={9cm 0 0 0}]{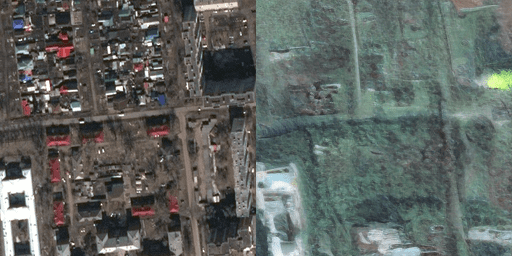} & 
       \includegraphics[width=0.13\textwidth, clip=true, trim={9cm 0 0 0}]{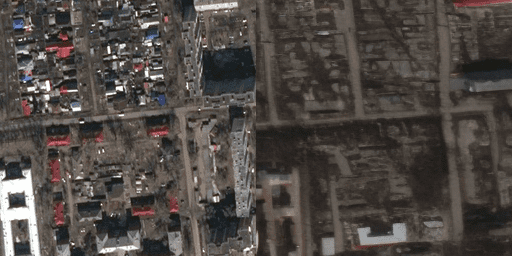} & 
       \includegraphics[width=0.13\textwidth, clip=true, trim={0 0 9cm 0}]{temporal_qualitative/ours/img2img_debris_or_rubble_3_4,5,6,0_gs=1.0_iters=20_backward_0.png} & 
       \includegraphics[width=0.13\textwidth, clip=true, trim={0 0 9cm 0}]{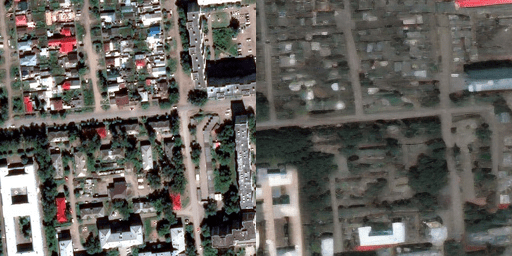} & 
       \includegraphics[width=0.13\textwidth, clip=true, trim={9cm 0 0 0}]{temporal_qualitative/ours/img2img_debris_or_rubble_3_4,5,6,12_gs=1.0_iters=20_forward_0.png} & 
       \includegraphics[width=0.13\textwidth, clip=true, trim={9cm 0 0 0}]{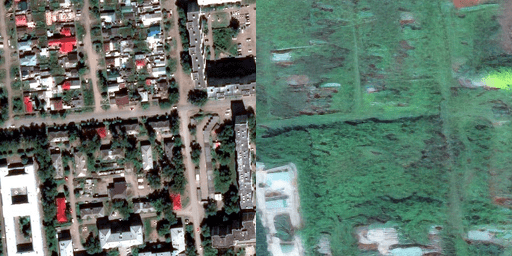}\\
       \includegraphics[width=0.13\textwidth, clip=true, trim={9cm 0 0 0}]{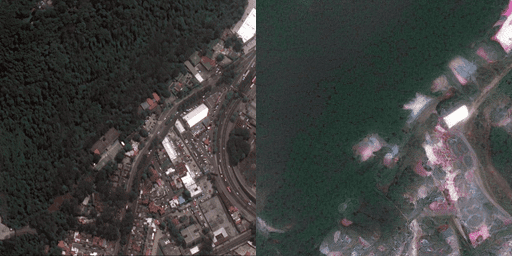} & 
       \includegraphics[width=0.13\textwidth, clip=true, trim={9cm 0 0 0}]{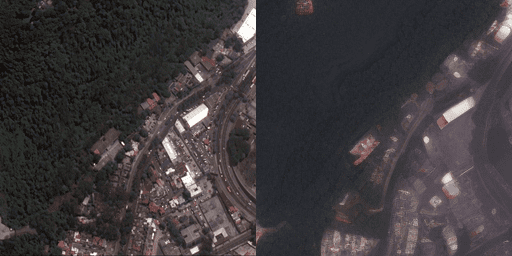} & 
       \includegraphics[width=0.13\textwidth, clip=true, trim={0 0 9cm 0}]{temporal_qualitative/ours/img2img_educational_institution_0_3,4,5,0_gs=1.0_iters=20_backward_0.png} & 
       \includegraphics[width=0.13\textwidth, clip=true, trim={0 0 9cm 0}]{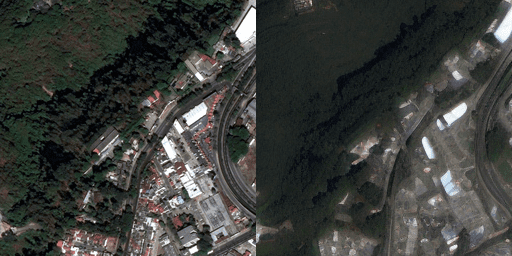} & 
       \includegraphics[width=0.13\textwidth, clip=true, trim={9cm 0 0 0}]{temporal_qualitative/ours/img2img_educational_institution_0_4,5,6,8_gs=1.0_iters=20_forward_0.png} & 
       \includegraphics[width=0.13\textwidth, clip=true, trim={9cm 0 0 0}]{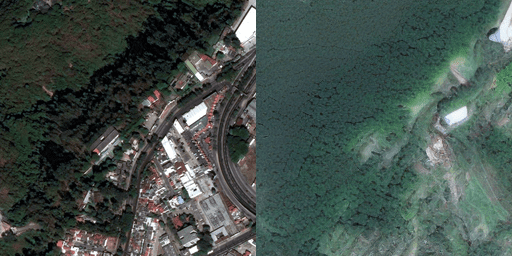}\\
       \includegraphics[width=0.13\textwidth, clip=true, trim={9cm 0 0 0}]{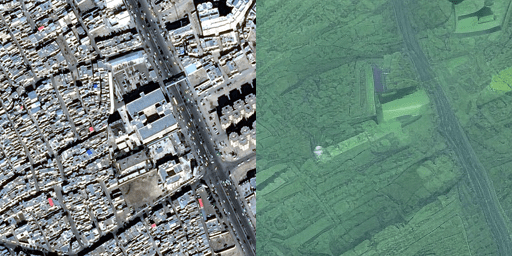} & 
       \includegraphics[width=0.13\textwidth, clip=true, trim={9cm 0 0 0}]{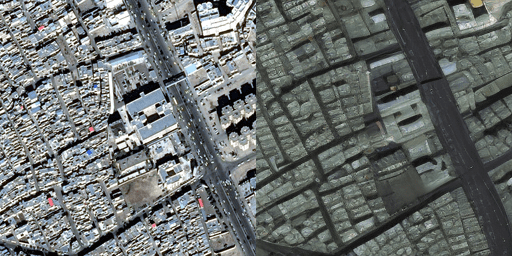} & 
       \includegraphics[width=0.13\textwidth, clip=true, trim={0 0 9cm 0}]{temporal_qualitative/ours/img2img_educational_institution_2_1,2,3,0_gs=1.0_iters=20_backward_0.png} & 
       \includegraphics[width=0.13\textwidth, clip=true, trim={0 0 9cm 0}]{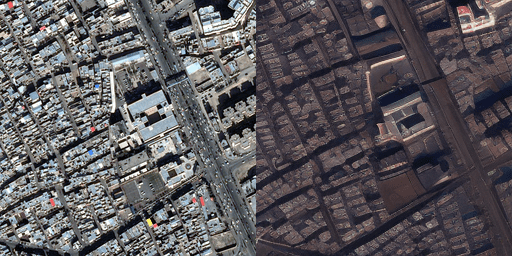} & 
       \includegraphics[width=0.13\textwidth, clip=true, trim={9cm 0 0 0}]{temporal_qualitative/ours/img2img_educational_institution_2_0,1,2,3_gs=1.0_iters=20_forward_0.png} & 
       \includegraphics[width=0.13\textwidth, clip=true, trim={9cm 0 0 0}]{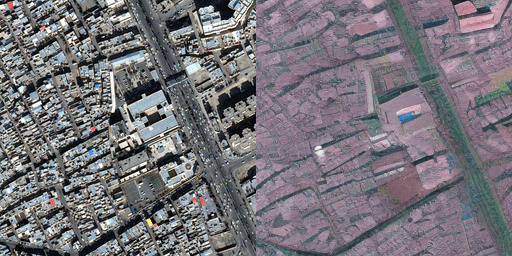}\\
       \includegraphics[width=0.13\textwidth, clip=true, trim={9cm 0 0 0}]{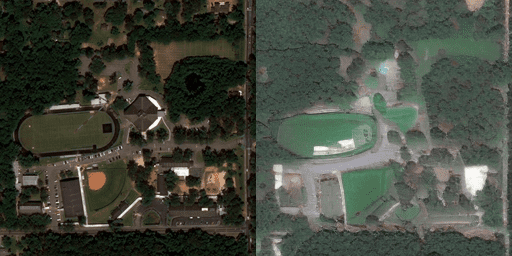} & 
       \includegraphics[width=0.13\textwidth, clip=true, trim={9cm 0 0 0}]{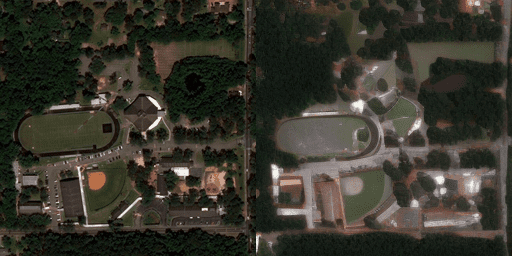} & 
       \includegraphics[width=0.13\textwidth, clip=true, trim={0 0 9cm 0}]{temporal_qualitative/ours/img2img_educational_institution_5_1,3,4,0_gs=1.0_iters=20_backward_0.png} & 
       \includegraphics[width=0.13\textwidth, clip=true, trim={0 0 9cm 0}]{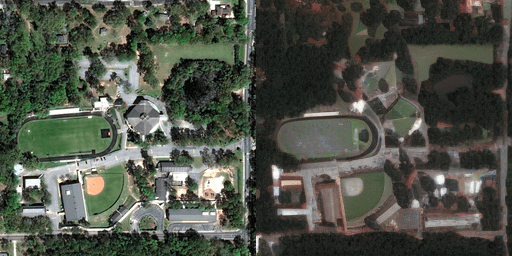} & 
       \includegraphics[width=0.13\textwidth, clip=true, trim={9cm 0 0 0}]{temporal_qualitative/ours/img2img_educational_institution_5_1,3,4,7_gs=1.0_iters=20_forward_0.png} & 
       \includegraphics[width=0.13\textwidth, clip=true, trim={9cm 0 0 0}]{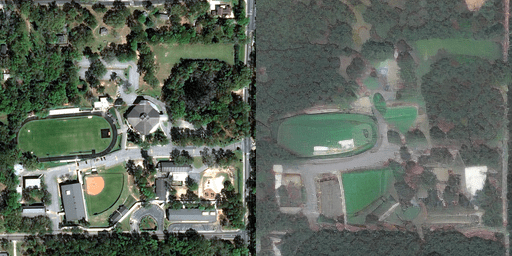}\\
       \includegraphics[width=0.13\textwidth, clip=true, trim={9cm 0 0 0}]{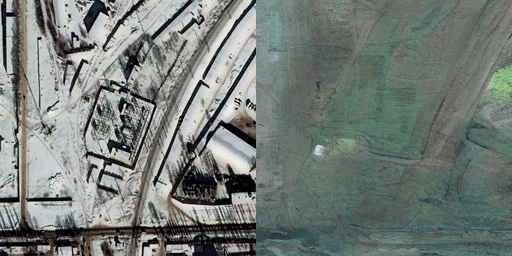} & 
       \includegraphics[width=0.13\textwidth, clip=true, trim={9cm 0 0 0}]{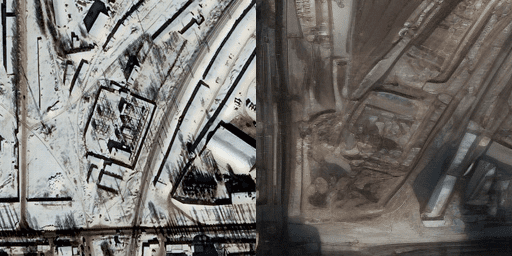} & 
       \includegraphics[width=0.13\textwidth, clip=true, trim={0 0 9cm 0}]{temporal_qualitative/ours/img2img_electric_substation_3_1,4,7,0_gs=1.0_iters=20_backward_0.png} & 
       \includegraphics[width=0.13\textwidth, clip=true, trim={0 0 9cm 0}]{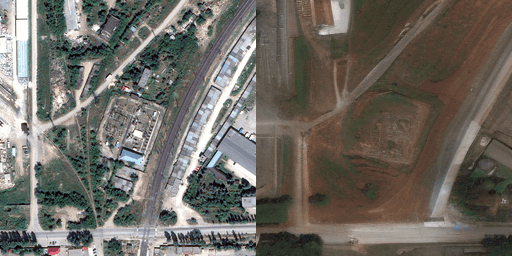} & 
       \includegraphics[width=0.13\textwidth, clip=true, trim={9cm 0 0 0}]{temporal_qualitative/ours/img2img_electric_substation_3_1,4,7,8_gs=1.0_iters=20_forward_0.png} & 
       \includegraphics[width=0.13\textwidth, clip=true, trim={9cm 0 0 0}]{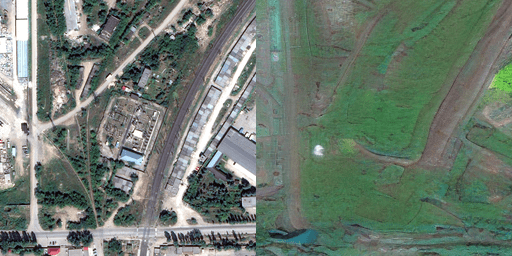}\\
       \includegraphics[width=0.13\textwidth, clip=true, trim={9cm 0 0 0}]{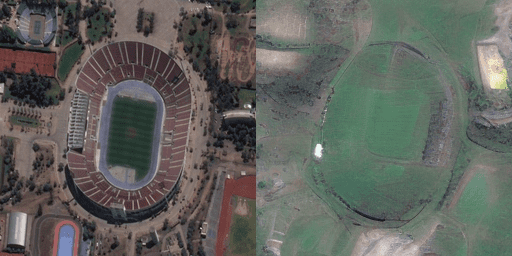} & 
       \includegraphics[width=0.13\textwidth, clip=true, trim={9cm 0 0 0}]{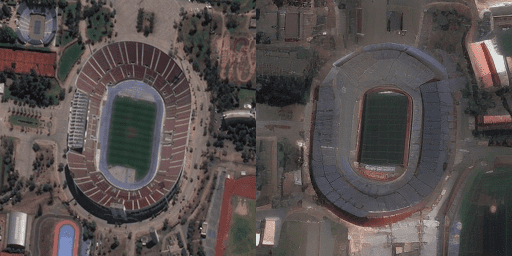} & 
       \includegraphics[width=0.13\textwidth, clip=true, trim={0 0 9cm 0}]{temporal_qualitative/ours/img2img_stadium_11_4,11,17,0_gs=1.0_iters=20_backward_0.png} & 
       \includegraphics[width=0.13\textwidth, clip=true, trim={0 0 9cm 0}]{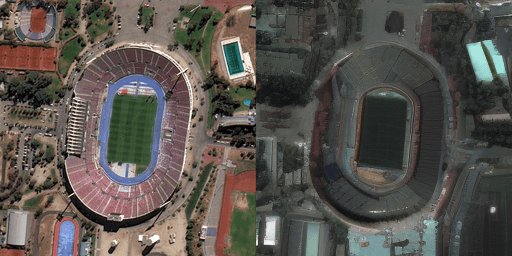} & 
       \includegraphics[width=0.13\textwidth, clip=true, trim={9cm 0 0 0}]{temporal_qualitative/ours/img2img_stadium_11_0,4,11,17_gs=1.0_iters=20_forward_0.png} & 
       \includegraphics[width=0.13\textwidth, clip=true, trim={9cm 0 0 0}]{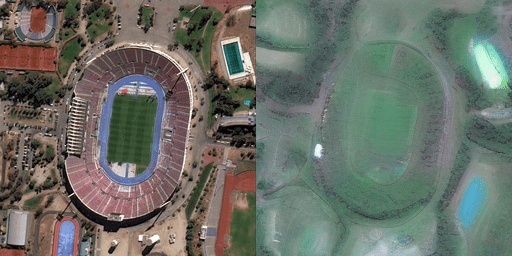}\\
       \includegraphics[width=0.13\textwidth, clip=true, trim={9cm 0 0 0}]{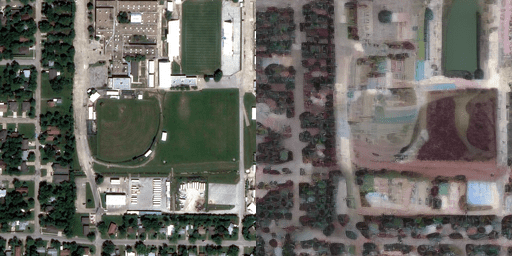} & 
       \includegraphics[width=0.13\textwidth, clip=true, trim={9cm 0 0 0}]{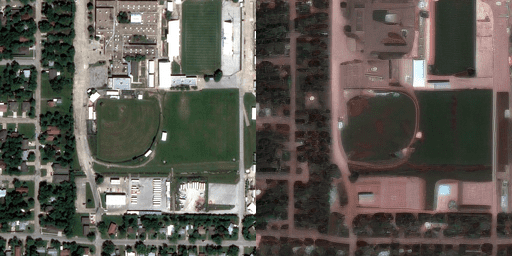} & 
       \includegraphics[width=0.13\textwidth, clip=true, trim={0 0 9cm 0}]{temporal_qualitative/ours/img2img_recreational_facility_2_2,4,6,0_gs=1.0_iters=20_backward_0.png} & 
       \includegraphics[width=0.13\textwidth, clip=true, trim={0 0 9cm 0}]{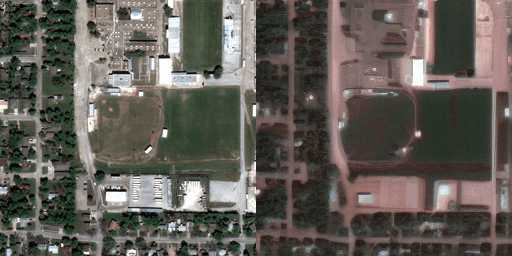} & 
       \includegraphics[width=0.13\textwidth, clip=true, trim={9cm 0 0 0}]{temporal_qualitative/ours/img2img_recreational_facility_2_2,4,6,8_gs=1.0_iters=20_forward_0.png} & 
       \includegraphics[width=0.13\textwidth, clip=true, trim={9cm 0 0 0}]{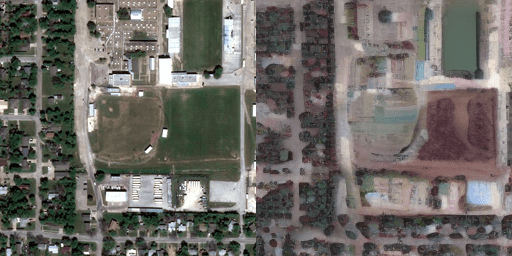}
\end{tabular}
\caption{}
\label{tab:appendix_temporal}
\end{figure*}

\begin{figure*}[ht]
    \centering
    \begin{tabular}{ll|l|l|ll}
    \multicolumn{3}{c|}{\textit{t' \textless t}}   & \multicolumn{3}{c}{\textit{t' \textgreater t}}\\
    \textbf{DiffusionSAT} & \textbf{Ours} & \textbf{G.T.} & \textbf{G.T} & \textbf{Ours} & \textbf{DiffusionSAT}\\
       \includegraphics[width=0.13\textwidth, clip=true, trim={9cm 0 0 0}]{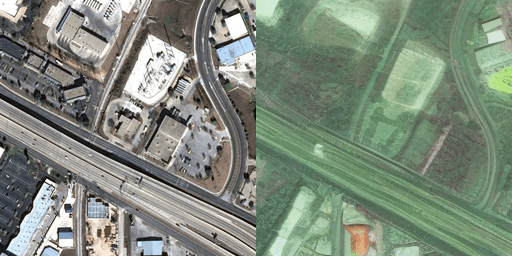} & 
       \includegraphics[width=0.13\textwidth, clip=true, trim={9cm 0 0 0}]{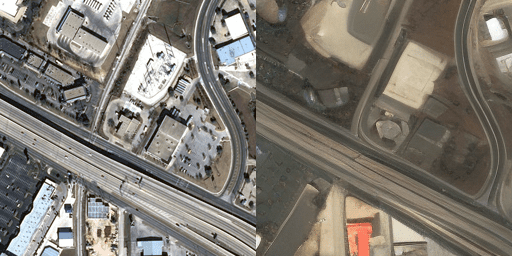} & 
       \includegraphics[width=0.13\textwidth, clip=true, trim={0 0 9cm 0}]{temporal_qualitative/ours/img2img_fire_station_8_1,3,8,0_gs=1.0_iters=20_backward_0.png} & 
       \includegraphics[width=0.13\textwidth, clip=true, trim={0 0 9cm 0}]{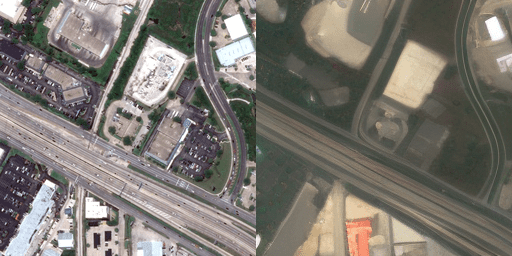} & 
       \includegraphics[width=0.13\textwidth, clip=true, trim={9cm 0 0 0}]{temporal_qualitative/ours/img2img_fire_station_8_1,3,8,9_gs=1.0_iters=20_forward_0.png} &
       \includegraphics[width=0.13\textwidth, clip=true, trim={9cm 0 0 0}]{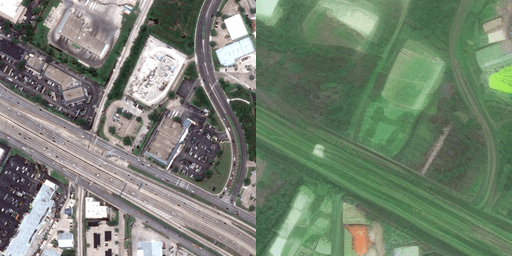}\\
       \includegraphics[width=0.13\textwidth, clip=true, trim={9cm 0 0 0}]{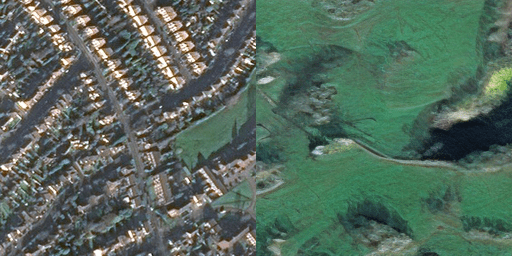} & 
       \includegraphics[width=0.13\textwidth, clip=true, trim={9cm 0 0 0}]{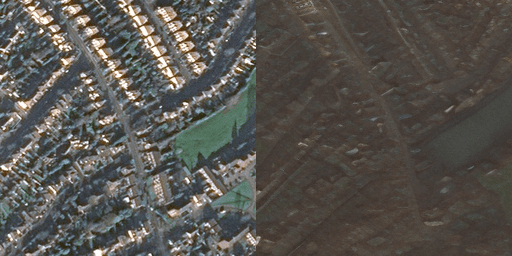} & 
       \includegraphics[width=0.13\textwidth, clip=true, trim={0 0 9cm 0}]{temporal_qualitative/ours/img2img_flooded_road_5_2,4,5,1_gs=1.0_iters=20_backward_0.png} & 
       \includegraphics[width=0.13\textwidth, clip=true, trim={0 0 9cm 0}]{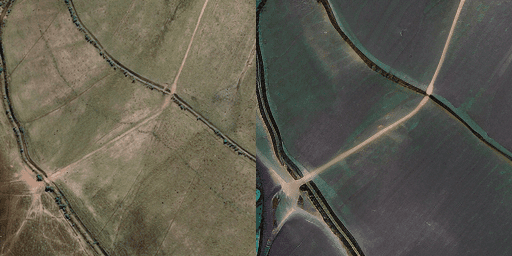} & 
       \includegraphics[width=0.13\textwidth, clip=true, trim={9cm 0 0 0}]{temporal_qualitative/ours/img2img_flooded_road_3_3,6,8,9_gs=1.0_iters=20_forward_0.png} & 
       \includegraphics[width=0.13\textwidth, clip=true, trim={9cm 0 0 0}]{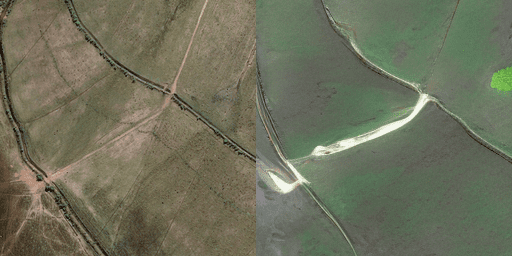}\\
       \includegraphics[width=0.13\textwidth, clip=true, trim={9cm 0 0 0}]{temporal_qualitative/ermon/img2img_flooded_road_5_2,4,5,1_gs=1.0_iters=20_backward_0.png} & 
       \includegraphics[width=0.13\textwidth, clip=true, trim={9cm 0 0 0}]{temporal_qualitative/ours/img2img_flooded_road_5_2,4,5,1_gs=1.0_iters=20_backward_0.png} & 
       \includegraphics[width=0.13\textwidth, clip=true, trim={0 0 9cm 0}]{temporal_qualitative/ours/img2img_flooded_road_5_2,4,5,1_gs=1.0_iters=20_backward_0.png} & 
       \includegraphics[width=0.13\textwidth, clip=true, trim={0 0 9cm 0}]{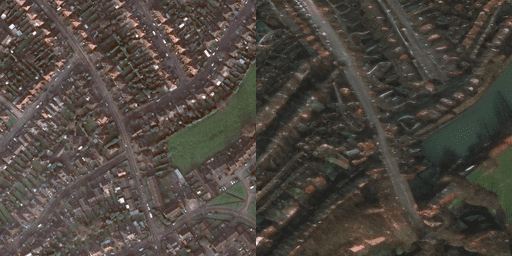} & 
       \includegraphics[width=0.13\textwidth, clip=true, trim={9cm 0 0 0}]{temporal_qualitative/ours/img2img_flooded_road_5_2,4,5,9_gs=1.0_iters=20_forward_0.png} & 
       \includegraphics[width=0.13\textwidth, clip=true, trim={9cm 0 0 0}]{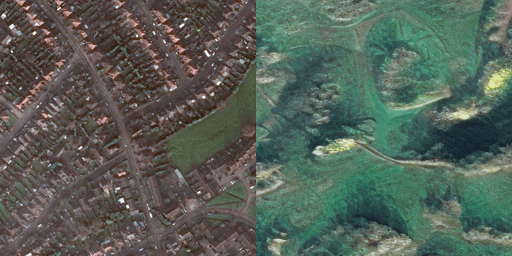}\\
       \includegraphics[width=0.13\textwidth, clip=true, trim={9cm 0 0 0}]{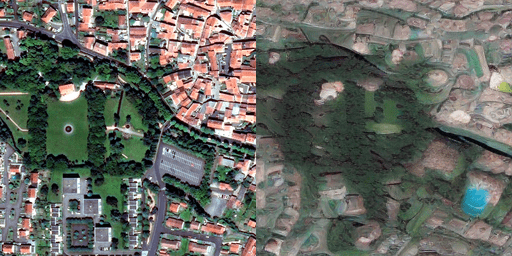} & 
       \includegraphics[width=0.13\textwidth, clip=true, trim={9cm 0 0 0}]{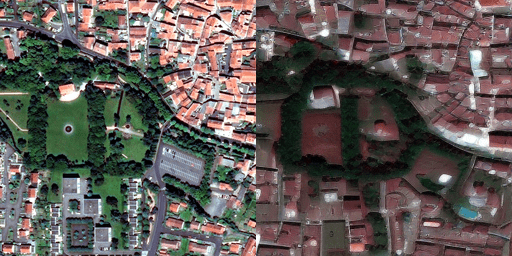} & 
       \includegraphics[width=0.13\textwidth, clip=true, trim={0 0 9cm 0}]{temporal_qualitative/ours/img2img_fountain_5_1,2,3,0_gs=1.0_iters=20_backward_0.png} & 
       \includegraphics[width=0.13\textwidth, clip=true, trim={0 0 9cm 0}]{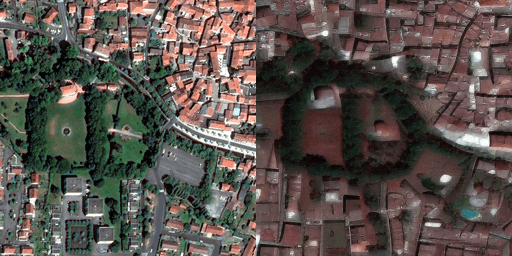} & 
       \includegraphics[width=0.13\textwidth, clip=true, trim={9cm 0 0 0}]{temporal_qualitative/ours/img2img_fountain_5_1,2,3,4_gs=1.0_iters=20_forward_0.png} & 
       \includegraphics[width=0.13\textwidth, clip=true, trim={9cm 0 0 0}]{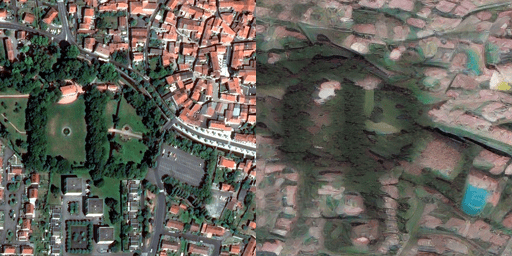}\\
       \includegraphics[width=0.13\textwidth, clip=true, trim={9cm 0 0 0}]{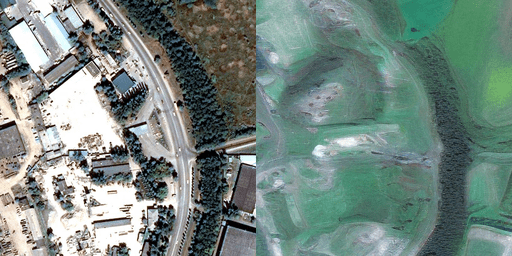} & 
       \includegraphics[width=0.13\textwidth, clip=true, trim={9cm 0 0 0}]{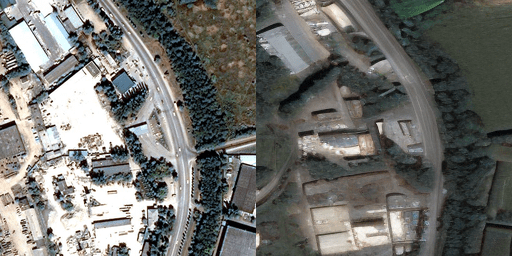} & 
       \includegraphics[width=0.13\textwidth, clip=true, trim={0 0 9cm 0}]{temporal_qualitative/ours/img2img_gas_station_0_5,10,17,0_gs=1.0_iters=20_backward_0.png} & 
       \includegraphics[width=0.13\textwidth, clip=true, trim={0 0 9cm 0}]{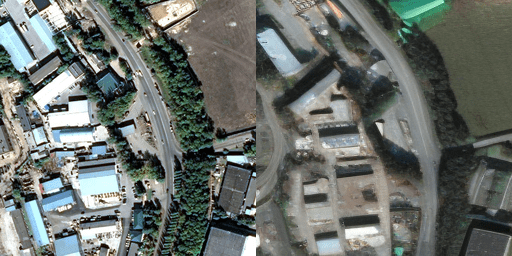} & 
       \includegraphics[width=0.13\textwidth, clip=true, trim={9cm 0 0 0}]{temporal_qualitative/ours/img2img_gas_station_0_5,10,17,19_gs=1.0_iters=20_forward_0.png} & 
       \includegraphics[width=0.13\textwidth, clip=true, trim={9cm 0 0 0}]{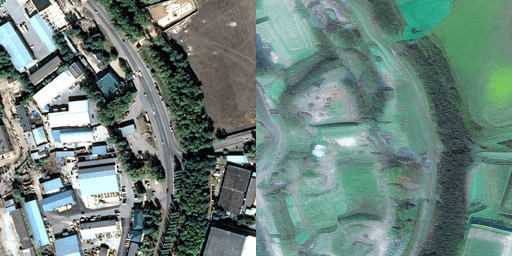}\\
       \includegraphics[width=0.13\textwidth, clip=true, trim={9cm 0 0 0}]{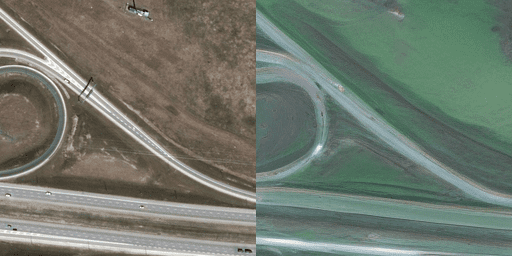} & 
       \includegraphics[width=0.13\textwidth, clip=true, trim={9cm 0 0 0}]{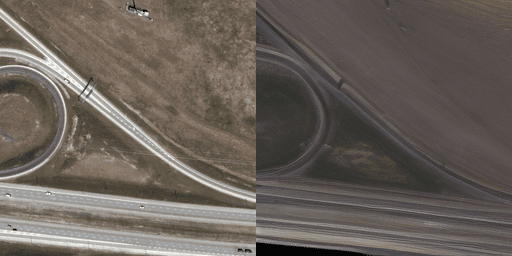} & 
       \includegraphics[width=0.13\textwidth, clip=true, trim={0 0 9cm 0}]{temporal_qualitative/ours/img2img_interchange_1_1,2,3,0_gs=1.0_iters=20_backward_0.png} & 
       \includegraphics[width=0.13\textwidth, clip=true, trim={0 0 9cm 0}]{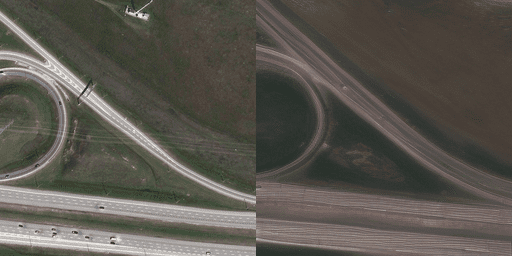} & 
       \includegraphics[width=0.13\textwidth, clip=true, trim={9cm 0 0 0}]{temporal_qualitative/ours/img2img_interchange_1_0,1,2,3_gs=1.0_iters=20_forward_0.png} & 
       \includegraphics[width=0.13\textwidth, clip=true, trim={9cm 0 0 0}]{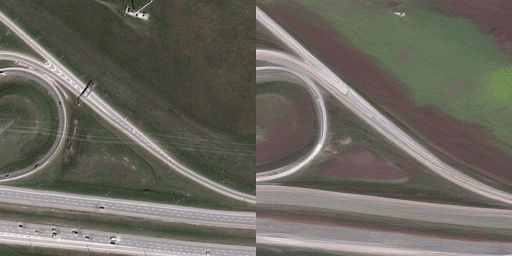}\\
       \includegraphics[width=0.13\textwidth, clip=true, trim={9cm 0 0 0}]{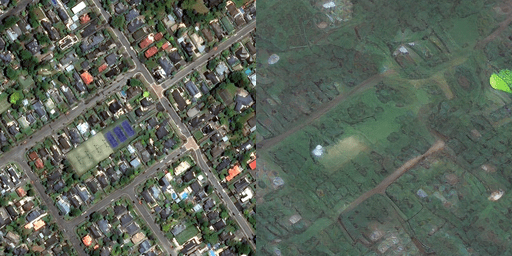} & 
       \includegraphics[width=0.13\textwidth, clip=true, trim={9cm 0 0 0}]{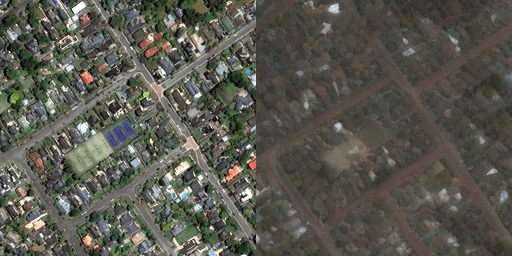} & 
       \includegraphics[width=0.13\textwidth, clip=true, trim={0 0 9cm 0}]{temporal_qualitative/ours/img2img_recreational_facility_1_1,2,3,0_gs=1.0_iters=20_backward_0.png} & 
       \includegraphics[width=0.13\textwidth, clip=true, trim={0 0 9cm 0}]{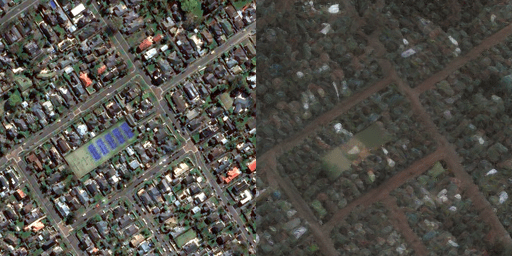} & 
       \includegraphics[width=0.13\textwidth, clip=true, trim={9cm 0 0 0}]{temporal_qualitative/ours/img2img_recreational_facility_1_1,2,3,4_gs=1.0_iters=20_forward_0.png} & 
       \includegraphics[width=0.13\textwidth, clip=true, trim={9cm 0 0 0}]{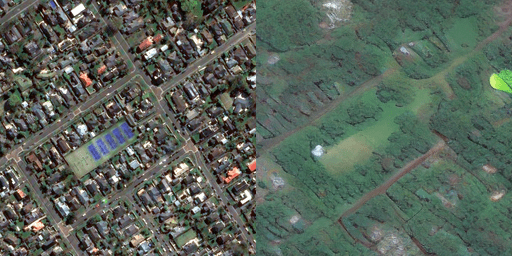}\\
       \includegraphics[width=0.13\textwidth, clip=true, trim={9cm 0 0 0}]{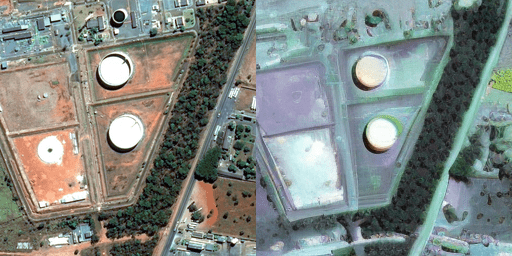} & 
       \includegraphics[width=0.13\textwidth, clip=true, trim={9cm 0 0 0}]{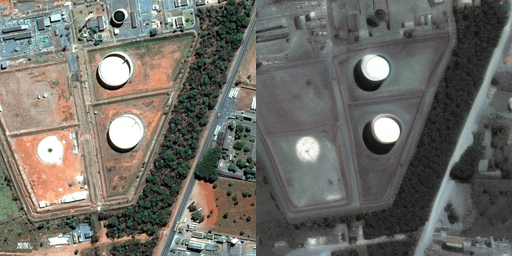} & 
       \includegraphics[width=0.13\textwidth, clip=true, trim={0 0 9cm 0}]{temporal_qualitative/ours/img2img_storage_tank_4_2,6,8,0_gs=1.0_iters=20_backward_0.png} & 
       \includegraphics[width=0.13\textwidth, clip=true, trim={0 0 9cm 0}]{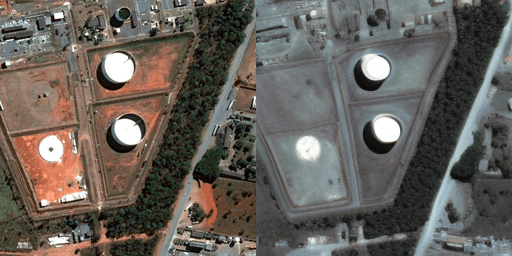} & 
       \includegraphics[width=0.13\textwidth, clip=true, trim={9cm 0 0 0}]{temporal_qualitative/ours/img2img_storage_tank_4_2,6,8,9_gs=1.0_iters=20_forward_0.png} & 
       \includegraphics[width=0.13\textwidth, clip=true, trim={9cm 0 0 0}]{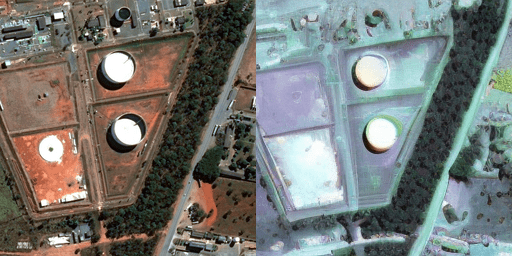}\\
       \includegraphics[width=0.13\textwidth, clip=true, trim={9cm 0 0 0}]{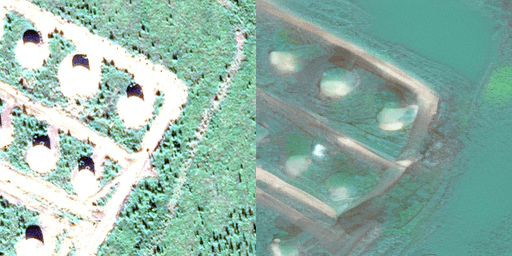} & 
       \includegraphics[width=0.13\textwidth, clip=true, trim={9cm 0 0 0}]{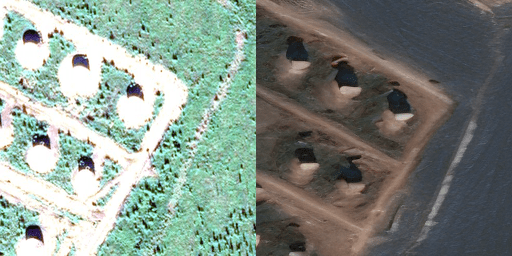} & 
       \includegraphics[width=0.13\textwidth, clip=true, trim={0 0 9cm 0}]{temporal_qualitative/ours/img2img_storage_tank_5_5,6,8,0_gs=1.0_iters=20_backward_0.png} & 
       \includegraphics[width=0.13\textwidth, clip=true, trim={0 0 9cm 0}]{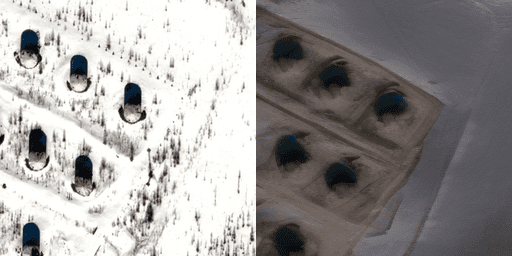} & 
       \includegraphics[width=0.13\textwidth, clip=true, trim={9cm 0 0 0}]{temporal_qualitative/ours/img2img_storage_tank_5_5,6,8,9_gs=1.0_iters=20_forward_0.png} & 
       \includegraphics[width=0.13\textwidth, clip=true, trim={9cm 0 0 0}]{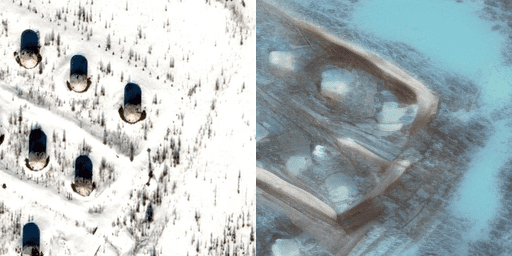}
\end{tabular}
\caption{}
\label{tab:appendix_temporal_b}
\end{figure*}

\end{document}